# CardioLoom: A Report-Guided Electrocardiogram Foundation Model for Cross-Domain Cardiovascular Phenotyping

**Ziqing Yu†[1], Yuhui Tao†[2,3], Jiayu Huo†[4], Lei Pan[1], Zilong Xiao[1], Juecheng Chen[1], Xiao Li[1], Jianxuan Li[1], You Zhou[5], Zhixing Li[1], Cong Wang[1], Beijian Zhang[1], Chen Chen[11], Hongyang Lu[6], Konstantinos Patlatzoglou[4], Daniel B. Kramer[7], Jonathan W. Waks[8], Yangang Su[1], Fu Siong Ng[4,9,10], Shuo Wang[2,3], Yixiu Liang[1,4], Junbo Ge[1]**

1 Department of Cardiology, Zhongshan Hospital Fudan University, Shanghai Institute of Cardiovascular Diseases, National Clinical Research Centre for Interventional Medicine, Shanghai, China

2 Digital Medical Research Center, School of Basic Medical Sciences, Fudan University, Shanghai, 200032, China.

3 Shanghai Key Laboratory of Medical Imaging Computing and Computer Assisted Intervention, Shanghai, 200032, China.

4 National Heart and Lung Institute, Imperial College London, Hammersmith Hospital, Du Cane Road, London W120HS, UK

5 Department of Cardiology, Shanghai Geriatric Medical Center, Shanghai, China

6 Cardiac Rhythm Management, Medtronic Technology Center, Medtronic (Shanghai) Ltd., Shanghai, China

7 Richard A. and Susan F. Smith Center for Outcomes Research in Cardiology, Beth Israel Deaconess Medical Center, Harvard Medical School, Boston MA USA

8 Harvard-Thorndike Electrophysiology Institute, Beth Israel Deaconess Medical Center, Harvard Medical School, Boston, MA, USA.

9 Department of Cardiology, Imperial College Healthcare NHS Trust, London, United Kingdom

10 Department of Cardiology, Chelsea and Westminster NHS Foundation Trust, London, United Kingdom

11 Department of Computer Science and Technology, University of Cambridge, Cambridge, UK

†: Equal contribution as first author

**Correspondence**:

Fu Siong Ng, Email: f.ng@imperial.ac.uk, Shuo Wang, Email: shuowang26@gmail.com, and Yixiu Liang, Email: liang.yixiu@zs-hospital.sh.cn

**Abstract**

Routine electrocardiogram (ECG) reports offer scalable supervision for cardiovascular representation learning. In this retrospective study, we pretrained CardioLoom, a report-guided ECG foundation model, on 2,781,289 ECG–report pairs from 1,298,359 patients. Downstream development used a separate cohort of 721,317 ECGs from 233,798 patients. Evaluation covered 89 targets, with external testing in nine cohorts. Across 45 internal ECG diagnoses, macro area under the precision–recall curve (AUPRC) was 0.624 versus 0.615 for DeepECG-SSL. Across 203 external ECG evaluations, corresponding means were 0.425 and 0.424. Using labels from 10% of training patients, CardioLoom achieved macro-AUPRC 0.559, exceeding supervised training with all labels (0.549). For heart failure with reduced ejection fraction, AUPRCs were 0.690, 0.641 and 0.677 in Zhongshan, Shiyuan and BIDMC. Cardiac amyloidosis screening yielded 4.5% positive predictive value using a threshold targeting 99% specificity in validation. Report-guided pretraining supported label-efficient ECG diagnosis, with variable transfer across echocardiographic phenotypes.

## Introduction

The electrocardiogram (ECG) records cardiac electrical activity and contains information about cardiac structure and function. Artificial intelligence–enabled ECG analysis has identified ventricular dysfunction and valvular disease that may not be apparent from conventional interpretation.[1,2,3] These applications could expand the role of a test already embedded in routine care. Developing a separate model for each condition, however, requires sufficient labeled examples. Obtaining these labels can be difficult when the reference diagnosis depends on imaging, longitudinal assessment, or uncommon disease.

Pretraining offers a way to share learned representations across these applications. ECG-FM, DeepECG-SSL, and ECGFounder learn from large ECG collections before adaptation to individual tasks.[4,5,6] Signal–language models extend this approach by linking waveforms to diagnostic reports produced during clinical care.[7,8,9] These reports associate waveform patterns with clinical findings and provide a readily available source of supervision before task-specific adaptation.

These advances raise a translational question: how far can a common ECG representation extend beyond electrical diagnosis across clinical settings? Routine ECG diagnoses, echocardiographic phenotypes, and clinically adjudicated rare diseases differ in their reference standards and ECG manifestations. Evaluating these domains together can identify which clinical features transfer across cohorts. Varying the training budget for routine ECG diagnoses can additionally establish where report-guided pretraining reduces dependence on task-specific labels.

We developed CardioLoom using paired ECGs and diagnostic reports and evaluated 89 cardiovascular targets with external testing in 9 cohorts. Comparisons with 2 pretrained ECG models and an architecture-matched supervised control assessed transfer across diagnoses and clinical settings. CardioLoom supported routine ECG diagnosis and selected echocardiographic phenotypes across cohorts, with its largest comparative gains observed in ECG tasks with limited labels.

## Results

### Study Population

The Zhongshan resource contained 3,559,279 ECGs from 1,558,654 patients. Pretraining used 2,781,289 ECG and report pairs from 1,298,359 patients, with 56,673 pairs from 26,497 patients for validation. Downstream training, validation, and test sets contained 505,171, 107,973, and 108,173 ECGs. The test set represented 35,049 patients (Fig. 1). External evaluation included 203 combinations of ECG diagnoses and cohorts across 7 cohorts, with 36 echocardiographic targets in Shiyuan and 17 in BIDMC. Cohort characteristics, target definitions, and training settings are provided in Supplementary Tables 1–17.

### Routine ECG Diagnoses Across Cohorts

With all downstream training labels, CardioLoom achieved an internal macro-AUPRC of 0.624 (95% CI, 0.616–0.634) across 45 ECG targets, compared with 0.615 for DeepECG-SSL, 0.608 for ECG-FM, and 0.549 for the supervised control (Fig. 2 and Supplementary Fig. 1a). The paired difference from the control was 0.075 (95% CI, 0.067–0.082). Across 203 evaluations in the external ECG cohorts, mean AUPRC was 0.425 for CardioLoom, 0.424 for DeepECG-SSL, 0.414 for ECG-FM, and 0.394 for the control. Thus, the separation from supervised training persisted externally, but the full-label difference between CardioLoom and DeepECG-SSL was small.

Clinical-category summaries showed the highest internal mean AUPRC for CardioLoom in arrhythmia (0.778), pacing (0.689), and structural findings (0.320). Pacing and structural findings also favored CardioLoom in the external summaries, while other category rankings varied by setting and model (Supplementary Fig. 1b).

Left ventricular hypertrophy illustrated a gain within a clinically familiar electrical phenotype. Internal AUPRC was 0.756 for CardioLoom, compared with 0.742 for DeepECG-SSL and 0.740 for ECG-FM. CardioLoom also had higher estimates than both pretrained comparators in MIMIC-IV-ECG and PTB-XL, whereas DeepECG-SSL ranked higher in Xiamen. Model rankings therefore varied by diagnosis and cohort (Fig. 2, Supplementary Fig. 2, and Supplementary Tables 18–25).

**Transfer to Echocardiographic Phenotypes**

Across all available echocardiographic targets, CardioLoom macro-AUPRC was 0.318 in Zhongshan, 0.185 in Shiyuan, and 0.402 in BIDMC. Corresponding estimates for DeepECG-SSL were 0.314, 0.179, and 0.399, and for ECG-FM were 0.313, 0.182, and 0.396. These means covered 39, 36, and 17 available targets, respectively (Fig. 3, Supplementary Fig. 1c, Supplementary Tables 18 and 26–27).

CardioLoom had the highest mean AUPRC for myocardial and shunt categories internally (0.322 and 0.252) and in the external summaries (0.439 and 0.433). ECG-FM had the highest valvular-category mean AUPRC in both settings (0.258 and 0.184, Supplementary Fig. 1d).

Heart failure with reduced ejection fraction (HFrEF), operationally defined by echocardiographic left ventricular ejection fraction (LVEF) below 40%, showed a consistent point-estimate ranking across the 3 clinical cohorts. CardioLoom AUPRCs were 0.690, 0.641, and 0.677, compared with 0.676, 0.616, and 0.665 for DeepECG-SSL and 0.670, 0.616, and 0.661 for ECG-FM. In BIDMC, the difference from ECG-FM was 0.017 (95% CI, 0.002–0.031, q=0.0425). The difference from DeepECG-SSL was 0.013 (95% CI, 0.001–0.025, q=0.0717), which was not statistically significant after multiple-comparison adjustment. All 3 pretrained models exceeded the supervised control, whose estimates were 0.563, 0.483, and 0.507.

Atrial enlargement provided a complementary structural phenotype. For left atrial enlargement, CardioLoom AUPRC was 0.476 in Zhongshan and 0.366 in Shiyuan, compared with 0.463 and 0.353 for DeepECG-SSL. This ranking reversed in BIDMC (0.630 vs 0.634). For biatrial enlargement in BIDMC, CardioLoom reached 0.651, exceeding DeepECG-SSL by 0.008 (95% CI, 0.006–0.011, q=0.0042) and ECG-FM by 0.017 (95% CI, 0.015–0.020, q=0.0034). Complete results are in Supplementary Fig. 3 and Supplementary Tables 18 and 26–27.

Performance also varied across myocardial, valvular, and shunt disease. For obstructive hypertrophic cardiomyopathy, CardioLoom had the highest internal estimate (0.429) but an AUPRC of 0.047 in Shiyuan, similar to ECG-FM (0.047). For mitral stenosis in Shiyuan, CardioLoom AUPRC was 0.180 compared with 0.275 for ECG-FM. For atrial septal defect in Zhongshan, it was 0.631 compared with 0.644 for DeepECG-SSL.

To assess how HFrEF discrimination translated into binary decisions, we held each model's Zhongshan validation-selected threshold fixed across cohorts. CardioLoom sensitivity was 68.2% (95% CI, 64.7%–71.4%) internally, 49.8% (47.8%–51.7%) in Shiyuan, and 43.9% (41.7%–46.0%) in BIDMC. Specificities were 98.7%, 99.6%, and 98.4%, respectively. At external prevalences of 1.83% and 10.31%, positive predictive values were 69.0% and 76.0%. High specificity was retained, but fewer than half of positive ECGs were identified in either external cohort (Fig. 4a–b, Supplementary Fig. 4a, Supplementary Tables 26–27).

**Learning With Fewer Task-Specific Labels**

The advantage of CardioLoom over the pretrained comparators was larger when downstream ECG labels were restricted (Fig. 5). Internal macro-AUPRC was 0.420 (95% CI, 0.413–0.428) with ECGs from 1% of training patients and 0.559 (0.551–0.568) with 10% of training patients, compared with 0.374 and 0.527 for DeepECG-SSL and 0.367 and 0.515 for ECG-FM. The difference from DeepECG-SSL was 0.046 at 1%, versus 0.009 at 100%. The supervised control achieved 0.206 and 0.366. CardioLoom with ECGs from 10% of training patients slightly exceeded the control trained with all labels (0.549).

The aggregate performance gap also narrowed with greater label availability externally. Across 203 mapped evaluations in the external ECG cohorts, mean AUPRCs were 0.330 and 0.401 for CardioLoom at the 1% and 10% patient budgets, compared with 0.314 and 0.394 for DeepECG-SSL. Cohort-specific differences included reversals (Fig. 5b–c, Supplementary Table 28).

**Ablation Study**

The report-aware configuration increased internal ECG macro-AUPRC over ordinary CLIP by 0.010 at 1% (95% CI, 0.004–0.016) and 0.013 at 10% (95% CI, 0.008–0.018). Both comparisons had q=0.001. At 100%, the difference was 0.003 (95% CI, −0.004 to 0.009, q=1.000). Across all 39 echocardiographic targets, macro-AUPRC was 0.318 for CardioLoom and 0.316 for ordinary CLIP (Supplementary Fig. 5 and Supplementary Table 29).

**Rare Cardiac Diseases and Screening Burden**

The internal test set contained 14 TC, 16 NCVM, 4 ARVC, 6 AC, and 115 CA-positive ECGs, representing 5, 3, 3, 3, and 11 patients. Prevalence ranged from 0.004% to 0.106%. Five-target macro-AUPRCs were 0.020 for CardioLoom, 0.022 for DeepECG-SSL, 0.021 for ECG-FM, and 0.008 for the supervised control. No consistent CardioLoom advantage emerged across all 5 conditions (Fig. 4, Supplementary Table 18).

At the validation threshold targeting 99% specificity, CardioLoom detected 57 of 115 CA-positive test ECGs and generated 1220 false-positive results, compared with 34 and 1241 for DeepECG-SSL. CardioLoom sensitivity was 49.6% (95% CI, 22.1%–75.0%), achieved specificity 98.9%, and positive predictive value 4.5%. Approximately 1898 ECGs were screened per detected positive ECG. CA AUPRC was 0.095 for CardioLoom and 0.108 for DeepECG-SSL (Fig. 4c–d, Supplementary Fig. 4b, Supplementary Tables 30–31).

**Representation and Attribution Analyses**

CardioLoom feature projections are shown by clinical category and diagnosis in Supplementary Fig. 6. Six clinician-selected integrated-gradient examples are presented in Supplementary Fig. 7. Positive attribution highlighted ST–T abnormalities in STEMI, premature QRS complexes and associated T waves in VPB, and rhythm disturbances in AF. For LBBB and RBBB, attribution aligned with characteristic QRS morphologies. In VVI pacing, highlighted regions included pacing spikes and paced QRS complexes.

## Discussion

CardioLoom supported cardiovascular assessment across routine ECG diagnoses and selected echocardiographic phenotypes in internal and external cohorts. This extends report-guided ECG pretraining to clinical targets defined by different reference standards within a common evaluation framework. The largest comparative gains occurred in ECG tasks with limited labels. Performance in valvular and shunt disease was mixed, and rare-disease screening yielded low predictive values. Together, these findings identify both the applications supported by a shared ECG representation and those requiring further disease-specific development.

ECG-FM and DeepECG have demonstrated the use of waveform pretraining for ECG interpretation and downstream clinical prediction.[4,5] ECGFounder explored supervised pretraining for transfer across several clinical domains,[6] while signal-language approaches incorporated routine reports.[7,8,9] CardioLoom examines how report-guided training transfers across targets defined by ECG interpretation, echocardiography, and clinical adjudication. These reference standards probe different aspects of the learned representation and connect its evaluation to the diagnostic information required in clinical care.

The ECG label-efficiency results suggest that diagnostic reports provide a useful starting representation before task-specific adaptation. ECG and report alignment exposes the model to diagnostic distinctions in routinely collected text, which may reduce the downstream supervision needed to learn those distinctions. The internal ablation showed clearer benefits at limited training budgets, while performance differences narrowed as more labels became available. This pattern places the clearest value of report-aware training in the early stages of adapting a model to ECG tasks with limited annotated examples.

Transfer beyond routine ECG interpretation was clearest for reduced ventricular function. Prior studies have evaluated AI-enabled ECG detection of low LVEF and structural heart disease, including prospective clinical assessment.[1,10,11] CardioLoom had the highest HFrEF AUPRC point estimates across the 3 clinical cohorts. Category summaries showed recurring relative rankings: CardioLoom led for pacing and structural ECG findings and for myocardial and shunt echocardiographic targets in both internal and external summaries, whereas ECG-FM led for valvular targets. These patterns support phenotype-specific model selection. At the validation-selected thresholds, external HFrEF sensitivity remained below 50% despite specificity above 98%, making the intended role within the echocardiography pathway central to threshold selection.

The rare-disease analysis illustrates why the number of ECGs alone is an incomplete measure of useful training and evaluation data. Repeated recordings from a few affected patients increase the record count without capturing a broader range of disease presentations. For CA, CardioLoom identified more positive ECGs than DeepECG-SSL at the validation-targeted high-specificity operating point, yet false-positive results substantially outnumbered detected positives. The low prevalence means that even a small false-positive rate can create a substantial evaluation burden. A useful next setting is a clinically enriched population in which ECG scores can be assessed alongside history, imaging, and biomarkers. Such evaluation would determine whether the additional signal improves the yield of further investigation in patients already being considered for diagnostic workup.

This study has several limitations. First, report-guided pretraining used a single institution, whose documentation practices may influence the learned representation. External evaluation was retrospective, and differences in reference standards, prevalence, and target coverage complicate comparisons across hospitals. Dataset independence also varied. Xiamen was affiliated with Zhongshan, and some public evaluation sources overlapped with comparator pretraining data. ECG-FM used MIMIC-IV and PhysioNet sources, while DeepECG-SSL used MIMIC-IV and provided CardioLoom's initialization.[4,5] Record-level exclusion from these pretraining resources was not established. Second, sparse-target comparisons with the supervised control reflect differences in both pretrained representations and downstream classifiers. Rare-disease evaluation included only a small number of patients with each diagnosis, some contributing multiple ECGs, and lacked external validation. These features limit the precision and generalizability of the rare-disease findings. Label-informed downstream sampling and reference-label availability also limit the interpretation of prevalence and predictive values outside the evaluated populations. Third, clinical impact was not measured, and the benefit of model-guided referral remains to be established in clinical practice.

CardioLoom transferred from routine diagnostic reports to ECG diagnoses and selected echocardiographic phenotypes across clinical cohorts. Its strongest comparative gains occurred in ECG tasks with limited labels. These results support further evaluation of shared ECG representations for defined diagnostic applications, while low rare-disease predictive values limit population screening.

## Methods

### Study design and cohorts

The Ethics Committee of Zhongshan Hospital, Fudan University, approved this study (B2024-497) and waived informed consent. Zhongshan ECGs and reports from 2013 through 2023 were divided by patients into separate pretraining and downstream development partitions. Each partition had validation data, and the downstream partition also had a held-out test set (Fig. 1 and Supplementary Tables 1–2).

Patients with ECHO-related or rare ECG labels were prioritized for downstream development. From a source population of 1,558,654 patients, 347,001 met these criteria and 233,798 were selected for the downstream population. Other patients were assigned to pretraining. Downstream patients were then separated into training, validation, and test sets before task-specific reference-label screening. Prevalence and predictive values describe the resulting evaluable populations.

External ECG cohorts comprised Zhongshan-affiliated Xiamen, MIMIC-IV-ECG, UK Biobank, Chapman, PTB-XL, Georgia, and CPSC2018. Shanghai Tenth People's Hospital (Shiyuan) and Beth Israel Deaconess Medical Center (BIDMC) provided echocardiography-linked evaluation. ECG and echocardiography matching used a 30-day window. Records without the required reference label were excluded from the corresponding analysis.

### Clinical targets and label harmonization

The 89 targets comprised 45 routine ECG diagnoses, 39 echocardiographic phenotypes, and 5 rare cardiac diseases. Rare targets were Takotsubo cardiomyopathy (TC), noncompaction of ventricular myocardium (NCVM), arrhythmogenic right ventricular cardiomyopathy (ARVC), alcoholic cardiomyopathy (AC), and cardiac amyloidosis (CA), based on integrated clinical assessment.

The 45 routine ECG targets covered ischemia, arrhythmia, conduction abnormalities, structural findings, pacing, and electrolyte abnormalities. The 39 echocardiography-related targets covered normal examinations, myocardial, valvular, shunt, vascular, pericardial, and arrhythmia-related categories. Clinical-category performance was the unweighted mean of task AUPRCs within each category and cohort. External category summaries then averaged the available cohort means with equal weight. The task assignments for downstream fitting are specified separately in Supplementary Table 17. In Supplementary Fig. 1, radial distances are scaled within each cohort or category. Printed values give absolute AUPRC.

External diagnostic labels were mapped to the Zhongshan definitions using the cohort-specific rules in Supplementary Tables 4–11. Xiamen shared the Zhongshan diagnostic terminology. Evaluation used only overlapping targets with the required observed reference labels. Echocardiography-related definitions used the available structured measurements, diagnostic fields, and report mappings. ECG and echocardiography pairs were restricted to a 30-day interval. The rare-disease reference diagnoses combined specialist assessment with available imaging, laboratory, and pathological information. Supplementary Tables 3–14 give task definitions, mapping criteria, and target-specific positive counts.

**ECG and report representations**

CardioLoom paired a DeepECG-SSL-initialized convolutional Transformer ECG encoder with a MedCPT text encoder.[5,12] Contrastive pretraining aligned ECG and report representations and encouraged consistency between representations of the same ECG, following the MERL framework.[7]

The ECG backbone produced 768-dimensional features. Each projection head used a 256-unit hidden layer with GELU activation to map features into a 256-dimensional contrastive space. Both projected embeddings were normalized to unit length before similarity calculation. The first 9 MedCPT Transformer layers were frozen, while the final 3 were fine-tuned. Downstream prediction used the ECG encoder and a task-specific model.

Cardiologist-authored diagnostic reports underwent senior physician review and GPT-4o-assisted English translation. Cardiologists reviewed 200 randomly sampled processed reports and found no clinically meaningful diagnostic discrepancies. For template matching, report strings underwent Unicode NFKC normalization and conversion to lowercase. Characters outside the ranges a–z and 0–9 were replaced by spaces, and repeated whitespace was collapsed. The normalized strings defined exact-template matches during pretraining.

**Template-aware sampling**

Let $t(i)$ identify the normalized report template paired with training ECG $i$, and let $f[t(i)]$ denote its frequency in the pretraining training split. ECG and report pairs were sampled with replacement using the following probability. This downweighted each record associated with a frequent template without making templates equally likely.

$$p_i = \frac{f[t(i)]^{-0.25}}{\sum_{k=1}^{N} f[t(k)]^{-0.25}} \tag{1}$$

$N$ denotes the number of training pairs. Template frequencies were calculated from the pretraining training split.

**Multiple-positive ECG and report alignment**

Let $e_i$ and $r_j$ be unit-normalized projected ECG and report embeddings in a batch of $B$ pairs. Similarity was $s_{ij} = e_i^{\top} r_j / \tau$ with $\tau = 0.07$. The matching set $M(i)$ contained every report $j$ in the batch with the same normalized template as the report paired with ECG $i$, including its paired report. Thus, identical standardized reports were treated as additional positive matches, rather than negatives, for that ECG. The ECG-to-report loss was:

$$L_{E \to R} = -\frac{1}{B} \sum_{i=1}^{B} log \frac{\sum_{j \in M(i)} exp(s_{ij})}{\sum_{j=1}^{B} exp(s_{ij})} \tag{2}$$

The reverse loss $L_{R \to E}$ used the transposed similarity matrix and the corresponding matching-template ECGs. The 2 directional losses were summed:

$$L_{CMA} = L_{E \to R} + L_{R \to E} \tag{3}$$

**Within-ECG alignment and total loss**

The unimodal alignment term $L_{UMA}$ applied bidirectional contrastive alignment to 2 dropout-perturbed representations of the same ECG. The same-record pair was the positive pair for this term. Dropout was 0.1, and the unimodal loss weight was 1.0. The total objective was:

$$L = L_{CMA} + L_{UMA} \tag{4}$$

Exact normalized-template matching and frequency-weighted sampling were introduced together in the final pretraining configuration. Supplementary Table 16 gives the optimization parameters. The convolutional extractor remained frozen, and the entire ECG backbone was frozen for the first 500 updates. The checkpoint at epoch 16 had the minimum validation loss and was retained.

**Model inputs and downstream adaptation**

Comparators were publicly released ECG-FM and DeepECG-SSL checkpoints and the same ECG architecture initialized randomly.[4,5]

The dataset interface provided 10-second ECGs at 500 Hz from 8 independent leads: I, II, and V1–V6. CardioLoom, DeepECG-SSL, and the supervised control reconstructed the 12-lead input and resampled it to 250 Hz, yielding a 12 × 2500 array. ECG-FM retained 500-Hz signals, applied lead-wise standardization, and averaged representations from two 5-second segments. Cohort-specific acquisition filtering was retained, including the predefined 60-Hz notch setting for BIDMC.

All models used the same task-specific partitions and labels. Most pretrained-model tasks adapted the Transformer and prediction head with the convolutional extractor frozen. For targets with few positive examples, we froze the entire pretrained encoder and fitted a linear SVM to its 768-dimensional features. This restricted task-specific fitting to the classifier parameters. The protocol applied to the 5 rare diseases and 6 echocardiographic targets: tetralogy of Fallot (TOF), Ebstein's anomaly (EA), dextrocardia (Dex), pulmonary regurgitation (PR), pulmonary stenosis (PS), and tricuspid stenosis (TS). All pretrained comparators used this protocol, and SVM outputs were raw decision margins. The supervised control trained the encoder and task head from random initialization.

End-to-end optimization used AdamW, a fixed learning rate of $5 \times 10^{-5}$, and batch size 128. The budget was 12,000 updates for full-label ECG classification and 4000 updates otherwise. Pretrained models used validation-based early stopping, while the supervised control completed the allocated budget. All checkpoints were selected by validation AUPRC. Supplementary Table 17 provides the remaining adaptation settings.

**ECG label efficiency**

Training budgets of 1%, 10%, and 100% referred to 1,637, 16,370, and 163,700 downstream training patients. These subsets contained 5,218, 50,043, and 505,171 ECGs, respectively. The original sampling procedure used a greedy step to cover available ECG diagnoses, followed by random filling of the patient quota. The 1% subset was nested within the 10% subset, and all training ECGs belonging to selected patients were retained. All models reused the same frozen index files.

## Internal ablation study

The ablation compared report-aware pretraining with ordinary CLIP using the same ECG and report resource, ECG backbone, and 20-epoch optimization schedule. Ordinary CLIP used diagonal ECG and report alignment without report-aware sampling or multiple-positive template matching. Each run used its own validation-selected checkpoint. Both models used the same downstream data partitions, target definitions, input processing, and task-specific adaptation protocols. Frozen-feature linear SVMs were used for the same sparse targets. Evaluation covered all 89 targets at 100% and the 45 ECG targets at 1% and 10% (Supplementary Table 29).

## Outcomes and statistical analysis

The primary metric was area under the precision-recall curve (AUPRC), selected because many targets were infrequent and class distributions were imbalanced. AUPRC was interpreted alongside target prevalence. Secondary metrics were area under the receiver operating characteristic curve (AUROC), sensitivity, specificity, F1 score, and Matthews correlation coefficient.

AUPRC was calculated by trapezoidal integration of the precision-recall curve. Task prevalence provided the reference for interpreting performance under class imbalance. Point estimates used individual ECG records. For paired bootstrap analysis, the same sampled units and their multiplicities were applied to all compared models. Repeated ECGs were sampled together by patient in Zhongshan, Xiamen, Shiyuan, BIDMC, MIMIC-IV-ECG, UK Biobank, PTB-XL, and Chapman. Georgia and CPSC2018 used individual ECG records because patient identifiers were unavailable.

We generated 1000 paired bootstrap replicates while keeping model checkpoints and decision thresholds fixed. Percentile 95% CIs used the 2.5th and 97.5th percentiles and required at least 950 valid replicates. For pairwise comparisons, we subtracted the observed metric difference from each bootstrap difference to form a distribution centered on zero. The P value was the proportion of valid centered differences at least as large in absolute value as the observed difference, with 1 added to the numerator and denominator. Benjamini–Hochberg adjustment was applied within each analysis and metric, separately for individual-target and macro comparisons.[13] Macro-AUPRC is the unweighted mean across the specified targets. Source Data identifies each comparator, adjustment set, set size, and availability status.

The ablation used the same paired patient-cluster bootstrap and Benjamini–Hochberg procedure. The 1000 shared draws sampled from the 35,049 internal test patients, retaining task-specific label eligibility. Each macro replicate required all targets in its stated set to be estimable. Full-budget families contained 89 task-level and 4 macro comparisons per metric. For the low-budget analyses, task-level correction combined the comparisons for 45 ECG targets at the 1% and 10% budgets. The 2 macro comparisons were corrected as a separate set. AUPRC and AUROC were adjusted separately. Nonestimable hypotheses entered adjustment with P=1 and were displayed with unavailable q values.

## Validation thresholds and clinical operating points

For each fitted model configuration and target, we evaluated each distinct score observed in the Zhongshan validation set as a candidate threshold. An ECG was classified as positive when its score was greater than or equal to that threshold. Precision and sensitivity were calculated at each candidate, and we selected the threshold maximizing their harmonic mean, the F1 score. If several candidates achieved the same maximum F1, the smallest of those thresholds was selected. Each selected threshold was applied unchanged to the internal test set and every available external cohort.

For the separate rare-disease operating-point analysis, we selected the candidate with the highest validation sensitivity among thresholds achieving at least 99% specificity. Equal-sensitivity candidates were resolved by choosing the smallest eligible threshold. If no observed-score candidate met the specificity target, a threshold above the maximum validation score classified every record as negative. The selected threshold was then fixed for internal testing. We reported achieved sensitivity and specificity, positive and negative predictive values, true-positive (TP) and false-positive (FP) ECG counts, and the number evaluated per detected positive ECG (N/TP) and per false-positive ECG (N/FP).

For CA, operating-point intervals and paired model differences used 1000 shared patient-cluster draws from the 35,049 internal test patients, with the validation-selected 99%-specificity thresholds held fixed. Pointwise percentile 95% intervals covered sensitivity, specificity, predictive values, false-positive and positive calls per 1000 evaluated ECGs, N/TP and N/FP (Supplementary Table 31 and Source Data).

**Representation visualization**

Exploratory t-distributed stochastic neighbor embedding (t-SNE) visualized CardioLoom features across clinical categories and selected diagnoses.[14] Integrated gradients characterized waveform contributions in correctly classified positive ECGs.[15]

We visualized pretrained CardioLoom features by broad clinical category and by diagnoses within selected categories. Macro-category eligibility required the record's positive labels to belong to one clinical category. Multiple positive diagnoses within that category were allowed. Each patient contributed at most one record to a macro-category panel. Patients eligible for several categories were assigned before sampling, prioritizing categories containing less frequent targets. Sampling then balanced categories and prioritized less frequent targets within each category.

Macro-category displays required at least 50 eligible patients per clinical category after eligibility screening and patient assignment. Four ECG categories met this requirement: arrhythmia, conduction, pacing, and structure. Ischemia and electrolyte abnormalities did not meet the requirement. Balancing to the smallest retained category yielded 185 records per category and 740 patient-distinct records. All 7 echocardiographic categories met the requirement, with 300 records per category and 2100 patient-distinct records.

Within-category projections included ECGs with exactly one positive label in the displayed clinical category. Positive labels in other categories were allowed. Each diagnosis required at least 150 eligible patients. We sampled 150 ECGs per diagnosis, with one ECG per patient within each diagnosis. Categories with at least 2 eligible diagnoses were displayed.

The selected 768-dimensional pretrained-backbone features were standardized by feature and normalized to unit length. Separate t-SNE fits used cosine distance, 2 dimensions, perplexity 30, PCA initialization, an automatically selected learning rate, and 1000 iterations.

### Integrated-gradient attribution

Integrated gradients were calculated for the task-specific positive-minus-negative logit of the fitted CardioLoom model. The reference input was an all-zero ECG. The path integral of the input gradients between this reference and the observed ECG was approximated using 128-point Gauss–Legendre quadrature. Attribution used the gradient of the fixed forward model, with the backbone's training-time gradient multiplier set to 1.0. This setting leaves forward predictions unchanged.

Candidate examples were ground-truth-positive ECGs classified as positive at the corresponding validation-selected threshold. Within each endpoint, candidates were ordered by predicted positive probability, and at most one ECG was retained per patient. Up to 100 eligible candidates were considered to obtain up to 12 accepted examples per endpoint. We checked numerical integration by comparing the sum of the attributions with the model-score change from the zero reference to the observed ECG. Examples were retained only when attributions were finite and nonzero and their relative integration error was at most 15%. This error was the absolute difference between the attribution sum and the score change, divided by the absolute score change plus $10^{-6}$.

The review set contained 149 accepted endpoint-specific examples across 13 endpoints. Patient uniqueness was enforced within each endpoint. Clinicians selected six examples for Supplementary Fig. 7. Positive-only displays retained attribution values contributing toward the target diagnosis and overlaid them on the 8 input waveforms. The color ceiling was the 99.5th percentile of absolute attribution within each example. Each lead was scaled separately for display, retaining the complete 10-second recording.

## Data availability

Hospital ECGs, reports, and linked clinical data are not publicly available because of patient privacy and institutional restrictions. Public datasets remain available under their providers' terms. Aggregate results are reported in Supplementary Tables 18–28 and 30.

## Code availability

The project repository is https://github.com/dt-yuhui/CardioLoom.

## Acknowledgments

The authors would like to acknowledge the expert technical assistance of the data center staff at Zhongshan Hospital, Fudan University. This work was supported by the National Natural Science Foundation of China (Grant G2021-046), Shanghai Pujiang Talent Program (24PJD014), and Fudan AI4S program (FudanX24A1056).

## Author contributions

Z.Y., Y.T., J.H., Y.L., J.G., F.S.N., S.W. designed the study, interpreted data, and wrote the manuscript. Z.Y., Y.T., J.H., L.P., Z.X., J.C., X.L., J.L., Y.Z., Z.L., C.W., B.Z., C.C. performed model development, statistical analyses, and prepared figures. H.L., K.P., D.B.K., J.W.W., Y.S. contributed external validation data and critically revised the manuscript. Y.L., J.G., F.S.N., S.W. supervised the project. All authors read and approved the final manuscript.

## Competing interests

Hongyang Lu is an employee of Medtronic. Medtronic did not sponsor this study. All other authors declared no conflicts of interest.

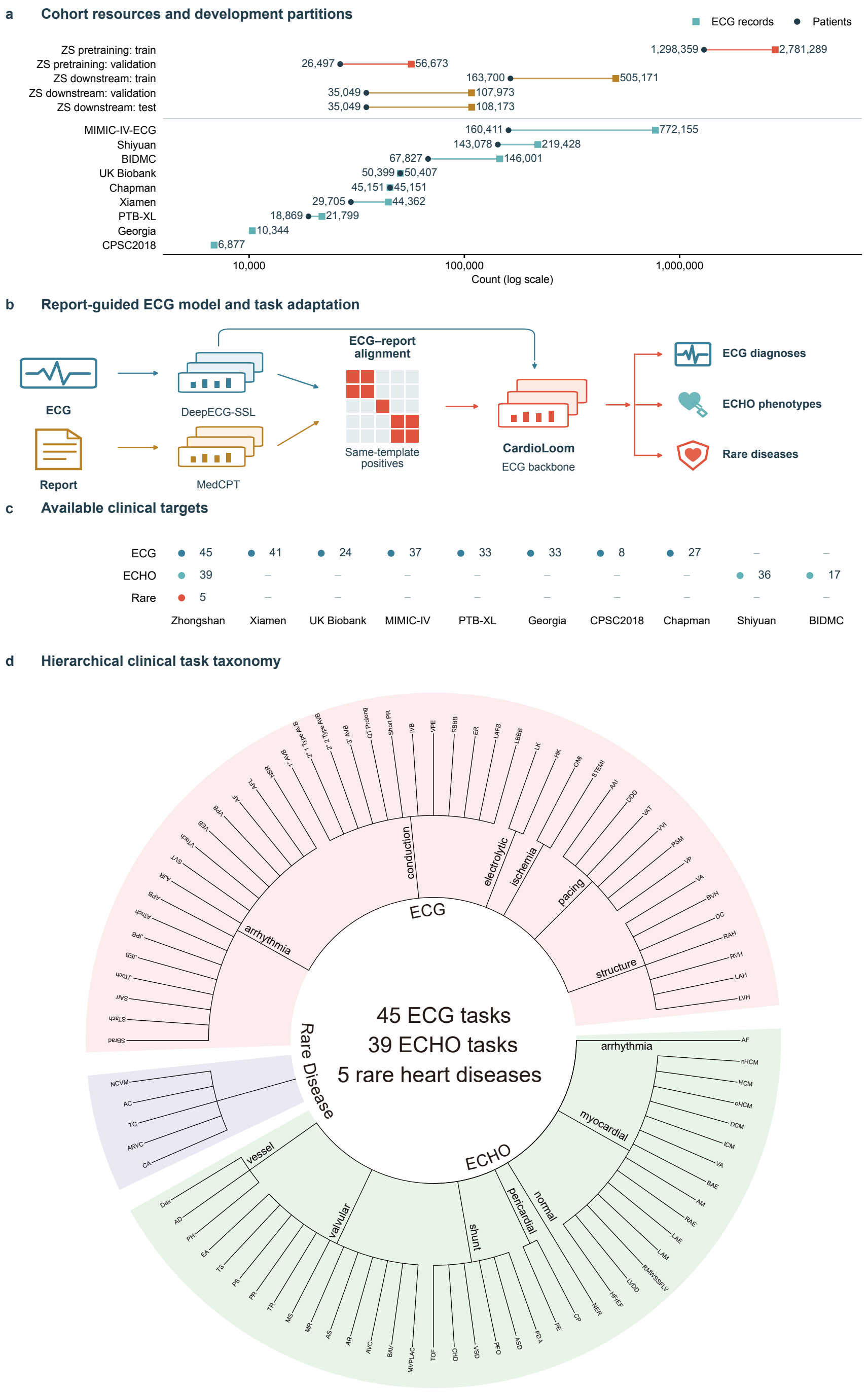

**Fig. 1. Cohort Resources, Model Development, and Clinical Task Taxonomy**

a, Cohort ECG and patient counts on a logarithmic axis. Squares indicate ECG records and circles patients. Patient counts are unavailable for Georgia and CPSC2018. b, Report-guided pretraining aligns a DeepECG-SSL-initialized ECG encoder with a MedCPT report encoder. The pretrained ECG backbone supports ECG-only downstream prediction. Architecture and training details are in the Methods. c, Available targets by domain and cohort. A dash indicates an unevaluated domain. d, Clinical-category hierarchy of the 45 ECG, 39 echocardiographic, and 5 rare-disease targets.

### ECG diagnoses

Scratch ECG-FM DeepECG-SSL CardioLoom | Prevalence

| Target | Scratch | ECG-FM | DeepECG-SSL | CardioLoom |
|---|---|---|---|---|
| OMI | 0.517* | 0.541 | 0.540 | **0.551** |
| STEMI | 0.440* | 0.509 | **0.540** | 0.503 |
| NSR | 0.996* | **0.997*** | **0.997*** | **0.997** |
| SBrad | 0.969* | 0.970* | 0.972* | **0.974** |
| STach | **0.971** | 0.970 | **0.971** | **0.971** |
| VPB | 0.958* | 0.969 | **0.971** | 0.970 |
| APB | 0.893* | 0.925* | 0.930* | **0.936** |
| SArr | 0.572* | 0.558* | 0.608 | **0.623** |
| AF | 0.927* | 0.940* | 0.943 | **0.947** |
| AFL | 0.828* | 0.876 | **0.886** | 0.883 |
| JPB | 0.613* | 0.658 | 0.677 | **0.682** |
| AJR | 0.209* | 0.303* | 0.324 | **0.339** |
| ATach | 0.518* | 0.637* | **0.710** | 0.693 |
| JTach | 0.468* | 0.601 | 0.569 | **0.605** |
| JEB | 0.572* | 0.669 | 0.706 | **0.712** |
| SVT | 0.798 | 0.836 | **0.843** | 0.840 |
| VEB | 0.442* | 0.517 | **0.637** | 0.581 |
| VTach | 0.543* | 0.611 | 0.667 | **0.701** |
| RBBB | 0.955 | **0.962*** | 0.955 | 0.954 |
| 1° AVB | 0.848* | **0.884** | 0.876 | 0.879 |
| QT Prolong | 0.242* | **0.305** | 0.294 | 0.291 |
| ER | 0.261* | 0.305 | 0.315 | **0.326** |
| LAFB | 0.482 | 0.531 | **0.545** | 0.510 |
| LBBB | 0.931* | 0.941 | 0.937 | **0.944** |
| IVB | 0.638* | **0.722** | 0.690 | 0.700 |
| Short PR | 0.235 | 0.257 | **0.300** | 0.271 |
| VPE | 0.766* | 0.847 | 0.850 | **0.857** |
| 3° AVB | 0.450* | 0.600 | 0.622 | **0.656** |
| 2° 1 Type AVB | 0.754* | 0.885 | 0.887 | **0.903** |
| 2° 2 Type AVB | 0.575* | 0.749 | **0.820** | 0.775 |
| LVH | 0.704* | 0.740* | 0.742* | **0.756** |
| LAH | 0.431 | 0.439 | 0.441 | **0.447** |
| RVH | 0.467* | **0.560** | 0.545 | 0.553 |
| VA | 0.229 | 0.236 | 0.228 | **0.258** |
| RAH | 0.142 | 0.101 | **0.157** | 0.153 |
| DC | 0.001 | 0.008 | 0.004 | **0.022** |
| BVH | **0.058** | 0.034 | 0.050 | 0.050 |
| VVI | 0.825* | 0.827* | 0.855 | **0.865** |
| VAT | 0.751* | **0.906** | 0.868 | 0.895 |
| DDD | 0.640* | **0.849** | 0.775 | 0.828 |
| AAI | 0.656* | 0.835 | **0.842** | 0.838 |
| VP | 0.145 | 0.198 | 0.098* | **0.222** |
| PSM | 0.263* | 0.412 | 0.347 | **0.489** |
| HK | 0.022* | **0.108** | 0.084 | 0.066 |
| LK | 0.015 | 0.042 | 0.063 | **0.070** |

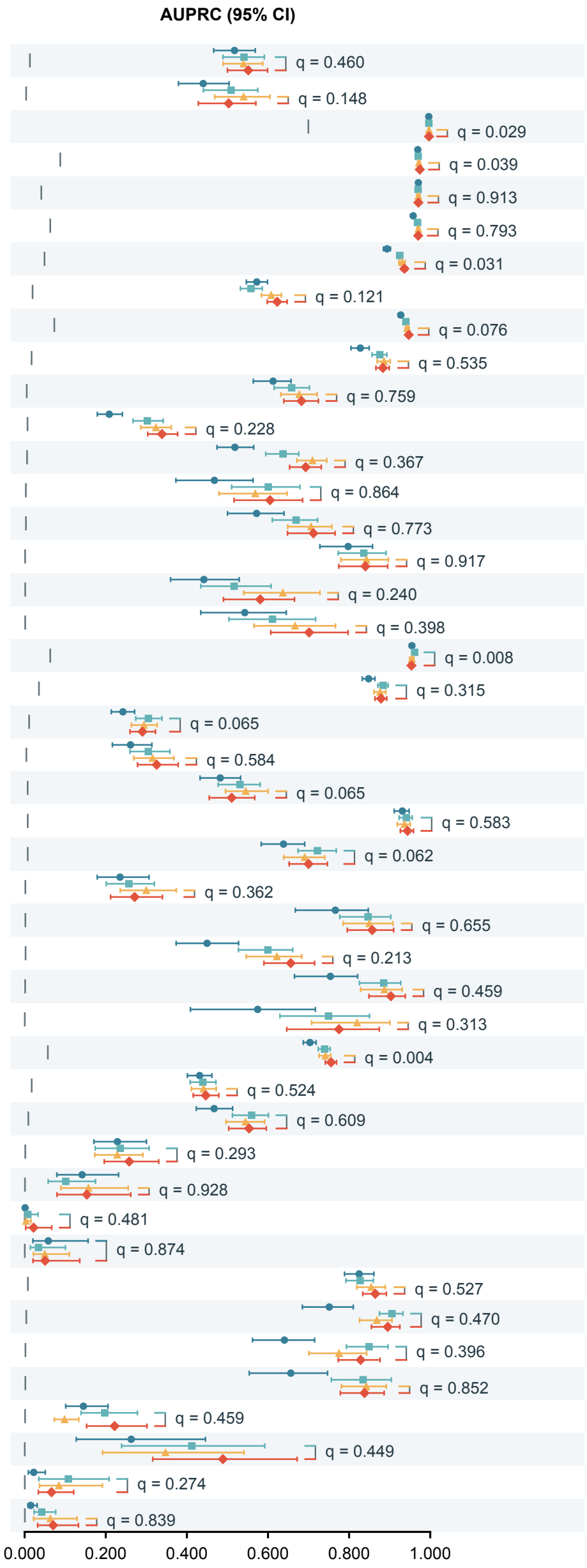

**Fig. 2. Model Comparison Across Routine ECG Diagnoses**

AUPRCs for 4 models across 45 ECG targets in the Zhongshan test cohort. Horizontal bars show 95% patient-bootstrap CIs, and gray ticks mark prevalence. Brackets compare CardioLoom with the highest-AUPRC baseline for each target. Endpoint colors identify the models. An asterisk denotes a paired difference with Benjamini–Hochberg-adjusted q<0.05 in either direction. Bold type marks the highest displayed estimate. Complete intervals and all 3 baseline comparisons are in Supplementary Fig. 2.

### Echocardiographic phenotypes

| Target | Scratch | ECG-FM | DeepECG-SSL | CardioLoom |
|---|---|---|---|---|
| NER | 0.850* | **0.878*** | 0.875* | 0.871 |
| LAE | 0.420* | 0.470 | 0.463* | **0.476** |
| HCM | 0.431* | 0.481 | **0.489** | 0.486 |
| nHCM | 0.346* | 0.386 | 0.390 | **0.391** |
| BAE | 0.405* | **0.504** | 0.494 | 0.503 |
| HFrEF | 0.563* | 0.670 | 0.676 | **0.690** |
| RMWSSFLV | 0.479* | 0.526 | 0.535 | **0.541** |
| DCM | 0.361* | 0.389 | **0.446** | 0.414 |
| oHCM | 0.213* | 0.369* | 0.411 | **0.429** |
| ICM | 0.201 | 0.217 | 0.208 | **0.243** |
| LVDD | 0.035 | 0.038 | **0.041** | 0.039 |
| VA | 0.163* | 0.200 | 0.206 | **0.232** |
| RAE | **0.017** | 0.014 | 0.016 | 0.016 |
| AM | 0.013 | 0.027 | **0.035** | 0.022 |
| LAM | 0.009 | 0.015 | 0.026 | **0.028** |
| AVC | 0.359* | 0.412 | 0.400* | **0.415** |
| TR | 0.528* | 0.579 | **0.590** | 0.588 |
| MR | 0.416* | **0.500*** | 0.483 | 0.478 |
| AR | 0.292* | **0.404** | 0.397 | 0.395 |
| MVPLAC | 0.072* | **0.092** | 0.086 | 0.088 |
| MS | 0.428* | **0.584** | 0.574 | 0.574 |
| AS | 0.200* | 0.348 | **0.361** | 0.341 |
| BAV | 0.046* | 0.113 | **0.115** | **0.115** |
| PS | **0.086** | 0.054 | 0.060 | 0.054 |
| EA | **0.143** | 0.004 | 0.014 | 0.023 |
| PR | 0.007 | 0.003 | 0.004 | **0.009** |
| TS | 0.003 | **0.006** | 0.002 | 0.002 |
| CHD | 0.342* | 0.455 | 0.451 | **0.460** |
| ASD | 0.483* | 0.622 | **0.644** | 0.631 |
| VSD | 0.108* | 0.262 | 0.205* | **0.274** |
| PFO | 0.014 | 0.021 | **0.026** | 0.025 |
| PDA | 0.058 | 0.092 | 0.070 | **0.123** |
| TOF | **0.003** | 0.000 | 0.001 | 0.000 |
| PH | 0.469* | **0.520** | 0.519 | 0.512 |
| AD | 0.256* | **0.307** | 0.302 | **0.307** |
| Dex | **0.308*** | 0.001 | 0.006 | 0.011 |
| PE | 0.255* | **0.310** | 0.302 | 0.294 |
| CP | 0.135* | **0.340** | 0.311 | 0.320 |
| AF | 0.979* | 0.992* | **0.993** | **0.993** |

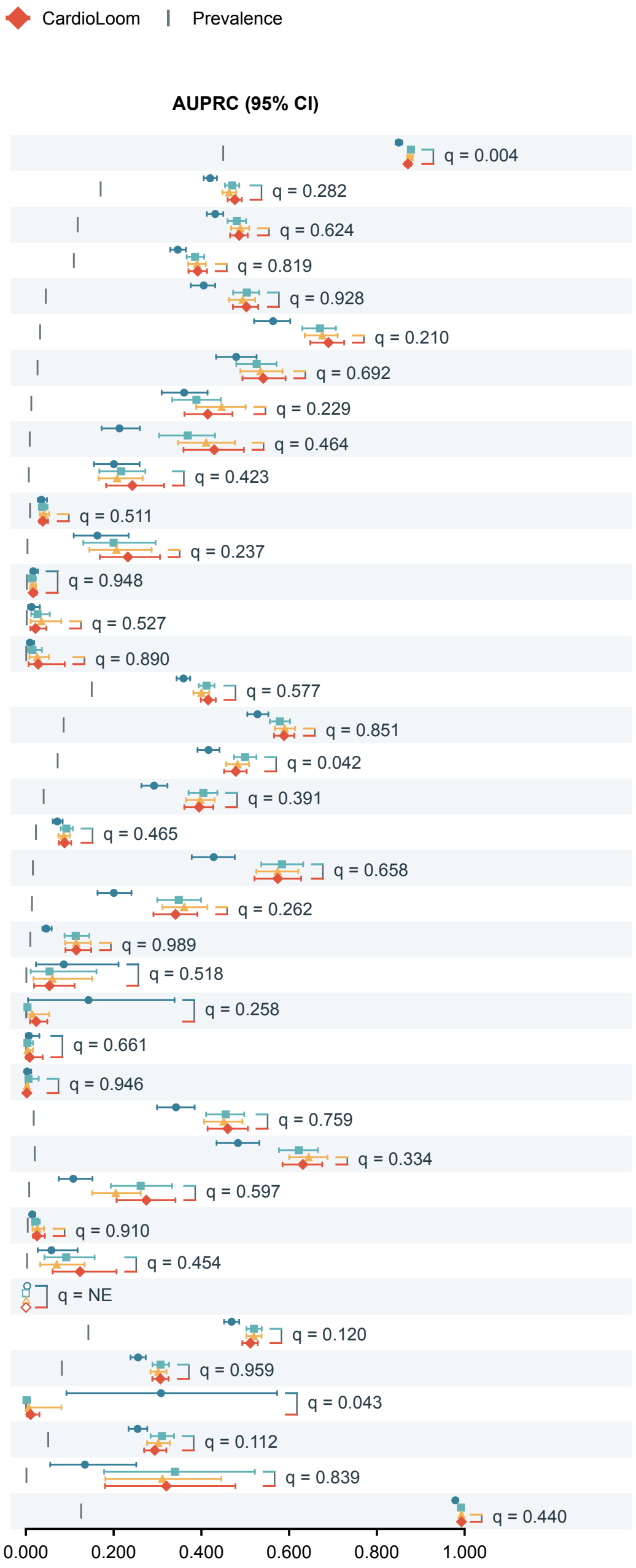

**Fig. 3. Model Comparison Across Echocardiographic Phenotypes**

AUPRCs for 4 models across 39 echocardiographic targets in the Zhongshan test cohort. Horizontal bars show 95% patient-bootstrap CIs, and gray ticks mark prevalence. Brackets compare CardioLoom with the highest-AUPRC baseline. Endpoint colors identify the models. An asterisk denotes a paired difference with Benjamini–Hochberg-adjusted q<0.05 in either direction. Bold type marks the highest displayed estimate. NE denotes a nonestimable comparison. Hollow markers indicate unavailable CIs. Complete intervals and all 3 baseline comparisons are in Supplementary Fig. 3.

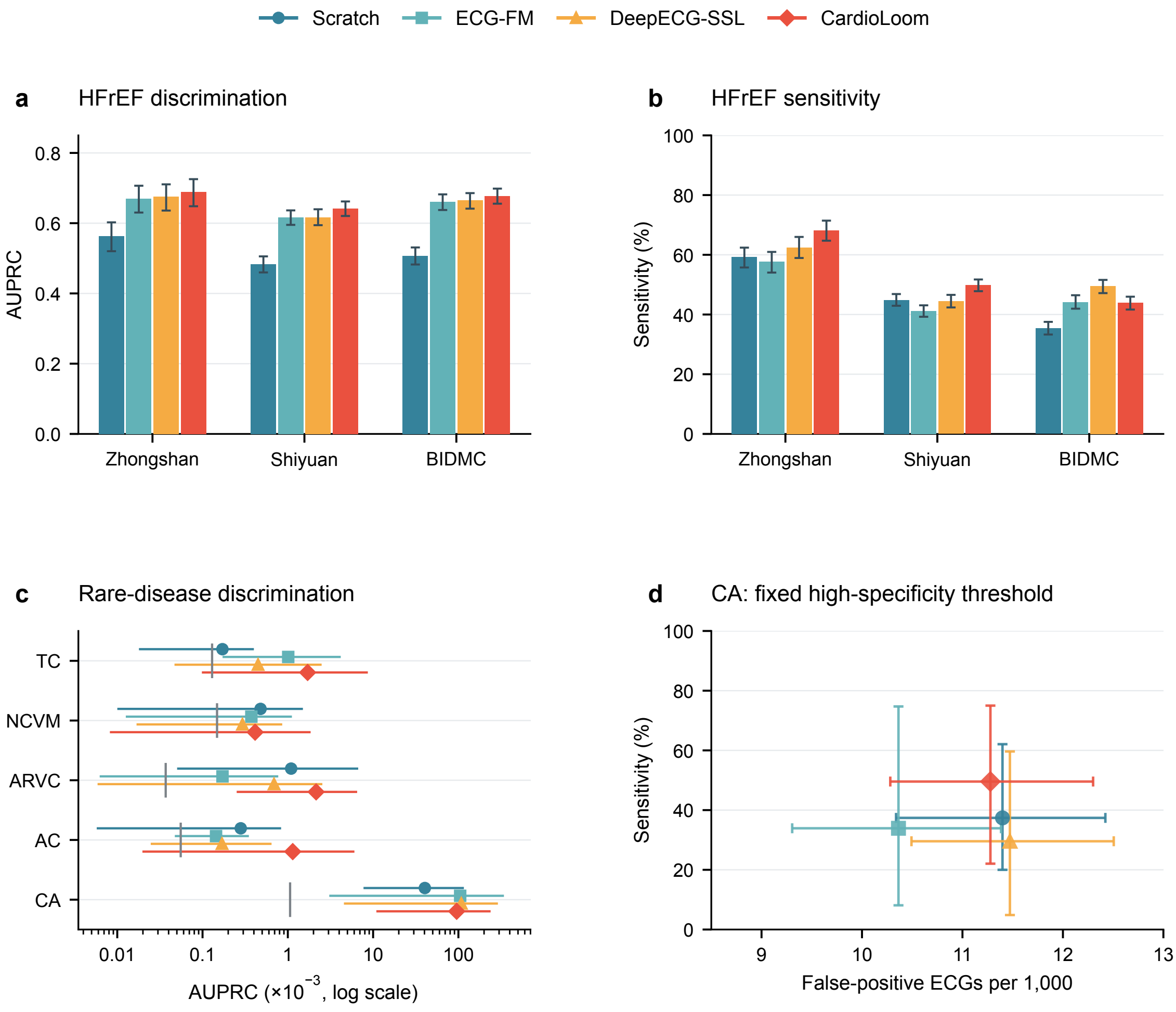

**Fig. 4. Discrimination and Fixed-Threshold Clinical Performance**

a–b, HFrEF AUPRC and sensitivity in Zhongshan, Shiyuan, and BIDMC. HFrEF denotes LVEF below 40%. Prevalences were 3.23%, 1.83%, and 10.31%, respectively. Sensitivity uses each model's Zhongshan validation-F1 threshold. c, AUPRCs for 5 rare diseases on a logarithmic axis in units of $10^{-3}$. Gray ticks mark prevalence. d, Cardiac amyloidosis (CA) sensitivity versus false-positive ECGs per 1000 evaluated ECGs at validation thresholds targeting 99% specificity. Error bars show 95% patient-bootstrap CIs. The rare-disease test set contained 108,173 ECGs, including 115 CA-positive ECGs from 11 patients. TC denotes Takotsubo cardiomyopathy and NCVM denotes noncompaction of ventricular myocardium. ARVC denotes arrhythmogenic right ventricular cardiomyopathy and AC denotes alcoholic cardiomyopathy.

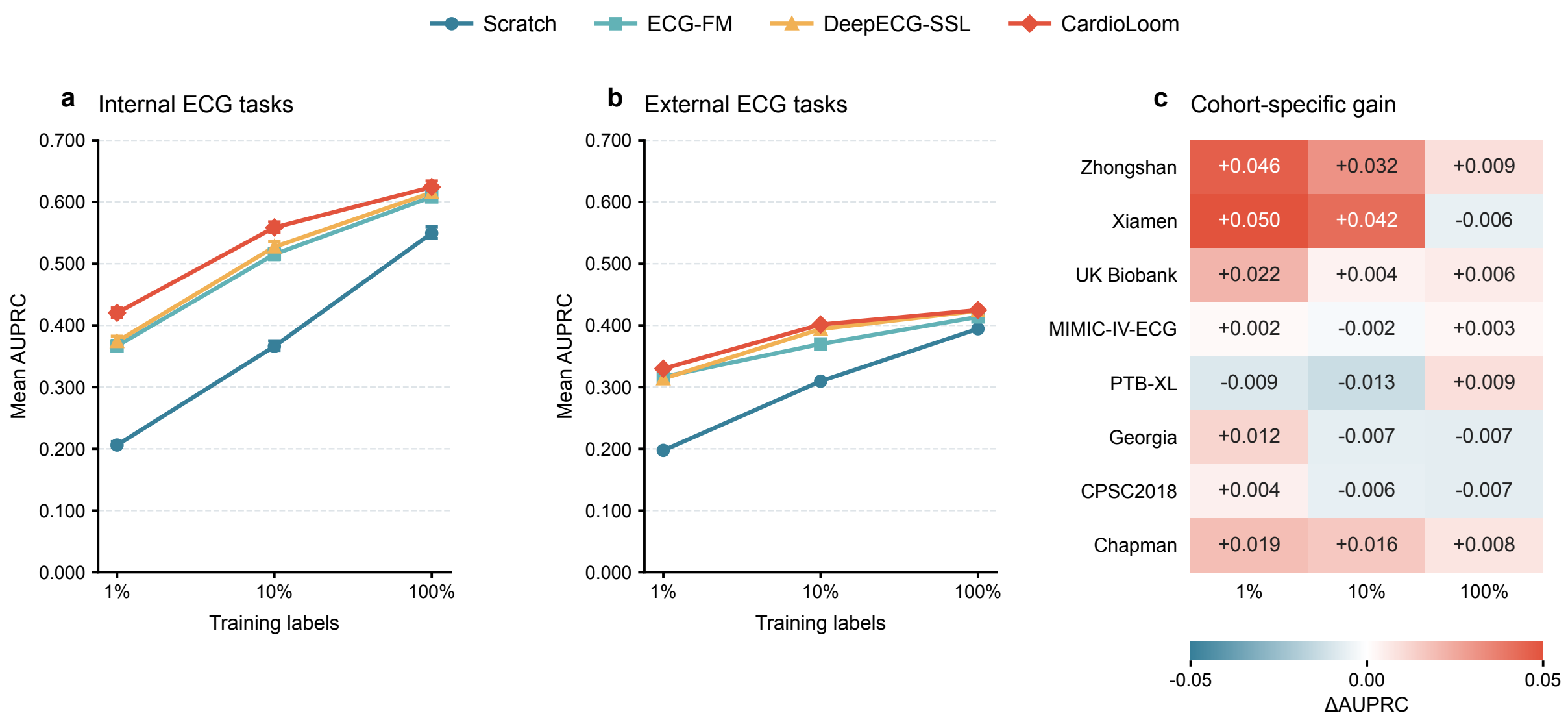

**Fig. 5. ECG Label Efficiency**

a–b, Mean AUPRC using ECGs from 1%, 10%, and 100% of Zhongshan training patients. A averages 45 internal targets, with 95% patient-bootstrap CIs. B averages 203 external combinations of tasks and cohorts. c, CardioLoom-minus-DeepECG-SSL mean AUPRC by cohort and training fraction.

# Supplementary Information

CardioLoom: A Report-Guided Electrocardiogram Foundation Model for Cross-Domain Cardiovascular Phenotyping

Supplementary Figs. 1–7 and Supplementary Tables 1–31. All Methods are provided in the main manuscript.

**List of Supplementary Figures**

**List of Supplementary Tables**

**Supplementary Fig. 1. Cohort and Clinical-Category Performance Summaries**

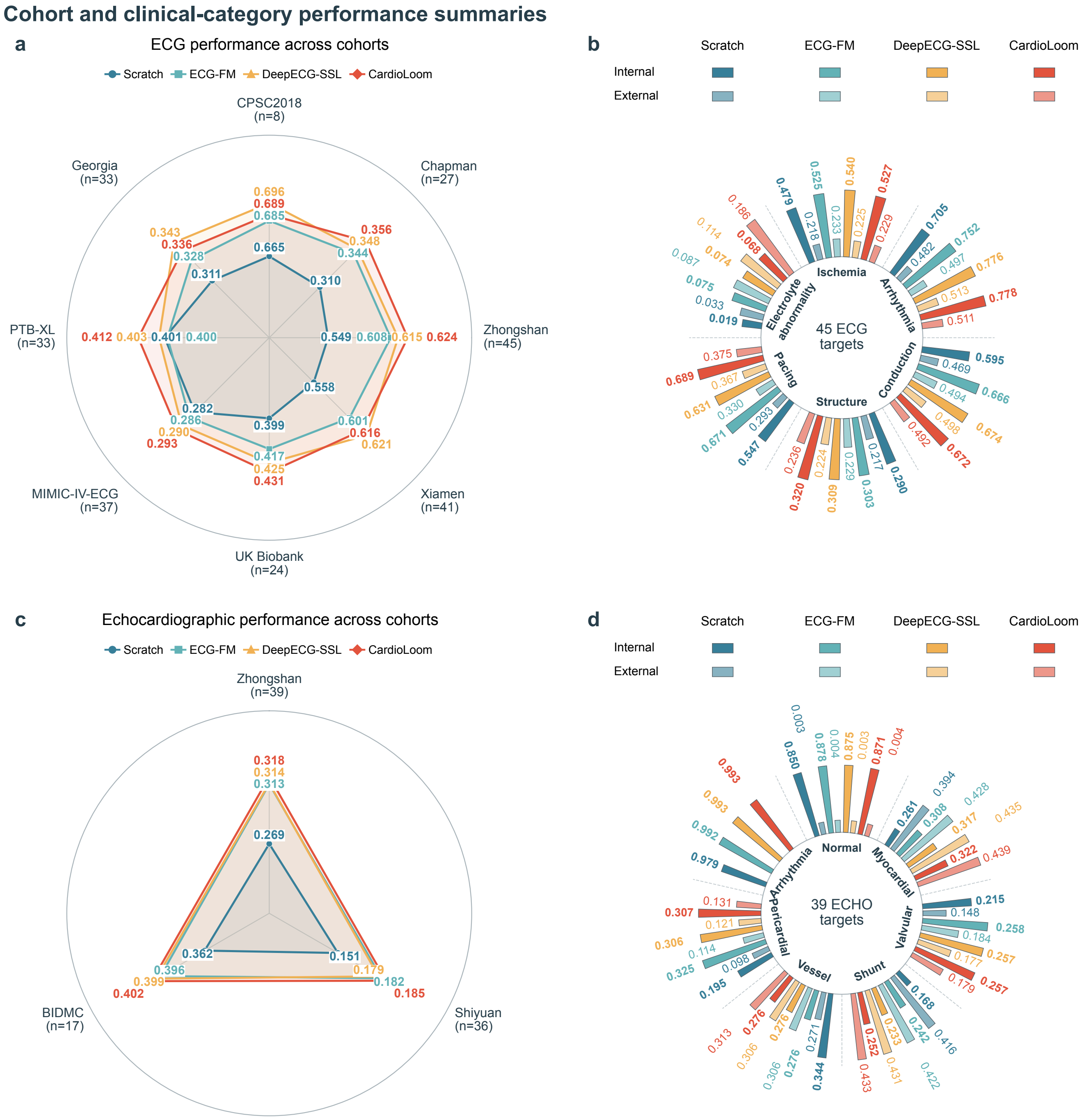

a and c, Cohort-specific mean AUPRC for ECG and echocardiographic targets. The label n indicates the number of available targets. b and d, Clinical-category mean AUPRC, with solid colors indicating the internal cohort and lighter colors the external summaries. External summaries average category means across available cohorts with equal weight. Printed values are AUPRCs.

**Supplementary Fig. 2. Detailed ECG Model Comparisons**

Zhongshan: ECG diagnoses

Scratch — ECG-FM — DeepECG-SSL — CardioLoom | Prevalence

| Target | Scratch | ECG-FM | DeepECG-SSL | CardioLoom |
|---|---|---|---|---|
| OMI | 0.517 (0.466–0.569) q = 0.036 * | 0.541 (0.489–0.591) q = 0.460 | 0.540 (0.490–0.587) q = 0.368 | **0.551 (0.500–0.599)** |
| STEMI | 0.440 (0.379–0.504) q = 0.006 * | 0.509 (0.441–0.575) q = 0.773 | **0.540 (0.469–0.605)** q = 0.148 | 0.503 (0.428–0.570) |
| NSR | 0.996 (0.996–0.997) q = 0.002 * | **0.997 (0.996–0.997)** q = 0.004 * | **0.997 (0.996–0.997)** q = 0.029 * | **0.997 (0.997–0.997)** |
| SBrad | 0.969 (0.966–0.972) q = 0.002 * | 0.970 (0.967–0.973) q = 0.004 * | 0.972 (0.969–0.975) q = 0.039 * | **0.974 (0.971–0.977)** |
| STach | **0.971 (0.967–0.975)** q = 0.980 | 0.970 (0.966–0.974) q = 0.676 | **0.971 (0.967–0.975)** q = 0.913 | **0.971 (0.966–0.975)** |
| VPB | 0.958 (0.953–0.962) q = 0.002 * | 0.969 (0.965–0.972) q = 0.324 | **0.971 (0.967–0.974)** q = 0.793 | 0.970 (0.967–0.974) |
| APB | 0.893 (0.885–0.902) q = 0.002 * | 0.925 (0.917–0.932) q = 0.004 * | 0.930 (0.923–0.938) q = 0.031 * | **0.936 (0.930–0.943)** |
| SArr | 0.572 (0.547–0.599) q = 0.002 * | 0.558 (0.532–0.586) q = 0.004 * | 0.608 (0.584–0.633) q = 0.121 | **0.623 (0.598–0.647)** |
| AF | 0.927 (0.921–0.933) q = 0.002 * | 0.940 (0.935–0.945) q = 0.004 * | 0.943 (0.939–0.948) q = 0.076 | **0.947 (0.942–0.951)** |
| AFL | 0.828 (0.805–0.850) q = 0.002 * | 0.876 (0.857–0.893) q = 0.082 | **0.886 (0.869–0.901)** q = 0.535 | 0.883 (0.866–0.899) |
| JPB | 0.613 (0.564–0.657) q = 0.002 * | 0.658 (0.615–0.702) q = 0.258 | 0.677 (0.632–0.721) q = 0.759 | **0.682 (0.639–0.724)** |
| AJR | 0.209 (0.179–0.241) q = 0.002 * | 0.303 (0.267–0.341) q = 0.004 * | 0.324 (0.286–0.361) q = 0.228 | **0.339 (0.304–0.377)** |
| ATach | 0.518 (0.474–0.565) q = 0.002 * | 0.637 (0.595–0.676) q = 0.004 * | **0.710 (0.671–0.745)** q = 0.367 | 0.693 (0.653–0.731) |
| JTach | 0.468 (0.373–0.563) q = 0.002 * | 0.601 (0.510–0.679) q = 0.864 | 0.569 (0.479–0.647) q = 0.115 | **0.605 (0.516–0.685)** |
| JEB | 0.572 (0.500–0.640) q = 0.002 * | 0.669 (0.611–0.722) q = 0.058 | 0.706 (0.649–0.758) q = 0.773 | **0.712 (0.648–0.766)** |
| SVT | 0.798 (0.728–0.858) q = 0.080 | 0.836 (0.773–0.891) q = 0.862 | **0.843 (0.780–0.897)** q = 0.917 | 0.840 (0.775–0.895) |
| VEB | 0.442 (0.360–0.529) q = 0.002 * | 0.517 (0.434–0.608) q = 0.118 | **0.637 (0.540–0.728)** q = 0.240 | 0.581 (0.490–0.665) |
| VTach | 0.543 (0.434–0.645) q = 0.002 * | 0.611 (0.504–0.717) q = 0.081 | 0.667 (0.566–0.767) q = 0.398 | **0.701 (0.607–0.797)** |
| RBBB | 0.955 (0.949–0.960) q = 0.539 | **0.962 (0.957–0.966)** q = 0.008 * | 0.955 (0.949–0.960) q = 0.597 | 0.954 (0.948–0.959) |
| 1° AVB | 0.848 (0.832–0.864) q = 0.002 * | **0.884 (0.871–0.896)** q = 0.315 | 0.876 (0.861–0.890) q = 0.577 | 0.879 (0.864–0.893) |
| QT Prolong | 0.242 (0.214–0.271) q = 0.002 * | **0.305 (0.274–0.338)** q = 0.065 | 0.294 (0.263–0.327) q = 0.743 | 0.291 (0.260–0.322) |
| ER | 0.261 (0.216–0.314) q = 0.002 * | 0.305 (0.260–0.358) q = 0.083 | 0.315 (0.269–0.368) q = 0.584 | **0.326 (0.278–0.379)** |
| LAFB | 0.482 (0.433–0.532) q = 0.133 | 0.531 (0.477–0.581) q = 0.293 | **0.545 (0.495–0.600)** q = 0.065 | 0.510 (0.455–0.567) |

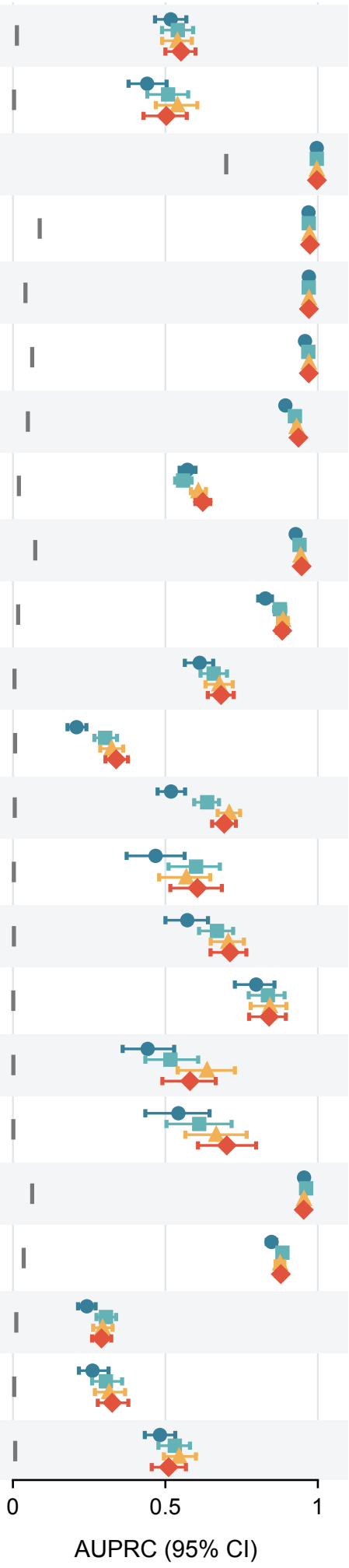

| Target | Scratch | ECG-FM | DeepECG-SSL | CardioLoom |
|---|---|---|---|---|
| LBBB | 0.931 (0.911–0.948) q = 0.048 * | 0.941 (0.923–0.956) q = 0.583 | 0.937 (0.919–0.951) q = 0.126 | **0.944 (0.926–0.959)** |
| IVB | 0.638 (0.583–0.691) q = 0.002 * | **0.722 (0.674–0.768)** q = 0.062 | 0.690 (0.639–0.740) q = 0.631 | 0.700 (0.652–0.746) |
| Short PR | 0.235 (0.179–0.307) q = 0.258 | 0.257 (0.201–0.320) q = 0.527 | **0.300 (0.236–0.374)** q = 0.362 | 0.271 (0.212–0.340) |
| VPE | 0.766 (0.668–0.847) q = 0.004 * | 0.847 (0.777–0.903) q = 0.703 | 0.850 (0.785–0.908) q = 0.655 | **0.857 (0.796–0.910)** |
| 3° AVB | 0.450 (0.374–0.527) q = 0.002 * | 0.600 (0.527–0.661) q = 0.062 | 0.622 (0.546–0.683) q = 0.213 | **0.656 (0.590–0.715)** |
| 2° 1 Type AVB | 0.754 (0.665–0.821) q = 0.002 * | 0.885 (0.826–0.927) q = 0.327 | 0.887 (0.829–0.931) q = 0.459 | **0.903 (0.849–0.938)** |
| 2° 2 Type AVB | 0.575 (0.409–0.717) q = 0.009 * | 0.749 (0.629–0.850) q = 0.667 | **0.820 (0.708–0.901)** q = 0.313 | 0.775 (0.646–0.874) |
| LVH | 0.704 (0.687–0.719) q = 0.002 * | 0.740 (0.724–0.753) q = 0.004 * | 0.742 (0.726–0.755) q = 0.004 * | **0.756 (0.741–0.769)** |
| LAH | 0.431 (0.401–0.462) q = 0.095 | 0.439 (0.408–0.472) q = 0.396 | 0.441 (0.411–0.472) q = 0.524 | **0.447 (0.416–0.479)** |
| RVH | 0.467 (0.423–0.513) q = 0.002 * | **0.560 (0.512–0.601)** q = 0.609 | 0.545 (0.497–0.591) q = 0.592 | 0.553 (0.503–0.596) |
| VA | 0.229 (0.171–0.300) q = 0.311 | 0.236 (0.174–0.307) q = 0.293 | 0.228 (0.173–0.291) q = 0.339 | **0.258 (0.196–0.331)** |
| RAH | 0.142 (0.080–0.231) q = 0.805 | 0.101 (0.058–0.175) q = 0.124 | **0.157 (0.089–0.255)** q = 0.928 | 0.153 (0.080–0.262) |
| DC | 0.001 (0.000–0.003) q = 0.204 | 0.008 (0.001–0.033) q = 0.481 | 0.004 (0.000–0.015) q = 0.339 | **0.022 (0.003–0.067)** |
| BVH | **0.058 (0.021–0.156)** q = 0.874 | 0.034 (0.014–0.101) q = 0.597 | 0.050 (0.022–0.110) q = 0.962 | 0.050 (0.021–0.136) |
| VVI | 0.825 (0.789–0.861) q = 0.020 * | 0.827 (0.792–0.860) q = 0.004 * | 0.855 (0.819–0.889) q = 0.527 | **0.865 (0.834–0.892)** |
| VAT | 0.751 (0.685–0.810) q = 0.002 * | **0.906 (0.875–0.933)** q = 0.470 | 0.868 (0.826–0.906) q = 0.062 | 0.895 (0.855–0.925) |
| DDD | 0.640 (0.562–0.715) q = 0.002 * | **0.849 (0.794–0.896)** q = 0.396 | 0.775 (0.701–0.843) q = 0.050 | 0.828 (0.773–0.876) |
| AAI | 0.656 (0.554–0.747) q = 0.002 * | 0.835 (0.756–0.904) q = 0.933 | **0.842 (0.781–0.891)** q = 0.852 | 0.838 (0.778–0.886) |
| VP | 0.145 (0.101–0.206) q = 0.067 | 0.198 (0.139–0.278) q = 0.459 | 0.098 (0.073–0.133) q = 0.012 * | **0.222 (0.153–0.302)** |
| PSM | 0.263 (0.127–0.446) q = 0.016 * | 0.412 (0.239–0.592) q = 0.449 | 0.347 (0.192–0.541) q = 0.099 | **0.489 (0.316–0.672)** |
| HK | 0.022 (0.009–0.051) q = 0.036 * | **0.108 (0.035–0.208)** q = 0.274 | 0.084 (0.037–0.191) q = 0.591 | 0.066 (0.034–0.121) |
| LK | 0.015 (0.008–0.031) q = 0.054 | 0.042 (0.024–0.077) q = 0.220 | 0.063 (0.022–0.129) q = 0.839 | **0.070 (0.032–0.133)** |

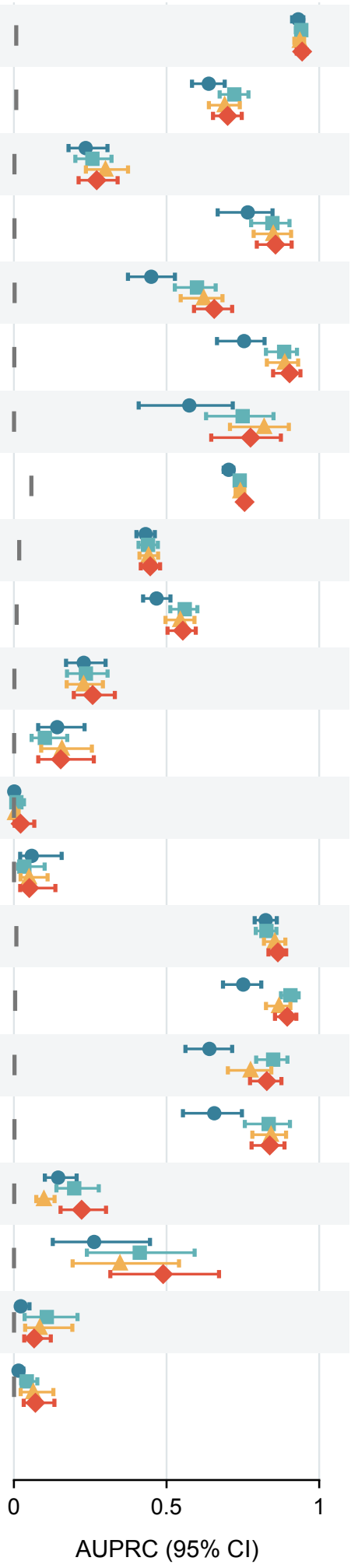

AUPRCs and 95% patient-bootstrap CIs for 4 models across 45 ECG targets in Zhongshan. Each baseline q value compares that model with CardioLoom using paired bootstrap tests and Benjamini–Hochberg adjustment. An asterisk denotes q<0.05 in either direction. Bold type marks the highest displayed estimate, and gray ticks mark prevalence.

**Supplementary Fig. 3. Detailed Echocardiographic Model Comparisons**

Zhongshan: echocardiographic phenotypes

Scratch ECG-FM DeepECG-SSL CardioLoom | Prevalence

| Target | Scratch | ECG-FM | DeepECG-SSL | CardioLoom |
|---|---|---|---|---|
| NER | 0.850 (0.843–0.858)<br>q = 0.002 * | **0.878 (0.872–0.884)**<br>q = 0.004 * | 0.875 (0.869–0.882)<br>q = 0.012 * | 0.871 (0.864–0.878) |
| LAE | 0.420 (0.405–0.435)<br>q = 0.002 * | 0.470 (0.454–0.486)<br>q = 0.282 | 0.463 (0.448–0.479)<br>q = 0.015 * | **0.476 (0.460–0.492)** |
| HCM | 0.431 (0.413–0.450)<br>q = 0.002 * | 0.481 (0.460–0.502)<br>q = 0.440 | **0.489 (0.468–0.509)**<br>q = 0.624 | 0.486 (0.465–0.506) |
| nHCM | 0.346 (0.329–0.365)<br>q = 0.002 * | 0.386 (0.367–0.406)<br>q = 0.426 | 0.390 (0.370–0.410)<br>q = 0.819 | **0.391 (0.371–0.413)** |
| BAE | 0.405 (0.376–0.431)<br>q = 0.002 * | **0.504 (0.472–0.532)**<br>q = 0.928 | 0.494 (0.463–0.523)<br>q = 0.396 | 0.503 (0.472–0.529) |
| HFrEF | 0.563 (0.520–0.602)<br>q = 0.002 * | 0.670 (0.630–0.707)<br>q = 0.157 | 0.676 (0.636–0.711)<br>q = 0.210 | **0.690 (0.648–0.725)** |
| RMWSSFLV | 0.479 (0.434–0.526)<br>q = 0.002 * | 0.526 (0.480–0.571)<br>q = 0.399 | 0.535 (0.489–0.585)<br>q = 0.692 | **0.541 (0.494–0.592)** |
| DCM | 0.361 (0.309–0.414)<br>q = 0.009 * | 0.389 (0.334–0.444)<br>q = 0.246 | **0.446 (0.388–0.501)**<br>q = 0.229 | 0.414 (0.361–0.471) |
| oHCM | 0.213 (0.172–0.260)<br>q = 0.002 * | 0.369 (0.304–0.431)<br>q = 0.022 * | 0.411 (0.347–0.477)<br>q = 0.464 | **0.429 (0.359–0.497)** |
| ICM | 0.201 (0.155–0.259)<br>q = 0.146 | 0.217 (0.168–0.272)<br>q = 0.423 | 0.208 (0.166–0.266)<br>q = 0.210 | **0.243 (0.183–0.315)** |
| LVDD | 0.035 (0.027–0.048)<br>q = 0.372 | 0.038 (0.030–0.049)<br>q = 0.855 | **0.041 (0.032–0.053)**<br>q = 0.511 | 0.039 (0.031–0.050) |
| VA | 0.163 (0.109–0.234)<br>q = 0.037 * | 0.200 (0.131–0.296)<br>q = 0.422 | 0.206 (0.145–0.286)<br>q = 0.237 | **0.232 (0.168–0.306)** |
| RAE | **0.017 (0.011–0.027)**<br>q = 0.948 | 0.014 (0.010–0.022)<br>q = 0.527 | 0.016 (0.011–0.023)<br>q = 0.839 | 0.016 (0.011–0.025) |
| AM | 0.013 (0.005–0.032)<br>q = 0.346 | 0.027 (0.012–0.054)<br>q = 0.509 | **0.035 (0.011–0.081)**<br>q = 0.527 | 0.022 (0.010–0.046) |
| LAM | 0.009 (0.003–0.019)<br>q = 0.320 | 0.015 (0.004–0.036)<br>q = 0.567 | 0.026 (0.008–0.052)<br>q = 0.890 | **0.028 (0.006–0.088)** |
| AVC | 0.359 (0.343–0.374)<br>q = 0.002 * | 0.412 (0.394–0.430)<br>q = 0.577 | 0.400 (0.382–0.418)<br>q = 0.004 * | **0.415 (0.399–0.433)** |
| TR | 0.528 (0.505–0.553)<br>q = 0.002 * | 0.579 (0.556–0.602)<br>q = 0.180 | **0.590 (0.567–0.614)**<br>q = 0.851 | 0.588 (0.566–0.612) |
| MR | 0.416 (0.391–0.441)<br>q = 0.002 * | **0.500 (0.474–0.525)**<br>q = 0.042 * | 0.483 (0.457–0.508)<br>q = 0.683 | 0.478 (0.452–0.503) |
| AR | 0.292 (0.263–0.323)<br>q = 0.002 * | **0.404 (0.371–0.436)**<br>q = 0.391 | 0.397 (0.365–0.430)<br>q = 0.774 | 0.395 (0.361–0.427) |
| MVPLAC | 0.072 (0.061–0.084)<br>q = 0.004 * | **0.092 (0.079–0.107)**<br>q = 0.465 | 0.086 (0.074–0.100)<br>q = 0.759 | 0.088 (0.075–0.103) |

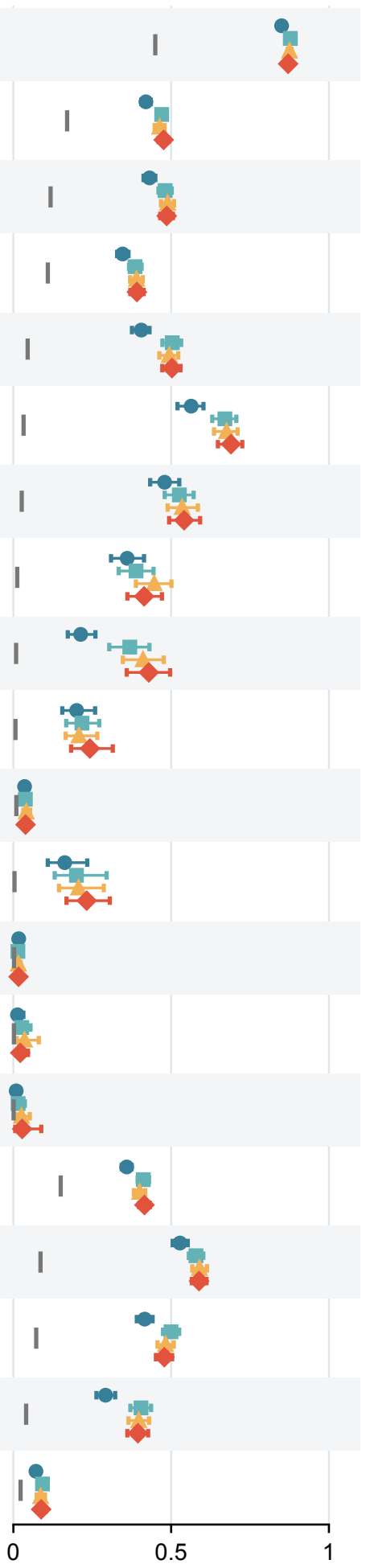

| Target | Scratch | ECG-FM | DeepECG-SSL | CardioLoom |
|---|---|---|---|---|
| MS | 0.428 (0.378–0.476)<br>q = 0.002 * | **0.584 (0.537–0.632)**<br>q = 0.658 | 0.574 (0.525–0.621)<br>q = 0.983 | 0.574 (0.521–0.627) |
| AS | 0.200 (0.163–0.241)<br>q = 0.002 * | 0.348 (0.299–0.399)<br>q = 0.727 | **0.361 (0.311–0.414)**<br>q = 0.262 | 0.341 (0.291–0.391) |
| BAV | 0.046 (0.038–0.059)<br>q = 0.002 * | 0.113 (0.088–0.144)<br>q = 0.910 | **0.115 (0.089–0.147)**<br>q = 0.989 | **0.115 (0.090–0.149)** |
| PS | **0.086 (0.023–0.211)**<br>q = 0.518 | 0.054 (0.011–0.161)<br>q = 0.993 | 0.060 (0.017–0.151)<br>q = 0.946 | 0.054 (0.019–0.111) |
| EA | **0.143 (0.005–0.339)**<br>q = 0.258 | 0.004 (0.001–0.008)<br>q = 0.224 | 0.014 (0.003–0.053)<br>q = 0.480 | 0.023 (0.009–0.049) |
| PR | 0.007 (0.000–0.031)<br>q = 0.661 | 0.003 (0.001–0.016)<br>q = 0.946 | 0.004 (0.000–0.016)<br>q = 0.946 | **0.009 (0.001–0.038)** |
| TS | 0.003 (0.000–0.012)<br>q = 0.851 | **0.006 (0.001–0.029)**<br>q = 0.946 | 0.002 (0.000–0.006)<br>q = 0.946 | 0.002 (0.000–0.007) |
| CHD | 0.342 (0.299–0.385)<br>q = 0.002 * | 0.455 (0.411–0.498)<br>q = 0.759 | 0.451 (0.406–0.493)<br>q = 0.527 | **0.460 (0.414–0.506)** |
| ASD | 0.483 (0.435–0.532)<br>q = 0.002 * | 0.622 (0.577–0.665)<br>q = 0.481 | **0.644 (0.600–0.688)**<br>q = 0.334 | 0.631 (0.585–0.675) |
| VSD | 0.108 (0.075–0.152)<br>q = 0.002 * | 0.262 (0.194–0.333)<br>q = 0.597 | 0.205 (0.151–0.262)<br>q = 0.004 * | **0.274 (0.207–0.341)** |
| PFO | 0.014 (0.010–0.021)<br>q = 0.116 | 0.021 (0.014–0.032)<br>q = 0.484 | **0.026 (0.015–0.041)**<br>q = 0.910 | 0.025 (0.016–0.043) |
| PDA | 0.058 (0.027–0.118)<br>q = 0.070 | 0.092 (0.042–0.157)<br>q = 0.454 | 0.070 (0.033–0.134)<br>q = 0.124 | **0.123 (0.061–0.206)** |
| TOF | **0.003 (NE)**<br>q = NE | 0.000 (NE)<br>q = NE | 0.001 (NE)<br>q = NE | 0.000 (NE) |
| PH | 0.469 (0.452–0.486)<br>q = 0.002 * | **0.520 (0.503–0.537)**<br>q = 0.120 | 0.519 (0.503–0.537)<br>q = 0.167 | 0.512 (0.493–0.529) |
| AD | 0.256 (0.239–0.273)<br>q = 0.002 * | **0.307 (0.289–0.326)**<br>q = 0.959 | 0.302 (0.284–0.320)<br>q = 0.422 | **0.307 (0.288–0.325)** |
| Dex | **0.308 (0.092–0.573)**<br>q = 0.043 * | 0.001 (0.000–0.010)<br>q = 0.358 | 0.006 (0.000–0.081)<br>q = 0.665 | 0.011 (0.002–0.031) |
| PE | 0.255 (0.234–0.277)<br>q = 0.002 * | **0.310 (0.284–0.338)**<br>q = 0.112 | 0.302 (0.276–0.328)<br>q = 0.454 | 0.294 (0.269–0.320) |
| CP | 0.135 (0.055–0.251)<br>q = 0.008 * | **0.340 (0.178–0.522)**<br>q = 0.839 | 0.311 (0.181–0.446)<br>q = 0.910 | 0.320 (0.180–0.478) |
| AF | 0.979 (0.977–0.981)<br>q = 0.002 * | 0.992 (0.991–0.993)<br>q = 0.004 * | **0.993 (0.992–0.994)**<br>q = 0.440 | **0.993 (0.992–0.994)** |

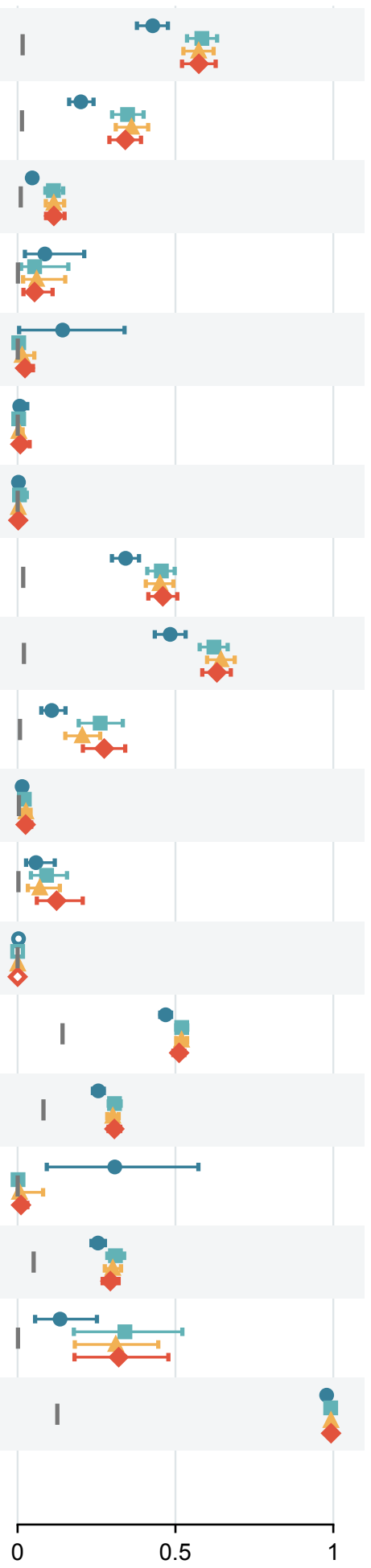

AUPRCs and 95% patient-bootstrap CIs for 4 models across 39 echocardiographic targets in Zhongshan. Each baseline q value compares that model with CardioLoom using paired bootstrap tests and Benjamini–Hochberg adjustment. An asterisk denotes q<0.05 in either direction. Bold type marks the highest displayed estimate, and gray ticks mark prevalence. NE denotes a nonestimable comparison. Hollow markers indicate unavailable CIs.

**Supplementary Fig. 4. Additional Clinical Operating Characteristics**

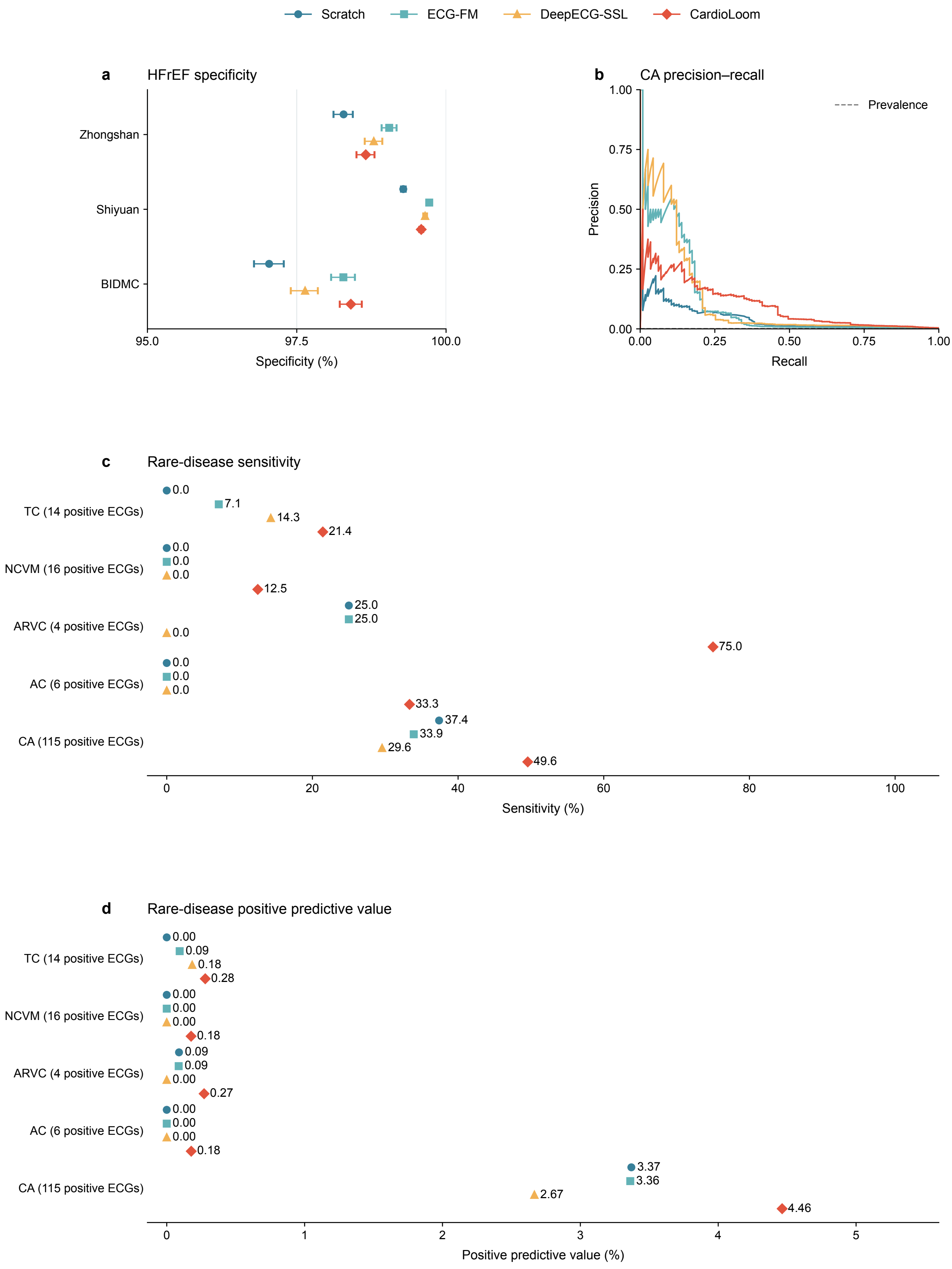

a, HFrEF specificity in Zhongshan, Shiyuan, and BIDMC at each model's Zhongshan validation-F1 threshold. Error bars show 95% patient-bootstrap CIs. b, Cardiac amyloidosis precision-recall curves, with prevalence as the reference line. c–d, Sensitivity and positive predictive value for 5 rare diseases at validation thresholds targeting 99% specificity. Positive counts refer to ECG records. CA intervals are reported in Supplementary Table 31.

**Supplementary Fig. 5. Internal Ablation Study of Report-Aware Pretraining**

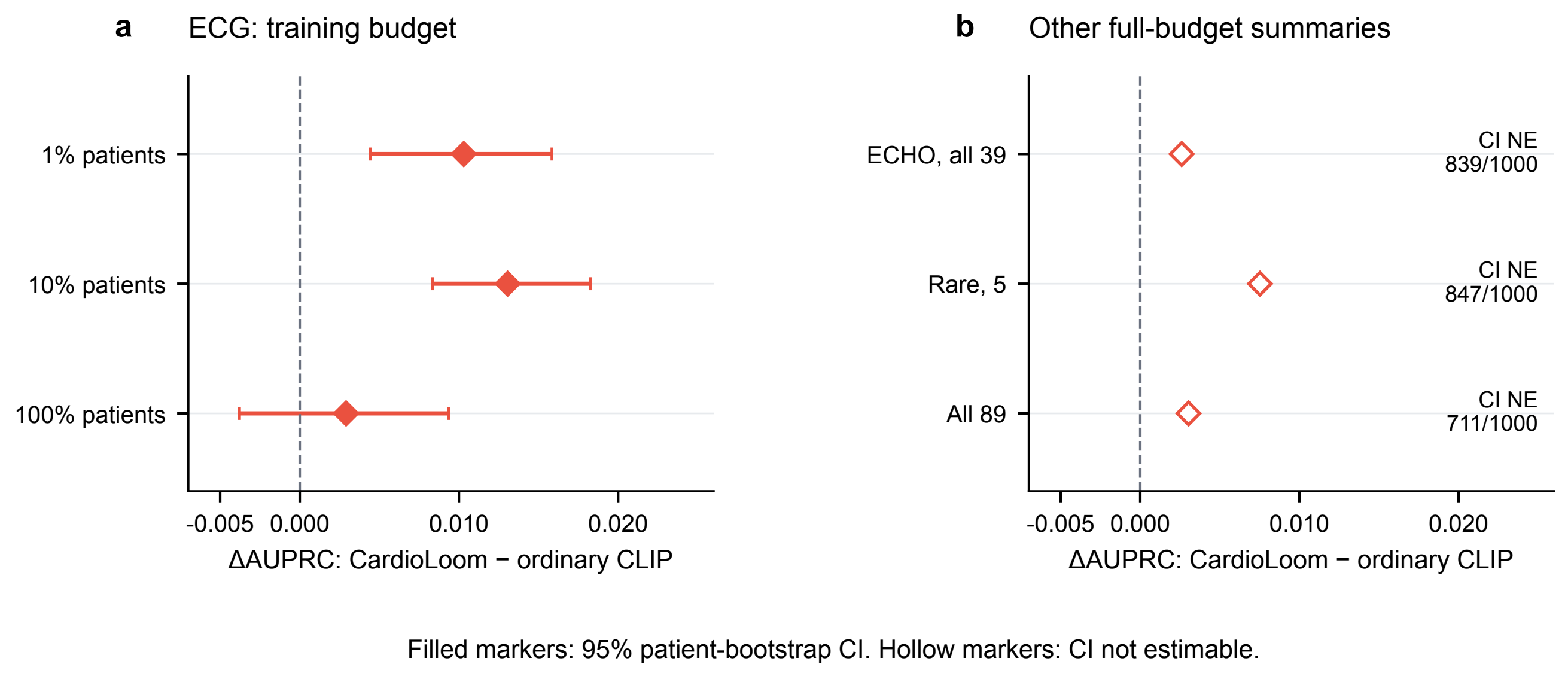

a, CardioLoom-minus-ordinary-CLIP macro-AUPRC across 45 ECG targets at 3 training-patient budgets. b, Full-budget differences across 39 echocardiographic targets, 5 rare diseases, and all 89 targets. Positive differences favor CardioLoom. Horizontal bars show 95% paired patient-bootstrap CIs. Hollow markers denote nonestimable intervals. The label n indicates the number of valid bootstrap replicates. Complete comparisons are in Supplementary Table 29.

**Supplementary Fig. 6. CardioLoom Representations Across Clinical Categories and Diagnoses**

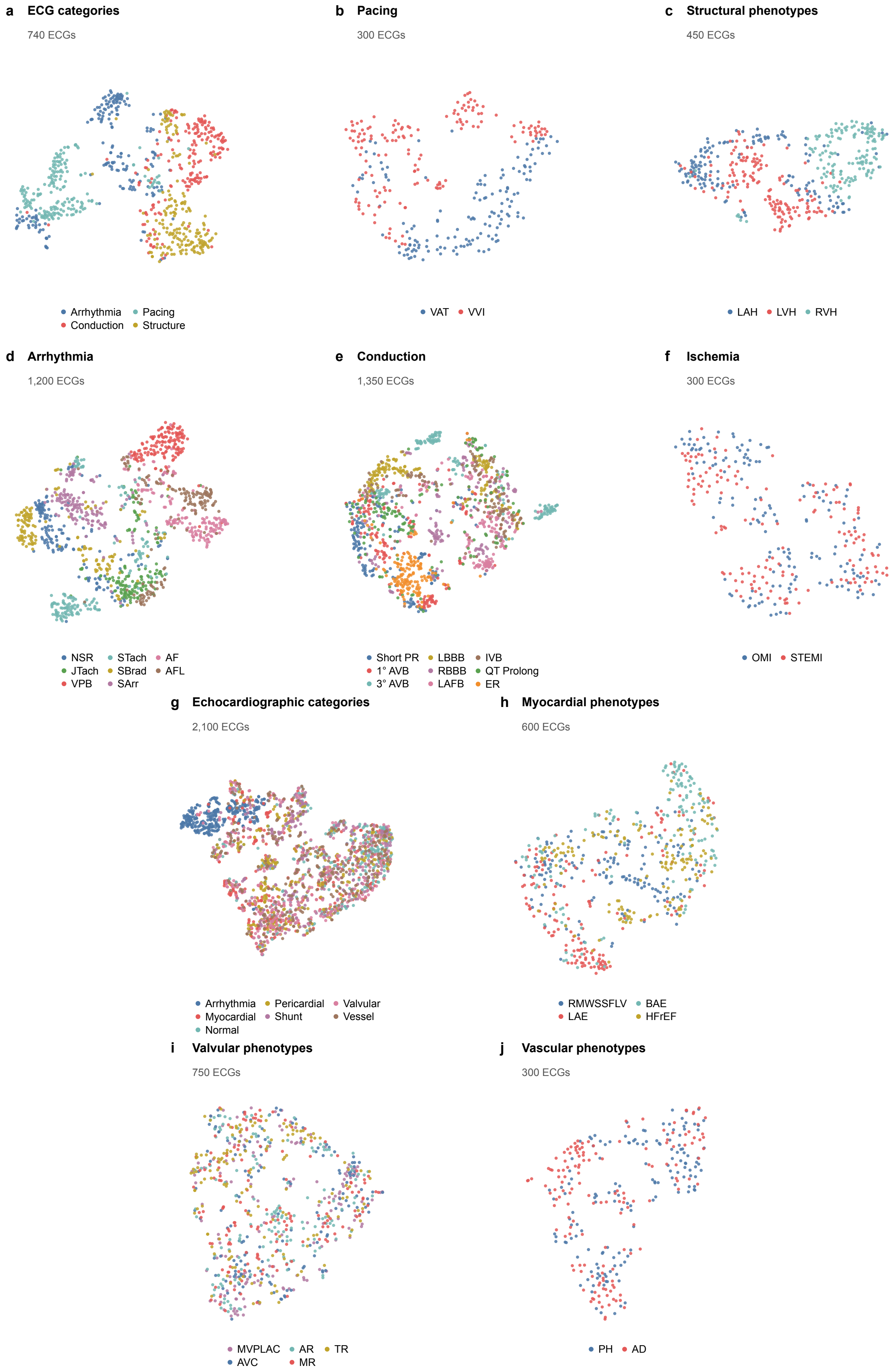

t-SNE projections of pretrained CardioLoom features. a and g, ECG and echocardiographic clinical categories. b–f, ECG diagnoses within pacing, structural, arrhythmia, conduction, and ischemia categories. h–j, Echocardiographic diagnoses within myocardial, valvular, and vascular categories. Each point represents one ECG, colored by category or diagnosis. Counts indicate ECG records. Sampling and projection methods are described in the Methods. Diagnostic abbreviations are defined in Supplementary Table 3.

**Supplementary Fig. 7. Integrated-Gradient Examples**

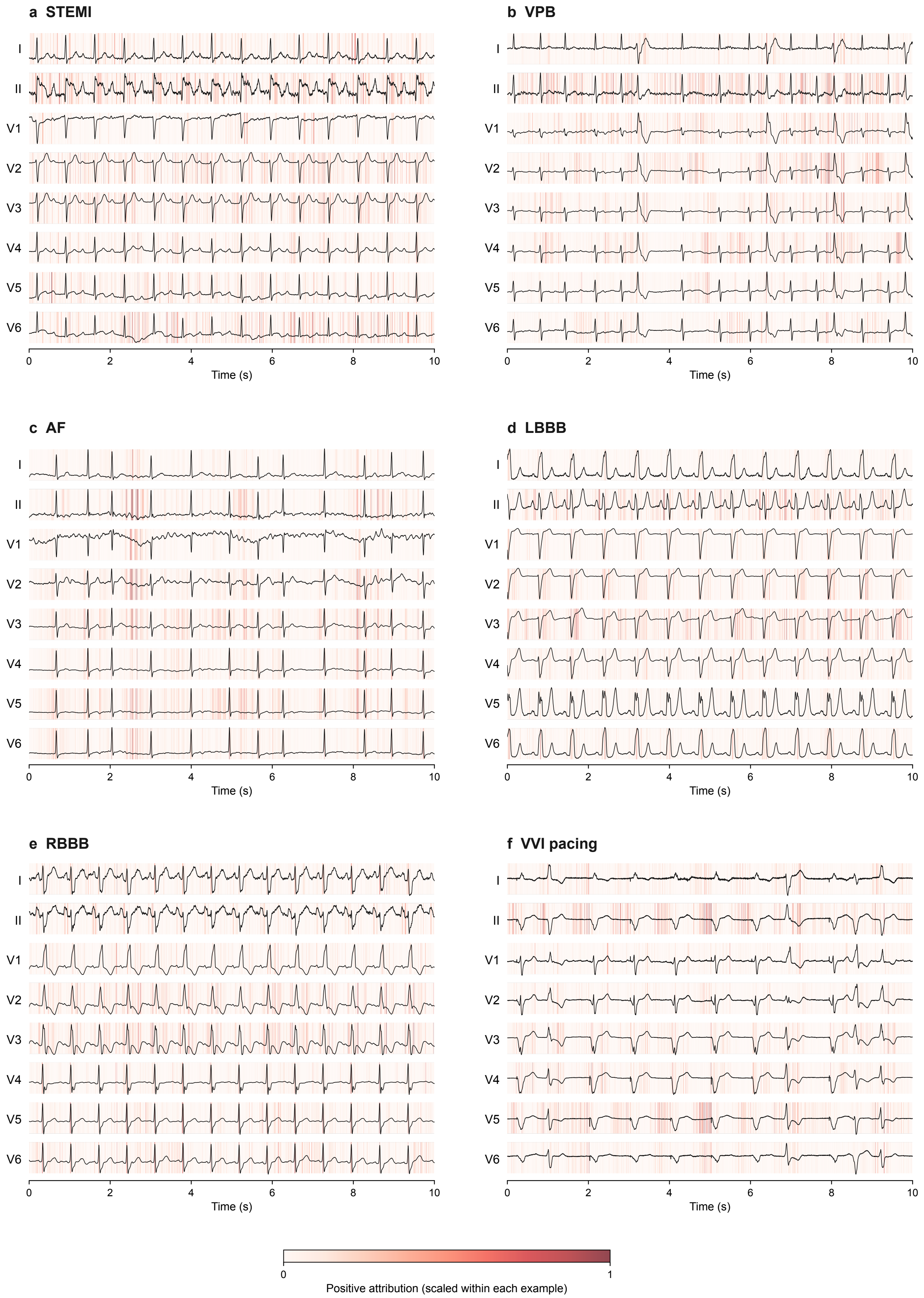

Positive integrated gradients are overlaid on eight input ECG leads from six clinician-selected true-positive examples. a, ST-elevation myocardial infarction (STEMI). b, Ventricular premature beats (VPB). c, Atrial fibrillation (AF). d, Left bundle branch block (LBBB). e, Right bundle branch block (RBBB). f, VVI pacing. Black traces show ECG waveforms. Red intensity represents positive attribution, scaled within each example. Waveforms are scaled separately by lead for display.

**Supplementary Table 1: Overview of data cohorts.**

| Cohort Category | Region (Country) | Cohort Name | Population count (unit) | # ECG | Source |
|---|---|---|---|---|---|
| Internal Development Cohort (Pretraining & Tuning) | Asia (China) | Zhongshan | 1,558,654 patients | 3,559,279 | Hospital |
| External evaluation cohorts | Asia (China) | Shiyuan | 143,078 patients | 219,428 | Hospital |
| | North America (USA) | BIDMC | 67,827 patients | 146,001 | Hospital |
| | Asia (China) | Xiamen | 29,705 patients | 44,362 | Hospital |
| | North America (USA) | MIMIC-IV-ECG | 160,411 patients | 772,155 | Public Repository |
| | Europe (United Kingdom) | UKB | 50,399 participants | 50,407 | Public Repository |
| | Asia (China) | Chapman | 45,151 patients | 45,151 | Public Repository |
| | Europe (Germany) | PTB-XL | 18,869 patients | 21,799 | Public Repository |
| | North America (USA) | Georgia | Not available | 10,344 | Public Repository |
| | Asia (China) | CPSC2018 | Not available | 6,877 | Public Repository |
| Source ECG total* | - | - | Not pooled* | 4,875,803 | - |

*Counts describe source inventories. Evaluable populations are reported by target. Hospital inventories were supplied by the contributing institutions. UK Biobank counts denote participants. Patient counts for the public Georgia and CPSC2018 subsets are unavailable. Xiamen is affiliated with Zhongshan.

**Supplementary Table 2: Detailed data partitioning of the Internal Development Cohort.**

| Phase | Subset | Allocation | # Patients | # ECG |
|---|---|---|---|---|
| Pretraining | Training | 98% | 1,298,359 | 2,781,289 |
| | Validation | 2% | 26,497 | 56,673 |
| | Total | 100% | 1,324,856 | 2,837,962 |
| Downstream Development | Training | 70% | 163,700 | 505,171 |
| | Validation | 15% | 35,049 | 107,973 |
| | Test | 15% | 35,049 | 108,173 |
| | Total | 100% | 233,798 | 721,317 |

Internal development partitions were separated at the patient level.

**Supplementary Table 3: Diagnostic tasks and positive ECG counts in the effective internal analysis populations.**

| Category | Abbr. | Full Diagnostic Name | Training n (%) | Validation n (%) | Test n (%) |
|---|---|---|---|---|---|
| **Tier1: ECG Interpretation** | | | | | |
| Ischemia | OMI[a] | Old Myocardial Infarction | 5,749 (1.14) | 1,281 (1.19) | 1,411 (1.30) |
| | STEMI[a] | Acute ST-Elevation Myocardial Infarction | 1,714 (0.34) | 402 (0.37) | 401 (0.37) |
| Arrhythmia | NSR[a] | Normal Sinus Rhythm | 352,913 (69.86) | 75,840 (70.24) | 75,675 (69.96) |
| | SBrad[a] | Sinus Bradycardia | 43,646 (8.64) | 9,408 (8.71) | 9,506 (8.79) |
| | STach[a] | Sinus Tachycardia | 21,343 (4.22) | 4,592 (4.25) | 4,450 (4.11) |
| | VPB[a] | Ventricular Premature Beat | 31,471 (6.23) | 6,815 (6.31) | 6,837 (6.32) |
| | APB[a] | Atrial Premature Beat | 24,913 (4.93) | 5,416 (5.02) | 5,308 (4.91) |
| | SArr[a] | Sinus Arrhythmia | 9,621 (1.90) | 2,134 (1.98) | 2,119 (1.96) |
| | AF[a] | Atrial Fibrillation | 38,074 (7.54) | 7,769 (7.20) | 7,919 (7.32) |
| | AFL[a] | Atrial Flutter | 8,328 (1.65) | 1,714 (1.59) | 1,853 (1.71) |
| | JPB[a] | Atrioventricular Junctional Premature Beat | 2,667 (0.53) | 572 (0.53) | 564 (0.52) |
| | AJR[a] | Accelerated Atrioventricular Junctional Rhythm | 3,454 (0.68) | 792 (0.73) | 769 (0.71) |
| | ATach[a] | Atrial Tachycardia | 3,074 (0.61) | 666 (0.62) | 649 (0.60) |
| | JTach[a] | Atrioventricular Junctional Tachycardia | 1,521 (0.30) | 326 (0.30) | 328 (0.30) |
| | JEB[a] | Atrioventricular Junctional Escape Beat | 1,727 (0.34) | 334 (0.31) | 353 (0.33) |
| | SVT[a] | Supraventricular Tachycardia | 664 (0.13) | 160 (0.15) | 126 (0.12) |
| | VEB[a] | Ventricular Escape Beat | 622 (0.12) | 118 (0.11) | 161 (0.15) |
| | VTach[a] | Ventricular Tachycardia | 619 (0.12) | 139 (0.13) | 149 (0.14) |
| Conduction | RBBB[a] | Right Bundle Branch Block | 32,175 (6.37) | 6,991 (6.47) | 6,832 (6.32) |
| Conduction | 1° AVB[a] | First-Degree Atrioventricular Block | 18,671 (3.70) | 4,266 (3.95) | 3,823 (3.53) |
| | QT Prolong[a] | Prolonged QT Interval | 5,963 (1.18) | 1,255 (1.16) | 1,198 (1.11) |
| | ER[a] | Early Repolarization | 2,328 (0.46) | 457 (0.42) | 459 (0.42) |
| | LAFB[a] | Left Anterior Fascicular Block | 3,895 (0.77) | 878 (0.81) | 807 (0.75) |
| | LBBB[a] | Left Bundle Branch Block | 4,481 (0.89) | 1,056 (0.98) | 828 (0.77) |
| | IVB[a] | Intraventricular Block | 3,601 (0.71) | 888 (0.82) | 835 (0.77) |

| Category | Abbr. | Full Diagnostic Name | Training n (%) | Validation n (%) | Test n (%) |
|---|---|---|---|---|---|
| | Short PR[a] | Short PR Interval | 1,010 (0.20) | 195 (0.18) | 192 (0.18) |
| | VPE[a] | Ventricular Pre-Excitation | 901 (0.18) | 184 (0.17) | 202 (0.19) |
| | 3° AVB[a] | Third-Degree Atrioventricular Block | 1,232 (0.24) | 200 (0.19) | 249 (0.23) |
| | 2° 1 Type AVB[a] | Second-Degree AV Block, Mobitz Type I | 675 (0.13) | 112 (0.10) | 136 (0.13) |
| | 2° 2 Type AVB[a] | Second-Degree AV Block, Mobitz Type II | 299 (0.06) | 55 (0.05) | 50 (0.05) |
| | LVH[a] | Left Ventricular Hypertrophy | 29,285 (5.80) | 6,528 (6.05) | 6,189 (5.72) |
| | LAH[a] | Left Atrial Hypertrophy | 9,110 (1.80) | 2,058 (1.91) | 1,883 (1.74) |
| | RVH[a] | Right Ventricular Hypertrophy | 3,879 (0.77) | 827 (0.77) | 984 (0.91) |
| Structure | VA[a] | Ventricular Aneurysm | 888 (0.18) | 212 (0.20) | 174 (0.16) |
| | RAH[a] | Right Atrial Hypertrophy | 407 (0.08) | 87 (0.08) | 89 (0.08) |
| | DC[a] | Dextrocardia | 143 (0.03) | 36 (0.03) | 24 (0.02) |
| | BVH[a] | Biventricular Hypertrophy | 186 (0.04) | 36 (0.03) | 44 (0.04) |
| | VVI[a] | VVI Pacing | 4,338 (0.86) | 798 (0.74) | 874 (0.81) |
| | VAT[a] | VAT Pacing | 2,219 (0.44) | 446 (0.41) | 443 (0.41) |
| Pacing | DDD[a] | DDD Pacing | 1,013 (0.20) | 244 (0.23) | 219 (0.20) |
| | AAI[a] | AAI Pacing | 875 (0.17) | 174 (0.16) | 196 (0.18) |
| | VP[a] | Ventricular Pacing | 477 (0.09) | 107 (0.10) | 117 (0.11) |
| Pacing | PSM[a] | Pacing/Sensing Malfunction | 193 (0.04) | 28 (0.03) | 35 (0.03) |
| Electrolyte abnormality | HK[a] | Hyperkalemia | 202 (0.04) | 43 (0.04) | 45 (0.04) |
| | LK[a] | Hypokalemia | 213 (0.04) | 47 (0.04) | 56 (0.05) |
| **Tier 2: ECHO Diagnosis** | | | | | |
| Normal | NER[b] | Normal Echocardiogram at Rest | 97,637 (44.40) | 20,773 (44.38) | 21,117 (44.97) |
| | LAE[b] | Left Atrial Enlargement | 37,878 (17.22) | 8,097 (17.30) | 7,997 (17.03) |
| | HCM[b] | Hypertrophic Cardiomyopathy | 27,086 (12.32) | 5,695 (12.17) | 5,538 (11.79) |
| Myocardial | nHCM[b] | Non-Obstructive Hypertrophic Cardiomyopathy | 25,126 (11.43) | 5,258 (11.23) | 5,130 (10.93) |
| | BAE[b] | Bi-Atrial Enlargement | 10,079 (4.58) | 2,128 (4.55) | 2,127 (4.53) |

| Category | Abbr. | Full Diagnostic Name | Training n (%) | Validation n (%) | Test n (%) |
|---|---|---|---|---|---|
| | HFrEF[c] | Reduced LVEF (<40%, operational HFrEF label) | 5,712 (3.17) | 1,176 (3.06) | 1,242 (3.23) |
| | RMWSSFLV[b] | Reduced Multi-Wall Segment Systolic Function of Left Ventricle | 5,889 (2.68) | 1,304 (2.79) | 1,244 (2.65) |
| | DCM[b] | Dilated Cardiomyopathy | 2,997 (1.36) | 615 (1.31) | 577 (1.23) |
| | oHCM[b] | Obstructive Hypertrophic Cardiomyopathy | 1,960 (0.89) | 437 (0.93) | 408 (0.87) |
| | ICM[b] | Ischemic Cardiomyopathy | 1,302 (0.59) | 321 (0.69) | 312 (0.66) |
| | LVDD[b] | Left Ventricular Diastolic Dysfunction | 1,910 (0.87) | 418 (0.89) | 443 (0.94) |
| | VA[b] | Ventricular Aneurysm | 590 (0.27) | 161 (0.34) | 169 (0.36) |
| | RAE[b] | Right Atrial Enlargement | 428 (0.19) | 83 (0.18) | 101 (0.22) |
| | AM[b] | Atrial Myxoma | 321 (0.15) | 60 (0.13) | 81 (0.17) |
| | LAM[b] | Left Atrial Mass | 155 (0.07) | 44 (0.09) | 36 (0.08) |
| | AVC[b] | Aortic Valve Calcification | 33,489 (15.23) | 7,232 (15.45) | 7,046 (15.01) |
| | TR[d] | Tricuspid Regurgitation | 19,123 (8.70) | 3,941 (8.36) | 4,071 (8.60) |
| | MR[d] | Mitral Regurgitation | 16,375 (7.45) | 3,521 (7.47) | 3,426 (7.24) |
| | AR[d] | Aortic Regurgitation | 9,135 (4.15) | 1,935 (4.10) | 1,914 (4.04) |
| | MVPLAC[b] | Mitral Valve Posterior Leaflet Annular Calcification | 5,011 (2.28) | 1,075 (2.30) | 1,074 (2.29) |
| Valvular | MS[d] | Mitral Stenosis | 4,018 (1.83) | 799 (1.70) | 757 (1.60) |
| | AS[d] | Aortic Stenosis | 2,983 (1.36) | 639 (1.36) | 653 (1.38) |
| | BAV[b] | Bicuspid Aortic Valve | 2,272 (1.03) | 512 (1.09) | 470 (1.00) |
| | PS[d] | Pulmonary Stenosis | 161 (0.07) | 40 (0.08) | 49 (0.10) |
| | EA[e] | Ebstein's Anomaly | 120 (0.05) | 36 (0.08) | 27 (0.06) |
| | PR[d] | Pulmonary Regurgitation | 99 (0.05) | 18 (0.04) | 21 (0.04) |
| | TS[d] | Tricuspid Stenosis | 36 (0.02) | 2 (0.00) | 12 (0.03) |
| | CHD[b] | Congenital Heart Disease | 4,082 (1.86) | 917 (1.96) | 836 (1.78) |
| Shunt | ASD[b] | Atrial Septal Defect | 4,508 (2.05) | 996 (2.13) | 944 (2.01) |
| | VSD[b] | Ventricular Septal Defect | 1,640 (0.75) | 358 (0.76) | 353 (0.75) |

| Category | Abbr. | Full Diagnostic Name | Training n (%) | Validation n (%) | Test n (%) |
|---|---|---|---|---|---|
| | PFO[b] | Patent Foramen Ovale | 1,001 (0.46) | 196 (0.42) | 198 (0.42) |
| | PDA[b] | Patent Ductus Arteriosus | 606 (0.28) | 118 (0.25) | 123 (0.26) |
| | TOF[e] | Tetralogy of Fallot | 6 (0.00) | 2 (0.00) | 2 (0.00) |
| | PH[b] | Pulmonary Hypertension | 31,494 (14.32) | 6,821 (14.57) | 6,684 (14.24) |
| Vessel | AD[b] | Aortic Dilation | 18,485 (8.41) | 3,998 (8.54) | 3,848 (8.20) |
| | Dex[e] | Dextrocardia | 99 (0.04) | 15 (0.03) | 15 (0.03) |
| Pericardial | PE[b] | Pericardial Effusion | 11,618 (5.28) | 2,471 (5.28) | 2,388 (5.09) |
| Pericardial | CP[b] | Constrictive Pericarditis | 248 (0.11) | 68 (0.15) | 57 (0.12) |
| Arrhythmia | AF[a] | Atrial Fibrillation | 64,960 (12.86) | 13,255 (12.28) | 13,615 (12.59) |
| **Tier 3: Rare Diseases** | | | | | |
| - | CA[a] | Cardiac Amyloidosis | 194 (0.04) | 85 (0.08) | 115 (0.11) |
| | ARVC[a] | Arrhythmogenic Right Ventricular Cardiomyopathy | 51 (0.01) | 7 (0.01) | 4 (0.00) |
| | TC[a] | Takotsubo Cardiomyopathy | 64 (0.01) | 17 (0.02) | 14 (0.01) |
| | AC[a] | Alcoholic Cardiomyopathy | 28 (0.01) | 2 (0.00) | 6 (0.01) |
| | NCVM[a] | Noncompaction of Ventricular Myocardium | 39 (0.01) | 1 (0.00) | 16 (0.01) |

Evaluable ECG records (N), training / validation / test:
a: ECG labels, standalone AF and rare diseases: 505,171 / 107,973 / 108,173.
b: Regular ECHO targets: 219,920 / 46,811 / 46,953.
c: HFrEF (LVEF <40%): 180,171 / 38,439 / 38,425.
d: AS, AR, TR, MR, MS, PR, PS and TS: 219,878 / 47,138 / 47,327.
e: EA, TOF and Dex: 222,423 / 47,335 / 47,476.

Values are positive ECG counts (percentages). Superscript letters identify the evaluable ECG denominator sets above the table. Multi-task models use records with observed labels for all modeled tasks. Single-task models use records with an observed endpoint label. Missing labels are excluded. HFrEF denotes LVEF <40%.

**Supplementary Table 4: Diagnostic label mapping criteria for the external Shiyuan test cohort.**

| Target Label (Zhongshan Ontology) | Mapping Criteria (English Translation [Original Chinese]) |
|---|---|
| NER | “Resting...No abnormalities seen” (静息...未见异常) |
| LAE | “Left Atrium...Enlarged” (左房…大) |
| HCM | “Thickening” (增厚) OR “Hypertrophy” (肥厚) |
| nHCM | “Non-obstructive” (非梗阻) |
| BAE | “Bi-Atrial...Enlarged” (双房...大) |
| HFrEF | † |
| RMWSSFLV | “LV...Systolic...Weakened” (左室...收缩...减弱) AND NOT “Non-global” (非整体) |
| DCM | “Dilated...Heart” (扩张...心) |
| oHCM | “Obstructive” (梗阻) AND NOT Negation terms (e.g., "No/Not/None" [未/不/无]) |
| LVDD | “LV Diastolic” (左室舒张) |
| VA | “Ventricular Aneurysm” (室壁瘤) |
| RAE | “Right Atrium...Enlarged” (右房...大) |
| AM | “Myxoma” (粘液瘤) |
| LAM | “Left Atrium...Occupying Lesion” (左房...占位) |
| AVC | “Aorta...Calcification” (主动脉...钙化) |
| TR | † |
| MR | † |
| AR | † |
| MVPLAC | “Mitral Valve...Calcification” (二尖瓣...钙化) |
| MS | † |
| AS | † |
| BAV | “Two-leaf...Aorta” (二叶...主动脉) |
| PS | † |
| EA | “Tricuspid downward displacement” (三尖瓣下移畸形) |
| PR | † |
| CHD | “Congenital Heart Disease” (先天性心脏病) |

| Target Label (Zhongshan Ontology) | Mapping Criteria (English Translation [Original Chinese]) |
|---|---|
| ASD | “Atrial Septal...Defect” (房间隔...缺损) |
| VSD | “Ventricular Septal...Defect” (室间隔...缺损) |
| PFO | “Patent Foramen Ovale” (卵圆孔未闭) |
| PDA | “Patent Ductus Arteriosus” (动脉导管未闭) |
| TOF | “Fallot” (法洛) |
| PH | “Pulmonary Artery...High Pressure” (肺动脉...高压) |
| AD | “Aorta...Widened” (主动脉...增宽) |
| Dex | “Dextrocardia” (右位心) |
| PE | “Pericardial Effusion” (心包积液) |
| CP | “Constrictive...Pericarditis” (缩窄...心包炎) |

†: Indicates labels retrieved directly from structured database fields (pre-existing diagnoses) within the Shiyuan dataset. All other labels were extracted via regular expression matching applied to the original clinical reports. Matching keywords are presented as English translations followed by the representative original Chinese terms (in parentheses).

**Supplementary Table 5: Diagnostic label mapping criteria for the external BIDMC test cohort.**

| Target Label (Zhongshan Ontology) | Mapping Criteria (Structured Variables & Thresholds) |
|---|---|
| LAE | la_size > 40 mm |
| BAE | ra_size AND la_size > 40 mm |
| HFrEF | biplane_lvef < 40 % |
| oHCM | lv_obstruction / lv_obstruction_valsalva |
| VA | lv_aneurysm |
| RAE | ra_size > 40 mm |
| TR | tricuspid_regurg |
| MR | mitral_regurg |
| AR | aortic_regurg |
| MS | mitral_stenosis |
| AS | aortic_stenosis |
| PS | pulm_stenosis |
| PR | pulm_regurg |
| VSD | vsd |
| PH | pasp_underestimated_tr > 40 mmHg |
| AD | ascending_diam > 37 mm |
| PE | pericardial_effusion |

The mapping criteria involve either applying clinical thresholds to quantitative measurements (e.g., chamber sizes) or identifying positive diagnostic flags in categorical fields.

**Supplementary Table 6: Diagnostic label mapping criteria for the external MIMIC-IV-ECG test cohort.**

| Target Label (Zhongshan Ontology) | Mapping Criteria (Original Diagnostic Terms) |
|---|---|
| OMI | old inferior myocardial infarction / old anterolateral myocardial infarction / old anterior myocardial infarction / old septal myocardial infarction / old lateral myocardial infarction / old posterior myocardial infarction / old anteroseptal myocardial infarction |
| STEMI | inferior myocardial infarction / anterior myocardial infarction / anteroseptal myocardial infarction / septal myocardial infarction / acute myocardial infarction / acute ST segment elevation myocardial infarction / lateral myocardial infarction / extensive myocardial infarction / inferior ST segment elevation / lateral ST segment elevation / anterolateral myocardial infarction / anteroseptal ST segment elevation / anterolateral ST segment elevation / inferior and lateral ST segment elevation / anterior ST segment elevation / extensive ST segment elevation / extensive ST elevation / septal ST segment elevation / acute inferior myocardial infarction / acute anterior myocardial infarction / acute lateral myocardial infarction / acute anterolateral myocardial infarction / inferior and anterior/septal ST segment elevation / acute posterior myocardial infarction / acute anteroseptal myocardial infarction / inferior and septal ST segment elevation |
| NSR | normal ECG |
| SBrad | sinus bradycardia / marked sinus bradycardia / irregular sinus bradycardia / slow sinus arrhythmia |
| STach | sinus tachycardia / irregular sinus tachycardia |
| VPB | ventricular premature complex / frequent ventricular premature complex / frequent multifocal ventricular premature complex / multifocal ventricular premature complex / interpolated ventricular premature complex / bigeminal ventricular premature complex / multiple premature ventricular complexes / multiform ventricular premature complexes / paired ventricular premature complexes / ventricular bigeminy / ventricular couplets / ventricular trigeminy / ventricular pre-excitation / multifocal interpolated ventricular premature complex |
| APB | atrial premature complex / bigeminal atrial premature complex / frequent atrial premature complex / occasional frequent atrial premature complex / atrial premature complexes in couplets |
| SArr | sinus arrhythmia |
| AF | artrial fibrillation |
| JPB | supraventricular premature beats / supraventricular bigeminy / aberrantly conducted supraventricular premature beats |
| AJR | accelerated junctional rhythm |
| ATach | ectopic atrial tachycardia / atrial tachycardia / sinus/ectopic atrial tachycardia |
| JTach | junctional tachycardia |
| SVT | supraventricular tachycardia |
| VEB | paroxysmal idioventricular rhythm |
| VTach | nonsustained ventricular tachycardia / ventricular tachycardia / unsustained ventricular tachycardia |
| RBBB | right bundle branch block / incomplete right bundle branch block / atypical right bundle branch block |
| 1° AVB | borderline first degree atrioventricular block / first degree atrioventricular block |

| **Target Label (Zhongshan Ontology)** | **Mapping Criteria (Original Diagnostic Terms)** |
|---|---|
| QT Prolong | prolonged QT interval / long QTc interval |
| ER | early repolarization |
| LAFB | left anterior fascicular block / anterior fascicular block |
| LBBB | left bundle branch block / incomplete left bundle branch block / atypical left bundle branch block |
| IVB | intraventricular conduction defect / non-specific intraventricular conduction delay / non-specific intraventricular conduction disturbance (block) / extensive intraventricular conduction defect |
| Short PR | short PR interval |
| VPE | Wolf-Parkinson-White syndrome |
| 3° AVB | complete atrioventricular block / third degree atrioventricular block |
| 2° 1 Type AVB | second degree atrioventricular block |
| 2° 2 Type AVB | 4:1 atrioventricular block / 3:1 atrioventricular block / second degree atrioventricular block / predominant 4:1 atrioventricular block / predominant 3:1 atrioventricular block |
| LVH | left ventricular hypertrophy / voltage criteria for left ventricular hypertrophy |
| LAH | left atrial enlargement / bi-atrial enlargement |
| RVH | right ventricular hypertrophy / bi-atrial enlargement |
| RAH | right atrial enlargement |
| DC | dextrocardia |
| BVH | ventricular hypertrophy / biventricular hypertrophy |
| VAT | dual chamber pacemaker |
| DDD | dual chamber pacemaker / atrioventricular sequential pacing pattern |
| HK | hyperkalemia |

The mapping schema aggregates diverse original diagnostic terms and synonym variations into the target standardized categories.

**Supplementary Table 7: Diagnostic label mapping criteria for the external UKB test cohort.**

| Target Label (Zhongshan Ontology) | Mapping Criteria |
|---|---|
| STEMI | "ACUTE MI" OR "acute...infarction" |
| NSR | "Normal ECG" OR "Normal Sinus Rhythm" |
| SBrad | "sinus bradycardia" |
| VPB | "premature ventricular" |
| APB | "premature atrial complexes" |
| SArr | "sinus arrhythmia" |
| AF | "Atrial fibrillation" |
| AFL | "Atrial flutter" |
| JPB | "premature supraventricular" OR "junctional pacemaker" |
| AJR | "Junctional rhythm" |
| JTach | "Junctional rhythm" |
| JEB | "Junctional rhythm" |
| RBBB | "Right bundle branch block" |
| 1° AVB | "1st degree AV block" |
| QT Prolong | "Prolonged QT" |
| ER | "Early repolarization" |
| LAFB | "Left anterior fascicular block" |
| LBBB | "Left bundle branch block" |
| IVB | "Nonspecific intraventricular" |
| Short PR | "short PR" |
| LVH | "Left ventricular hypertrophy" |
| LAH | "Left atrial enlargement" OR "Biatrial enlargement" |
| RVH | "Right ventricular hypertrophy" |
| RAH | Right atrial enlargement" OR "Biatrial enlargement" |

All labels were extracted via regular expression matching applied to the original clinical reports.

**Supplementary Table 8: Diagnostic label mapping criteria for the external Chapman test cohort.**

| Target Label (Zhongshan Ontology) | Mapping Criteria (Original Diagnostic Terms) |
|---|---|
| STEMI | MI (myocardial infarction) / MIBW (myocardial infraction in back wall) / MIFW (Myocardial infraction in the front wall) / MILW (Myocardial infraction in the lower wall) / MISW (Myocardial infraction in the side wall) |
| SBrad | SB (Sinus Bradycardia) |
| STach | ST (Sinus Tachycardia) |
| VPB | VB (ventricular bigeminy) / VPB (ventricular premature beat) / VPE (ventricular preexcitation) |
| APB | ABI (atrial bigeminy) / APB (atrial premature beats) |
| SArr | SA (Sinus Irregularity) / SAAWR (Sinus Atrium to Atrial Wandering Rhythm) |
| AF | AFIB (Atrial Fibrillation) |
| AFL | AF (Atrial Flutter) |
| JPB | JPT (junctional premature beat) |
| ATach | AT (Atrial Tachycardia) |
| JEB | JEB (junctional escape beat) |
| SVT | SVT (Supraventricular Tachycardia) / AVNRT (Atrioventricular Node Reentrant Tachycardia) / AVRT (Atrioventricular Reentrant Tachycardia) |
| VEB | VEB (ventricular escape beat) / VET (ventricular escape trigeminy) |
| RBBB | RBBB (right bundle branch block) |
| 1° AVB | 1AVB (1 degree atrioventricular block) / AVB (atrioventricular block) |
| QT Prolong | QTIE (QT interval extension) |
| ER | ERV (Early repolarization of the ventricles) |
| LAFB | LFBBB (left front bundle branch block) |
| IVB | IVB (Intraventricular block) |
| VPE | WPW (Wolf-Parkinson-White syndrome) |
| 3° AVB | 3AVB (3 degree atrioventricular block) / AVB (atrioventricular block) |
| 2° 1 Type AVB | 2AVB (2 degree atrioventricular block) / 2AVB1 (2 degree atrioventricular block(Type one)) / AVB (atrioventricular block) |
| 2° 2 Type AVB | 2AVB (2 degree atrioventricular block) / 2AVB2 (2 degree atrioventricular block(Type two)) / AVB (atrioventricular block) |
| LVH | LVH (left ventricle hypertrophy) |

| **Target Label (Zhongshan Ontology)** | **Mapping Criteria (Original Diagnostic Terms)** |
|---|---|
| RVH | RVH (right ventricle hypertrophy) |
| RAH | RAH (right atrial hypertrophy) |
| LK | UW (U wave) |

The mapping schema aggregates diverse original diagnostic terms and synonym variations into the target standardized categories.

**Supplementary Table 9: Diagnostic label mapping criteria for the external PTBXL test cohort.**

| Target Label (Zhongshan Ontology) | Mapping Criteria (Original Diagnostic Terms) |
|---|---|
| STEMI | IMI (inferior myocardial infarction) / ASMI (anteroseptal myocardial infarction) / ILMI (inferolateral myocardial infarction) / AMI (anterior myocardial infarction) / ALMI (anterolateral myocardial infarction) / LMI (lateral myocardial infarction) / IPLMI (inferoposterolateral myocardial infarction) / IPMI (inferoposterior myocardial infarction) / PMI (posterior myocardial infarction) |
| NSR | NORM (normal ECG) |
| SBrad | SBRAD (sinus bradycardia) |
| STach | STACH (sinus tachycardia) |
| VPB | PVC (ventricular premature complex) |
| APB | PAC (atrial premature complex) |
| SArr | SARRH (sinus arrhythmia) |
| AF | AFIB (atrial fibrillation) |
| AFL | AFLT (atrial flutter) |
| AJR | SVTAC (supraventricular tachycardia) |
| ATach | SVTAC (supraventricular tachycardia) |
| JTach | SVTAC (supraventricular tachycardia) |
| SVT | SVARR (supraventricular arrhythmia) / PSVT (paroxysmal supraventricular tachycardia) |
| RBBB | IRBBB (incomplete right bundle branch block) / CRBBB (complete right bundle branch block) |
| 1° AVB | 1AVB (first degree AV block) |
| QT Prolong | LNGQT (long QT-interval) |
| LAFB | LAFB (left anterior fascicular block) |
| LBBB | ILBBB (incomplete left bundle branch block) / CLBBB (complete left bundle branch block) |
| IVB | IVCD (non-specific intraventricular conduction disturbance (block)) |
| VPE | WPW (Wolf-Parkinson-White syndrome) |
| 3° AVB | 3AVB (third degree AV block) |
| 2° 1 Type AVB | 2AVB (second degree AV block) |
| 2° 2 Type AVB | 2AVB (second degree AV block) |
| LVH | LVH (left ventricular hypertrophy) / VCLVH (voltage criteria (QRS) for left ventricular hypertrophy) |
| LAH | LAO/LAE (left atrial overload/enlargement) |

| Target Label (Zhongshan Ontology) | Mapping Criteria (Original Diagnostic Terms) |
|---|---|
| RVH | RVH (right ventricular hypertrophy) |
| VA | ANEUR (ST-T changes compatible with ventricular aneurysm) |
| RAH | RAO/RAE (right atrial overload/enlargement) |
| VVI | PACE (normal functioning artificial pacemaker) |
| VAT | PACE (normal functioning artificial pacemaker) |
| DDD | PACE (normal functioning artificial pacemaker) |
| AAI | PACE (normal functioning artificial pacemaker) |
| VP | PACE (normal functioning artificial pacemaker) |

The mapping schema aggregates diverse original diagnostic terms and synonym variations into the target standardized categories.

**Supplementary Table 10: Diagnostic label mapping criteria for the external Georgia test cohort.**

| Target Label (Zhongshan Ontology) | Mapping Criteria (Original Diagnostic Terms) |
|---|---|
| STEMI | acute myocardial infarction / anterior myocardial infarction / myocardial infarction |
| SBrad | Bradycardia (Brady) / sinus bradycardia (SB) |
| STach | sinus tachycardia (STach) |
| VPB | paired ventricular premature complexes (VPVC) / premature ventricular contractions (PVC) / ventricular premature beats (VPB) / ventricular bigeminy (VBIG) / ventricular ectopics (VEB) / ventricular pre-excitation (VPEx) / ventricular trigeminy (VTrig) |
| APB | atrial bigeminy (AB) / premature atrial contraction (PAC) |
| SArr | sinus arrhythmia (SA) |
| AF | atrial fibrillation (AF) / chronic atrial fibrillation (CAF) / paroxysmal atrial fibrillation (PAF) / rapid atrial fibrillation (RAF) |
| AFL | atrial flutter (AFL) |
| JPB | junctional premature complex (JPC) / supraventricular bigeminy (SVB) / supraventricular premature beats (SVPB) |
| AJR | accelerated junctional rhythm (AJR) |
| ATach | atrial tachycardia (ATach) |
| JTach | junctional tachycardia (JTach) |
| JEB | junctional escape (JE) |
| SVT | paroxysmal supraventricular tachycardia (PSVT) / supraventricular tachycardia (SVT) |
| VEB | ventricular escape beat (VEsB) / ventricular escape rhythm (VEsR) |
| RBBB | bundle branch block (BBB) / complete right bundle branch block (CRBBB) / incomplete right bundle branch block (IRBBB) / right bundle branch block (RBBB) |
| 1° AVB | 1st degree av block (IAVB) / av block (AVB) |
| QT Prolong | prolonged qt interval (LQT) |
| ER | early repolarization (ERe) |
| LAFB | left anterior fascicular block (LAnFB) |
| LBBB | bundle branch block (BBB) / incomplete left bundle branch block (ILBBB) / left bundle branch block (LBBB) |
| IVB | diffuse intraventricular block (DIB) / nonspecific intraventricular conduction disorder (NSIVCB) |
| Short PR | shortened pr interval (SPRI) |
| VPE | wolff parkinson white pattern (WPW) |
| 3° AVB | av block (AVB) / complete heart block (CHB) |

| Target Label (Zhongshan Ontology) | Mapping Criteria (Original Diagnostic Terms) |
|---|---|
| 2° 1 Type AVB | 2nd degree av block (IIAVB) / av block (AVB) / mobitz type i wenckebach atrioventricular block (MoI) |
| 2° 2 Type AVB | 2nd degree av block (IIAVB) / av block (AVB) |
| LVH | left ventricular hypertrophy (LVH) / ventricular hypertrophy (VH) |
| LAH | atrial hypertrophy (AH) / left atrial enlargement (LAE) / left atrial hypertrophy (LAH) |
| RVH | right ventricular hypertrophy (RVH) / ventricular hypertrophy (VH) |
| RAH | atrial hypertrophy (AH) / right atrial hypertrophy (RAH) / |
| AAI | atrial pacing pattern (AP) |
| VP | ventricular pacing pattern (VPP) |

The mapping schema aggregates diverse original diagnostic terms and synonym variations into the target standardized categories.

**Supplementary Table 11: Diagnostic label mapping criteria for the external CPSC2018 test cohort.**

| Target Label (Zhongshan Ontology) | Mapping Criteria (Original Diagnostic Terms) |
|---|---|
| STEMI | ST-segment elevated (STE) |
| NSR | Normal |
| VPB | Premature ventricular contraction (PVC) |
| APB | Premature atrial contraction (PAC) |
| AF | Atrial fibrillation (AF) |
| RBBB | Right bundle branch block (RBBB) |
| 1° AVB | First-degree atrioventricular block (I-AVB) |
| LBBB | Left bundle branch block (LBBB) |

The mapping schema aggregates diverse original diagnostic terms and synonym variations into the target standardized categories.

**Supplementary Table 12: Availability of diagnostic tasks across cohorts.**

| Cohort | Tier 1: ECG Interpretation | Tier 2: ECHO Diagnosis | Tier 3: Rare Disease Diagnosis |
|---|---|---|---|
| Zhongshan | √ | √ | √ |
| Shiyuan | - | √ | - |
| BIDMC | - | √ | - |
| Xiamen | √ | - | - |
| MIMIC-IV-ECG | √ | - | - |
| UKB | √ | - | - |
| Chapman | √ | - | - |
| PTB-XL | √ | - | - |
| Georgia | √ | - | - |
| CPSC2018 | √ | - | - |

√: Indicates that validated ground truth labels for this task tier were available and mapped to the standardized ontology. -: Indicates that the specific ground truth labels were unavailable and the cohort was not utilized for this specific tier.

**Supplementary Table 13: Prevalence of ECG diagnostic labels in External Test Cohorts.**

| Disease | Xiamen<br>N = 44,362<br>n (%) | MIMIC-IV-ECG<br>N = 762,853<br>n (%) | UKB<br>N = 50,407<br>n (%) | Chapman<br>N = 45,053<br>n (%) | PTBXL<br>N = 21,799<br>n (%) | Georgia<br>N = 10,338<br>n (%) | CPSC2018<br>N = 6,877<br>n (%) |
|---|---|---|---|---|---|---|---|
| OMI | 47 (0.11) | 14,581 (1.91) | — | — | — | — | — |
| STEMI | 432 (0.97) | 166,317 (21.80) | 220 (0.44) | 123 (0.27) | 5,288 (24.26) | 7 (0.07) | 220 (3.20) |
| NSR | 34,336 (77.40) | 117,266 (15.37) | 47,890 (95.01) | — | 9,514 (43.64) | — | 918 (13.35) |
| SBrad | 5,396 (12.16) | 97,656 (12.80) | 22,546 (44.73) | 16,525 (36.68) | 637 (2.92) | 1,682 (16.27) | — |
| STach | 1,394 (3.14) | 67,450 (8.84) | — | 7,236 (16.06) | 826 (3.79) | 1,259 (12.18) | — |
| VPB | 1,277 (2.88) | 50,962 (6.68) | 1,409 (2.80) | 309 (0.69) | 1,143 (5.24) | 399 (3.86) | 700 (10.18) |
| APB | 1,003 (2.26) | 34,983 (4.59) | 1,049 (2.08) | 3 (0.01) | 398 (1.83) | 638 (6.17) | 616 (8.96) |
| SArr | 1,522 (3.43) | 25,668 (3.36) | 2,377 (4.72) | 2,535 (5.63) | 772 (3.54) | 454 (4.39) | — |
| AF | 427 (0.96) | 78,363 (10.27) | 772 (1.53) | 1,780 (3.95) | 1,514 (6.95) | 570 (5.51) | 1,221 (17.75) |
| AFL | 114 (0.26) | — | 65 (0.13) | 8,046 (17.86) | 73 (0.33) | 186 (1.80) | — |
| JPB | 250 (0.56) | 2,195 (0.29) | 453 (0.90) | 11 (0.02) | — | 1 (0.01) | — |
| AJR | — | 3,030 (0.40) | 184 (0.37) | — | 27 (0.12) | 19 (0.18) | — |
| ATach | 145 (0.33) | 1,884 (0.25) | — | 296 (0.66) | 27 (0.12) | 28 (0.27) | — |
| JTach | 114 (0.26) | 168 (0.02) | 184 (0.37) | — | 27 (0.12) | 4 (0.04) | — |
| JEB | 16 (0.04) | — | 184 (0.37) | 75 (0.17) | — | 5 (0.05) | — |
| SVT | 16 (0.04) | 939 (0.12) | — | 749 (1.66) | 179 (0.82) | 32 (0.31) | — |
| VEB | 15 (0.03) | 583 (0.08) | — | 63 (0.14) | — | 1 (0.01) | — |
| VTach | 11 (0.02) | 675 (0.09) | — | — | — | — | — |
| RBBB | 1,274 (2.87) | 63,641 (8.34) | 2,519 (5.00) | 649 (1.44) | 1,658 (7.61) | 992 (9.60) | 1,857 (27.00) |
| 1° AVB | 985 (2.22) | 57,497 (7.54) | 2,954 (5.86) | 1,380 (3.06) | 793 (3.64) | 843 (8.15) | 722 (10.50) |
| QT Prolong | 780 (1.76) | 38,296 (5.02) | 510 (1.01) | 394 (0.87) | 117 (0.54) | 1,390 (13.45) | — |
| ER | 366 (0.83) | 3,912 (0.51) | 435 (0.86) | 365 (0.81) | — | 140 (1.35) | — |
| LAFB | 98 (0.22) | 47,563 (6.23) | 756 (1.50) | 240 (0.53) | 1,623 (7.45) | 180 (1.74) | — |
| LBBB | 139 (0.31) | 28,535 (3.74) | 501 (0.99) | — | 613 (2.81) | 327 (3.16) | 236 (3.43) |

| Disease | Xiamen N = 44,362 n (%) | MIMIC-IV-ECG N = 762,853 n (%) | UKB N = 50,407 n (%) | Chapman N = 45,053 n (%) | PTBXL N = 21,799 n (%) | Georgia N = 10,338 n (%) | CPSC2018 N = 6,877 n (%) |
|---|---|---|---|---|---|---|---|
| IVB | 57 (0.13) | 31,683 (4.15) | 308 (0.61) | 770 (1.71) | 787 (3.61) | 203 (1.96) | — |
| Short PR | 80 (0.18) | 11,012 (1.44) | 437 (0.87) | — | — | 2 (0.02) | — |
| VPE | 82 (0.18) | 499 (0.07) | — | 72 (0.16) | 79 (0.36) | 2 (0.02) | — |
| 3° AVB | 2 (0.00) | 204 (0.03) | — | 320 (0.71) | 16 (0.07) | 82 (0.79) | — |
| 2° 1 Type AVB | 11 (0.02) | 298 (0.04) | — | 339 (0.75) | 14 (0.06) | 96 (0.93) | — |
| 2° 2 Type AVB | 5 (0.01) | 1,534 (0.20) | — | 308 (0.68) | 14 (0.06) | 96 (0.93) | — |
| LVH | 1,781 (4.01) | 71,267 (9.34) | 1,033 (2.05) | 645 (1.43) | 2,354 (10.80) | 1,264 (12.23) | — |
| LAH | 1,225 (2.76) | 16,785 (2.20) | 3,278 (6.50) | — | 426 (1.95) | 930 (9.00) | — |
| RVH | 192 (0.43) | 7,910 (1.04) | 88 (0.17) | 110 (0.24) | 126 (0.58) | 71 (0.69) | — |
| VA | 19 (0.04) | — | — | — | 104 (0.48) | — | — |
| RAH | 49 (0.11) | 330 (0.04) | 112 (0.22) | 36 (0.08) | 99 (0.45) | 141 (1.36) | — |
| DC | 7 (0.02) | 218 (0.03) | — | — | — | — | — |
| BVH | — | 17,772 (2.33) | — | — | — | — | — |
| VVI | 44 (0.10) | — | — | — | 294 (1.35) | — | — |
| VAT | 32 (0.07) | 239 (0.03) | — | — | 294 (1.35) | — | — |
| DDD | 3 (0.01) | 288 (0.04) | — | — | 294 (1.35) | — | — |
| AAI | 25 (0.06) | — | — | — | 294 (1.35) | 52 (0.50) | — |
| VP | — | — | — | — | 294 (1.35) | 45 (0.44) | — |
| PSM | 3 (0.01) | — | — | — | — | — | — |
| HK | 5 (0.01) | 1,650 (0.22) | — | — | — | — | — |
| LK | — | — | — | 136 (0.30) | — | — | — |

Values are positive ECG counts (percentages). Each column heading gives the number of evaluable ECG records for every available target in that cohort. An em dash denotes an unavailable target.

**Supplementary Table 14: Prevalence of ECHO diagnostic labels in External Test Cohorts.**

| Disease | Shiyuan | BIDMC |
|---|---|---|
| NER | 440/219,248 (0.20%) | — |
| LAE | 25,834/219,248 (11.78%) | 69,199/131,987 (52.43%) |
| HCM | 17,740/219,248 (8.09%) | — |
| nHCM | 17,258/219,248 (7.87%) | — |
| BAE | 3,631/219,248 (1.66%) | 39,303/111,424 (35.27%) |
| HFrEF | 4,005/219,248 (1.83%) | 2,385/23,144 (10.31%) |
| RMWSSFLV | 5/219,248 (0.00%) | — |
| DCM | 3/219,248 (0.00%) | — |
| oHCM | 565/219,248 (0.26%) | 3,208/114,022 (2.81%) |
| ICM | — | — |
| LVDD | 148,641/219,248 (67.80%) | — |
| VA | 756/219,248 (0.34%) | 682/762 (89.50%) |
| RAE | 218/219,248 (0.10%) | 50,221/112,952 (44.46%) |
| AM | 30/219,248 (0.01%) | — |
| LAM | 55/219,248 (0.03%) | — |
| AVC | 23,654/219,248 (10.79%) | — |
| TR | 7,382/217,016 (3.40%) | 15,537/135,178 (11.49%) |
| MR | 5,244/217,016 (2.42%) | 15,062/134,961 (11.16%) |
| AR | 3,539/217,016 (1.63%) | 3,000/54,345 (5.52%) |
| MVPLAC | 3,322/219,248 (1.52%) | — |
| MS | 287/217,016 (0.13%) | 500/23,318 (2.14%) |
| AS | 547/217,016 (0.25%) | 7,266/16,860 (43.10%) |
| BAV | 249/219,248 (0.11%) | — |
| PS | 31/217,016 (0.01%) | 70/38,016 (0.18%) |
| EA | 10/219,248 (0.00%) | — |
| PR | 163/217,016 (0.08%) | 2,040/102,298 (1.99%) |

| Disease | Shiyuan | BIDMC |
|---|---|---|
| TS | — | — |
| CHD | 746/219,248 (0.34%) | — |
| ASD | 657/219,248 (0.30%) | — |
| VSD | 139/219,248 (0.06%) | 254/345 (73.62%) |
| PFO | 697/219,248 (0.32%) | — |
| PDA | 65/219,248 (0.03%) | — |
| TOF | 27/219,248 (0.01%) | — |
| PH | 26,621/219,248 (12.14%) | 15,988/109,440 (14.61%) |
| AD | 27,990/219,248 (12.77%) | 19,699/118,242 (16.66%) |
| Dex | 78/219,248 (0.04%) | — |
| PE | 11,795/219,248 (5.38%) | 1,677/28,686 (5.85%) |
| CP | 4/219,248 (0.00%) | — |
| AF | — | — |

Values are positive ECGs/evaluable ECGs (%) in the final external evaluation. An em dash denotes an unavailable target. Denominators are specific to the analysed cohort and endpoint. HFrEF denotes LVEF <40%.

**Supplementary Table 15: Global performance overview (AUPRC) of the four final models across diagnostic tiers and cohorts.**

| Cohort | Scratch CNN–Transformer | ECG-FM | DeepECG-SSL | CardioLoom |
|---|---|---|---|---|
| **Tier 1: ECG Interpretation** | | | | |
| Zhongshan | 0.549±0.296 | 0.608±0.300 | 0.615±0.300 | **0.624±0.295** |
| Xiamen | 0.558±0.315 | 0.601±0.292 | **0.621±0.285** | 0.616±0.292 |
| MIMIC-IV-ECG | 0.282±0.316 | 0.286±0.314 | 0.290±0.317 | **0.293±0.317** |
| UKB | 0.399±0.352 | 0.417±0.352 | 0.425±0.351 | **0.431±0.351** |
| Chapman | 0.310±0.306 | 0.344±0.313 | 0.348±0.308 | **0.356±0.312** |
| PTB-XL | 0.401±0.309 | 0.400±0.304 | 0.403±0.305 | **0.412±0.305** |
| Georgia | 0.311±0.309 | 0.328±0.310 | **0.343±0.313** | 0.336±0.319 |
| CPSC2018 | 0.665±0.295 | 0.685±0.295 | **0.696±0.288** | 0.689±0.292 |
| Average | 0.434 | 0.459 | 0.468 | **0.470** |
| **Tier 2: ECHO Diagnosis** | | | | |
| Zhongshan | 0.269±0.232 | 0.313±0.259 | 0.314±0.260 | **0.318±0.258** |
| Shiyuan | 0.151±0.189 | 0.182±0.207 | 0.179±0.209 | **0.185±0.209** |
| BIDMC | 0.362±0.281 | 0.396±0.279 | 0.399±0.286 | **0.402±0.284** |
| Average | 0.261 | 0.297 | 0.297 | **0.302** |
| **Tier 3: Rare Disease Diagnosis** | | | | |
| Zhongshan | 0.008 | 0.021 | **0.022** | 0.020 |

Values are mean±SD across task-level AUPRCs within each cohort. Averages give equal weight to the displayed cohorts. Bold type identifies the highest displayed mean.

**Supplementary Table 16: Training hyperparameters for CardioLoom pretraining.**

| Hyperparameter | Value |
|---|---|
| ECG encoder | DeepECG-SSL convolutional Transformer initialization with 768-dimensional features |
| Dataset input | 8 leads (I, II, V1–V6) at 500 Hz for 10 s, followed by deterministic 12-lead reconstruction and resampling to 250 Hz |
| Report sampling | Sampling with replacement weighted by normalized-template frequency $f^{-0.25}$ and deterministic DDP sharding |
| Cross-modal loss | Symmetric multi-positive InfoNCE with temperature 0.07 |
| Uni-modal loss | ECG dropout-view contrastive loss with dropout 0.1 and weight 1.0 |
| Epochs | 20 |
| Selected checkpoint | Minimum validation-loss checkpoint at epoch 16 |
| Hardware | 2 × NVIDIA A100-80GB |
| Precision | Bfloat16 |
| Batch size | 700 per GPU (global 1,400) |
| Learning rates | Projection/text learning rate $5\times10^{-5}$ and ECG backbone learning rate $5\times10^{-6}$. The backbone was frozen for the first 500 updates. |
| Optimizer | AdamW with $\beta$=(0.9,0.98), $\varepsilon=1\times10^{-6}$, and weight decay 0.01 |
| Schedule | 500-update linear warm-up followed by global-batch-adjusted cosine warm restarts |
| Gradient clipping | Global norm 5 |
| Random seed | 42 |

Pretraining used two NVIDIA A100-80GB GPUs and bfloat16 arithmetic for up to 20 epochs. The reported checkpoint was selected by minimum validation loss under the report-aware multi-positive objective.

**Supplementary Table 17: Final downstream adaptation protocols and hyperparameters.**

| Component | Final protocol |
|---|---|
| Common ECG / ECHO / valve | Transformer and prediction-head fine-tuning with the pretrained convolutional extractor frozen |
| AF / LVEF | Single-task adaptation |
| Low-positive / rare, pretrained models | Frozen 768-dimensional features with a linear SVM for the 5 rare diseases and echocardiographic TOF, EA, Dex, PR, PS and TS. Outputs are raw decision margins |
| Low-positive / rare, Scratch | Task-specific end-to-end training from random initialization |
| Batch size | 128 |
| Pretrained-model optimizer | AdamW with fixed LR $5\times10^{-5}$, weight decay $5\times10^{-3}$, $\beta$=(0.9,0.98), and $\varepsilon=1\times10^{-8}$ |
| Scratch optimizer | AdamW with fixed LR $5\times10^{-5}$ for every task, weight decay $5\times10^{-3}$, $\beta$=(0.9,0.98), and $\varepsilon=1\times10^{-8}$. No scheduler or LR search was used. |
| Updates | ECG 100%: 12,000 updates. Other end-to-end tasks: 4,000 updates. Scratch completed its allocated budget. |
| Validation / selection | Every 200 updates (1000 for full-budget ECG). Checkpoints were selected by validation AUPRC. |
| Gradient clipping | Global norm 5 |
| Classification threshold | Observed validation score maximizing F1, with the smallest threshold resolving ties. The selected threshold was fixed externally. |
| Primary metric | Full-split trapezoidal AUPRC |

Batch size 128 and AdamW LR $5 \times 10^{-5}$ apply to end-to-end adaptation. Pretrained models use early-stopping patience five. Scratch completes its allocated budget. The frozen-feature SVM protocol instead uses batch size 256, up to 1,000 hinge-loss SGD updates and validation every 200 updates. SVM class balancing is applied through sampling only, and raw decision margins are retained. Exact thresholds are selected from Zhongshan validation scores and fixed for test and external evaluation.

**Supplementary Table 18: Detailed task-specific performance comparison on the internal Zhongshan test cohort.**

| Disease | Model | AUPRC | AUROC | Sensitivity | Specificity | F1 Score | MCC |
|---|---|---|---|---|---|---|---|
| **Tier 1: ECG Interpretation** | | | | | | | |
| OMI | Scratch CNN–Transformer | 0.517 (0.466–0.569) | 0.980 (0.976–0.983) | 0.553 (0.502–0.603) | **0.993 (0.992–0.994)** | 0.532 (0.489–0.575) | 0.526 (0.484–0.570) |
| | ECG-FM | 0.541 (0.489–0.591) | **0.983 (0.979–0.986)** | 0.578 (0.532–0.625) | **0.993 (0.993–0.994)** | 0.558 (0.517–0.597) | 0.552 (0.511–0.592) |
| | DeepECG-SSL | 0.540 (0.490–0.587) | **0.983 (0.979–0.986)** | 0.632 (0.593–0.672) | 0.991 (0.990–0.992) | 0.554 (0.517–0.591) | 0.552 (0.516–0.588) |
| | CardioLoom | **0.551 (0.500–0.599)** | 0.982 (0.978–0.986) | **0.637 (0.593–0.679)** | 0.992 (0.991–0.993) | **0.566 (0.528–0.602)** | **0.563 (0.525–0.599)** |
| STEMI | Scratch CNN–Transformer | 0.440 (0.379–0.504) | **0.989 (0.984–0.993)** | 0.596 (0.535–0.655) | 0.997 (0.996–0.997) | 0.482 (0.428–0.532) | 0.489 (0.437–0.538) |
| | ECG-FM | 0.509 (0.441–0.575) | 0.986 (0.978–0.992) | 0.651 (0.589–0.705) | 0.997 (0.997–0.998) | **0.535 (0.480–0.582)** | **0.542 (0.487–0.587)** |
| | DeepECG-SSL | **0.540 (0.469–0.605)** | **0.989 (0.983–0.993)** | **0.693 (0.623–0.751)** | 0.996 (0.996–0.997) | 0.518 (0.458–0.569) | 0.533 (0.476–0.582) |
| | CardioLoom | 0.503 (0.428–0.570) | 0.988 (0.980–0.993) | 0.501 (0.437–0.567) | **0.998 (0.998–0.999)** | 0.514 (0.455–0.567) | 0.512 (0.453–0.566) |
| NSR | Scratch CNN–Transformer | 0.996 (0.996–0.997) | 0.993 (0.992–0.993) | **0.982 (0.981–0.983)** | 0.934 (0.930–0.937) | 0.977 (0.976–0.978) | 0.922 (0.919–0.925) |
| | ECG-FM | **0.997 (0.996–0.997)** | 0.993 (0.992–0.993) | **0.982 (0.981–0.983)** | 0.935 (0.932–0.938) | 0.977 (0.976–0.978) | 0.923 (0.920–0.926) |
| | DeepECG-SSL | **0.997 (0.996–0.997)** | 0.993 (0.993–0.994) | **0.982 (0.981–0.983)** | 0.938 (0.935–0.941) | **0.978 (0.977–0.979)** | 0.925 (0.922–0.928) |
| | CardioLoom | **0.997 (0.997–0.997)** | **0.994 (0.993–0.994)** | 0.979 (0.978–0.980) | **0.946 (0.943–0.949)** | **0.978 (0.977–0.979)** | **0.927 (0.924–0.929)** |
| SBrad | Scratch CNN–Transformer | 0.969 (0.966–0.972) | **0.997 (0.996–0.997)** | 0.927 (0.921–0.932) | **0.994 (0.993–0.994)** | 0.930 (0.925–0.934) | 0.923 (0.918–0.927) |
| | ECG-FM | 0.970 (0.967–0.973) | **0.997 (0.996–0.997)** | 0.933 (0.928–0.939) | 0.992 (0.991–0.993) | 0.926 (0.922–0.930) | 0.919 (0.914–0.923) |
| | DeepECG-SSL | 0.972 (0.969–0.975) | **0.997 (0.996–0.997)** | 0.942 (0.937–0.947) | 0.992 (0.992–0.993) | 0.932 (0.928–0.936) | 0.926 (0.921–0.930) |
| | CardioLoom | **0.974 (0.971–0.977)** | **0.997 (0.997–0.997)** | **0.950 (0.945–0.954)** | 0.992 (0.991–0.992) | **0.933 (0.930–0.937)** | **0.927 (0.923–0.931)** |

| Disease | Model | AUPRC | AUROC | Sensitivity | Specificity | F1 Score | MCC |
|---|---|---|---|---|---|---|---|
| STach | Scratch CNN–Transformer | **0.971 (0.967–0.975)** | 0.997 (0.997–0.998) | 0.922 (0.913–0.930) | **0.997 (0.997–0.998)** | 0.928 (0.922–0.934) | 0.925 (0.918–0.931) |
| | ECG-FM | 0.970 (0.966–0.974) | 0.997 (0.997–0.998) | **0.942 (0.935–0.949)** | 0.996 (0.996–0.997) | 0.930 (0.924–0.936) | 0.927 (0.921–0.933) |
| | DeepECG-SSL | **0.971 (0.967–0.975)** | **0.998 (0.997–0.998)** | 0.924 (0.916–0.932) | **0.997 (0.997–0.998)** | 0.932 (0.926–0.938) | 0.930 (0.923–0.936) |
| | CardioLoom | **0.971 (0.966–0.975)** | **0.998 (0.997–0.998)** | 0.937 (0.929–0.944) | **0.997 (0.997–0.997)** | **0.935 (0.929–0.941)** | **0.932 (0.926–0.938)** |
| VPB | Scratch CNN–Transformer | 0.958 (0.953–0.962) | 0.991 (0.989–0.992) | 0.909 (0.902–0.916) | 0.996 (0.995–0.996) | 0.920 (0.915–0.925) | 0.915 (0.910–0.921) |
| | ECG-FM | 0.969 (0.965–0.972) | **0.993 (0.991–0.994)** | 0.927 (0.921–0.933) | **0.998 (0.997–0.998)** | 0.944 (0.940–0.949) | 0.941 (0.936–0.945) |
| | DeepECG-SSL | **0.971 (0.967–0.974)** | 0.992 (0.991–0.993) | 0.933 (0.926–0.939) | 0.997 (0.997–0.998) | **0.946 (0.941–0.950)** | 0.942 (0.938–0.947) |
| | CardioLoom | 0.970 (0.967–0.974) | **0.993 (0.991–0.994)** | **0.937 (0.931–0.943)** | 0.997 (0.997–0.997) | **0.946 (0.942–0.951)** | **0.943 (0.938–0.947)** |
| APB | Scratch CNN–Transformer | 0.893 (0.885–0.902) | 0.987 (0.986–0.989) | 0.840 (0.829–0.851) | 0.991 (0.991–0.992) | 0.837 (0.828–0.846) | 0.828 (0.819–0.837) |
| | ECG-FM | 0.925 (0.917–0.932) | 0.991 (0.989–0.992) | 0.882 (0.871–0.891) | 0.993 (0.993–0.994) | 0.878 (0.870–0.885) | 0.872 (0.864–0.879) |
| | DeepECG-SSL | 0.930 (0.923–0.938) | 0.991 (0.990–0.993) | **0.887 (0.878–0.897)** | 0.994 (0.994–0.995) | 0.887 (0.879–0.894) | 0.881 (0.873–0.888) |
| | CardioLoom | **0.936 (0.930–0.943)** | **0.992 (0.990–0.993)** | 0.883 (0.874–0.892) | **0.995 (0.994–0.995)** | **0.890 (0.883–0.896)** | **0.884 (0.877–0.891)** |
| SArr | Scratch CNN–Transformer | 0.572 (0.547–0.599) | 0.971 (0.968–0.974) | 0.576 (0.552–0.599) | **0.991 (0.990–0.991)** | 0.565 (0.544–0.585) | 0.556 (0.535–0.576) |
| | ECG-FM | 0.558 (0.532–0.586) | 0.969 (0.965–0.972) | 0.583 (0.560–0.608) | 0.989 (0.988–0.989) | 0.540 (0.521–0.561) | 0.532 (0.513–0.553) |
| | DeepECG-SSL | 0.608 (0.584–0.633) | 0.976 (0.973–0.978) | **0.642 (0.620–0.665)** | 0.990 (0.989–0.991) | **0.597 (0.578–0.615)** | **0.590 (0.571–0.608)** |
| | CardioLoom | **0.623 (0.598–0.647)** | **0.979 (0.977–0.982)** | 0.633 (0.611–0.655) | 0.990 (0.990–0.991) | **0.597 (0.578–0.615)** | 0.589 (0.570–0.608) |
| AF | Scratch CNN–Transformer | 0.927 (0.921–0.933) | 0.993 (0.993–0.994) | 0.876 (0.868–0.884) | 0.988 (0.987–0.989) | 0.863 (0.856–0.869) | 0.852 (0.845–0.859) |
| | ECG-FM | 0.940 (0.935–0.945) | 0.994 (0.994–0.995) | 0.887 (0.879–0.894) | **0.989 (0.989–0.990)** | 0.877 (0.871–0.883) | 0.868 (0.861–0.874) |
| | DeepECG-SSL | 0.943 (0.939–0.948) | **0.995 (0.994–0.995)** | 0.898 (0.890–0.905) | **0.989 (0.988–0.990)** | 0.881 (0.874–0.887) | 0.871 (0.864–0.878) |
| | CardioLoom | **0.947 (0.942–0.951)** | **0.995 (0.994–0.995)** | **0.915 (0.908–0.922)** | 0.988 (0.987–0.989) | **0.884 (0.878–0.890)** | **0.875 (0.869–0.881)** |

| Disease | Model | AUPRC | AUROC | Sensitivity | Specificity | F1 Score | MCC |
|---|---|---|---|---|---|---|---|
| AFL | Scratch CNN–Transformer | 0.828 (0.805–0.850) | 0.993 (0.991–0.994) | 0.737 (0.708–0.767) | 0.996 (0.996–0.997) | 0.759 (0.736–0.780) | 0.755 (0.732–0.776) |
| | ECG-FM | 0.876 (0.857–0.893) | **0.996 (0.995–0.997)** | 0.811 (0.786–0.832) | **0.997 (0.996–0.997)** | 0.810 (0.790–0.827) | 0.806 (0.787–0.824) |
| | DeepECG-SSL | **0.886 (0.869–0.901)** | **0.996 (0.994–0.997)** | 0.775 (0.751–0.800) | **0.997 (0.997–0.998)** | 0.804 (0.785–0.822) | 0.801 (0.783–0.820) |
| | CardioLoom | 0.883 (0.866–0.899) | **0.996 (0.995–0.997)** | **0.813 (0.789–0.837)** | **0.997 (0.996–0.997)** | **0.814 (0.794–0.831)** | **0.810 (0.791–0.828)** |
| JPB | Scratch CNN–Transformer | 0.613 (0.564–0.657) | 0.974 (0.964–0.982) | 0.539 (0.497–0.584) | **0.999 (0.999–0.999)** | 0.609 (0.571–0.647) | 0.613 (0.575–0.649) |
| | ECG-FM | 0.658 (0.615–0.702) | 0.977 (0.968–0.985) | **0.686 (0.648–0.724)** | 0.998 (0.998–0.998) | 0.667 (0.635–0.698) | 0.665 (0.633–0.696) |
| | DeepECG-SSL | 0.677 (0.632–0.721) | **0.979 (0.970–0.986)** | 0.652 (0.611–0.692) | **0.999 (0.998–0.999)** | **0.680 (0.646–0.712)** | **0.679 (0.645–0.711)** |
| | CardioLoom | **0.682 (0.639–0.724)** | 0.978 (0.969–0.986) | 0.631 (0.591–0.671) | **0.999 (0.999–0.999)** | 0.674 (0.639–0.705) | 0.674 (0.639–0.705) |
| AJR | Scratch CNN–Transformer | 0.209 (0.179–0.241) | 0.904 (0.893–0.916) | 0.328 (0.292–0.363) | 0.994 (0.994–0.995) | 0.309 (0.279–0.338) | 0.304 (0.274–0.334) |
| | ECG-FM | 0.303 (0.267–0.341) | 0.922 (0.911–0.932) | 0.406 (0.369–0.441) | 0.995 (0.994–0.995) | 0.382 (0.353–0.412) | 0.378 (0.349–0.409) |
| | DeepECG-SSL | 0.324 (0.286–0.361) | 0.921 (0.910–0.931) | 0.377 (0.339–0.415) | **0.996 (0.996–0.997)** | 0.393 (0.361–0.425) | 0.389 (0.358–0.422) |
| | CardioLoom | **0.339 (0.304–0.377)** | **0.930 (0.919–0.940)** | **0.420 (0.385–0.456)** | 0.995 (0.995–0.996) | **0.406 (0.377–0.437)** | **0.402 (0.373–0.433)** |
| ATach | Scratch CNN–Transformer | 0.518 (0.474–0.565) | 0.979 (0.974–0.984) | 0.550 (0.504–0.596) | 0.996 (0.996–0.997) | 0.516 (0.482–0.549) | 0.513 (0.479–0.547) |
| | ECG-FM | 0.637 (0.595–0.676) | 0.986 (0.982–0.990) | 0.653 (0.615–0.692) | 0.997 (0.997–0.997) | 0.605 (0.571–0.637) | 0.604 (0.570–0.635) |
| | DeepECG-SSL | **0.710 (0.671–0.745)** | **0.989 (0.985–0.993)** | **0.730 (0.691–0.767)** | 0.997 (0.997–0.998) | 0.663 (0.631–0.692) | 0.664 (0.632–0.692) |
| | CardioLoom | 0.693 (0.653–0.731) | **0.989 (0.985–0.992)** | 0.675 (0.633–0.716) | **0.998 (0.998–0.998)** | **0.675 (0.642–0.706)** | **0.673 (0.640–0.704)** |
| JTach | Scratch CNN–Transformer | 0.468 (0.373–0.563) | **0.994 (0.992–0.995)** | 0.463 (0.384–0.538) | 0.998 (0.998–0.999) | 0.465 (0.390–0.535) | 0.463 (0.389–0.534) |
| | ECG-FM | 0.601 (0.510–0.679) | **0.994 (0.992–0.996)** | 0.515 (0.436–0.591) | **0.999 (0.999–0.999)** | **0.568 (0.496–0.635)** | **0.570 (0.499–0.636)** |
| | DeepECG-SSL | 0.569 (0.479–0.647) | 0.993 (0.988–0.996) | 0.576 (0.503–0.643) | **0.999 (0.998–0.999)** | 0.563 (0.494–0.625) | 0.562 (0.492–0.624) |
| | CardioLoom | **0.605 (0.516–0.685)** | **0.994 (0.991–0.996)** | **0.622 (0.548–0.687)** | 0.998 (0.998–0.998) | 0.554 (0.489–0.613) | 0.556 (0.491–0.614) |

| Disease | Model | AUPRC | AUROC | Sensitivity | Specificity | F1 Score | MCC |
|---|---|---|---|---|---|---|---|
| JEB | Scratch CNN–Transformer | 0.572 (0.500–0.640) | 0.995 (0.994–0.996) | 0.538 (0.478–0.595) | **0.999 (0.998–0.999)** | 0.548 (0.497–0.599) | 0.547 (0.496–0.598) |
| | ECG-FM | 0.669 (0.611–0.722) | **0.997 (0.996–0.998)** | 0.632 (0.575–0.688) | **0.999 (0.998–0.999)** | 0.621 (0.572–0.669) | 0.620 (0.571–0.668) |
| | DeepECG-SSL | 0.706 (0.649–0.758) | **0.997 (0.996–0.998)** | 0.637 (0.582–0.694) | **0.999 (0.999–0.999)** | 0.653 (0.606–0.697) | 0.652 (0.606–0.696) |
| | CardioLoom | **0.712 (0.648–0.766)** | **0.997 (0.996–0.998)** | **0.714 (0.667–0.765)** | **0.999 (0.998–0.999)** | **0.666 (0.622–0.709)** | **0.666 (0.623–0.709)** |
| SVT | Scratch CNN–Transformer | 0.798 (0.728–0.858) | 0.997 (0.991–1.000) | 0.627 (0.533–0.725) | **1.000 (1.000–1.000)** | 0.690 (0.614–0.766) | 0.693 (0.619–0.767) |
| | ECG-FM | 0.836 (0.773–0.891) | **0.999 (0.999–1.000)** | 0.698 (0.609–0.786) | **1.000 (1.000–1.000)** | 0.733 (0.660–0.798) | 0.734 (0.663–0.799) |
| | DeepECG-SSL | **0.843 (0.780–0.897)** | 0.996 (0.989–1.000) | **0.841 (0.778–0.902)** | **1.000 (0.999–1.000)** | 0.779 (0.722–0.833) | 0.781 (0.726–0.835) |
| | CardioLoom | 0.840 (0.775–0.895) | **0.999 (0.997–1.000)** | 0.738 (0.651–0.819) | **1.000 (1.000–1.000)** | **0.782 (0.719–0.838)** | **0.783 (0.723–0.839)** |
| VEB | Scratch CNN–Transformer | 0.442 (0.360–0.529) | 0.994 (0.989–0.997) | 0.509 (0.433–0.586) | **0.999 (0.999–0.999)** | 0.498 (0.429–0.559) | 0.498 (0.429–0.560) |
| | ECG-FM | 0.517 (0.434–0.608) | 0.994 (0.988–0.998) | **0.752 (0.688–0.815)** | **0.999 (0.998–0.999)** | 0.567 (0.504–0.628) | 0.584 (0.524–0.643) |
| | DeepECG-SSL | **0.637 (0.540–0.728)** | 0.996 (0.993–0.998) | 0.696 (0.623–0.774) | **0.999 (0.999–0.999)** | **0.610 (0.541–0.670)** | 0.614 (0.545–0.673) |
| | CardioLoom | 0.581 (0.490–0.665) | **0.997 (0.993–0.999)** | 0.727 (0.654–0.799) | **0.999 (0.999–0.999)** | 0.609 (0.548–0.667) | **0.617 (0.558–0.673)** |
| VTach | Scratch CNN–Transformer | 0.543 (0.434–0.645) | 0.995 (0.987–0.998) | 0.463 (0.376–0.548) | **1.000 (0.999–1.000)** | 0.517 (0.429–0.591) | 0.520 (0.430–0.595) |
| | ECG-FM | 0.611 (0.504–0.717) | 0.996 (0.991–0.999) | **0.631 (0.546–0.709)** | **1.000 (0.999–1.000)** | 0.662 (0.580–0.727) | 0.662 (0.581–0.728) |
| | DeepECG-SSL | 0.667 (0.566–0.767) | 0.997 (0.994–0.999) | 0.604 (0.517–0.682) | **1.000 (1.000–1.000)** | 0.664 (0.584–0.730) | 0.667 (0.587–0.734) |
| | CardioLoom | **0.701 (0.607–0.797)** | **0.998 (0.996–0.999)** | 0.597 (0.503–0.688) | **1.000 (1.000–1.000)** | **0.667 (0.582–0.738)** | **0.671 (0.587–0.742)** |
| RBBB | Scratch CNN–Transformer | 0.955 (0.949–0.960) | 0.996 (0.995–0.996) | **0.891 (0.881–0.900)** | 0.991 (0.990–0.992) | 0.880 (0.871–0.888) | 0.872 (0.863–0.880) |
| | ECG-FM | **0.962 (0.957–0.966)** | **0.997 (0.996–0.997)** | 0.877 (0.865–0.888) | **0.993 (0.993–0.994)** | **0.887 (0.879–0.895)** | **0.880 (0.871–0.888)** |
| | DeepECG-SSL | 0.955 (0.949–0.960) | 0.995 (0.995–0.996) | 0.876 (0.863–0.886) | 0.992 (0.992–0.993) | 0.881 (0.871–0.889) | 0.873 (0.863–0.882) |
| | CardioLoom | 0.954 (0.948–0.959) | 0.995 (0.995–0.996) | 0.870 (0.859–0.881) | **0.993 (0.992–0.993)** | 0.879 (0.870–0.886) | 0.871 (0.862–0.879) |

| Disease | Model | AUPRC | AUROC | Sensitivity | Specificity | F1 Score | MCC |
|---|---|---|---|---|---|---|---|
| 1° AVB | Scratch CNN–Transformer | 0.848 (0.832–0.864) | 0.990 (0.989–0.991) | 0.802 (0.783–0.819) | 0.990 (0.989–0.991) | 0.771 (0.755–0.785) | 0.762 (0.746–0.777) |
| | ECG-FM | **0.884 (0.871–0.896)** | **0.993 (0.991–0.994)** | 0.810 (0.793–0.825) | **0.993 (0.992–0.994)** | **0.810 (0.796–0.822)** | **0.803 (0.789–0.816)** |
| | DeepECG-SSL | 0.876 (0.861–0.890) | 0.992 (0.991–0.993) | **0.853 (0.838–0.866)** | 0.990 (0.989–0.991) | 0.805 (0.792–0.818) | 0.798 (0.786–0.812) |
| | CardioLoom | 0.879 (0.864–0.893) | **0.993 (0.992–0.994)** | 0.825 (0.809–0.841) | 0.992 (0.991–0.993) | 0.809 (0.796–0.822) | 0.802 (0.789–0.815) |
| QT Prolong | Scratch CNN–Transformer | 0.242 (0.214–0.271) | 0.933 (0.928–0.939) | 0.299 (0.267–0.328) | 0.992 (0.992–0.993) | 0.302 (0.273–0.329) | 0.294 (0.266–0.321) |
| | ECG-FM | **0.305 (0.274–0.338)** | 0.949 (0.944–0.954) | **0.412 (0.379–0.445)** | 0.990 (0.989–0.991) | **0.355 (0.328–0.382)** | **0.350 (0.324–0.378)** |
| | DeepECG-SSL | 0.294 (0.263–0.327) | 0.948 (0.943–0.954) | 0.409 (0.376–0.440) | 0.990 (0.989–0.991) | **0.355 (0.329–0.378)** | **0.350 (0.324–0.373)** |
| | CardioLoom | 0.291 (0.260–0.322) | **0.951 (0.946–0.956)** | 0.324 (0.292–0.355) | **0.993 (0.993–0.994)** | 0.337 (0.307–0.364) | 0.330 (0.300–0.357) |
| ER | Scratch CNN–Transformer | 0.261 (0.216–0.314) | 0.970 (0.964–0.977) | 0.303 (0.258–0.347) | **0.998 (0.998–0.998)** | 0.341 (0.296–0.385) | 0.341 (0.295–0.385) |
| | ECG-FM | 0.305 (0.260–0.358) | 0.974 (0.968–0.980) | 0.410 (0.359–0.460) | 0.996 (0.996–0.997) | **0.362 (0.319–0.406)** | **0.361 (0.318–0.405)** |
| | DeepECG-SSL | 0.315 (0.269–0.368) | **0.978 (0.971–0.983)** | **0.423 (0.376–0.470)** | 0.996 (0.996–0.997) | 0.360 (0.322–0.401) | **0.361 (0.324–0.401)** |
| | CardioLoom | **0.326 (0.278–0.379)** | **0.978 (0.974–0.983)** | 0.305 (0.259–0.353) | **0.998 (0.998–0.999)** | 0.355 (0.306–0.402) | 0.357 (0.310–0.405) |
| LAFB | Scratch CNN–Transformer | 0.482 (0.433–0.532) | 0.990 (0.989–0.992) | 0.559 (0.510–0.606) | 0.995 (0.995–0.996) | 0.508 (0.472–0.543) | 0.507 (0.471–0.542) |
| | ECG-FM | 0.531 (0.477–0.581) | 0.989 (0.987–0.991) | **0.586 (0.535–0.631)** | **0.996 (0.995–0.996)** | 0.538 (0.500–0.573) | 0.536 (0.497–0.572) |
| | DeepECG-SSL | **0.545 (0.495–0.600)** | **0.992 (0.991–0.994)** | 0.575 (0.527–0.624) | **0.996 (0.995–0.996)** | **0.543 (0.506–0.580)** | **0.540 (0.503–0.577)** |
| | CardioLoom | 0.510 (0.455–0.567) | 0.989 (0.987–0.992) | 0.553 (0.504–0.599) | **0.996 (0.995–0.996)** | 0.523 (0.486–0.559) | 0.520 (0.483–0.556) |
| LBBB | Scratch CNN–Transformer | 0.931 (0.911–0.948) | **0.999 (0.998–0.999)** | 0.876 (0.847–0.903) | **0.999 (0.999–0.999)** | 0.867 (0.843–0.889) | 0.866 (0.842–0.889) |
| | ECG-FM | 0.941 (0.923–0.956) | **0.999 (0.999–1.000)** | 0.864 (0.836–0.891) | **0.999 (0.999–0.999)** | 0.873 (0.850–0.894) | 0.872 (0.849–0.894) |
| | DeepECG-SSL | 0.937 (0.919–0.951) | **0.999 (0.999–0.999)** | **0.894 (0.868–0.920)** | **0.999 (0.998–0.999)** | 0.864 (0.840–0.887) | 0.864 (0.839–0.886) |
| | CardioLoom | **0.944 (0.926–0.959)** | **0.999 (0.999–1.000)** | 0.871 (0.842–0.899) | **0.999 (0.999–0.999)** | **0.877 (0.853–0.899)** | **0.876 (0.852–0.898)** |

| Disease | Model | AUPRC | AUROC | Sensitivity | Specificity | F1 Score | MCC |
|---|---|---|---|---|---|---|---|
| IVB | Scratch CNN–Transformer | 0.638 (0.583–0.691) | 0.992 (0.990–0.993) | 0.607 (0.556–0.656) | **0.997 (0.996–0.997)** | 0.600 (0.554–0.644) | 0.597 (0.551–0.640) |
| | ECG-FM | **0.722 (0.674–0.768)** | **0.995 (0.993–0.996)** | 0.672 (0.630–0.715) | **0.997 (0.997–0.998)** | 0.662 (0.625–0.699) | 0.659 (0.622–0.697) |
| | DeepECG-SSL | 0.690 (0.639–0.740) | 0.994 (0.993–0.996) | 0.668 (0.625–0.715) | **0.997 (0.996–0.998)** | 0.656 (0.616–0.694) | 0.654 (0.614–0.692) |
| | CardioLoom | 0.700 (0.652–0.746) | **0.995 (0.994–0.996)** | **0.734 (0.691–0.775)** | 0.996 (0.996–0.997) | **0.663 (0.629–0.698)** | **0.663 (0.629–0.699)** |
| Short PR | Scratch CNN–Transformer | 0.235 (0.179–0.307) | 0.985 (0.979–0.989) | 0.297 (0.227–0.360) | **0.999 (0.999–0.999)** | 0.313 (0.248–0.370) | 0.313 (0.248–0.371) |
| | ECG-FM | 0.257 (0.201–0.320) | **0.990 (0.985–0.993)** | **0.448 (0.376–0.513)** | 0.998 (0.998–0.998) | 0.347 (0.295–0.395) | **0.355 (0.302–0.404)** |
| | DeepECG-SSL | **0.300 (0.236–0.374)** | **0.990 (0.986–0.994)** | 0.323 (0.256–0.391) | **0.999 (0.999–0.999)** | **0.349 (0.284–0.408)** | 0.349 (0.285–0.409) |
| | CardioLoom | 0.271 (0.212–0.340) | 0.986 (0.977–0.993) | 0.297 (0.232–0.361) | **0.999 (0.999–0.999)** | 0.311 (0.249–0.369) | 0.311 (0.248–0.371) |
| VPE | Scratch CNN–Transformer | 0.766 (0.668–0.847) | 0.992 (0.986–0.996) | 0.678 (0.569–0.770) | **1.000 (1.000–1.000)** | 0.757 (0.680–0.822) | 0.762 (0.691–0.825) |
| | ECG-FM | 0.847 (0.777–0.903) | **0.997 (0.995–0.999)** | **0.772 (0.677–0.859)** | **1.000 (1.000–1.000)** | **0.804 (0.736–0.862)** | **0.804 (0.739–0.862)** |
| | DeepECG-SSL | 0.850 (0.785–0.908) | **0.997 (0.995–0.999)** | 0.738 (0.647–0.839) | **1.000 (1.000–1.000)** | 0.801 (0.733–0.864) | **0.804 (0.740–0.865)** |
| | CardioLoom | **0.857 (0.796–0.910)** | 0.996 (0.992–0.999) | 0.718 (0.613–0.821) | **1.000 (1.000–1.000)** | 0.786 (0.713–0.852) | 0.789 (0.720–0.853) |
| 3° AVB | Scratch CNN–Transformer | 0.450 (0.374–0.527) | 0.996 (0.996–0.997) | 0.651 (0.586–0.715) | 0.997 (0.996–0.997) | 0.426 (0.377–0.475) | 0.452 (0.406–0.500) |
| | ECG-FM | 0.600 (0.527–0.661) | 0.997 (0.996–0.998) | **0.671 (0.608–0.733)** | 0.998 (0.998–0.998) | 0.536 (0.486–0.585) | 0.546 (0.497–0.593) |
| | DeepECG-SSL | 0.622 (0.546–0.683) | **0.998 (0.997–0.999)** | 0.598 (0.539–0.654) | **0.999 (0.998–0.999)** | 0.538 (0.481–0.588) | 0.540 (0.483–0.589) |
| | CardioLoom | **0.656 (0.590–0.715)** | **0.998 (0.998–0.999)** | 0.550 (0.483–0.617) | **0.999 (0.999–0.999)** | **0.578 (0.524–0.632)** | **0.578 (0.525–0.632)** |
| 2° 1 Type AVB | Scratch CNN–Transformer | 0.754 (0.665–0.821) | 0.997 (0.995–0.999) | 0.669 (0.584–0.744) | **1.000 (1.000–1.000)** | 0.717 (0.635–0.780) | 0.718 (0.636–0.780) |
| | ECG-FM | 0.885 (0.826–0.927) | **0.999 (0.998–1.000)** | 0.684 (0.592–0.762) | **1.000 (1.000–1.000)** | 0.798 (0.729–0.853) | 0.810 (0.747–0.861) |
| | DeepECG-SSL | 0.887 (0.829–0.931) | **0.999 (0.997–1.000)** | 0.735 (0.651–0.811) | **1.000 (1.000–1.000)** | 0.813 (0.747–0.862) | 0.817 (0.754–0.864) |
| | CardioLoom | **0.903 (0.849–0.938)** | **0.999 (0.998–1.000)** | **0.750 (0.673–0.820)** | **1.000 (1.000–1.000)** | **0.829 (0.773–0.877)** | **0.834 (0.780–0.880)** |

| Disease | Model | AUPRC | AUROC | Sensitivity | Specificity | F1 Score | MCC |
|---|---|---|---|---|---|---|---|
| 2° 2 Type AVB | Scratch CNN–Transformer | 0.575 (0.409–0.717) | 0.999 (0.999–1.000) | 0.440 (0.309–0.579) | **1.000 (1.000–1.000)** | 0.518 (0.371–0.635) | 0.526 (0.382–0.643) |
| | ECG-FM | 0.749 (0.629–0.850) | 0.999 (0.998–1.000) | **0.740 (0.600–0.860)** | **1.000 (1.000–1.000)** | 0.718 (0.602–0.817) | 0.719 (0.604–0.820) |
| | DeepECG-SSL | **0.820 (0.708–0.901)** | **1.000 (1.000–1.000)** | 0.640 (0.490–0.768) | **1.000 (1.000–1.000)** | **0.744 (0.619–0.842)** | **0.754 (0.637–0.844)** |
| | CardioLoom | 0.775 (0.646–0.874) | 0.999 (0.998–1.000) | 0.680 (0.537–0.805) | **1.000 (1.000–1.000)** | 0.701 (0.582–0.796) | 0.701 (0.586–0.797) |
| LVH | Scratch CNN–Transformer | 0.704 (0.687–0.719) | 0.964 (0.962–0.966) | 0.653 (0.637–0.669) | 0.976 (0.975–0.978) | 0.639 (0.626–0.652) | 0.617 (0.604–0.630) |
| | ECG-FM | 0.740 (0.724–0.753) | 0.969 (0.967–0.971) | 0.672 (0.656–0.686) | **0.980 (0.978–0.981)** | 0.670 (0.657–0.682) | 0.650 (0.637–0.662) |
| | DeepECG-SSL | 0.742 (0.726–0.755) | 0.971 (0.969–0.972) | **0.694 (0.681–0.708)** | 0.978 (0.976–0.979) | 0.674 (0.661–0.685) | 0.654 (0.640–0.665) |
| | CardioLoom | **0.756 (0.741–0.769)** | **0.972 (0.970–0.974)** | 0.686 (0.672–0.700) | **0.980 (0.979–0.981)** | **0.682 (0.670–0.694)** | **0.663 (0.650–0.675)** |
| LAH | Scratch CNN–Transformer | 0.431 (0.401–0.462) | 0.968 (0.965–0.971) | 0.535 (0.506–0.561) | **0.987 (0.986–0.988)** | 0.469 (0.444–0.491) | 0.462 (0.438–0.484) |
| | ECG-FM | 0.439 (0.408–0.472) | 0.970 (0.968–0.973) | 0.548 (0.519–0.575) | 0.985 (0.984–0.986) | 0.461 (0.437–0.484) | 0.456 (0.432–0.479) |
| | DeepECG-SSL | 0.441 (0.411–0.472) | 0.968 (0.965–0.971) | **0.549 (0.522–0.577)** | 0.985 (0.984–0.986) | 0.460 (0.438–0.483) | 0.455 (0.433–0.478) |
| | CardioLoom | **0.447 (0.416–0.479)** | **0.971 (0.969–0.973)** | **0.549 (0.520–0.576)** | 0.986 (0.985–0.987) | **0.470 (0.446–0.493)** | **0.465 (0.441–0.488)** |
| RVH | Scratch CNN–Transformer | 0.467 (0.423–0.513) | 0.985 (0.982–0.988) | 0.584 (0.543–0.624) | 0.993 (0.992–0.993) | 0.488 (0.455–0.523) | 0.489 (0.455–0.524) |
| | ECG-FM | **0.560 (0.512–0.601)** | **0.987 (0.983–0.990)** | **0.627 (0.584–0.667)** | 0.994 (0.994–0.995) | **0.559 (0.522–0.593)** | **0.558 (0.522–0.592)** |
| | DeepECG-SSL | 0.545 (0.497–0.591) | **0.987 (0.984–0.990)** | 0.545 (0.498–0.592) | **0.996 (0.995–0.996)** | 0.544 (0.506–0.581) | 0.540 (0.501–0.577) |
| | CardioLoom | 0.553 (0.503–0.596) | **0.987 (0.983–0.989)** | 0.560 (0.514–0.602) | **0.996 (0.995–0.996)** | 0.554 (0.512–0.590) | 0.550 (0.508–0.586) |
| VA | Scratch CNN–Transformer | 0.229 (0.171–0.300) | 0.992 (0.989–0.995) | 0.425 (0.320–0.521) | **0.998 (0.998–0.999)** | 0.330 (0.250–0.401) | 0.337 (0.256–0.408) |
| | ECG-FM | 0.236 (0.174–0.307) | **0.993 (0.991–0.996)** | 0.460 (0.357–0.553) | **0.998 (0.997–0.998)** | 0.317 (0.244–0.382) | 0.332 (0.259–0.399) |
| | DeepECG-SSL | 0.228 (0.173–0.291) | **0.993 (0.989–0.996)** | **0.523 (0.426–0.609)** | **0.998 (0.997–0.998)** | **0.364 (0.287–0.431)** | **0.381 (0.306–0.448)** |
| | CardioLoom | **0.258 (0.196–0.331)** | **0.993 (0.990–0.995)** | 0.431 (0.337–0.522) | **0.998 (0.998–0.999)** | 0.331 (0.260–0.397) | 0.339 (0.269–0.406) |

| Disease | Model | AUPRC | AUROC | Sensitivity | Specificity | F1 Score | MCC |
|---|---|---|---|---|---|---|---|
| RAH | Scratch CNN–Transformer | 0.142 (0.080–0.231) | 0.974 (0.953–0.987) | **0.258 (0.155–0.368)** | 0.999 (0.999–1.000) | **0.245 (0.150–0.333)** | **0.244 (0.150–0.334)** |
| | ECG-FM | 0.101 (0.058–0.175) | 0.966 (0.950–0.980) | 0.191 (0.110–0.281) | 0.999 (0.999–0.999) | 0.175 (0.105–0.248) | 0.175 (0.105–0.248) |
| | DeepECG-SSL | **0.157 (0.089–0.255)** | 0.978 (0.963–0.989) | 0.247 (0.162–0.345) | 0.999 (0.999–0.999) | 0.228 (0.153–0.304) | 0.228 (0.153–0.306) |
| | CardioLoom | 0.153 (0.080–0.262) | **0.980 (0.971–0.988)** | 0.157 (0.069–0.242) | **1.000 (1.000–1.000)** | 0.207 (0.094–0.299) | 0.218 (0.102–0.315) |
| DC | Scratch CNN–Transformer | 0.001 (0.000–0.003) | 0.703 (0.518–0.837) | **0.000 (0.000–0.000)** | **1.000 (1.000–1.000)** | **0.000 (0.000–0.000)** | **0.000 (0.000–0.000)** |
| | ECG-FM | 0.008 (0.001–0.033) | **0.853 (0.770–0.918)** | **0.000 (0.000–0.000)** | **1.000 (1.000–1.000)** | **0.000 (0.000–0.000)** | **0.000 (0.000–0.000)** |
| | DeepECG-SSL | 0.004 (0.000–0.015) | 0.831 (0.692–0.931) | **0.000 (0.000–0.000)** | **1.000 (1.000–1.000)** | **0.000 (0.000–0.000)** | **0.000 (0.000–0.000)** |
| | CardioLoom | **0.022 (0.003–0.067)** | 0.810 (0.641–0.942) | **0.000 (0.000–0.000)** | **1.000 (1.000–1.000)** | **0.000 (0.000–0.000)** | **0.000 (0.000–0.000)** |
| BVH | Scratch CNN–Transformer | **0.058 (0.021–0.156)** | 0.975 (0.962–0.986) | 0.136 (0.045–0.239) | **1.000 (0.999–1.000)** | **0.132 (0.044–0.225)** | 0.132 (0.044–0.225) |
| | ECG-FM | 0.034 (0.014–0.101) | 0.976 (0.961–0.986) | 0.045 (0.000–0.111) | **1.000 (1.000–1.000)** | 0.069 (0.000–0.167) | 0.080 (0.000–0.200) |
| | DeepECG-SSL | 0.050 (0.022–0.110) | **0.977 (0.959–0.989)** | 0.182 (0.067–0.314) | 0.999 (0.999–1.000) | 0.128 (0.048–0.219) | **0.134 (0.050–0.225)** |
| | CardioLoom | 0.050 (0.021–0.136) | 0.974 (0.953–0.987) | **0.227 (0.098–0.382)** | 0.999 (0.999–0.999) | 0.114 (0.049–0.194) | 0.131 (0.057–0.217) |
| VVI | Scratch CNN–Transformer | 0.825 (0.789–0.861) | 0.997 (0.996–0.998) | 0.850 (0.816–0.882) | **0.998 (0.998–0.998)** | 0.810 (0.783–0.836) | 0.809 (0.782–0.836) |
| | ECG-FM | 0.827 (0.792–0.860) | 0.998 (0.997–0.999) | 0.927 (0.907–0.944) | **0.998 (0.997–0.998)** | 0.837 (0.813–0.859) | 0.839 (0.816–0.861) |
| | DeepECG-SSL | 0.855 (0.819–0.889) | **0.999 (0.999–0.999)** | **0.946 (0.929–0.962)** | **0.998 (0.997–0.998)** | 0.852 (0.828–0.872) | **0.855 (0.833–0.874)** |
| | CardioLoom | **0.865 (0.834–0.892)** | **0.999 (0.999–0.999)** | 0.918 (0.896–0.938) | **0.998 (0.998–0.998)** | **0.854 (0.833–0.874)** | **0.855 (0.834–0.875)** |
| VAT | Scratch CNN–Transformer | 0.751 (0.685–0.810) | 0.995 (0.992–0.997) | 0.664 (0.597–0.721) | 0.999 (0.999–0.999) | 0.708 (0.646–0.757) | 0.709 (0.645–0.758) |
| | ECG-FM | **0.906 (0.875–0.933)** | **0.999 (0.998–0.999)** | 0.837 (0.796–0.877) | **1.000 (0.999–1.000)** | **0.866 (0.833–0.894)** | **0.866 (0.833–0.894)** |
| | DeepECG-SSL | 0.868 (0.826–0.906) | 0.997 (0.996–0.999) | 0.813 (0.764–0.858) | 0.999 (0.999–1.000) | 0.826 (0.779–0.865) | 0.825 (0.778–0.865) |
| | CardioLoom | 0.895 (0.855–0.925) | 0.998 (0.996–0.999) | **0.840 (0.790–0.882)** | 0.999 (0.999–1.000) | 0.839 (0.795–0.873) | 0.838 (0.795–0.872) |

| Disease | Model | AUPRC | AUROC | Sensitivity | Specificity | F1 Score | MCC |
|---|---|---|---|---|---|---|---|
| DDD | Scratch CNN–Transformer | 0.640 (0.562–0.715) | 0.995 (0.989–0.998) | 0.639 (0.567–0.706) | 0.999 (0.999–0.999) | 0.603 (0.534–0.668) | 0.604 (0.535–0.668) |
| | ECG-FM | **0.849 (0.794–0.896)** | 0.998 (0.996–1.000) | **0.813 (0.753–0.871)** | 0.999 (0.999–1.000) | **0.786 (0.741–0.827)** | **0.786 (0.741–0.827)** |
| | DeepECG-SSL | 0.775 (0.701–0.843) | **0.999 (0.998–0.999)** | 0.671 (0.601–0.746) | **1.000 (1.000–1.000)** | 0.739 (0.680–0.794) | 0.742 (0.684–0.798) |
| | CardioLoom | 0.828 (0.773–0.876) | **0.999 (0.998–0.999)** | 0.708 (0.640–0.781) | **1.000 (1.000–1.000)** | 0.756 (0.704–0.807) | 0.757 (0.706–0.807) |
| AAI | Scratch CNN–Transformer | 0.656 (0.554–0.747) | 0.989 (0.981–0.994) | 0.500 (0.396–0.606) | **1.000 (1.000–1.000)** | 0.616 (0.515–0.704) | 0.633 (0.540–0.713) |
| | ECG-FM | 0.835 (0.756–0.904) | **0.996 (0.991–0.999)** | **0.821 (0.753–0.881)** | **1.000 (0.999–1.000)** | **0.815 (0.757–0.865)** | **0.815 (0.757–0.866)** |
| | DeepECG-SSL | **0.842 (0.781–0.891)** | **0.996 (0.992–0.999)** | 0.750 (0.680–0.823) | **1.000 (1.000–1.000)** | 0.795 (0.739–0.844) | 0.796 (0.740–0.844) |
| | CardioLoom | 0.838 (0.778–0.886) | 0.994 (0.988–0.999) | 0.704 (0.625–0.782) | **1.000 (1.000–1.000)** | 0.782 (0.726–0.836) | 0.786 (0.732–0.838) |
| VP | Scratch CNN–Transformer | 0.145 (0.101–0.206) | 0.988 (0.980–0.995) | 0.282 (0.193–0.367) | **0.998 (0.998–0.999)** | 0.198 (0.132–0.258) | 0.206 (0.139–0.268) |
| | ECG-FM | 0.198 (0.139–0.278) | **0.993 (0.989–0.996)** | **0.513 (0.405–0.613)** | 0.997 (0.996–0.998) | 0.240 (0.183–0.296) | 0.283 (0.222–0.342) |
| | DeepECG-SSL | 0.098 (0.073–0.133) | 0.991 (0.985–0.995) | 0.197 (0.117–0.283) | **0.998 (0.998–0.999)** | 0.136 (0.084–0.195) | 0.141 (0.088–0.200) |
| | CardioLoom | **0.222 (0.153–0.302)** | 0.990 (0.981–0.996) | 0.453 (0.360–0.554) | **0.998 (0.998–0.999)** | **0.297 (0.236–0.360)** | **0.315 (0.251–0.376)** |
| PSM | Scratch CNN–Transformer | 0.263 (0.127–0.446) | 0.989 (0.972–0.998) | 0.371 (0.209–0.571) | **1.000 (1.000–1.000)** | 0.342 (0.205–0.490) | 0.343 (0.206–0.494) |
| | ECG-FM | 0.412 (0.239–0.592) | **0.998 (0.995–0.999)** | 0.286 (0.130–0.485) | **1.000 (1.000–1.000)** | 0.377 (0.187–0.560) | 0.398 (0.208–0.577) |
| | DeepECG-SSL | 0.347 (0.192–0.541) | 0.995 (0.987–0.999) | 0.457 (0.281–0.667) | **1.000 (1.000–1.000)** | 0.421 (0.274–0.561) | 0.422 (0.276–0.568) |
| | CardioLoom | **0.489 (0.316–0.672)** | 0.996 (0.992–0.999) | **0.514 (0.345–0.706)** | **1.000 (1.000–1.000)** | **0.537 (0.390–0.676)** | **0.538 (0.394–0.677)** |
| HK | Scratch CNN–Transformer | 0.022 (0.009–0.051) | 0.918 (0.866–0.959) | 0.067 (0.000–0.146) | **1.000 (1.000–1.000)** | 0.079 (0.000–0.162) | 0.080 (0.000–0.165) |
| | ECG-FM | **0.108 (0.035–0.208)** | 0.945 (0.902–0.982) | 0.200 (0.091–0.325) | **1.000 (0.999–1.000)** | **0.191 (0.091–0.300)** | **0.191 (0.092–0.300)** |
| | DeepECG-SSL | 0.084 (0.037–0.191) | 0.976 (0.947–0.992) | **0.222 (0.105–0.357)** | 0.999 (0.999–0.999) | 0.145 (0.071–0.229) | 0.154 (0.074–0.244) |
| | CardioLoom | 0.066 (0.034–0.121) | **0.982 (0.959–0.995)** | 0.111 (0.027–0.208) | **1.000 (1.000–1.000)** | 0.115 (0.028–0.209) | 0.115 (0.028–0.212) |

| Disease | Model | AUPRC | AUROC | Sensitivity | Specificity | F1 Score | MCC |
|---|---|---|---|---|---|---|---|
| LK | Scratch CNN–Transformer | 0.015 (0.008–0.031) | 0.940 (0.907–0.964) | 0.196 (0.109–0.302) | 0.996 (0.996–0.997) | 0.047 (0.025–0.075) | 0.071 (0.038–0.110) |
| | ECG-FM | 0.042 (0.024–0.077) | 0.965 (0.947–0.980) | 0.214 (0.111–0.333) | **0.999 (0.998–0.999)** | 0.113 (0.059–0.170) | 0.127 (0.066–0.192) |
| | DeepECG-SSL | 0.063 (0.022–0.129) | 0.981 (0.974–0.988) | 0.286 (0.164–0.415) | 0.998 (0.997–0.998) | 0.102 (0.056–0.150) | 0.132 (0.076–0.189) |
| | CardioLoom | **0.070 (0.032–0.133)** | **0.983 (0.975–0.990)** | **0.375 (0.250–0.508)** | 0.997 (0.997–0.998) | **0.121 (0.077–0.168)** | **0.163 (0.108–0.219)** |
| **Tier 2: ECHO Diagnosis** | | | | | | | |
| NER | Scratch CNN–Transformer | 0.850 (0.843–0.858) | 0.888 (0.884–0.892) | 0.860 (0.855–0.867) | 0.745 (0.738–0.752) | 0.792 (0.785–0.799) | 0.603 (0.594–0.613) |
| | ECG-FM | **0.878 (0.872–0.884)** | **0.908 (0.904–0.911)** | 0.886 (0.881–0.892) | 0.757 (0.750–0.764) | **0.812 (0.805–0.818)** | **0.641 (0.632–0.651)** |
| | DeepECG-SSL | 0.875 (0.869–0.882) | 0.906 (0.902–0.909) | **0.890 (0.885–0.896)** | 0.745 (0.738–0.752) | 0.808 (0.802–0.815) | 0.634 (0.625–0.643) |
| | CardioLoom | 0.871 (0.864–0.878) | 0.903 (0.899–0.907) | 0.857 (0.851–0.864) | **0.781 (0.775–0.788)** | 0.807 (0.801–0.814) | 0.635 (0.626–0.645) |
| LAE | Scratch CNN–Transformer | 0.420 (0.405–0.435) | 0.792 (0.785–0.799) | 0.621 (0.608–0.634) | 0.784 (0.778–0.790) | 0.465 (0.454–0.475) | 0.337 (0.325–0.349) |
| | ECG-FM | 0.470 (0.454–0.486) | 0.822 (0.816–0.828) | 0.615 (0.602–0.627) | **0.829 (0.823–0.834)** | 0.502 (0.491–0.514) | 0.387 (0.375–0.399) |
| | DeepECG-SSL | 0.463 (0.448–0.479) | 0.821 (0.815–0.827) | 0.628 (0.615–0.642) | 0.817 (0.811–0.823) | 0.499 (0.487–0.510) | 0.382 (0.370–0.395) |
| | CardioLoom | **0.476 (0.460–0.492)** | **0.824 (0.818–0.830)** | **0.670 (0.657–0.683)** | 0.797 (0.791–0.803) | **0.504 (0.493–0.516)** | **0.390 (0.378–0.402)** |
| HCM | Scratch CNN–Transformer | 0.431 (0.413–0.450) | 0.824 (0.816–0.831) | **0.556 (0.537–0.573)** | 0.868 (0.863–0.872) | 0.437 (0.421–0.451) | 0.354 (0.338–0.369) |
| | ECG-FM | 0.481 (0.460–0.502) | **0.846 (0.838–0.852)** | 0.539 (0.522–0.558) | 0.898 (0.894–0.902) | 0.468 (0.453–0.483) | 0.390 (0.374–0.406) |
| | DeepECG-SSL | **0.489 (0.468–0.509)** | 0.843 (0.836–0.851) | 0.526 (0.506–0.543) | **0.904 (0.900–0.908)** | **0.469 (0.452–0.485)** | **0.393 (0.374–0.409)** |
| | CardioLoom | 0.486 (0.465–0.506) | 0.843 (0.836–0.850) | 0.549 (0.532–0.568) | 0.891 (0.887–0.895) | 0.465 (0.449–0.480) | 0.387 (0.370–0.403) |
| nHCM | Scratch CNN–Transformer | 0.346 (0.329–0.365) | 0.809 (0.801–0.817) | **0.547 (0.528–0.565)** | 0.854 (0.849–0.859) | 0.400 (0.386–0.414) | 0.319 (0.304–0.334) |
| | ECG-FM | 0.386 (0.367–0.406) | **0.830 (0.823–0.838)** | 0.538 (0.519–0.557) | 0.878 (0.873–0.882) | **0.425 (0.410–0.440)** | **0.348 (0.332–0.364)** |
| | DeepECG-SSL | 0.390 (0.370–0.410) | 0.828 (0.820–0.836) | 0.521 (0.500–0.540) | **0.885 (0.881–0.889)** | 0.424 (0.408–0.439) | 0.347 (0.330–0.363) |
| | CardioLoom | **0.391 (0.371–0.413)** | 0.829 (0.821–0.836) | 0.522 (0.504–0.542) | **0.885 (0.880–0.889)** | 0.424 (0.410–0.439) | 0.346 (0.331–0.363) |

| Disease | Model | AUPRC | AUROC | Sensitivity | Specificity | F1 Score | MCC |
|---|---|---|---|---|---|---|---|
| BAE | Scratch CNN–Transformer | 0.405 (0.376–0.431) | 0.921 (0.914–0.927) | 0.520 (0.492–0.546) | 0.965 (0.963–0.967) | 0.461 (0.439–0.482) | 0.436 (0.413–0.457) |
| | ECG-FM | **0.504 (0.472–0.532)** | **0.938 (0.932–0.944)** | **0.576 (0.549–0.604)** | 0.970 (0.968–0.972) | **0.522 (0.497–0.543)** | **0.499 (0.474–0.522)** |
| | DeepECG-SSL | 0.494 (0.463–0.523) | 0.935 (0.929–0.941) | 0.559 (0.532–0.586) | **0.971 (0.969–0.974)** | 0.518 (0.495–0.539) | 0.495 (0.471–0.517) |
| | CardioLoom | 0.503 (0.472–0.529) | **0.938 (0.932–0.943)** | **0.576 (0.547–0.602)** | 0.968 (0.966–0.970) | 0.513 (0.490–0.534) | 0.491 (0.467–0.512) |
| HFrEF | Scratch CNN–Transformer | 0.563 (0.520–0.602) | 0.958 (0.950–0.965) | 0.593 (0.558–0.624) | 0.983 (0.981–0.984) | 0.563 (0.529–0.591) | 0.549 (0.515–0.577) |
| | ECG-FM | 0.670 (0.630–0.707) | 0.972 (0.965–0.977) | 0.577 (0.540–0.610) | **0.990 (0.989–0.992)** | 0.620 (0.589–0.648) | 0.610 (0.579–0.639) |
| | DeepECG-SSL | 0.676 (0.636–0.711) | **0.973 (0.966–0.979)** | 0.625 (0.589–0.660) | 0.988 (0.986–0.989) | 0.629 (0.599–0.658) | 0.617 (0.586–0.647) |
| | CardioLoom | **0.690 (0.648–0.725)** | **0.973 (0.966–0.978)** | **0.682 (0.647–0.714)** | 0.987 (0.985–0.988) | **0.654 (0.626–0.681)** | **0.643 (0.614–0.671)** |
| RMWSSFLV | Scratch CNN–Transformer | 0.479 (0.434–0.526) | 0.952 (0.944–0.959) | 0.488 (0.446–0.530) | **0.989 (0.987–0.990)** | 0.512 (0.475–0.549) | 0.500 (0.462–0.537) |
| | ECG-FM | 0.526 (0.480–0.571) | 0.960 (0.952–0.966) | 0.551 (0.512–0.591) | 0.986 (0.984–0.987) | 0.532 (0.499–0.566) | 0.520 (0.486–0.554) |
| | DeepECG-SSL | 0.535 (0.489–0.585) | 0.962 (0.954–0.968) | **0.570 (0.527–0.610)** | 0.986 (0.985–0.988) | 0.551 (0.516–0.586) | 0.539 (0.503–0.574) |
| | CardioLoom | **0.541 (0.494–0.592)** | **0.963 (0.955–0.969)** | 0.543 (0.501–0.586) | **0.989 (0.987–0.990)** | **0.554 (0.518–0.590)** | **0.542 (0.506–0.579)** |
| DCM | Scratch CNN–Transformer | 0.361 (0.309–0.414) | 0.963 (0.955–0.970) | 0.515 (0.464–0.562) | 0.989 (0.987–0.990) | 0.424 (0.381–0.465) | 0.422 (0.381–0.463) |
| | ECG-FM | 0.389 (0.334–0.444) | 0.973 (0.967–0.979) | 0.549 (0.498–0.603) | 0.989 (0.988–0.991) | 0.457 (0.415–0.498) | 0.456 (0.413–0.497) |
| | DeepECG-SSL | **0.446 (0.388–0.501)** | 0.976 (0.970–0.982) | **0.558 (0.505–0.607)** | 0.991 (0.990–0.992) | 0.483 (0.439–0.522) | **0.481 (0.436–0.519)** |
| | CardioLoom | 0.414 (0.361–0.471) | **0.978 (0.973–0.983)** | 0.511 (0.462–0.558) | **0.993 (0.992–0.994)** | **0.487 (0.442–0.530)** | **0.481 (0.435–0.524)** |
| oHCM | Scratch CNN–Transformer | 0.213 (0.172–0.260) | 0.948 (0.937–0.958) | 0.326 (0.267–0.387) | 0.993 (0.992–0.994) | 0.300 (0.254–0.346) | 0.295 (0.248–0.341) |
| | ECG-FM | 0.369 (0.304–0.431) | 0.964 (0.953–0.974) | 0.412 (0.353–0.471) | **0.995 (0.995–0.996)** | 0.426 (0.374–0.474) | 0.422 (0.369–0.469) |
| | DeepECG-SSL | 0.411 (0.347–0.477) | 0.965 (0.954–0.975) | **0.556 (0.496–0.619)** | 0.992 (0.991–0.993) | 0.459 (0.412–0.503) | 0.460 (0.414–0.505) |
| | CardioLoom | **0.429 (0.359–0.497)** | **0.967 (0.956–0.977)** | 0.510 (0.451–0.571) | **0.995 (0.994–0.996)** | **0.484 (0.433–0.531)** | **0.480 (0.429–0.527)** |

| Disease | Model | AUPRC | AUROC | Sensitivity | Specificity | F1 Score | MCC |
|---|---|---|---|---|---|---|---|
| ICM | Scratch CNN–Transformer | 0.201 (0.155–0.259) | 0.962 (0.943–0.976) | **0.429 (0.359–0.503)** | 0.990 (0.989–0.992) | 0.297 (0.249–0.349) | 0.306 (0.257–0.359) |
| | ECG-FM | 0.217 (0.168–0.272) | **0.970 (0.956–0.979)** | 0.337 (0.273–0.398) | **0.994 (0.993–0.995)** | 0.298 (0.244–0.351) | 0.295 (0.240–0.349) |
| | DeepECG-SSL | 0.208 (0.166–0.266) | 0.968 (0.950–0.979) | 0.410 (0.340–0.483) | 0.990 (0.989–0.992) | 0.289 (0.238–0.343) | 0.297 (0.246–0.350) |
| | CardioLoom | **0.243 (0.183–0.315)** | **0.970 (0.952–0.982)** | 0.426 (0.344–0.507) | 0.992 (0.991–0.993) | **0.326 (0.263–0.388)** | **0.329 (0.267–0.392)** |
| LVDD | Scratch CNN–Transformer | 0.035 (0.027–0.048) | 0.772 (0.746–0.799) | 0.108 (0.075–0.143) | **0.982 (0.981–0.984)** | 0.073 (0.050–0.094) | 0.064 (0.040–0.088) |
| | ECG-FM | 0.038 (0.030–0.049) | 0.793 (0.769–0.818) | 0.172 (0.130–0.219) | 0.975 (0.973–0.977) | 0.090 (0.068–0.114) | 0.088 (0.063–0.115) |
| | DeepECG-SSL | **0.041 (0.032–0.053)** | **0.806 (0.783–0.829)** | 0.176 (0.129–0.222) | 0.979 (0.977–0.980) | **0.103 (0.075–0.130)** | **0.100 (0.071–0.129)** |
| | CardioLoom | 0.039 (0.031–0.050) | 0.803 (0.779–0.828) | **0.192 (0.144–0.244)** | 0.971 (0.969–0.973) | 0.090 (0.067–0.115) | 0.091 (0.064–0.121) |
| VA | Scratch CNN–Transformer | 0.163 (0.109–0.234) | 0.968 (0.951–0.983) | 0.266 (0.171–0.383) | 0.997 (0.996–0.997) | 0.245 (0.160–0.330) | 0.243 (0.160–0.330) |
| | ECG-FM | 0.200 (0.131–0.296) | 0.968 (0.948–0.983) | **0.373 (0.253–0.523)** | 0.996 (0.996–0.997) | **0.311 (0.225–0.395)** | **0.313 (0.226–0.398)** |
| | DeepECG-SSL | 0.206 (0.145–0.286) | 0.967 (0.949–0.982) | 0.260 (0.169–0.366) | **0.998 (0.997–0.998)** | 0.272 (0.187–0.352) | 0.269 (0.186–0.352) |
| | CardioLoom | **0.232 (0.168–0.306)** | **0.977 (0.963–0.987)** | 0.320 (0.214–0.444) | 0.997 (0.996–0.997) | 0.283 (0.206–0.359) | 0.282 (0.204–0.361) |
| RAE | Scratch CNN–Transformer | **0.017 (0.011–0.027)** | 0.871 (0.833–0.905) | 0.129 (0.065–0.202) | **0.990 (0.988–0.991)** | **0.043 (0.022–0.068)** | 0.053 (0.026–0.085) |
| | ECG-FM | 0.014 (0.010–0.022) | 0.882 (0.845–0.911) | 0.149 (0.082–0.227) | 0.982 (0.981–0.984) | 0.032 (0.017–0.049) | 0.046 (0.022–0.073) |
| | DeepECG-SSL | 0.016 (0.011–0.023) | 0.886 (0.852–0.914) | **0.307 (0.214–0.404)** | 0.970 (0.968–0.972) | 0.041 (0.027–0.056) | **0.075 (0.049–0.100)** |
| | CardioLoom | 0.016 (0.011–0.025) | **0.889 (0.856–0.916)** | 0.109 (0.050–0.174) | 0.988 (0.987–0.989) | 0.033 (0.015–0.055) | 0.041 (0.016–0.069) |
| AM | Scratch CNN–Transformer | 0.013 (0.005–0.032) | 0.742 (0.674–0.808) | 0.037 (0.000–0.080) | **0.999 (0.999–0.999)** | 0.045 (0.000–0.095) | 0.045 (-0.001–0.096) |
| | ECG-FM | 0.027 (0.012–0.054) | 0.809 (0.756–0.853) | **0.062 (0.014–0.122)** | **0.999 (0.999–0.999)** | 0.077 (0.017–0.145) | 0.078 (0.016–0.150) |
| | DeepECG-SSL | **0.035 (0.011–0.081)** | 0.800 (0.754–0.844) | **0.062 (0.011–0.127)** | **0.999 (0.999–1.000)** | **0.083 (0.016–0.165)** | **0.087 (0.016–0.174)** |
| | CardioLoom | 0.022 (0.010–0.046) | **0.823 (0.778–0.864)** | **0.062 (0.013–0.119)** | **0.999 (0.998–0.999)** | 0.070 (0.015–0.131) | 0.070 (0.013–0.132) |

| Disease | Model | AUPRC | AUROC | Sensitivity | Specificity | F1 Score | MCC |
|---|---|---|---|---|---|---|---|
| LAM | Scratch CNN–Transformer | 0.009 (0.003–0.019) | 0.853 (0.758–0.928) | 0.083 (0.000–0.192) | 0.997 (0.996–0.997) | 0.032 (0.000–0.072) | 0.039 (-0.001–0.087) |
| | ECG-FM | 0.015 (0.004–0.036) | 0.876 (0.792–0.936) | 0.083 (0.000–0.161) | 0.998 (0.997–0.998) | 0.043 (0.000–0.089) | 0.048 (-0.001–0.097) |
| | DeepECG-SSL | 0.026 (0.008–0.052) | **0.903 (0.842–0.956)** | **0.111 (0.023–0.220)** | **0.999 (0.999–0.999)** | **0.085 (0.020–0.165)** | **0.087 (0.019–0.169)** |
| | CardioLoom | **0.028 (0.006–0.088)** | 0.880 (0.801–0.940) | 0.083 (0.000–0.222) | **0.999 (0.999–1.000)** | 0.082 (0.000–0.196) | 0.081 (-0.001–0.200) |
| AVC | Scratch CNN–Transformer | 0.359 (0.343–0.374) | 0.787 (0.780–0.795) | 0.626 (0.611–0.641) | 0.772 (0.766–0.778) | 0.429 (0.417–0.441) | 0.314 (0.300–0.326) |
| | ECG-FM | 0.412 (0.394–0.430) | **0.816 (0.810–0.823)** | 0.632 (0.618–0.647) | 0.807 (0.802–0.813) | **0.464 (0.451–0.476)** | **0.358 (0.344–0.371)** |
| | DeepECG-SSL | 0.400 (0.382–0.418) | 0.811 (0.803–0.817) | 0.614 (0.598–0.628) | **0.809 (0.803–0.815)** | 0.455 (0.442–0.467) | 0.346 (0.332–0.360) |
| | CardioLoom | **0.415 (0.399–0.433)** | **0.816 (0.810–0.823)** | **0.641 (0.627–0.654)** | 0.800 (0.794–0.806) | 0.462 (0.450–0.475) | 0.356 (0.343–0.370) |
| TR | Scratch CNN–Transformer | 0.528 (0.505–0.553) | 0.899 (0.892–0.906) | 0.587 (0.565–0.609) | **0.947 (0.945–0.951)** | 0.547 (0.529–0.566) | 0.503 (0.483–0.522) |
| | ECG-FM | 0.579 (0.556–0.602) | 0.913 (0.906–0.918) | **0.650 (0.629–0.670)** | 0.937 (0.934–0.941) | 0.561 (0.544–0.577) | 0.520 (0.501–0.536) |
| | DeepECG-SSL | **0.590 (0.567–0.614)** | **0.917 (0.910–0.922)** | 0.638 (0.616–0.659) | 0.946 (0.943–0.950) | **0.577 (0.560–0.594)** | **0.536 (0.518–0.555)** |
| | CardioLoom | 0.588 (0.566–0.612) | 0.916 (0.910–0.922) | 0.633 (0.612–0.653) | **0.947 (0.943–0.950)** | 0.576 (0.557–0.592) | 0.534 (0.515–0.552) |
| MR | Scratch CNN–Transformer | 0.416 (0.391–0.441) | 0.891 (0.884–0.898) | 0.540 (0.517–0.561) | **0.941 (0.938–0.944)** | 0.471 (0.451–0.489) | 0.428 (0.408–0.447) |
| | ECG-FM | **0.500 (0.474–0.525)** | 0.913 (0.906–0.919) | 0.617 (0.594–0.640) | 0.939 (0.936–0.943) | **0.515 (0.496–0.534)** | 0.478 (0.460–0.498) |
| | DeepECG-SSL | 0.483 (0.457–0.508) | 0.913 (0.907–0.920) | **0.629 (0.604–0.651)** | 0.936 (0.933–0.940) | **0.515 (0.496–0.532)** | **0.479 (0.459–0.497)** |
| | CardioLoom | 0.478 (0.452–0.503) | **0.914 (0.908–0.920)** | 0.627 (0.603–0.648) | 0.936 (0.932–0.939) | 0.512 (0.491–0.529) | 0.476 (0.455–0.494) |
| AR | Scratch CNN–Transformer | 0.292 (0.263–0.323) | 0.860 (0.848–0.870) | 0.351 (0.324–0.380) | 0.972 (0.970–0.974) | 0.348 (0.325–0.375) | 0.321 (0.296–0.348) |
| | ECG-FM | **0.404 (0.371–0.436)** | **0.890 (0.880–0.900)** | **0.432 (0.404–0.462)** | **0.977 (0.975–0.979)** | **0.438 (0.413–0.463)** | **0.414 (0.389–0.441)** |
| | DeepECG-SSL | 0.397 (0.365–0.430) | **0.890 (0.880–0.899)** | 0.420 (0.390–0.448) | 0.976 (0.974–0.978) | 0.422 (0.394–0.447) | 0.398 (0.370–0.423) |
| | CardioLoom | 0.395 (0.361–0.427) | **0.890 (0.880–0.899)** | 0.420 (0.390–0.448) | **0.977 (0.975–0.978)** | 0.425 (0.398–0.450) | 0.401 (0.374–0.427) |

| Disease | Model | AUPRC | AUROC | Sensitivity | Specificity | F1 Score | MCC |
|---|---|---|---|---|---|---|---|
| MVPLAC | Scratch CNN–Transformer | 0.072 (0.061–0.084) | 0.786 (0.770–0.803) | 0.240 (0.206–0.274) | 0.945 (0.942–0.947) | 0.134 (0.114–0.155) | 0.117 (0.095–0.139) |
| | ECG-FM | **0.092 (0.079–0.107)** | **0.827 (0.809–0.842)** | **0.312 (0.276–0.348)** | 0.945 (0.942–0.948) | **0.170 (0.147–0.193)** | **0.160 (0.136–0.184)** |
| | DeepECG-SSL | 0.086 (0.074–0.100) | 0.821 (0.806–0.836) | 0.276 (0.237–0.317) | 0.942 (0.939–0.946) | 0.148 (0.126–0.168) | 0.135 (0.110–0.158) |
| | CardioLoom | 0.088 (0.075–0.103) | 0.816 (0.800–0.832) | 0.282 (0.247–0.320) | **0.946 (0.943–0.949)** | 0.157 (0.136–0.178) | 0.145 (0.121–0.168) |
| MS | Scratch CNN–Transformer | 0.428 (0.378–0.476) | 0.957 (0.948–0.965) | 0.417 (0.369–0.466) | 0.993 (0.992–0.994) | 0.452 (0.410–0.490) | 0.446 (0.404–0.483) |
| | ECG-FM | **0.584 (0.537–0.632)** | **0.979 (0.974–0.984)** | 0.543 (0.495–0.595) | **0.995 (0.994–0.996)** | **0.583 (0.544–0.626)** | **0.579 (0.539–0.622)** |
| | DeepECG-SSL | 0.574 (0.525–0.621) | 0.975 (0.969–0.979) | **0.563 (0.516–0.614)** | 0.993 (0.992–0.994) | 0.562 (0.525–0.602) | 0.555 (0.517–0.595) |
| | CardioLoom | 0.574 (0.521–0.627) | 0.975 (0.969–0.980) | 0.518 (0.467–0.569) | 0.994 (0.993–0.995) | 0.556 (0.508–0.598) | 0.551 (0.504–0.593) |
| AS | Scratch CNN–Transformer | 0.200 (0.163–0.241) | 0.897 (0.882–0.912) | 0.375 (0.328–0.424) | 0.977 (0.975–0.978) | 0.246 (0.214–0.278) | 0.248 (0.214–0.280) |
| | ECG-FM | 0.348 (0.299–0.399) | **0.940 (0.930–0.950)** | **0.386 (0.341–0.432)** | 0.991 (0.990–0.992) | **0.386 (0.345–0.427)** | **0.378 (0.337–0.419)** |
| | DeepECG-SSL | **0.361 (0.311–0.414)** | 0.938 (0.927–0.949) | 0.326 (0.285–0.372) | **0.995 (0.994–0.995)** | 0.380 (0.337–0.425) | **0.378 (0.335–0.424)** |
| | CardioLoom | 0.341 (0.291–0.391) | 0.937 (0.926–0.948) | 0.368 (0.317–0.417) | 0.992 (0.991–0.993) | 0.382 (0.336–0.425) | 0.374 (0.328–0.418) |
| BAV | Scratch CNN–Transformer | 0.046 (0.038–0.059) | 0.771 (0.745–0.797) | 0.151 (0.120–0.187) | 0.982 (0.980–0.983) | 0.101 (0.080–0.126) | 0.095 (0.072–0.119) |
| | ECG-FM | 0.113 (0.088–0.144) | **0.845 (0.824–0.866)** | 0.168 (0.135–0.206) | 0.993 (0.992–0.994) | 0.181 (0.147–0.220) | 0.174 (0.140–0.213) |
| | DeepECG-SSL | **0.115 (0.089–0.147)** | 0.832 (0.810–0.854) | **0.196 (0.155–0.234)** | 0.991 (0.991–0.992) | 0.192 (0.154–0.229) | 0.184 (0.145–0.221) |
| | CardioLoom | **0.115 (0.090–0.149)** | 0.839 (0.819–0.859) | 0.177 (0.143–0.215) | **0.994 (0.993–0.994)** | **0.194 (0.161–0.232)** | **0.188 (0.154–0.226)** |
| PS | Scratch CNN–Transformer | **0.086 (0.023–0.211)** | 0.801 (0.662–0.908) | 0.184 (0.060–0.322) | 0.999 (0.999–0.999) | **0.186 (0.058–0.314)** | 0.185 (0.058–0.314) |
| | ECG-FM | 0.054 (0.011–0.161) | 0.875 (0.786–0.939) | 0.143 (0.024–0.281) | 0.999 (0.999–1.000) | 0.157 (0.026–0.306) | 0.157 (0.026–0.310) |
| | DeepECG-SSL | 0.060 (0.017–0.151) | **0.894 (0.802–0.958)** | **0.347 (0.161–0.537)** | 0.997 (0.997–0.998) | 0.168 (0.068–0.288) | **0.195 (0.082–0.321)** |
| | CardioLoom | 0.054 (0.019–0.111) | 0.890 (0.785–0.964) | 0.020 (0.000–0.068) | **1.000 (1.000–1.000)** | 0.032 (0.000–0.107) | 0.039 (-0.001–0.128) |

| Disease | Model | AUPRC | AUROC | Sensitivity | Specificity | F1 Score | MCC |
|---|---|---|---|---|---|---|---|
| EA | Scratch CNN–Transformer | **0.143 (0.005–0.339)** | 0.855 (0.747–0.939) | **0.185 (0.000–0.333)** | **1.000 (1.000–1.000)** | **0.244 (0.000–0.417)** | **0.257 (0.000–0.444)** |
| | ECG-FM | 0.004 (0.001–0.008) | 0.831 (0.734–0.937) | 0.000 (0.000–0.000) | 0.998 (0.998–0.999) | 0.000 (0.000–0.000) | -0.001 (-0.001–-0.001) |
| | DeepECG-SSL | 0.014 (0.003–0.053) | 0.852 (0.737–0.943) | 0.111 (0.000–0.263) | 0.999 (0.999–0.999) | 0.082 (0.000–0.179) | 0.084 (-0.001–0.188) |
| | CardioLoom | 0.023 (0.009–0.049) | **0.933 (0.863–0.986)** | 0.037 (0.000–0.138) | 0.999 (0.999–1.000) | 0.038 (0.000–0.129) | 0.037 (-0.001–0.131) |
| PR | Scratch CNN–Transformer | 0.007 (0.000–0.031) | 0.651 (0.431–0.840) | 0.048 (0.000–0.143) | 0.999 (0.999–0.999) | **0.027 (0.000–0.081)** | 0.029 (-0.001–0.089) |
| | ECG-FM | 0.003 (0.001–0.016) | 0.703 (0.492–0.936) | 0.000 (0.000–0.000) | **1.000 (1.000–1.000)** | 0.000 (0.000–0.000) | 0.000 (0.000–0.000) |
| | DeepECG-SSL | 0.004 (0.000–0.016) | 0.644 (0.447–0.892) | 0.000 (0.000–0.000) | **1.000 (1.000–1.000)** | 0.000 (0.000–0.000) | 0.000 (0.000–0.000) |
| | CardioLoom | **0.009 (0.001–0.038)** | **0.746 (0.576–0.942)** | **0.143 (0.000–0.357)** | 0.995 (0.995–0.996) | 0.025 (0.000–0.064) | **0.043 (-0.001–0.107)** |
| TS | Scratch CNN–Transformer | 0.003 (0.000–0.012) | 0.813 (0.634–0.959) | 0.083 (0.000–0.222) | 0.997 (0.997–0.998) | 0.014 (0.000–0.044) | 0.025 (-0.001–0.069) |
| | ECG-FM | **0.006 (0.001–0.029)** | 0.824 (0.606–0.967) | **0.167 (0.000–0.429)** | 0.997 (0.997–0.998) | **0.029 (0.000–0.078)** | **0.051 (-0.001–0.124)** |
| | DeepECG-SSL | 0.002 (0.000–0.006) | **0.883 (0.696–0.972)** | 0.000 (0.000–0.000) | **0.999 (0.999–1.000)** | 0.000 (0.000–0.000) | 0.000 (-0.001–0.000) |
| | CardioLoom | 0.002 (0.000–0.007) | 0.842 (0.670–0.964) | 0.000 (0.000–0.000) | 0.997 (0.996–0.997) | 0.000 (0.000–0.000) | -0.001 (-0.001–0.000) |
| CHD | Scratch CNN–Transformer | 0.342 (0.299–0.385) | 0.873 (0.853–0.890) | 0.449 (0.403–0.490) | 0.987 (0.985–0.988) | 0.410 (0.372–0.443) | 0.400 (0.362–0.434) |
| | ECG-FM | 0.455 (0.411–0.498) | 0.900 (0.882–0.916) | 0.496 (0.451–0.538) | **0.990 (0.989–0.991)** | **0.485 (0.449–0.518)** | **0.475 (0.440–0.509)** |
| | DeepECG-SSL | 0.451 (0.406–0.493) | **0.906 (0.889–0.924)** | 0.518 (0.475–0.560) | 0.989 (0.987–0.990) | 0.483 (0.446–0.518) | 0.474 (0.437–0.509) |
| | CardioLoom | **0.460 (0.414–0.506)** | 0.897 (0.878–0.914) | **0.528 (0.481–0.570)** | 0.988 (0.987–0.990) | 0.484 (0.448–0.517) | **0.475 (0.439–0.509)** |
| ASD | Scratch CNN–Transformer | 0.483 (0.435–0.532) | 0.906 (0.889–0.922) | 0.497 (0.454–0.538) | 0.991 (0.990–0.992) | 0.511 (0.472–0.548) | 0.502 (0.462–0.539) |
| | ECG-FM | 0.622 (0.577–0.665) | 0.933 (0.918–0.946) | 0.570 (0.525–0.615) | 0.993 (0.992–0.994) | 0.600 (0.561–0.637) | 0.594 (0.554–0.630) |
| | DeepECG-SSL | **0.644 (0.600–0.688)** | **0.936 (0.922–0.949)** | **0.576 (0.530–0.620)** | **0.994 (0.993–0.995)** | **0.617 (0.577–0.653)** | **0.611 (0.572–0.647)** |
| | CardioLoom | 0.631 (0.585–0.675) | 0.930 (0.914–0.945) | 0.570 (0.526–0.614) | 0.993 (0.992–0.994) | 0.599 (0.560–0.637) | 0.592 (0.553–0.630) |

| Disease | Model | AUPRC | AUROC | Sensitivity | Specificity | F1 Score | MCC |
|---|---|---|---|---|---|---|---|
| VSD | Scratch CNN–Transformer | 0.108 (0.075–0.152) | 0.847 (0.817–0.875) | 0.227 (0.174–0.280) | 0.993 (0.992–0.994) | 0.206 (0.157–0.256) | 0.201 (0.151–0.252) |
| | ECG-FM | 0.262 (0.194–0.333) | 0.888 (0.859–0.912) | **0.357 (0.283–0.423)** | 0.993 (0.993–0.994) | 0.320 (0.261–0.377) | **0.316 (0.256–0.373)** |
| | DeepECG-SSL | 0.205 (0.151–0.262) | 0.890 (0.862–0.911) | 0.212 (0.158–0.273) | **0.997 (0.997–0.998)** | 0.272 (0.209–0.337) | 0.279 (0.215–0.346) |
| | CardioLoom | **0.274 (0.207–0.341)** | **0.896 (0.870–0.917)** | 0.334 (0.268–0.395) | 0.994 (0.993–0.995) | **0.321 (0.262–0.377)** | **0.316 (0.257–0.372)** |
| PFO | Scratch CNN–Transformer | 0.014 (0.010–0.021) | 0.762 (0.720–0.799) | 0.015 (0.000–0.034) | **0.996 (0.996–0.997)** | 0.016 (0.000–0.036) | 0.012 (-0.004–0.032) |
| | ECG-FM | 0.021 (0.014–0.032) | 0.796 (0.751–0.835) | 0.126 (0.076–0.180) | 0.987 (0.986–0.988) | 0.060 (0.035–0.087) | 0.064 (0.035–0.093) |
| | DeepECG-SSL | **0.026 (0.015–0.041)** | 0.790 (0.743–0.831) | 0.045 (0.019–0.081) | 0.994 (0.993–0.995) | 0.036 (0.014–0.062) | 0.032 (0.010–0.059) |
| | CardioLoom | 0.025 (0.016–0.043) | **0.805 (0.759–0.841)** | **0.187 (0.122–0.256)** | 0.981 (0.979–0.982) | **0.065 (0.040–0.092)** | **0.077 (0.045–0.111)** |
| PDA | Scratch CNN–Transformer | 0.058 (0.027–0.118) | 0.850 (0.803–0.892) | 0.163 (0.073–0.264) | 0.998 (0.998–0.999) | 0.179 (0.084–0.286) | 0.178 (0.083–0.288) |
| | ECG-FM | 0.092 (0.042–0.157) | 0.871 (0.824–0.910) | 0.154 (0.085–0.226) | 0.998 (0.998–0.999) | 0.178 (0.101–0.255) | 0.178 (0.102–0.257) |
| | DeepECG-SSL | 0.070 (0.033–0.134) | **0.890 (0.856–0.920)** | 0.122 (0.058–0.194) | **0.999 (0.998–0.999)** | 0.153 (0.074–0.237) | 0.157 (0.075–0.242) |
| | CardioLoom | **0.123 (0.061–0.206)** | 0.877 (0.838–0.912) | **0.171 (0.097–0.253)** | **0.999 (0.998–0.999)** | **0.204 (0.118–0.291)** | **0.206 (0.118–0.296)** |
| TOF | Scratch CNN–Transformer | **0.003 (NE)** | 0.723 (NE) | 0.000 (NE) | **0.999 (NE)** | **0.000 (NE)** | 0.000 (NE) |
| | ECG-FM | 0.000 (NE) | 0.832 (NE) | **1.000 (NE)** | 0.168 (NE) | **0.000 (NE)** | 0.003 (NE) |
| | DeepECG-SSL | 0.001 (NE) | **0.931 (NE)** | **1.000 (NE)** | 0.710 (NE) | **0.000 (NE)** | 0.010 (NE) |
| | CardioLoom | 0.000 (NE) | 0.907 (NE) | **1.000 (NE)** | 0.785 (NE) | **0.000 (NE)** | **0.012 (NE)** |
| PH | Scratch CNN–Transformer | 0.469 (0.452–0.486) | 0.831 (0.825–0.838) | 0.594 (0.578–0.609) | 0.863 (0.858–0.869) | 0.491 (0.478–0.505) | 0.398 (0.383–0.413) |
| | ECG-FM | **0.520 (0.503–0.537)** | **0.857 (0.851–0.863)** | 0.616 (0.602–0.630) | **0.880 (0.875–0.884)** | 0.527 (0.515–0.539) | **0.442 (0.428–0.455)** |
| | DeepECG-SSL | 0.519 (0.503–0.537) | 0.856 (0.850–0.862) | 0.627 (0.612–0.641) | 0.876 (0.871–0.881) | **0.528 (0.515–0.541)** | **0.442 (0.429–0.457)** |
| | CardioLoom | 0.512 (0.493–0.529) | 0.851 (0.845–0.857) | **0.639 (0.624–0.654)** | 0.865 (0.860–0.870) | 0.521 (0.508–0.534) | 0.435 (0.420–0.449) |

| Disease | Model | AUPRC | AUROC | Sensitivity | Specificity | F1 Score | MCC |
|---|---|---|---|---|---|---|---|
| AD | Scratch CNN–Transformer | 0.256 (0.239–0.273) | 0.797 (0.789–0.806) | **0.534 (0.515–0.554)** | 0.839 (0.834–0.844) | 0.320 (0.305–0.334) | 0.259 (0.245–0.275) |
| | ECG-FM | **0.307 (0.289–0.326)** | **0.829 (0.821–0.837)** | 0.443 (0.424–0.463) | **0.910 (0.906–0.914)** | **0.362 (0.345–0.377)** | 0.299 (0.283–0.316) |
| | DeepECG-SSL | 0.302 (0.284–0.320) | 0.824 (0.816–0.832) | 0.513 (0.492–0.533) | 0.881 (0.877–0.885) | 0.361 (0.344–0.375) | **0.302 (0.285–0.317)** |
| | CardioLoom | **0.307 (0.288–0.325)** | 0.827 (0.819–0.835) | 0.521 (0.500–0.541) | 0.877 (0.873–0.882) | 0.360 (0.345–0.374) | **0.302 (0.286–0.317)** |
| Dex | Scratch CNN–Transformer | **0.308 (0.092–0.573)** | 0.839 (0.713–0.939) | **0.200 (0.000–0.429)** | **1.000 (1.000–1.000)** | **0.333 (0.000–0.600)** | **0.447 (0.000–0.655)** |
| | ECG-FM | 0.001 (0.000–0.010) | 0.683 (0.537–0.785) | 0.000 (0.000–0.000) | **1.000 (1.000–1.000)** | 0.000 (0.000–0.000) | 0.000 (0.000–0.000) |
| | DeepECG-SSL | 0.006 (0.000–0.081) | 0.678 (0.498–0.816) | 0.067 (0.000–0.231) | **1.000 (1.000–1.000)** | 0.077 (0.000–0.250) | 0.078 (0.000–0.258) |
| | CardioLoom | 0.011 (0.002–0.031) | **0.861 (0.701–0.965)** | 0.000 (0.000–0.000) | **1.000 (1.000–1.000)** | 0.000 (0.000–0.000) | 0.000 (0.000–0.000) |
| PE | Scratch CNN–Transformer | 0.255 (0.234–0.277) | 0.844 (0.835–0.853) | 0.452 (0.427–0.477) | 0.930 (0.927–0.933) | 0.328 (0.309–0.347) | 0.294 (0.275–0.314) |
| | ECG-FM | **0.310 (0.284–0.338)** | **0.871 (0.863–0.880)** | 0.443 (0.418–0.468) | 0.948 (0.945–0.950) | **0.366 (0.345–0.386)** | **0.332 (0.311–0.352)** |
| | DeepECG-SSL | 0.302 (0.276–0.328) | 0.863 (0.854–0.871) | **0.485 (0.460–0.511)** | 0.934 (0.931–0.937) | 0.357 (0.337–0.377) | 0.326 (0.305–0.346) |
| | CardioLoom | 0.294 (0.269–0.320) | 0.864 (0.855–0.872) | 0.394 (0.369–0.419) | **0.955 (0.952–0.957)** | 0.352 (0.331–0.375) | 0.316 (0.294–0.338) |
| CP | Scratch CNN–Transformer | 0.135 (0.055–0.251) | 0.966 (0.933–0.983) | 0.316 (0.167–0.481) | 0.998 (0.998–0.999) | 0.234 (0.113–0.357) | 0.241 (0.119–0.366) |
| | ECG-FM | **0.340 (0.178–0.522)** | 0.970 (0.941–0.989) | **0.351 (0.203–0.515)** | **1.000 (0.999–1.000)** | **0.400 (0.241–0.539)** | **0.403 (0.246–0.542)** |
| | DeepECG-SSL | 0.311 (0.181–0.446) | **0.981 (0.966–0.991)** | **0.351 (0.225–0.500)** | 0.999 (0.999–1.000) | 0.388 (0.253–0.500) | 0.390 (0.253–0.500) |
| | CardioLoom | 0.320 (0.180–0.478) | 0.979 (0.959–0.992) | 0.333 (0.192–0.520) | **1.000 (0.999–1.000)** | 0.396 (0.243–0.532) | 0.402 (0.249–0.542) |
| AF | Scratch CNN–Transformer | 0.979 (0.977–0.981) | 0.996 (0.996–0.997) | 0.934 (0.929–0.939) | 0.991 (0.990–0.992) | 0.934 (0.930–0.938) | 0.925 (0.920–0.929) |
| | ECG-FM | 0.992 (0.991–0.993) | **0.998 (0.998–0.999)** | 0.963 (0.959–0.966) | **0.994 (0.993–0.995)** | 0.961 (0.958–0.964) | 0.955 (0.952–0.958) |
| | DeepECG-SSL | **0.993 (0.992–0.994)** | **0.998 (0.998–0.999)** | 0.968 (0.964–0.971) | **0.994 (0.993–0.995)** | 0.963 (0.960–0.966) | 0.958 (0.954–0.961) |
| | CardioLoom | **0.993 (0.992–0.994)** | **0.998 (0.998–0.999)** | **0.969 (0.965–0.972)** | **0.994 (0.993–0.995)** | **0.964 (0.961–0.967)** | **0.959 (0.955–0.962)** |

| Disease | Model | AUPRC | AUROC | Sensitivity | Specificity | F1 Score | MCC |
|---|---|---|---|---|---|---|---|
| **Tier 3: Rare Diseases** | | | | | | | |
| CA | Scratch CNN–Transformer | 0.040 (0.008–0.116) | 0.917 (0.863–0.950) | 0.078 (0.000–0.220) | 0.999 (0.999–1.000) | 0.097 (0.000–0.203) | 0.100 (-0.001–0.207) |
| | ECG-FM | 0.105 (0.003–0.342) | 0.896 (0.825–0.967) | 0.174 (0.000–0.429) | **1.000 (0.999–1.000)** | **0.216 (0.000–0.436)** | **0.222 (-0.001–0.439)** |
| | DeepECG-SSL | **0.108 (0.005–0.291)** | 0.954 (0.911–0.975) | 0.209 (0.011–0.495) | 0.998 (0.998–0.999) | 0.148 (0.008–0.302) | 0.154 (0.007–0.319) |
| | CardioLoom | 0.095 (0.011–0.239) | **0.965 (0.943–0.986)** | **0.217 (0.050–0.471)** | 0.999 (0.999–0.999) | 0.192 (0.040–0.342) | 0.193 (0.041–0.350) |
| ARVC | Scratch CNN–Transformer | 0.001 (0.000–0.007) | 0.886 (0.778–0.998) | **0.250 (0.000–1.000)** | 0.995 (0.994–0.995) | **0.003 (0.000–0.011)** | **0.020 (-0.001–0.060)** |
| | ECG-FM | 0.000 (0.000–0.001) | 0.722 (0.256–0.914) | 0.000 (0.000–0.000) | 0.994 (0.993–0.995) | 0.000 (0.000–0.000) | 0.000 (-0.001–0.000) |
| | DeepECG-SSL | 0.001 (0.000–0.003) | 0.793 (0.212–0.988) | 0.000 (0.000–0.000) | 0.995 (0.994–0.995) | 0.000 (0.000–0.000) | 0.000 (-0.001–0.000) |
| | CardioLoom | **0.002 (0.000–0.007)** | **0.991 (0.981–0.995)** | 0.000 (0.000–0.000) | **1.000 (1.000–1.000)** | 0.000 (0.000–0.000) | 0.000 (0.000–0.000) |
| TC | Scratch CNN–Transformer | 0.000 (0.000–0.000) | 0.638 (0.547–0.661) | 0.000 (0.000–0.000) | **1.000 (1.000–1.000)** | 0.000 (0.000–0.000) | 0.000 (0.000–0.000) |
| | ECG-FM | 0.001 (0.000–0.004) | 0.833 (0.781–0.959) | 0.071 (0.000–0.400) | 0.999 (0.998–0.999) | **0.013 (0.000–0.043)** | **0.022 (-0.001–0.079)** |
| | DeepECG-SSL | 0.000 (0.000–0.002) | 0.663 (0.532–0.977) | 0.000 (0.000–0.000) | **1.000 (0.999–1.000)** | 0.000 (0.000–0.000) | 0.000 (0.000–0.000) |
| | CardioLoom | **0.002 (0.000–0.009)** | **0.842 (0.770–0.911)** | **0.214 (0.000–0.607)** | 0.978 (0.976–0.979) | 0.002 (0.000–0.007) | 0.015 (-0.002–0.036) |
| AC | Scratch CNN–Transformer | 0.000 (0.000–0.001) | 0.827 (0.538–0.971) | **0.667 (0.000–1.000)** | 0.943 (0.940–0.945) | 0.001 (0.000–0.004) | **0.020 (-0.001–0.037)** |
| | ECG-FM | 0.000 (0.000–0.000) | 0.783 (0.726–0.958) | 0.000 (0.000–0.000) | 0.994 (0.994–0.995) | 0.000 (0.000–0.000) | -0.001 (-0.001–0.000) |
| | DeepECG-SSL | 0.000 (0.000–0.001) | 0.792 (0.630–0.985) | 0.000 (0.000–0.000) | **0.997 (0.997–0.997)** | 0.000 (0.000–0.000) | 0.000 (-0.001–0.000) |
| | CardioLoom | **0.001 (0.000–0.006)** | **0.921 (0.764–0.996)** | 0.167 (0.000–0.500) | 0.994 (0.993–0.995) | **0.003 (0.000–0.010)** | 0.016 (-0.001–0.046) |

| Disease | Model | AUPRC | AUROC | Sensitivity | Specificity | F1 Score | MCC |
|---|---|---|---|---|---|---|---|
| NCVM | Scratch CNN–Transformer | **0.000 (0.000–0.002)** | **0.845 (0.626–0.933)** | 0.000 (0.000–0.000) | **0.994 (0.993–0.995)** | 0.000 (0.000–0.000) | -0.001 (-0.002–0.000) |
| | ECG-FM | **0.000 (0.000–0.001)** | 0.794 (0.632–0.969) | 0.062 (0.000–1.000) | 0.961 (0.959–0.963) | 0.000 (0.000–0.001) | 0.001 (-0.004–0.022) |
| | DeepECG-SSL | **0.000 (0.000–0.001)** | 0.757 (0.724–0.849) | **0.375 (0.000–1.000)** | 0.820 (0.816–0.825) | **0.001 (0.000–0.002)** | 0.006 (-0.002–0.011) |
| | CardioLoom | **0.000 (0.000–0.002)** | 0.745 (0.435–0.995) | 0.188 (0.000–1.000) | 0.956 (0.953–0.958) | **0.001 (0.000–0.003)** | **0.008 (-0.001–0.021)** |

Values are estimates (95% cluster-bootstrap CIs). NE indicates fewer than 950 valid replicates among 1000. Sensitivity, specificity, F1 score, and MCC use the Zhongshan validation-F1 threshold. Bold type marks the highest displayed estimate within each disease.

**Supplementary Table 19: Detailed task-specific performance comparison on the external Xiamen cohort.**

| Disease | Model | AUPRC | AUROC | Sensitivity | Specificity | F1 Score | MCC |
|---|---|---|---|---|---|---|---|
| OMI | Scratch CNN–Transformer | 0.230 (0.083–0.419) | 0.994 (0.989–0.998) | 0.723 (0.469–0.915) | **0.995 (0.993–0.996)** | 0.223 (0.110–0.355) | 0.307 (0.177–0.430) |
| | ECG-FM | **0.297 (0.121–0.502)** | **0.995 (0.991–0.999)** | **0.809 (0.611–0.969)** | **0.995 (0.993–0.996)** | **0.235 (0.123–0.365)** | **0.332 (0.212–0.455)** |
| | DeepECG-SSL | 0.179 (0.076–0.331) | **0.995 (0.991–0.998)** | **0.809 (0.611–0.969)** | 0.994 (0.992–0.995) | 0.206 (0.105–0.323) | 0.307 (0.196–0.422) |
| | CardioLoom | 0.198 (0.082–0.376) | **0.995 (0.991–0.998)** | **0.809 (0.654–0.943)** | 0.994 (0.993–0.995) | 0.226 (0.118–0.345) | 0.324 (0.217–0.428) |
| STEMI | Scratch CNN–Transformer | **0.071 (0.048–0.121)** | **0.879 (0.864–0.893)** | 0.039 (0.009–0.090) | **0.999 (0.999–0.999)** | 0.070 (0.017–0.152) | 0.106 (0.027–0.215) |
| | ECG-FM | 0.056 (0.035–0.099) | 0.832 (0.816–0.850) | 0.044 (0.012–0.092) | **0.999 (0.998–0.999)** | 0.075 (0.021–0.152) | 0.101 (0.027–0.203) |
| | DeepECG-SSL | 0.057 (0.036–0.103) | 0.858 (0.844–0.872) | **0.046 (0.013–0.096)** | **0.999 (0.998–0.999)** | **0.078 (0.023–0.156)** | 0.103 (0.030–0.202) |
| | CardioLoom | 0.049 (0.027–0.097) | 0.773 (0.750–0.798) | 0.039 (0.010–0.082) | **0.999 (0.999–1.000)** | 0.071 (0.018–0.142) | **0.118 (0.033–0.220)** |
| NSR | Scratch CNN–Transformer | 0.994 (0.994–0.995) | **0.984 (0.982–0.985)** | **0.991 (0.990–0.992)** | 0.791 (0.781–0.801) | 0.966 (0.964–0.968) | 0.842 (0.834–0.849) |
| | ECG-FM | 0.993 (0.992–0.994) | 0.979 (0.978–0.981) | 0.990 (0.989–0.991) | 0.793 (0.783–0.803) | 0.966 (0.964–0.967) | 0.840 (0.832–0.847) |
| | DeepECG-SSL | **0.995 (0.994–0.995)** | **0.984 (0.982–0.985)** | **0.991 (0.990–0.992)** | 0.785 (0.775–0.795) | 0.965 (0.963–0.966) | 0.836 (0.829–0.843) |
| | CardioLoom | 0.994 (0.994–0.995) | 0.983 (0.982–0.985) | 0.988 (0.986–0.989) | **0.810 (0.801–0.819)** | **0.967 (0.965–0.968)** | **0.846 (0.839–0.853)** |
| SBrad | Scratch CNN–Transformer | 0.981 (0.978–0.985) | **0.998 (0.997–0.998)** | 0.953 (0.947–0.959) | **0.993 (0.992–0.993)** | **0.950 (0.946–0.954)** | **0.943 (0.938–0.948)** |
| | ECG-FM | 0.979 (0.975–0.982) | 0.997 (0.997–0.998) | 0.957 (0.952–0.962) | 0.989 (0.988–0.990) | 0.940 (0.936–0.945) | 0.932 (0.927–0.937) |
| | DeepECG-SSL | 0.978 (0.975–0.983) | 0.997 (0.997–0.998) | 0.957 (0.952–0.963) | 0.992 (0.991–0.993) | 0.948 (0.944–0.953) | 0.941 (0.937–0.946) |
| | CardioLoom | **0.984 (0.982–0.987)** | **0.998 (0.998–0.998)** | **0.974 (0.970–0.979)** | 0.989 (0.988–0.990) | **0.950 (0.946–0.954)** | **0.943 (0.938–0.948)** |
| STach | Scratch CNN–Transformer | 0.975 (0.966–0.983) | **0.999 (0.999–1.000)** | 0.945 (0.931–0.957) | **0.999 (0.998–0.999)** | 0.950 (0.940–0.958) | 0.948 (0.938–0.957) |
| | ECG-FM | **0.979 (0.972–0.986)** | **0.999 (0.999–1.000)** | **0.971 (0.962–0.979)** | 0.998 (0.997–0.998) | 0.948 (0.939–0.957) | 0.947 (0.937–0.956) |
| | DeepECG-SSL | 0.978 (0.970–0.986) | **0.999 (0.999–1.000)** | 0.950 (0.938–0.961) | 0.998 (0.998–0.999) | 0.951 (0.943–0.960) | 0.950 (0.941–0.958) |
| | CardioLoom | 0.978 (0.971–0.986) | **0.999 (0.999–1.000)** | 0.964 (0.954–0.974) | 0.998 (0.998–0.999) | **0.954 (0.946–0.963)** | **0.953 (0.944–0.962)** |

| Disease | Model | AUPRC | AUROC | Sensitivity | Specificity | F1 Score | MCC |
|---|---|---|---|---|---|---|---|
| VPB | Scratch CNN–Transformer | 0.823 (0.793–0.851) | 0.995 (0.994–0.997) | 0.943 (0.924–0.960) | 0.992 (0.991–0.993) | 0.854 (0.839–0.871) | 0.853 (0.838–0.870) |
| | ECG-FM | 0.815 (0.789–0.840) | 0.995 (0.994–0.996) | 0.952 (0.935–0.967) | **0.993 (0.992–0.993)** | 0.864 (0.848–0.880) | 0.863 (0.848–0.879) |
| | DeepECG-SSL | 0.813 (0.782–0.843) | 0.995 (0.994–0.996) | 0.948 (0.927–0.966) | **0.993 (0.992–0.993)** | 0.863 (0.846–0.878) | 0.862 (0.846–0.877) |
| | CardioLoom | **0.848 (0.821–0.874)** | **0.996 (0.994–0.997)** | **0.956 (0.937–0.972)** | **0.993 (0.992–0.993)** | **0.866 (0.851–0.882)** | **0.866 (0.851–0.881)** |
| APB | Scratch CNN–Transformer | 0.801 (0.773–0.828) | 0.993 (0.991–0.995) | 0.857 (0.836–0.879) | 0.990 (0.989–0.991) | 0.749 (0.728–0.771) | 0.749 (0.729–0.770) |
| | ECG-FM | 0.755 (0.722–0.788) | 0.992 (0.990–0.994) | 0.886 (0.867–0.906) | **0.992 (0.991–0.993)** | 0.788 (0.769–0.808) | 0.788 (0.769–0.808) |
| | DeepECG-SSL | 0.804 (0.775–0.834) | 0.993 (0.991–0.995) | 0.890 (0.871–0.908) | 0.991 (0.990–0.992) | 0.784 (0.762–0.805) | 0.784 (0.764–0.805) |
| | CardioLoom | **0.828 (0.799–0.854)** | **0.995 (0.993–0.996)** | **0.903 (0.885–0.922)** | **0.992 (0.991–0.993)** | **0.797 (0.776–0.818)** | **0.798 (0.777–0.818)** |
| SArr | Scratch CNN–Transformer | 0.763 (0.741–0.785) | 0.982 (0.980–0.985) | 0.701 (0.678–0.722) | 0.988 (0.987–0.989) | 0.692 (0.673–0.710) | 0.681 (0.662–0.699) |
| | ECG-FM | 0.678 (0.652–0.703) | 0.971 (0.967–0.974) | 0.661 (0.636–0.685) | 0.986 (0.985–0.987) | 0.641 (0.621–0.661) | 0.629 (0.607–0.649) |
| | DeepECG-SSL | 0.760 (0.737–0.783) | 0.984 (0.981–0.986) | 0.708 (0.685–0.731) | 0.990 (0.989–0.991) | 0.708 (0.689–0.727) | 0.698 (0.678–0.717) |
| | CardioLoom | **0.780 (0.759–0.800)** | **0.986 (0.984–0.988)** | **0.716 (0.695–0.738)** | **0.991 (0.990–0.992)** | **0.725 (0.707–0.744)** | **0.716 (0.697–0.735)** |
| AF | Scratch CNN–Transformer | 0.917 (0.889–0.942) | **0.999 (0.997–0.999)** | 0.920 (0.892–0.947) | **0.998 (0.998–0.998)** | 0.867 (0.842–0.891) | 0.867 (0.842–0.891) |
| | ECG-FM | 0.922 (0.891–0.949) | **0.999 (0.998–1.000)** | 0.944 (0.920–0.965) | **0.998 (0.998–0.999)** | **0.886 (0.863–0.907)** | **0.886 (0.863–0.908)** |
| | DeepECG-SSL | **0.942 (0.922–0.960)** | **0.999 (0.998–1.000)** | 0.948 (0.925–0.970) | **0.998 (0.998–0.999)** | 0.883 (0.860–0.905) | 0.884 (0.861–0.906) |
| | CardioLoom | 0.933 (0.902–0.960) | **0.999 (0.999–1.000)** | **0.970 (0.951–0.986)** | **0.998 (0.997–0.998)** | 0.877 (0.852–0.902) | 0.880 (0.856–0.903) |
| AFL | Scratch CNN–Transformer | 0.766 (0.662–0.853) | 0.999 (0.998–0.999) | 0.754 (0.656–0.840) | **0.999 (0.999–0.999)** | 0.708 (0.621–0.780) | 0.708 (0.622–0.780) |
| | ECG-FM | 0.814 (0.712–0.902) | **1.000 (0.999–1.000)** | **0.825 (0.742–0.897)** | **0.999 (0.999–1.000)** | **0.783 (0.705–0.847)** | **0.784 (0.707–0.847)** |
| | DeepECG-SSL | 0.810 (0.719–0.890) | 0.999 (0.999–1.000) | 0.772 (0.673–0.862) | **0.999 (0.999–1.000)** | 0.752 (0.672–0.821) | 0.752 (0.673–0.820) |
| | CardioLoom | **0.834 (0.751–0.901)** | 0.999 (0.999–1.000) | 0.789 (0.698–0.868) | **0.999 (0.999–1.000)** | 0.759 (0.680–0.825) | 0.759 (0.683–0.827) |

| Disease | Model | AUPRC | AUROC | Sensitivity | Specificity | F1 Score | MCC |
|---|---|---|---|---|---|---|---|
| JPB | Scratch CNN–Transformer | 0.707 (0.643–0.766) | 0.996 (0.993–0.997) | 0.568 (0.502–0.629) | **0.999 (0.999–0.999)** | 0.659 (0.600–0.709) | 0.666 (0.610–0.715) |
| | ECG-FM | 0.751 (0.680–0.812) | 0.996 (0.995–0.998) | **0.656 (0.596–0.714)** | **0.999 (0.999–0.999)** | **0.715 (0.664–0.764)** | 0.716 (0.666–0.764) |
| | DeepECG-SSL | 0.759 (0.702–0.810) | 0.996 (0.994–0.997) | 0.592 (0.531–0.658) | **0.999 (0.999–1.000)** | 0.692 (0.638–0.742) | 0.700 (0.649–0.747) |
| | CardioLoom | **0.779 (0.726–0.829)** | **0.997 (0.995–0.998)** | 0.616 (0.557–0.677) | **0.999 (0.999–1.000)** | 0.710 (0.660–0.757) | **0.717 (0.669–0.763)** |
| ATach | Scratch CNN–Transformer | 0.388 (0.310–0.479) | 0.988 (0.982–0.993) | 0.476 (0.380–0.571) | 0.998 (0.998–0.998) | 0.457 (0.379–0.529) | 0.455 (0.377–0.530) |
| | ECG-FM | 0.501 (0.415–0.591) | 0.993 (0.991–0.996) | 0.524 (0.433–0.613) | **0.999 (0.998–0.999)** | 0.543 (0.464–0.615) | 0.542 (0.462–0.614) |
| | DeepECG-SSL | **0.584 (0.496–0.668)** | 0.995 (0.993–0.997) | **0.593 (0.511–0.669)** | 0.998 (0.998–0.999) | 0.566 (0.498–0.628) | 0.565 (0.498–0.627) |
| | CardioLoom | 0.567 (0.482–0.661) | **0.996 (0.994–0.997)** | 0.538 (0.445–0.623) | **0.999 (0.999–0.999)** | **0.580 (0.502–0.651)** | **0.580 (0.503–0.652)** |
| JTach | Scratch CNN–Transformer | 0.438 (0.337–0.559) | 0.986 (0.977–0.993) | 0.281 (0.186–0.395) | **1.000 (0.999–1.000)** | 0.398 (0.286–0.512) | 0.436 (0.333–0.541) |
| | ECG-FM | 0.524 (0.394–0.645) | 0.981 (0.965–0.992) | 0.272 (0.193–0.368) | **1.000 (1.000–1.000)** | 0.390 (0.294–0.485) | 0.432 (0.340–0.519) |
| | DeepECG-SSL | 0.500 (0.364–0.640) | 0.973 (0.951–0.988) | 0.360 (0.233–0.474) | **1.000 (0.999–1.000)** | 0.474 (0.328–0.590) | 0.499 (0.360–0.609) |
| | CardioLoom | **0.551 (0.410–0.682)** | **0.991 (0.984–0.995)** | **0.439 (0.309–0.561)** | 0.999 (0.999–1.000) | **0.529 (0.392–0.643)** | **0.540 (0.408–0.652)** |
| JEB | Scratch CNN–Transformer | **0.054 (0.013–0.139)** | **0.989 (0.980–0.995)** | 0.188 (0.000–0.429) | **1.000 (0.999–1.000)** | 0.154 (0.000–0.343) | 0.156 (0.000–0.353) |
| | ECG-FM | 0.036 (0.007–0.115) | 0.978 (0.955–0.993) | 0.188 (0.000–0.429) | 0.999 (0.999–1.000) | 0.122 (0.000–0.290) | 0.130 (-0.001–0.299) |
| | DeepECG-SSL | 0.035 (0.004–0.128) | 0.960 (0.923–0.984) | 0.188 (0.000–0.429) | **1.000 (0.999–1.000)** | **0.162 (0.000–0.364)** | 0.163 (0.000–0.370) |
| | CardioLoom | 0.049 (0.010–0.149) | 0.979 (0.955–0.994) | **0.250 (0.000–0.500)** | 0.999 (0.999–1.000) | 0.151 (0.000–0.314) | **0.164 (0.000–0.340)** |
| SVT | Scratch CNN–Transformer | **0.929 (0.801–1.000)** | **1.000 (1.000–1.000)** | 0.750 (0.533–0.950) | **1.000 (1.000–1.000)** | 0.828 (0.666–0.960) | 0.832 (0.668–0.961) |
| | ECG-FM | 0.911 (0.770–1.000) | **1.000 (1.000–1.000)** | 0.812 (0.615–1.000) | **1.000 (1.000–1.000)** | **0.897 (0.762–1.000)** | **0.901 (0.784–1.000)** |
| | DeepECG-SSL | 0.871 (0.691–1.000) | 0.999 (0.999–1.000) | **0.875 (0.692–1.000)** | **1.000 (1.000–1.000)** | 0.848 (0.696–0.966) | 0.849 (0.703–0.966) |
| | CardioLoom | 0.870 (0.699–0.991) | **1.000 (0.999–1.000)** | 0.750 (0.533–0.950) | **1.000 (1.000–1.000)** | 0.828 (0.636–0.957) | 0.832 (0.653–0.957) |

| Disease | Model | AUPRC | AUROC | Sensitivity | Specificity | F1 Score | MCC |
|---|---|---|---|---|---|---|---|
| VEB | Scratch CNN–Transformer | 0.713 (0.458–0.916) | **0.997 (0.992–1.000)** | **0.600 (0.350–0.857)** | **1.000 (1.000–1.000)** | **0.750 (0.518–0.923)** | **0.775 (0.591–0.926)** |
| | ECG-FM | 0.547 (0.270–0.789) | 0.992 (0.977–1.000) | 0.533 (0.273–0.800) | **1.000 (1.000–1.000)** | 0.444 (0.205–0.650) | 0.451 (0.218–0.666) |
| | DeepECG-SSL | **0.725 (0.493–0.907)** | 0.991 (0.971–1.000) | **0.600 (0.353–0.846)** | **1.000 (1.000–1.000)** | 0.600 (0.375–0.815) | 0.600 (0.378–0.816) |
| | CardioLoom | 0.701 (0.460–0.903) | 0.988 (0.961–1.000) | **0.600 (0.353–0.846)** | **1.000 (1.000–1.000)** | 0.545 (0.320–0.750) | 0.548 (0.333–0.754) |
| VTach | Scratch CNN–Transformer | 0.274 (0.078–0.625) | 0.999 (0.998–1.000) | 0.455 (0.143–0.750) | **1.000 (1.000–1.000)** | 0.476 (0.166–0.710) | 0.477 (0.167–0.714) |
| | ECG-FM | **0.624 (0.350–0.946)** | **1.000 (1.000–1.000)** | **0.727 (0.444–1.000)** | **1.000 (1.000–1.000)** | **0.667 (0.400–0.857)** | **0.669 (0.408–0.866)** |
| | DeepECG-SSL | 0.449 (0.165–0.823) | 0.997 (0.993–1.000) | 0.455 (0.154–0.778) | **1.000 (1.000–1.000)** | 0.556 (0.222–0.800) | 0.570 (0.252–0.804) |
| | CardioLoom | 0.560 (0.288–0.923) | **1.000 (1.000–1.000)** | 0.636 (0.300–0.917) | **1.000 (1.000–1.000)** | **0.667 (0.353–0.883)** | 0.667 (0.364–0.888) |
| RBBB | Scratch CNN–Transformer | 0.968 (0.957–0.978) | **0.999 (0.999–0.999)** | **0.935 (0.917–0.952)** | 0.997 (0.996–0.998) | 0.919 (0.903–0.934) | 0.917 (0.900–0.932) |
| | ECG-FM | **0.974 (0.965–0.981)** | **0.999 (0.999–0.999)** | 0.912 (0.892–0.931) | **0.998 (0.997–0.998)** | **0.921 (0.906–0.936)** | **0.919 (0.903–0.934)** |
| | DeepECG-SSL | 0.969 (0.958–0.978) | 0.998 (0.997–0.999) | 0.911 (0.892–0.930) | 0.997 (0.997–0.998) | 0.909 (0.893–0.924) | 0.906 (0.890–0.922) |
| | CardioLoom | 0.965 (0.954–0.974) | **0.999 (0.998–0.999)** | 0.904 (0.883–0.925) | **0.998 (0.997–0.998)** | 0.912 (0.895–0.927) | 0.909 (0.892–0.925) |
| 1° AVB | Scratch CNN–Transformer | 0.732 (0.693–0.775) | 0.991 (0.988–0.993) | 0.707 (0.673–0.742) | 0.994 (0.993–0.995) | 0.712 (0.684–0.741) | 0.706 (0.678–0.735) |
| | ECG-FM | **0.767 (0.730–0.803)** | **0.993 (0.990–0.995)** | 0.686 (0.647–0.724) | **0.995 (0.994–0.996)** | 0.723 (0.694–0.751) | 0.718 (0.689–0.747) |
| | DeepECG-SSL | 0.763 (0.727–0.799) | **0.993 (0.992–0.995)** | **0.762 (0.731–0.797)** | 0.994 (0.993–0.995) | **0.753 (0.727–0.780)** | **0.748 (0.721–0.775)** |
| | CardioLoom | 0.765 (0.730–0.801) | **0.993 (0.991–0.995)** | 0.706 (0.669–0.741) | **0.995 (0.994–0.996)** | 0.735 (0.705–0.763) | 0.730 (0.700–0.758) |
| QT Prolong | Scratch CNN–Transformer | 0.354 (0.312–0.397) | 0.955 (0.949–0.960) | 0.177 (0.146–0.210) | 0.998 (0.997–0.998) | 0.271 (0.230–0.315) | 0.315 (0.273–0.357) |
| | ECG-FM | **0.461 (0.415–0.508)** | **0.969 (0.965–0.974)** | **0.263 (0.225–0.307)** | 0.997 (0.997–0.998) | **0.373 (0.327–0.422)** | **0.404 (0.360–0.450)** |
| | DeepECG-SSL | 0.435 (0.393–0.477) | 0.965 (0.959–0.970) | 0.209 (0.178–0.246) | 0.998 (0.997–0.998) | 0.310 (0.270–0.354) | 0.349 (0.309–0.390) |
| | CardioLoom | 0.454 (0.410–0.501) | **0.969 (0.964–0.974)** | 0.183 (0.156–0.217) | **0.999 (0.998–0.999)** | 0.290 (0.253–0.332) | 0.351 (0.313–0.391) |

| Disease | Model | AUPRC | AUROC | Sensitivity | Specificity | F1 Score | MCC |
|---|---|---|---|---|---|---|---|
| ER | Scratch CNN–Transformer | 0.539 (0.481–0.595) | 0.988 (0.986–0.991) | 0.486 (0.433–0.542) | 0.997 (0.996–0.997) | 0.517 (0.469–0.567) | 0.514 (0.466–0.565) |
| | ECG-FM | 0.590 (0.530–0.648) | 0.991 (0.989–0.993) | 0.626 (0.570–0.678) | 0.995 (0.994–0.996) | **0.566 (0.521–0.610)** | **0.565 (0.521–0.609)** |
| | DeepECG-SSL | **0.605 (0.549–0.660)** | **0.992 (0.990–0.994)** | **0.631 (0.579–0.683)** | 0.995 (0.994–0.995) | 0.555 (0.511–0.599) | 0.555 (0.511–0.600) |
| | CardioLoom | 0.583 (0.523–0.637) | 0.990 (0.988–0.992) | 0.451 (0.398–0.506) | **0.998 (0.998–0.999)** | 0.541 (0.491–0.589) | 0.549 (0.499–0.597) |
| LAFB | Scratch CNN–Transformer | **0.458 (0.330–0.554)** | 0.995 (0.992–0.997) | 0.582 (0.446–0.703) | **0.998 (0.998–0.999)** | 0.479 (0.372–0.558) | 0.485 (0.378–0.568) |
| | ECG-FM | 0.403 (0.309–0.505) | 0.995 (0.993–0.997) | **0.684 (0.562–0.808)** | 0.997 (0.996–0.998) | 0.467 (0.375–0.550) | 0.491 (0.405–0.575) |
| | DeepECG-SSL | 0.418 (0.332–0.512) | **0.997 (0.996–0.998)** | 0.663 (0.537–0.773) | **0.998 (0.997–0.998)** | **0.514 (0.417–0.600)** | **0.526 (0.426–0.611)** |
| | CardioLoom | 0.418 (0.319–0.517) | 0.995 (0.993–0.997) | 0.592 (0.482–0.710) | **0.998 (0.997–0.998)** | 0.443 (0.367–0.518) | 0.456 (0.380–0.532) |
| LBBB | Scratch CNN–Transformer | **0.987 (0.972–0.997)** | **1.000 (1.000–1.000)** | 0.950 (0.903–0.986) | **1.000 (1.000–1.000)** | 0.946 (0.911–0.974) | 0.946 (0.912–0.974) |
| | ECG-FM | 0.979 (0.955–0.995) | **1.000 (1.000–1.000)** | 0.950 (0.915–0.982) | **1.000 (1.000–1.000)** | 0.950 (0.918–0.976) | 0.949 (0.918–0.976) |
| | DeepECG-SSL | 0.972 (0.925–0.997) | **1.000 (1.000–1.000)** | **0.964 (0.924–0.993)** | **1.000 (1.000–1.000)** | 0.947 (0.907–0.977) | 0.947 (0.908–0.977) |
| | CardioLoom | **0.987 (0.972–0.997)** | **1.000 (1.000–1.000)** | 0.957 (0.914–0.992) | **1.000 (1.000–1.000)** | **0.953 (0.919–0.981)** | **0.953 (0.919–0.981)** |
| IVB | Scratch CNN–Transformer | **0.709 (0.553–0.842)** | 0.998 (0.996–0.999) | **0.667 (0.518–0.829)** | **1.000 (0.999–1.000)** | **0.655 (0.522–0.765)** | **0.655 (0.523–0.768)** |
| | ECG-FM | 0.642 (0.482–0.787) | **0.999 (0.998–1.000)** | 0.579 (0.391–0.795) | **1.000 (0.999–1.000)** | 0.589 (0.442–0.724) | 0.589 (0.446–0.724) |
| | DeepECG-SSL | 0.616 (0.468–0.754) | **0.999 (0.998–0.999)** | 0.509 (0.327–0.714) | **1.000 (0.999–1.000)** | 0.542 (0.390–0.676) | 0.543 (0.393–0.679) |
| | CardioLoom | 0.669 (0.528–0.801) | **0.999 (0.999–1.000)** | 0.649 (0.458–0.852) | 0.999 (0.999–1.000) | 0.607 (0.474–0.734) | 0.607 (0.475–0.739) |
| Short PR | Scratch CNN–Transformer | 0.142 (0.098–0.207) | 0.983 (0.974–0.991) | 0.287 (0.186–0.397) | 0.998 (0.997–0.998) | 0.229 (0.154–0.308) | 0.232 (0.156–0.312) |
| | ECG-FM | **0.230 (0.154–0.319)** | **0.993 (0.991–0.995)** | **0.438 (0.333–0.552)** | 0.997 (0.996–0.997) | 0.276 (0.204–0.349) | **0.295 (0.220–0.370)** |
| | DeepECG-SSL | 0.226 (0.146–0.319) | 0.991 (0.986–0.995) | 0.263 (0.167–0.358) | **0.999 (0.998–0.999)** | 0.256 (0.168–0.347) | 0.255 (0.167–0.346) |
| | CardioLoom | 0.210 (0.139–0.304) | 0.991 (0.988–0.994) | 0.338 (0.241–0.448) | 0.998 (0.998–0.999) | **0.284 (0.201–0.375)** | 0.286 (0.205–0.378) |

| Disease | Model | AUPRC | AUROC | Sensitivity | Specificity | F1 Score | MCC |
|---|---|---|---|---|---|---|---|
| VPE | Scratch CNN–Transformer | 0.693 (0.578–0.802) | 0.994 (0.990–0.997) | 0.573 (0.427–0.714) | **1.000 (1.000–1.000)** | 0.696 (0.569–0.800) | 0.713 (0.602–0.804) |
| | ECG-FM | 0.796 (0.678–0.893) | **0.998 (0.995–0.999)** | 0.707 (0.579–0.823) | **1.000 (1.000–1.000)** | 0.806 (0.708–0.881) | 0.813 (0.727–0.884) |
| | DeepECG-SSL | 0.821 (0.719–0.924) | **0.998 (0.995–1.000)** | 0.732 (0.602–0.845) | **1.000 (1.000–1.000)** | **0.816 (0.725–0.889)** | **0.822 (0.739–0.891)** |
| | CardioLoom | **0.845 (0.741–0.931)** | 0.995 (0.987–1.000) | **0.793 (0.663–0.897)** | **1.000 (1.000–1.000)** | 0.813 (0.722–0.881) | 0.812 (0.728–0.882) |
| 3° AVB | Scratch CNN–Transformer | **1.000 (NE)** | **1.000 (NE)** | **1.000 (NE)** | **1.000 (NE)** | 0.308 (NE) | 0.426 (NE) |
| | ECG-FM | **1.000 (NE)** | **1.000 (NE)** | **1.000 (NE)** | **1.000 (NE)** | 0.250 (NE) | 0.378 (NE) |
| | DeepECG-SSL | **1.000 (NE)** | **1.000 (NE)** | **1.000 (NE)** | **1.000 (NE)** | **0.444 (NE)** | **0.534 (NE)** |
| | CardioLoom | 0.417 (NE) | **1.000 (NE)** | **1.000 (NE)** | **1.000 (NE)** | 0.333 (NE) | 0.447 (NE) |
| 2° 1 Type AVB | Scratch CNN–Transformer | 0.565 (0.159–0.866) | 0.994 (0.978–1.000) | 0.545 (0.176–0.889) | **1.000 (1.000–1.000)** | 0.545 (0.182–0.800) | 0.545 (0.195–0.804) |
| | ECG-FM | 0.523 (0.196–0.826) | 0.999 (0.999–1.000) | 0.364 (0.100–0.715) | **1.000 (1.000–1.000)** | 0.400 (0.125–0.667) | 0.402 (0.126–0.680) |
| | DeepECG-SSL | **0.671 (0.233–0.965)** | 0.999 (0.997–1.000) | **0.818 (0.500–1.000)** | **1.000 (1.000–1.000)** | **0.667 (0.364–0.857)** | **0.678 (0.404–0.866)** |
| | CardioLoom | 0.617 (0.261–0.925) | **1.000 (0.999–1.000)** | 0.727 (0.333–1.000) | **1.000 (1.000–1.000)** | 0.640 (0.286–0.857) | 0.645 (0.316–0.866) |
| 2° 2 Type AVB | Scratch CNN–Transformer | 0.504 (0.042–1.000) | **1.000 (0.999–1.000)** | 0.400 (0.000–1.000) | **1.000 (1.000–1.000)** | 0.500 (0.000–0.800) | 0.516 (0.000–0.816) |
| | ECG-FM | 0.697 (0.163–1.000) | 0.993 (0.978–1.000) | **0.800 (0.333–1.000)** | **1.000 (1.000–1.000)** | 0.571 (0.222–0.825) | 0.596 (0.266–0.833) |
| | DeepECG-SSL | 0.618 (0.126–1.000) | 0.996 (0.988–1.000) | 0.600 (0.000–1.000) | **1.000 (1.000–1.000)** | 0.545 (0.000–0.833) | 0.548 (0.000–0.833) |
| | CardioLoom | **0.708 (0.250–1.000)** | 0.993 (0.977–1.000) | 0.600 (0.000–1.000) | **1.000 (1.000–1.000)** | **0.600 (0.000–0.889)** | **0.600 (0.000–0.894)** |
| LVH | Scratch CNN–Transformer | 0.583 (0.553–0.610) | 0.954 (0.949–0.958) | **0.547 (0.519–0.574)** | 0.981 (0.980–0.983) | **0.549 (0.524–0.570)** | 0.530 (0.505–0.552) |
| | ECG-FM | 0.585 (0.555–0.616) | 0.952 (0.947–0.956) | 0.482 (0.451–0.512) | 0.987 (0.986–0.988) | 0.538 (0.511–0.565) | 0.525 (0.499–0.553) |
| | DeepECG-SSL | **0.610 (0.579–0.639)** | **0.958 (0.954–0.962)** | 0.464 (0.433–0.494) | **0.990 (0.989–0.991)** | 0.544 (0.516–0.571) | **0.537 (0.509–0.564)** |
| | CardioLoom | 0.606 (0.577–0.634) | 0.957 (0.953–0.961) | 0.478 (0.447–0.508) | 0.988 (0.987–0.990) | 0.545 (0.517–0.570) | 0.534 (0.507–0.559) |

| Disease | Model | AUPRC | AUROC | Sensitivity | Specificity | F1 Score | MCC |
|---|---|---|---|---|---|---|---|
| LAH | Scratch CNN–Transformer | **0.557 (0.510–0.605)** | 0.966 (0.960–0.972) | 0.322 (0.283–0.362) | **0.996 (0.996–0.997)** | 0.443 (0.400–0.486) | 0.469 (0.427–0.508) |
| | ECG-FM | 0.553 (0.504–0.601) | 0.965 (0.960–0.970) | **0.381 (0.339–0.423)** | 0.994 (0.994–0.995) | **0.484 (0.440–0.525)** | **0.492 (0.450–0.532)** |
| | DeepECG-SSL | 0.535 (0.494–0.574) | 0.964 (0.958–0.969) | 0.296 (0.266–0.327) | **0.996 (0.995–0.997)** | 0.411 (0.377–0.442) | 0.437 (0.404–0.467) |
| | CardioLoom | 0.545 (0.503–0.589) | **0.967 (0.961–0.971)** | 0.337 (0.300–0.373) | **0.996 (0.995–0.996)** | 0.453 (0.411–0.490) | 0.472 (0.433–0.509) |
| RVH | Scratch CNN–Transformer | 0.556 (0.471–0.637) | 0.996 (0.993–0.997) | 0.719 (0.649–0.789) | 0.996 (0.995–0.997) | 0.544 (0.485–0.599) | 0.559 (0.504–0.610) |
| | ECG-FM | **0.664 (0.577–0.741)** | **0.998 (0.997–0.998)** | **0.802 (0.739–0.856)** | 0.997 (0.996–0.998) | **0.646 (0.582–0.701)** | **0.657 (0.598–0.707)** |
| | DeepECG-SSL | 0.597 (0.504–0.687) | 0.996 (0.994–0.997) | 0.547 (0.464–0.624) | **0.999 (0.998–0.999)** | 0.595 (0.523–0.660) | 0.596 (0.525–0.660) |
| | CardioLoom | 0.658 (0.573–0.736) | 0.997 (0.995–0.998) | 0.620 (0.535–0.695) | 0.998 (0.998–0.999) | 0.626 (0.559–0.689) | 0.625 (0.557–0.688) |
| VA | Scratch CNN–Transformer | 0.341 (0.115–0.553) | **0.999 (0.997–1.000)** | 0.632 (0.364–0.875) | **0.999 (0.999–1.000)** | 0.381 (0.176–0.563) | 0.415 (0.213–0.587) |
| | ECG-FM | 0.410 (0.149–0.765) | **0.999 (0.998–1.000)** | **0.737 (0.417–1.000)** | **0.999 (0.999–0.999)** | **0.394 (0.167–0.578)** | **0.445 (0.231–0.615)** |
| | DeepECG-SSL | **0.435 (0.176–0.728)** | **0.999 (0.998–1.000)** | **0.737 (0.407–1.000)** | **0.999 (0.999–0.999)** | 0.378 (0.169–0.545) | 0.433 (0.218–0.585) |
| | CardioLoom | 0.311 (0.101–0.546) | **0.999 (0.998–1.000)** | **0.737 (0.407–1.000)** | **0.999 (0.998–0.999)** | 0.350 (0.146–0.513) | 0.411 (0.203–0.564) |
| RAH | Scratch CNN–Transformer | 0.134 (0.073–0.207) | **0.991 (0.986–0.995)** | 0.224 (0.119–0.364) | 0.999 (0.999–0.999) | 0.200 (0.105–0.312) | 0.200 (0.106–0.312) |
| | ECG-FM | **0.239 (0.117–0.403)** | 0.984 (0.966–0.994) | 0.388 (0.231–0.551) | 0.999 (0.998–0.999) | **0.311 (0.192–0.417)** | **0.317 (0.196–0.423)** |
| | DeepECG-SSL | 0.230 (0.119–0.364) | **0.991 (0.984–0.996)** | **0.408 (0.256–0.544)** | 0.999 (0.998–0.999) | 0.308 (0.200–0.400) | **0.317 (0.209–0.410)** |
| | CardioLoom | 0.224 (0.118–0.358) | 0.990 (0.980–0.996) | 0.224 (0.118–0.343) | **1.000 (1.000–1.000)** | 0.297 (0.165–0.419) | 0.314 (0.174–0.433) |
| DC | Scratch CNN–Transformer | **0.001 (0.000–0.002)** | **0.827 (0.705–0.949)** | **0.000 (0.000–0.000)** | **1.000 (1.000–1.000)** | **0.000 (0.000–0.000)** | **0.000 (0.000–0.000)** |
| | ECG-FM | 0.000 (0.000–0.001) | 0.589 (0.204–0.957) | **0.000 (0.000–0.000)** | **1.000 (1.000–1.000)** | **0.000 (0.000–0.000)** | **0.000 (0.000–0.000)** |
| | DeepECG-SSL | **0.001 (0.000–0.002)** | 0.785 (0.598–0.964) | **0.000 (0.000–0.000)** | **1.000 (1.000–1.000)** | **0.000 (0.000–0.000)** | **0.000 (0.000–0.000)** |
| | CardioLoom | **0.001 (0.000–0.003)** | 0.813 (0.625–0.981) | **0.000 (0.000–0.000)** | **1.000 (1.000–1.000)** | **0.000 (0.000–0.000)** | **0.000 (0.000–0.000)** |

| Disease | Model | AUPRC | AUROC | Sensitivity | Specificity | F1 Score | MCC |
|---|---|---|---|---|---|---|---|
| VVI | Scratch CNN–Transformer | 0.806 (0.616–0.933) | 0.999 (0.997–1.000) | 0.659 (0.469–0.822) | **1.000 (1.000–1.000)** | 0.753 (0.586–0.873) | 0.761 (0.608–0.875) |
| | ECG-FM | 0.791 (0.556–0.905) | 0.999 (0.999–1.000) | 0.750 (0.580–0.868) | **1.000 (1.000–1.000)** | 0.795 (0.643–0.880) | 0.796 (0.646–0.881) |
| | DeepECG-SSL | 0.863 (0.710–0.959) | 0.999 (0.997–1.000) | **0.864 (0.739–0.956)** | **1.000 (1.000–1.000)** | **0.894 (0.795–0.954)** | **0.895 (0.797–0.954)** |
| | CardioLoom | **0.915 (0.800–0.977)** | **1.000 (0.999–1.000)** | 0.841 (0.704–0.936) | **1.000 (1.000–1.000)** | 0.892 (0.792–0.953) | 0.893 (0.795–0.954) |
| VAT | Scratch CNN–Transformer | 0.543 (0.238–0.873) | 0.992 (0.982–1.000) | 0.469 (0.182–0.829) | **1.000 (1.000–1.000)** | 0.556 (0.250–0.800) | 0.565 (0.280–0.801) |
| | ECG-FM | 0.798 (0.557–1.000) | 0.990 (0.975–1.000) | 0.656 (0.317–1.000) | **1.000 (1.000–1.000)** | 0.764 (0.471–0.966) | 0.774 (0.536–0.966) |
| | DeepECG-SSL | **0.825 (0.613–1.000)** | **0.995 (0.988–1.000)** | 0.750 (0.500–1.000) | **1.000 (1.000–1.000)** | 0.857 (0.667–1.000) | 0.866 (0.707–1.000) |
| | CardioLoom | 0.823 (0.586–1.000) | **0.995 (0.986–1.000)** | **0.781 (0.526–1.000)** | **1.000 (1.000–1.000)** | **0.877 (0.690–1.000)** | **0.884 (0.725–1.000)** |
| DDD | Scratch CNN–Transformer | 0.137 (NE) | **1.000 (NE)** | **1.000 (NE)** | **1.000 (NE)** | 0.400 (NE) | 0.500 (NE) |
| | ECG-FM | 0.332 (NE) | **1.000 (NE)** | **1.000 (NE)** | **1.000 (NE)** | **0.500 (NE)** | **0.577 (NE)** |
| | DeepECG-SSL | **0.519 (NE)** | **1.000 (NE)** | 0.667 (NE) | **1.000 (NE)** | 0.400 (NE) | 0.436 (NE) |
| | CardioLoom | 0.496 (NE) | **1.000 (NE)** | **1.000 (NE)** | **1.000 (NE)** | 0.462 (NE) | 0.548 (NE) |
| AAI | Scratch CNN–Transformer | 0.663 (0.227–0.872) | **0.996 (0.984–1.000)** | 0.440 (0.083–0.657) | **1.000 (1.000–1.000)** | 0.611 (0.154–0.793) | 0.663 (0.288–0.811) |
| | ECG-FM | 0.765 (0.337–0.928) | 0.995 (0.984–1.000) | 0.680 (0.250–0.889) | **1.000 (1.000–1.000)** | 0.791 (0.400–0.930) | 0.801 (0.433–0.931) |
| | DeepECG-SSL | 0.887 (0.627–0.983) | 0.976 (0.879–1.000) | **0.760 (0.454–0.940)** | **1.000 (1.000–1.000)** | **0.864 (0.625–0.969)** | **0.872 (0.674–0.970)** |
| | CardioLoom | **0.942 (0.748–1.000)** | 0.990 (0.948–1.000) | **0.760 (0.455–0.923)** | **1.000 (1.000–1.000)** | 0.844 (0.583–0.952) | 0.850 (0.612–0.953) |
| PSM | Scratch CNN–Transformer | 0.018 (NE) | 0.981 (NE) | 0.000 (NE) | **1.000 (NE)** | 0.000 (NE) | 0.000 (NE) |
| | ECG-FM | 0.029 (NE) | 0.908 (NE) | 0.000 (NE) | **1.000 (NE)** | 0.000 (NE) | 0.000 (NE) |
| | DeepECG-SSL | **0.338 (NE)** | **0.988 (NE)** | **0.333 (NE)** | **1.000 (NE)** | **0.500 (NE)** | **0.577 (NE)** |
| | CardioLoom | 0.089 (NE) | 0.979 (NE) | **0.333 (NE)** | **1.000 (NE)** | 0.400 (NE) | 0.408 (NE) |

| Disease | Model | AUPRC | AUROC | Sensitivity | Specificity | F1 Score | MCC |
|---|---|---|---|---|---|---|---|
| HK | Scratch CNN–Transformer | 0.059 (0.006–0.186) | 0.999 (0.996–1.000) | 0.400 (0.000–1.000) | **1.000 (0.999–1.000)** | 0.148 (0.000–0.348) | 0.191 (0.000–0.383) |
| | ECG-FM | 0.223 (0.058–0.609) | **1.000 (1.000–1.000)** | **0.800 (0.500–1.000)** | **1.000 (0.999–1.000)** | 0.333 (0.105–0.552) | 0.410 (0.208–0.589) |
| | DeepECG-SSL | 0.280 (0.074–0.796) | **1.000 (1.000–1.000)** | **0.800 (0.000–1.000)** | **1.000 (0.999–1.000)** | **0.348 (0.000–0.627)** | **0.422 (0.000–0.668)** |
| | CardioLoom | **0.491 (0.071–0.850)** | **1.000 (1.000–1.000)** | **0.800 (0.000–1.000)** | **1.000 (0.999–1.000)** | 0.333 (0.000–0.595) | 0.410 (0.000–0.645) |

Values are estimates (95% cluster-bootstrap CIs). NE indicates fewer than 950 valid replicates among 1000. Sensitivity, specificity, F1 score, and MCC use the Zhongshan validation-F1 threshold. Bold type marks the highest displayed estimate within each disease.

**Supplementary Table 20: Detailed task-specific performance comparison on the external MIMIC-IV-ECG cohort.**

| Disease | Model | AUPRC | AUROC | Sensitivity | Specificity | F1 Score | MCC |
|---|---|---|---|---|---|---|---|
| OMI | Scratch CNN–Transformer | 0.061 (0.058–0.064) | 0.792 (0.787–0.797) | **0.085 (0.077–0.093)** | 0.981 (0.980–0.982) | **0.082 (0.075–0.090)** | **0.064 (0.057–0.071)** |
| | ECG-FM | **0.064 (0.060–0.068)** | **0.801 (0.797–0.806)** | 0.050 (0.043–0.057) | **0.990 (0.989–0.990)** | 0.063 (0.055–0.072) | 0.052 (0.044–0.060) |
| | DeepECG-SSL | 0.058 (0.055–0.060) | 0.781 (0.776–0.786) | 0.080 (0.073–0.087) | 0.981 (0.980–0.981) | 0.077 (0.070–0.084) | 0.058 (0.051–0.065) |
| | CardioLoom | 0.061 (0.058–0.065) | 0.786 (0.780–0.791) | 0.072 (0.064–0.080) | 0.984 (0.983–0.985) | 0.076 (0.068–0.085) | 0.059 (0.052–0.068) |
| STEMI | Scratch CNN–Transformer | 0.536 (0.530–0.542) | 0.794 (0.792–0.796) | 0.019 (0.018–0.021) | 0.999 (0.999–0.999) | 0.038 (0.035–0.040) | 0.111 (0.106–0.115) |
| | ECG-FM | 0.558 (0.552–0.564) | 0.791 (0.788–0.793) | 0.018 (0.017–0.020) | **1.000 (0.999–1.000)** | 0.036 (0.034–0.039) | 0.113 (0.108–0.117) |
| | DeepECG-SSL | 0.544 (0.538–0.550) | 0.783 (0.781–0.786) | **0.026 (0.024–0.027)** | 0.999 (0.999–0.999) | **0.050 (0.046–0.053)** | **0.130 (0.125–0.135)** |
| | CardioLoom | **0.573 (0.567–0.579)** | **0.796 (0.793–0.798)** | 0.013 (0.012–0.014) | **1.000 (1.000–1.000)** | 0.025 (0.023–0.027) | 0.095 (0.090–0.099) |
| NSR | Scratch CNN–Transformer | 0.371 (0.367–0.376) | 0.795 (0.793–0.797) | 0.973 (0.972–0.974) | 0.448 (0.445–0.451) | **0.388 (0.385–0.392)** | **0.313 (0.310–0.315)** |
| | ECG-FM | 0.349 (0.345–0.354) | 0.777 (0.775–0.779) | 0.959 (0.957–0.960) | **0.454 (0.451–0.457)** | 0.386 (0.383–0.390) | 0.305 (0.302–0.308) |
| | DeepECG-SSL | **0.403 (0.399–0.408)** | **0.799 (0.797–0.801)** | 0.970 (0.969–0.971) | 0.447 (0.444–0.450) | 0.387 (0.383–0.390) | 0.309 (0.307–0.312) |
| | CardioLoom | 0.395 (0.390–0.399) | 0.794 (0.792–0.796) | **0.975 (0.974–0.976)** | 0.442 (0.439–0.445) | 0.386 (0.383–0.390) | 0.310 (0.308–0.313) |
| SBrad | Scratch CNN–Transformer | 0.893 (0.890–0.897) | **0.989 (0.989–0.989)** | 0.917 (0.915–0.919) | **0.979 (0.978–0.979)** | 0.889 (0.887–0.891) | 0.873 (0.871–0.875) |
| | ECG-FM | 0.883 (0.880–0.886) | 0.987 (0.986–0.987) | 0.868 (0.865–0.870) | **0.979 (0.978–0.980)** | 0.863 (0.861–0.865) | 0.843 (0.841–0.845) |
| | DeepECG-SSL | 0.898 (0.896–0.901) | 0.988 (0.987–0.988) | 0.931 (0.929–0.933) | 0.974 (0.974–0.975) | 0.884 (0.882–0.886) | 0.867 (0.865–0.869) |
| | CardioLoom | **0.905 (0.902–0.907)** | **0.989 (0.988–0.989)** | **0.944 (0.942–0.946)** | 0.974 (0.974–0.975) | **0.890 (0.888–0.892)** | **0.875 (0.873–0.877)** |
| STach | Scratch CNN–Transformer | **0.929 (0.926–0.931)** | 0.985 (0.985–0.986) | 0.838 (0.834–0.841) | **0.994 (0.994–0.994)** | 0.882 (0.880–0.884) | 0.873 (0.870–0.875) |
| | ECG-FM | 0.924 (0.921–0.926) | 0.985 (0.984–0.986) | **0.849 (0.845–0.852)** | 0.993 (0.993–0.993) | **0.884 (0.881–0.886)** | 0.874 (0.871–0.876) |
| | DeepECG-SSL | 0.927 (0.925–0.930) | **0.988 (0.988–0.988)** | 0.839 (0.836–0.843) | **0.994 (0.994–0.994)** | **0.884 (0.882–0.886)** | **0.875 (0.873–0.877)** |
| | CardioLoom | 0.920 (0.917–0.923) | 0.984 (0.984–0.985) | 0.842 (0.838–0.845) | 0.993 (0.993–0.994) | 0.882 (0.879–0.884) | 0.872 (0.869–0.874) |

| Disease | Model | AUPRC | AUROC | Sensitivity | Specificity | F1 Score | MCC |
|---|---|---|---|---|---|---|---|
| VPB | Scratch CNN–Transformer | 0.738 (0.731–0.745) | **0.966 (0.965–0.967)** | 0.743 (0.738–0.749) | **0.984 (0.983–0.984)** | 0.754 (0.749–0.758) | 0.736 (0.731–0.741) |
| | ECG-FM | 0.760 (0.753–0.766) | 0.963 (0.962–0.964) | 0.769 (0.764–0.775) | 0.983 (0.982–0.983) | 0.765 (0.761–0.770) | 0.749 (0.744–0.753) |
| | DeepECG-SSL | 0.748 (0.742–0.754) | 0.952 (0.951–0.954) | 0.776 (0.770–0.781) | 0.982 (0.981–0.982) | 0.765 (0.760–0.769) | 0.748 (0.743–0.752) |
| | CardioLoom | **0.765 (0.759–0.770)** | 0.961 (0.960–0.962) | **0.786 (0.781–0.792)** | 0.982 (0.981–0.982) | **0.770 (0.766–0.774)** | **0.753 (0.749–0.758)** |
| APB | Scratch CNN–Transformer | **0.622 (0.615–0.628)** | **0.946 (0.944–0.948)** | **0.804 (0.798–0.809)** | 0.965 (0.965–0.966) | **0.637 (0.633–0.642)** | **0.631 (0.627–0.636)** |
| | ECG-FM | 0.601 (0.595–0.608) | 0.931 (0.929–0.934) | 0.799 (0.794–0.804) | 0.963 (0.962–0.964) | 0.623 (0.618–0.628) | 0.618 (0.613–0.622) |
| | DeepECG-SSL | 0.606 (0.599–0.613) | 0.936 (0.934–0.938) | 0.792 (0.787–0.797) | 0.966 (0.966–0.967) | 0.636 (0.632–0.641) | 0.629 (0.624–0.633) |
| | CardioLoom | 0.611 (0.605–0.618) | 0.941 (0.939–0.943) | 0.774 (0.769–0.780) | **0.967 (0.966–0.968)** | 0.629 (0.624–0.633) | 0.620 (0.615–0.624) |
| SArr | Scratch CNN–Transformer | 0.269 (0.262–0.275) | 0.855 (0.852–0.858) | 0.259 (0.253–0.265) | **0.984 (0.984–0.985)** | 0.303 (0.297–0.309) | 0.287 (0.281–0.293) |
| | ECG-FM | 0.276 (0.270–0.282) | 0.865 (0.862–0.868) | 0.321 (0.315–0.328) | 0.980 (0.979–0.980) | 0.339 (0.333–0.344) | 0.317 (0.312–0.323) |
| | DeepECG-SSL | 0.313 (0.306–0.319) | 0.890 (0.888–0.893) | **0.371 (0.364–0.377)** | 0.978 (0.978–0.979) | **0.372 (0.367–0.379)** | **0.351 (0.345–0.357)** |
| | CardioLoom | **0.315 (0.308–0.321)** | **0.906 (0.904–0.908)** | 0.335 (0.329–0.342) | 0.981 (0.980–0.981) | 0.355 (0.349–0.362) | 0.335 (0.328–0.341) |
| AF | Scratch CNN–Transformer | 0.766 (0.760–0.772) | 0.961 (0.960–0.961) | 0.487 (0.481–0.494) | 0.988 (0.988–0.989) | 0.613 (0.607–0.620) | 0.606 (0.600–0.612) |
| | ECG-FM | 0.745 (0.738–0.752) | 0.953 (0.952–0.954) | 0.466 (0.459–0.473) | **0.991 (0.991–0.992)** | 0.604 (0.598–0.611) | 0.605 (0.600–0.612) |
| | DeepECG-SSL | **0.789 (0.783–0.795)** | 0.960 (0.959–0.961) | **0.506 (0.499–0.513)** | 0.989 (0.988–0.990) | **0.631 (0.625–0.638)** | **0.623 (0.617–0.629)** |
| | CardioLoom | 0.781 (0.774–0.787) | **0.962 (0.961–0.963)** | 0.497 (0.490–0.504) | 0.990 (0.989–0.990) | 0.626 (0.620–0.633) | 0.621 (0.615–0.627) |
| JPB | Scratch CNN–Transformer | **0.028 (0.026–0.030)** | **0.920 (0.914–0.924)** | 0.032 (0.024–0.040) | **0.998 (0.998–0.998)** | 0.035 (0.027–0.044) | 0.033 (0.025–0.042) |
| | ECG-FM | 0.021 (0.019–0.024) | 0.857 (0.849–0.865) | 0.060 (0.049–0.070) | 0.995 (0.995–0.996) | **0.046 (0.038–0.053)** | 0.043 (0.035–0.051) |
| | DeepECG-SSL | 0.021 (0.020–0.023) | 0.890 (0.883–0.896) | **0.062 (0.052–0.072)** | 0.995 (0.995–0.995) | **0.046 (0.038–0.053)** | **0.044 (0.036–0.051)** |
| | CardioLoom | 0.023 (0.022–0.025) | 0.903 (0.896–0.909) | 0.038 (0.030–0.046) | 0.996 (0.996–0.997) | 0.033 (0.027–0.041) | 0.030 (0.024–0.038) |

| Disease | Model | AUPRC | AUROC | Sensitivity | Specificity | F1 Score | MCC |
|---|---|---|---|---|---|---|---|
| AJR | Scratch CNN–Transformer | 0.017 (0.016–0.019) | 0.818 (0.810–0.827) | 0.054 (0.046–0.063) | 0.987 (0.986–0.987) | 0.025 (0.021–0.029) | 0.023 (0.018–0.027) |
| | ECG-FM | **0.024 (0.022–0.026)** | **0.845 (0.837–0.853)** | **0.077 (0.067–0.088)** | 0.990 (0.990–0.991) | **0.044 (0.038–0.050)** | **0.043 (0.036–0.050)** |
| | DeepECG-SSL | 0.018 (0.016–0.019) | 0.788 (0.779–0.798) | 0.048 (0.041–0.056) | **0.992 (0.991–0.992)** | 0.031 (0.026–0.036) | 0.027 (0.022–0.033) |
| | CardioLoom | 0.018 (0.016–0.020) | 0.814 (0.805–0.822) | 0.060 (0.051–0.071) | 0.991 (0.991–0.992) | 0.037 (0.031–0.043) | 0.034 (0.028–0.042) |
| ATach | Scratch CNN–Transformer | 0.009 (0.008–0.010) | 0.848 (0.841–0.855) | 0.022 (0.015–0.030) | **0.992 (0.991–0.992)** | 0.010 (0.007–0.014) | 0.008 (0.004–0.012) |
| | ECG-FM | 0.008 (0.007–0.009) | 0.790 (0.779–0.802) | **0.041 (0.030–0.051)** | 0.990 (0.990–0.990) | **0.016 (0.012–0.020)** | **0.015 (0.010–0.020)** |
| | DeepECG-SSL | **0.013 (0.012–0.014)** | **0.870 (0.862–0.877)** | 0.035 (0.025–0.044) | 0.991 (0.991–0.991) | 0.015 (0.011–0.019) | 0.014 (0.009–0.019) |
| | CardioLoom | 0.007 (0.007–0.008) | 0.779 (0.767–0.789) | 0.024 (0.017–0.032) | **0.992 (0.992–0.992)** | 0.012 (0.008–0.015) | 0.009 (0.005–0.014) |
| JTach | Scratch CNN–Transformer | 0.007 (0.005–0.010) | 0.947 (0.929–0.963) | 0.232 (0.162–0.300) | 0.996 (0.995–0.996) | 0.022 (0.014–0.030) | 0.051 (0.035–0.067) |
| | ECG-FM | **0.012 (0.008–0.017)** | 0.945 (0.925–0.964) | 0.238 (0.163–0.318) | **0.998 (0.997–0.998)** | **0.039 (0.025–0.055)** | **0.070 (0.047–0.096)** |
| | DeepECG-SSL | 0.008 (0.006–0.011) | **0.966 (0.953–0.978)** | 0.202 (0.137–0.267) | 0.996 (0.996–0.996) | 0.021 (0.013–0.028) | 0.046 (0.030–0.061) |
| | CardioLoom | 0.010 (0.006–0.014) | 0.937 (0.914–0.956) | **0.274 (0.192–0.349)** | 0.996 (0.996–0.996) | 0.029 (0.019–0.040) | 0.064 (0.044–0.084) |
| SVT | Scratch CNN–Transformer | 0.487 (0.446–0.527) | 0.991 (0.988–0.994) | 0.347 (0.314–0.384) | **1.000 (1.000–1.000)** | 0.450 (0.414–0.487) | 0.470 (0.436–0.505) |
| | ECG-FM | **0.505 (0.467–0.543)** | 0.985 (0.981–0.989) | 0.481 (0.445–0.521) | 0.999 (0.999–1.000) | 0.506 (0.473–0.539) | 0.506 (0.473–0.539) |
| | DeepECG-SSL | 0.457 (0.418–0.497) | **0.992 (0.989–0.993)** | **0.557 (0.518–0.597)** | 0.999 (0.999–0.999) | 0.489 (0.459–0.520) | 0.492 (0.461–0.523) |
| | CardioLoom | 0.503 (0.465–0.540) | 0.982 (0.978–0.986) | 0.504 (0.465–0.544) | 0.999 (0.999–1.000) | **0.519 (0.487–0.552)** | **0.519 (0.486–0.552)** |
| VEB | Scratch CNN–Transformer | 0.009 (0.007–0.011) | **0.929 (0.920–0.937)** | 0.034 (0.019–0.053) | **0.998 (0.998–0.999)** | 0.022 (0.012–0.033) | 0.022 (0.012–0.035) |
| | ECG-FM | **0.011 (0.009–0.017)** | 0.925 (0.915–0.933) | **0.051 (0.032–0.075)** | **0.998 (0.998–0.998)** | **0.028 (0.018–0.041)** | **0.030 (0.019–0.045)** |
| | DeepECG-SSL | **0.011 (0.009–0.014)** | 0.922 (0.912–0.931) | 0.048 (0.030–0.067) | **0.998 (0.998–0.998)** | **0.028 (0.017–0.039)** | **0.030 (0.018–0.042)** |
| | CardioLoom | 0.009 (0.008–0.011) | 0.923 (0.912–0.931) | 0.039 (0.023–0.057) | **0.998 (0.998–0.998)** | 0.023 (0.014–0.033) | 0.024 (0.014–0.036) |

| Disease | Model | AUPRC | AUROC | Sensitivity | Specificity | F1 Score | MCC |
|---|---|---|---|---|---|---|---|
| VTach | Scratch CNN–Transformer | 0.109 (0.087–0.136) | 0.967 (0.961–0.973) | 0.246 (0.212–0.284) | **0.999 (0.999–0.999)** | 0.191 (0.164–0.219) | 0.195 (0.168–0.224) |
| | ECG-FM | **0.128 (0.105–0.159)** | 0.963 (0.955–0.971) | **0.293 (0.252–0.334)** | **0.999 (0.999–0.999)** | **0.239 (0.207–0.273)** | **0.243 (0.210–0.276)** |
| | DeepECG-SSL | 0.127 (0.104–0.154) | **0.973 (0.967–0.978)** | 0.277 (0.239–0.316) | **0.999 (0.999–0.999)** | 0.222 (0.191–0.254) | 0.226 (0.194–0.258) |
| | CardioLoom | 0.122 (0.099–0.149) | 0.961 (0.953–0.969) | 0.281 (0.242–0.320) | **0.999 (0.999–0.999)** | **0.239 (0.206–0.271)** | 0.241 (0.208–0.273) |
| RBBB | Scratch CNN–Transformer | 0.894 (0.889–0.899) | 0.980 (0.978–0.981) | **0.876 (0.870–0.881)** | 0.969 (0.968–0.970) | 0.789 (0.783–0.796) | 0.773 (0.766–0.779) |
| | ECG-FM | 0.885 (0.880–0.890) | 0.980 (0.979–0.981) | 0.853 (0.846–0.858) | **0.975 (0.974–0.975)** | 0.800 (0.794–0.806) | 0.782 (0.776–0.788) |
| | DeepECG-SSL | 0.889 (0.884–0.894) | 0.978 (0.977–0.979) | **0.876 (0.870–0.881)** | 0.966 (0.965–0.967) | 0.779 (0.773–0.785) | 0.762 (0.756–0.768) |
| | CardioLoom | **0.901 (0.897–0.906)** | **0.984 (0.983–0.985)** | **0.876 (0.870–0.881)** | 0.973 (0.972–0.974) | **0.807 (0.801–0.813)** | **0.791 (0.785–0.797)** |
| 1° AVB | Scratch CNN–Transformer | 0.621 (0.613–0.628) | 0.947 (0.946–0.948) | 0.623 (0.616–0.630) | 0.968 (0.967–0.969) | 0.619 (0.613–0.625) | 0.588 (0.582–0.594) |
| | ECG-FM | 0.621 (0.614–0.629) | **0.955 (0.953–0.956)** | 0.598 (0.591–0.605) | **0.973 (0.972–0.973)** | 0.618 (0.612–0.624) | 0.588 (0.582–0.594) |
| | DeepECG-SSL | **0.624 (0.616–0.632)** | **0.955 (0.954–0.956)** | **0.691 (0.684–0.697)** | 0.964 (0.963–0.965) | **0.648 (0.643–0.654)** | **0.619 (0.613–0.625)** |
| | CardioLoom | 0.619 (0.611–0.627) | 0.954 (0.953–0.955) | 0.633 (0.625–0.640) | 0.968 (0.967–0.969) | 0.625 (0.619–0.630) | 0.594 (0.587–0.600) |
| QT Prolong | Scratch CNN–Transformer | 0.311 (0.304–0.320) | 0.875 (0.872–0.878) | 0.208 (0.202–0.215) | 0.986 (0.986–0.987) | 0.285 (0.277–0.292) | 0.282 (0.274–0.289) |
| | ECG-FM | 0.331 (0.322–0.339) | 0.868 (0.865–0.871) | 0.212 (0.205–0.220) | 0.989 (0.988–0.989) | 0.298 (0.290–0.306) | 0.303 (0.295–0.311) |
| | DeepECG-SSL | **0.388 (0.380–0.397)** | **0.896 (0.894–0.898)** | **0.250 (0.243–0.258)** | 0.989 (0.989–0.990) | **0.344 (0.336–0.353)** | **0.350 (0.343–0.358)** |
| | CardioLoom | 0.308 (0.300–0.316) | 0.849 (0.846–0.852) | 0.150 (0.144–0.156) | **0.993 (0.993–0.994)** | 0.234 (0.226–0.243) | 0.267 (0.258–0.275) |
| ER | Scratch CNN–Transformer | 0.040 (0.036–0.044) | 0.744 (0.730–0.759) | 0.033 (0.027–0.039) | **0.999 (0.999–0.999)** | 0.051 (0.041–0.060) | 0.058 (0.047–0.068) |
| | ECG-FM | 0.042 (0.039–0.046) | 0.762 (0.748–0.776) | 0.072 (0.063–0.083) | 0.997 (0.997–0.997) | 0.086 (0.076–0.097) | 0.084 (0.073–0.095) |
| | DeepECG-SSL | **0.055 (0.050–0.060)** | **0.828 (0.817–0.838)** | **0.137 (0.124–0.149)** | 0.994 (0.994–0.995) | **0.122 (0.112–0.133)** | **0.118 (0.107–0.129)** |
| | CardioLoom | 0.047 (0.043–0.052) | 0.791 (0.778–0.803) | 0.068 (0.059–0.079) | 0.997 (0.997–0.998) | 0.087 (0.076–0.098) | 0.087 (0.075–0.098) |

| Disease | Model | AUPRC | AUROC | Sensitivity | Specificity | F1 Score | MCC |
|---|---|---|---|---|---|---|---|
| LAFB | Scratch CNN–Transformer | 0.656 (0.644–0.666) | 0.956 (0.955–0.958) | 0.295 (0.283–0.305) | 0.995 (0.995–0.996) | 0.432 (0.419–0.445) | 0.471 (0.460–0.482) |
| | ECG-FM | 0.669 (0.657–0.680) | 0.953 (0.951–0.955) | **0.298 (0.286–0.311)** | 0.996 (0.996–0.997) | **0.440 (0.426–0.454)** | 0.484 (0.472–0.496) |
| | DeepECG-SSL | **0.686 (0.674–0.695)** | **0.959 (0.958–0.960)** | 0.272 (0.261–0.283) | **0.997 (0.997–0.997)** | 0.413 (0.400–0.426) | 0.466 (0.455–0.478) |
| | CardioLoom | 0.652 (0.640–0.664) | 0.941 (0.938–0.943) | 0.296 (0.284–0.308) | 0.996 (0.996–0.997) | 0.439 (0.425–0.453) | **0.485 (0.473–0.496)** |
| LBBB | Scratch CNN–Transformer | 0.810 (0.799–0.820) | 0.975 (0.973–0.976) | 0.553 (0.537–0.567) | **0.998 (0.997–0.998)** | 0.686 (0.672–0.697) | **0.698 (0.687–0.709)** |
| | ECG-FM | 0.811 (0.801–0.821) | 0.975 (0.974–0.977) | 0.486 (0.470–0.501) | **0.998 (0.998–0.998)** | 0.635 (0.620–0.648) | 0.658 (0.646–0.669) |
| | DeepECG-SSL | 0.803 (0.793–0.813) | 0.974 (0.972–0.975) | **0.559 (0.543–0.574)** | 0.997 (0.997–0.998) | **0.687 (0.674–0.699)** | 0.697 (0.686–0.708) |
| | CardioLoom | **0.816 (0.805–0.825)** | **0.976 (0.975–0.978)** | 0.549 (0.533–0.564) | **0.998 (0.997–0.998)** | 0.682 (0.669–0.694) | 0.694 (0.683–0.705) |
| IVB | Scratch CNN–Transformer | 0.301 (0.289–0.314) | 0.875 (0.871–0.878) | 0.218 (0.207–0.231) | **0.990 (0.989–0.990)** | 0.299 (0.287–0.313) | 0.304 (0.293–0.317) |
| | ECG-FM | 0.333 (0.320–0.347) | 0.879 (0.875–0.883) | 0.228 (0.217–0.241) | **0.990 (0.990–0.991)** | 0.315 (0.302–0.329) | 0.323 (0.310–0.336) |
| | DeepECG-SSL | 0.336 (0.323–0.350) | 0.892 (0.888–0.896) | 0.246 (0.234–0.258) | 0.989 (0.988–0.990) | 0.328 (0.315–0.341) | 0.328 (0.316–0.341) |
| | CardioLoom | **0.352 (0.339–0.366)** | **0.902 (0.899–0.905)** | **0.296 (0.283–0.310)** | 0.987 (0.987–0.988) | **0.373 (0.361–0.387)** | **0.367 (0.355–0.380)** |
| Short PR | Scratch CNN–Transformer | 0.155 (0.145–0.165) | 0.835 (0.830–0.841) | 0.095 (0.087–0.103) | 0.998 (0.998–0.998) | 0.154 (0.142–0.166) | 0.190 (0.178–0.203) |
| | ECG-FM | 0.146 (0.137–0.155) | 0.843 (0.837–0.848) | 0.068 (0.061–0.075) | 0.998 (0.998–0.998) | 0.114 (0.104–0.125) | 0.151 (0.138–0.163) |
| | DeepECG-SSL | **0.199 (0.188–0.212)** | **0.850 (0.843–0.855)** | **0.113 (0.103–0.122)** | 0.998 (0.998–0.998) | **0.182 (0.169–0.195)** | **0.226 (0.213–0.240)** |
| | CardioLoom | 0.160 (0.150–0.171) | 0.797 (0.791–0.804) | 0.051 (0.046–0.058) | **0.999 (0.999–0.999)** | 0.092 (0.083–0.103) | 0.149 (0.137–0.164) |
| VPE | Scratch CNN–Transformer | 0.126 (0.080–0.178) | 0.875 (0.856–0.894) | 0.248 (0.184–0.311) | 0.999 (0.999–1.000) | 0.236 (0.179–0.293) | 0.236 (0.178–0.293) |
| | ECG-FM | 0.137 (0.085–0.195) | 0.880 (0.859–0.898) | **0.267 (0.198–0.337)** | **1.000 (1.000–1.000)** | 0.284 (0.220–0.345) | 0.284 (0.220–0.346) |
| | DeepECG-SSL | 0.115 (0.072–0.164) | **0.882 (0.861–0.902)** | 0.259 (0.192–0.328) | **1.000 (0.999–1.000)** | 0.267 (0.203–0.329) | 0.267 (0.203–0.329) |
| | CardioLoom | **0.150 (0.099–0.203)** | 0.874 (0.853–0.894) | **0.267 (0.198–0.338)** | **1.000 (1.000–1.000)** | **0.295 (0.226–0.361)** | **0.296 (0.227–0.360)** |

| Disease | Model | AUPRC | AUROC | Sensitivity | Specificity | F1 Score | MCC |
|---|---|---|---|---|---|---|---|
| 3° AVB | Scratch CNN–Transformer | 0.024 (0.015–0.038) | **0.955 (0.939–0.968)** | **0.422 (0.351–0.490)** | 0.993 (0.992–0.993) | 0.030 (0.024–0.037) | 0.080 (0.066–0.095) |
| | ECG-FM | 0.032 (0.020–0.051) | 0.931 (0.912–0.947) | 0.348 (0.272–0.421) | 0.997 (0.997–0.997) | 0.055 (0.041–0.070) | 0.101 (0.077–0.125) |
| | DeepECG-SSL | 0.048 (0.026–0.082) | 0.917 (0.893–0.937) | 0.412 (0.335–0.490) | 0.997 (0.997–0.997) | 0.063 (0.049–0.078) | **0.118 (0.094–0.141)** |
| | CardioLoom | **0.050 (0.027–0.085)** | 0.925 (0.902–0.945) | 0.289 (0.215–0.362) | **0.998 (0.998–0.998)** | **0.067 (0.048–0.086)** | 0.104 (0.077–0.132) |
| 2° 1 Type AVB | Scratch CNN–Transformer | 0.022 (0.013–0.038) | **0.798 (0.767–0.824)** | 0.208 (0.153–0.265) | 0.998 (0.998–0.998) | 0.076 (0.055–0.100) | 0.098 (0.071–0.126) |
| | ECG-FM | **0.031 (0.020–0.049)** | 0.790 (0.763–0.814) | 0.285 (0.223–0.344) | **0.999 (0.998–0.999)** | **0.120 (0.093–0.147)** | **0.146 (0.115–0.177)** |
| | DeepECG-SSL | 0.028 (0.018–0.042) | 0.755 (0.718–0.786) | **0.295 (0.238–0.355)** | 0.998 (0.998–0.999) | 0.109 (0.086–0.134) | 0.140 (0.111–0.169) |
| | CardioLoom | 0.028 (0.018–0.044) | 0.721 (0.680–0.754) | 0.292 (0.228–0.353) | 0.998 (0.998–0.999) | 0.105 (0.082–0.128) | 0.136 (0.106–0.165) |
| 2° 2 Type AVB | Scratch CNN–Transformer | **0.004 (0.003–0.004)** | **0.593 (0.577–0.610)** | 0.003 (0.001–0.005) | **0.999 (0.999–0.999)** | 0.004 (0.001–0.008) | 0.004 (0.000–0.009) |
| | ECG-FM | 0.003 (0.002–0.003) | 0.480 (0.464–0.496) | **0.005 (0.002–0.009)** | **0.999 (0.999–0.999)** | **0.008 (0.003–0.013)** | **0.007 (0.002–0.014)** |
| | DeepECG-SSL | **0.004 (0.003–0.005)** | 0.569 (0.552–0.586) | 0.003 (0.001–0.007) | **0.999 (0.999–0.999)** | 0.005 (0.001–0.011) | 0.005 (0.000–0.012) |
| | CardioLoom | **0.004 (0.003–0.005)** | 0.513 (0.496–0.531) | 0.003 (0.001–0.006) | **0.999 (0.999–0.999)** | 0.005 (0.001–0.010) | 0.005 (0.000–0.010) |
| LVH | Scratch CNN–Transformer | 0.313 (0.303–0.321) | 0.753 (0.748–0.757) | 0.147 (0.140–0.154) | 0.987 (0.986–0.987) | 0.231 (0.221–0.241) | 0.247 (0.237–0.256) |
| | ECG-FM | 0.317 (0.308–0.326) | **0.758 (0.753–0.762)** | 0.140 (0.133–0.147) | 0.988 (0.988–0.988) | 0.223 (0.213–0.232) | 0.244 (0.234–0.253) |
| | DeepECG-SSL | 0.327 (0.318–0.337) | 0.738 (0.733–0.743) | **0.166 (0.159–0.174)** | 0.987 (0.986–0.987) | **0.257 (0.247–0.267)** | **0.272 (0.262–0.282)** |
| | CardioLoom | **0.342 (0.332–0.351)** | 0.756 (0.751–0.760) | 0.119 (0.113–0.126) | **0.993 (0.993–0.994)** | 0.202 (0.191–0.212) | 0.254 (0.243–0.263) |
| LAH | Scratch CNN–Transformer | 0.074 (0.070–0.078) | 0.805 (0.801–0.809) | **0.308 (0.297–0.318)** | 0.926 (0.925–0.928) | 0.134 (0.130–0.140) | 0.127 (0.122–0.134) |
| | ECG-FM | 0.103 (0.097–0.109) | 0.819 (0.815–0.823) | 0.271 (0.261–0.283) | **0.960 (0.959–0.961)** | **0.178 (0.171–0.185)** | **0.164 (0.157–0.171)** |
| | DeepECG-SSL | 0.074 (0.070–0.079) | 0.791 (0.786–0.795) | 0.279 (0.268–0.290) | 0.934 (0.932–0.935) | 0.132 (0.127–0.137) | 0.121 (0.115–0.127) |
| | CardioLoom | **0.105 (0.099–0.111)** | **0.831 (0.828–0.835)** | 0.275 (0.264–0.285) | 0.957 (0.956–0.958) | 0.173 (0.166–0.180) | 0.159 (0.153–0.166) |

| Disease | Model | AUPRC | AUROC | Sensitivity | Specificity | F1 Score | MCC |
|---|---|---|---|---|---|---|---|
| RVH | Scratch CNN–Transformer | **0.077 (0.067–0.088)** | 0.842 (0.834–0.850) | **0.109 (0.095–0.126)** | 0.993 (0.993–0.994) | **0.124 (0.108–0.141)** | **0.117 (0.101–0.134)** |
| | ECG-FM | 0.065 (0.057–0.073) | 0.815 (0.806–0.825) | 0.086 (0.075–0.098) | 0.994 (0.994–0.995) | 0.105 (0.094–0.119) | 0.101 (0.089–0.114) |
| | DeepECG-SSL | 0.075 (0.067–0.084) | **0.846 (0.837–0.854)** | 0.069 (0.058–0.079) | **0.996 (0.995–0.996)** | 0.092 (0.079–0.105) | 0.091 (0.078–0.106) |
| | CardioLoom | 0.072 (0.064–0.081) | 0.832 (0.824–0.841) | 0.073 (0.063–0.084) | **0.996 (0.995–0.996)** | 0.099 (0.087–0.114) | 0.100 (0.087–0.116) |
| RAH | Scratch CNN–Transformer | 0.016 (0.010–0.026) | **0.901 (0.881–0.919)** | **0.052 (0.026–0.080)** | **1.000 (1.000–1.000)** | **0.057 (0.028–0.087)** | **0.057 (0.028–0.088)** |
| | ECG-FM | 0.008 (0.004–0.015) | 0.820 (0.791–0.845) | 0.042 (0.017–0.077) | **1.000 (1.000–1.000)** | 0.043 (0.018–0.076) | 0.043 (0.018–0.076) |
| | DeepECG-SSL | 0.010 (0.005–0.021) | 0.871 (0.849–0.891) | 0.042 (0.017–0.075) | 0.999 (0.999–1.000) | 0.037 (0.015–0.061) | 0.037 (0.015–0.062) |
| | CardioLoom | **0.017 (0.009–0.029)** | 0.863 (0.838–0.886) | 0.015 (0.000–0.040) | **1.000 (1.000–1.000)** | 0.027 (0.000–0.068) | 0.046 (0.000–0.097) |
| DC | Scratch CNN–Transformer | **0.010 (0.006–0.018)** | **0.910 (0.890–0.928)** | **0.050 (0.024–0.082)** | **1.000 (1.000–1.000)** | 0.047 (0.022–0.075) | 0.047 (0.022–0.075) |
| | ECG-FM | **0.010 (0.004–0.024)** | 0.778 (0.740–0.816) | 0.041 (0.014–0.075) | **1.000 (1.000–1.000)** | **0.052 (0.018–0.093)** | **0.053 (0.018–0.097)** |
| | DeepECG-SSL | 0.007 (0.004–0.012) | 0.883 (0.854–0.910) | 0.000 (0.000–0.000) | **1.000 (1.000–1.000)** | 0.000 (0.000–0.000) | 0.000 (0.000–0.000) |
| | CardioLoom | 0.004 (0.002–0.008) | 0.805 (0.769–0.840) | 0.000 (0.000–0.000) | **1.000 (1.000–1.000)** | 0.000 (0.000–0.000) | 0.000 (0.000–0.000) |
| BVH | Scratch CNN–Transformer | 0.067 (0.064–0.070) | 0.770 (0.765–0.775) | 0.000 (0.000–0.000) | **1.000 (1.000–1.000)** | 0.000 (0.000–0.001) | 0.003 (-0.001–0.008) |
| | ECG-FM | 0.063 (0.060–0.066) | 0.753 (0.748–0.758) | 0.000 (0.000–0.000) | **1.000 (1.000–1.000)** | 0.000 (0.000–0.000) | 0.005 (0.000–0.012) |
| | DeepECG-SSL | 0.068 (0.064–0.071) | 0.748 (0.741–0.753) | 0.000 (0.000–0.001) | **1.000 (1.000–1.000)** | **0.001 (0.000–0.001)** | 0.005 (0.001–0.009) |
| | CardioLoom | **0.075 (0.071–0.079)** | **0.773 (0.767–0.778)** | **0.001 (0.000–0.001)** | **1.000 (1.000–1.000)** | **0.001 (0.000–0.002)** | **0.011 (0.004–0.018)** |
| VAT | Scratch CNN–Transformer | 0.003 (0.002–0.004) | 0.924 (0.914–0.933) | 0.113 (0.066–0.165) | 0.992 (0.992–0.993) | 0.009 (0.005–0.012) | 0.021 (0.012–0.031) |
| | ECG-FM | **0.009 (0.007–0.012)** | 0.976 (0.970–0.982) | 0.151 (0.095–0.217) | **0.993 (0.993–0.994)** | 0.014 (0.008–0.019) | 0.032 (0.019–0.045) |
| | DeepECG-SSL | 0.003 (0.002–0.005) | 0.883 (0.868–0.897) | 0.142 (0.088–0.207) | 0.989 (0.988–0.990) | 0.008 (0.005–0.011) | 0.022 (0.013–0.032) |
| | CardioLoom | **0.009 (0.007–0.011)** | **0.981 (0.978–0.984)** | **0.218 (0.150–0.292)** | 0.991 (0.991–0.992) | **0.015 (0.010–0.020)** | **0.040 (0.027–0.053)** |

| Disease | Model | AUPRC | AUROC | Sensitivity | Specificity | F1 Score | MCC |
|---|---|---|---|---|---|---|---|
| DDD | Scratch CNN–Transformer | 0.068 (0.050–0.089) | 0.994 (0.992–0.995) | 0.753 (0.688–0.823) | 0.994 (0.994–0.995) | 0.090 (0.073–0.109) | 0.189 (0.167–0.212) |
| | ECG-FM | 0.077 (0.058–0.100) | 0.992 (0.988–0.995) | 0.847 (0.776–0.915) | 0.994 (0.993–0.994) | 0.094 (0.078–0.113) | 0.205 (0.183–0.226) |
| | DeepECG-SSL | 0.052 (0.041–0.064) | 0.992 (0.988–0.995) | 0.830 (0.770–0.888) | 0.995 (0.995–0.996) | 0.112 (0.092–0.135) | 0.223 (0.198–0.247) |
| | CardioLoom | **0.107 (0.077–0.162)** | **0.996 (0.995–0.997)** | **0.872 (0.821–0.920)** | **0.996 (0.995–0.996)** | **0.129 (0.105–0.155)** | **0.245 (0.218–0.271)** |
| HK | Scratch CNN–Transformer | **0.011 (0.009–0.014)** | **0.779 (0.756–0.800)** | 0.007 (0.003–0.012) | **1.000 (1.000–1.000)** | 0.013 (0.006–0.021) | 0.019 (0.008–0.031) |
| | ECG-FM | 0.004 (0.003–0.006) | 0.574 (0.541–0.602) | 0.005 (0.002–0.008) | **1.000 (1.000–1.000)** | 0.009 (0.003–0.015) | 0.014 (0.005–0.025) |
| | DeepECG-SSL | 0.008 (0.007–0.011) | 0.726 (0.691–0.754) | **0.015 (0.009–0.023)** | 0.999 (0.999–0.999) | **0.023 (0.014–0.034)** | **0.026 (0.015–0.039)** |
| | CardioLoom | 0.007 (0.006–0.009) | 0.705 (0.681–0.729) | 0.004 (0.001–0.008) | **1.000 (1.000–1.000)** | 0.008 (0.002–0.014) | 0.013 (0.004–0.024) |

Values are estimates (95% cluster-bootstrap CIs). NE indicates fewer than 950 valid replicates among 1000. Sensitivity, specificity, F1 score, and MCC use the Zhongshan validation-F1 threshold. Bold type marks the highest displayed estimate within each disease.

**Supplementary Table 21: Detailed task-specific performance comparison on the external UKB cohort.**

| Disease | Model | AUPRC | AUROC | Sensitivity | Specificity | F1 Score | MCC |
|---|---|---|---|---|---|---|---|
| STEMI | Scratch CNN–Transformer | 0.018 (0.012–0.031) | 0.717 (0.680–0.757) | 0.000 (0.000–0.000) | **1.000 (1.000–1.000)** | 0.000 (0.000–0.000) | -0.001 (-0.001–0.000) |
| | ECG-FM | 0.021 (0.014–0.037) | **0.759 (0.727–0.792)** | **0.005 (0.000–0.015)** | **1.000 (1.000–1.000)** | **0.009 (0.000–0.029)** | **0.047 (0.000–0.114)** |
| | DeepECG-SSL | 0.018 (0.011–0.037) | 0.706 (0.670–0.745) | **0.005 (0.000–0.014)** | **1.000 (1.000–1.000)** | **0.009 (0.000–0.028)** | 0.039 (-0.001–0.102) |
| | CardioLoom | **0.023 (0.015–0.041)** | 0.733 (0.696–0.772) | 0.000 (0.000–0.000) | **1.000 (1.000–1.000)** | 0.000 (0.000–0.000) | 0.000 (-0.001–0.000) |
| NSR | Scratch CNN–Transformer | **0.979 (0.977–0.980)** | 0.684 (0.676–0.690) | 0.506 (0.502–0.511) | **0.880 (0.868–0.893)** | 0.669 (0.665–0.673) | **0.168 (0.162–0.175)** |
| | ECG-FM | 0.978 (0.977–0.979) | 0.678 (0.671–0.686) | **0.521 (0.516–0.525)** | 0.858 (0.845–0.872) | **0.681 (0.678–0.685)** | 0.165 (0.158–0.172) |
| | DeepECG-SSL | 0.978 (0.977–0.979) | 0.677 (0.670–0.684) | 0.520 (0.515–0.525) | 0.865 (0.852–0.878) | **0.681 (0.677–0.685)** | **0.168 (0.161–0.175)** |
| | CardioLoom | 0.978 (0.977–0.980) | **0.685 (0.678–0.693)** | 0.514 (0.510–0.519) | 0.871 (0.858–0.885) | 0.676 (0.672–0.680) | **0.168 (0.161–0.175)** |
| SBrad | Scratch CNN–Transformer | **0.992 (0.991–0.993)** | **0.995 (0.994–0.995)** | 0.950 (0.947–0.952) | **0.973 (0.971–0.975)** | 0.958 (0.956–0.960) | 0.924 (0.921–0.928) |
| | ECG-FM | 0.991 (0.991–0.992) | 0.994 (0.993–0.994) | 0.964 (0.961–0.966) | 0.964 (0.962–0.966) | **0.960 (0.958–0.962)** | **0.927 (0.924–0.931)** |
| | DeepECG-SSL | 0.990 (0.989–0.991) | 0.993 (0.993–0.994) | 0.967 (0.965–0.969) | 0.960 (0.957–0.962) | 0.959 (0.957–0.961) | 0.925 (0.922–0.928) |
| | CardioLoom | 0.991 (0.990–0.992) | 0.994 (0.993–0.994) | **0.974 (0.972–0.976)** | 0.951 (0.948–0.953) | 0.957 (0.956–0.959) | 0.922 (0.919–0.926) |
| VPB | Scratch CNN–Transformer | 0.660 (0.633–0.688) | **0.967 (0.962–0.972)** | 0.651 (0.625–0.674) | 0.994 (0.994–0.995) | 0.702 (0.681–0.721) | 0.697 (0.676–0.716) |
| | ECG-FM | 0.693 (0.667–0.720) | **0.967 (0.962–0.971)** | 0.664 (0.639–0.688) | **0.995 (0.994–0.995)** | 0.718 (0.697–0.737) | 0.713 (0.692–0.732) |
| | DeepECG-SSL | 0.656 (0.629–0.683) | 0.941 (0.933–0.949) | **0.678 (0.654–0.701)** | 0.994 (0.994–0.995) | 0.722 (0.702–0.740) | 0.716 (0.696–0.735) |
| | CardioLoom | **0.709 (0.682–0.734)** | **0.967 (0.962–0.972)** | 0.676 (0.652–0.700) | 0.994 (0.994–0.995) | **0.724 (0.705–0.743)** | **0.718 (0.699–0.738)** |
| APB | Scratch CNN–Transformer | 0.440 (0.408–0.474) | **0.952 (0.946–0.958)** | 0.648 (0.621–0.679) | 0.979 (0.978–0.980) | 0.494 (0.471–0.518) | 0.495 (0.473–0.518) |
| | ECG-FM | 0.424 (0.393–0.457) | 0.922 (0.911–0.932) | 0.644 (0.615–0.674) | **0.985 (0.984–0.986)** | 0.543 (0.518–0.567) | 0.539 (0.514–0.563) |
| | DeepECG-SSL | **0.496 (0.462–0.532)** | 0.945 (0.938–0.953) | **0.687 (0.660–0.718)** | **0.985 (0.983–0.986)** | **0.569 (0.545–0.592)** | **0.567 (0.543–0.591)** |
| | CardioLoom | 0.477 (0.444–0.513) | 0.946 (0.938–0.953) | 0.663 (0.634–0.694) | 0.984 (0.983–0.985) | 0.548 (0.525–0.571) | 0.545 (0.522–0.569) |

| Disease | Model | AUPRC | AUROC | Sensitivity | Specificity | F1 Score | MCC |
|---|---|---|---|---|---|---|---|
| SArr | Scratch CNN–Transformer | 0.501 (0.481–0.522) | 0.878 (0.870–0.886) | 0.288 (0.270–0.306) | 0.995 (0.995–0.996) | 0.416 (0.395–0.437) | 0.450 (0.429–0.469) |
| | ECG-FM | 0.479 (0.458–0.500) | 0.886 (0.879–0.894) | 0.266 (0.249–0.283) | 0.995 (0.994–0.995) | 0.387 (0.366–0.408) | 0.420 (0.399–0.441) |
| | DeepECG-SSL | 0.616 (0.596–0.638) | 0.919 (0.913–0.926) | **0.398 (0.378–0.417)** | 0.996 (0.995–0.996) | **0.535 (0.514–0.554)** | **0.556 (0.538–0.575)** |
| | CardioLoom | **0.625 (0.606–0.646)** | **0.951 (0.947–0.955)** | 0.319 (0.301–0.337) | **0.997 (0.996–0.997)** | 0.460 (0.440–0.480) | 0.501 (0.483–0.520) |
| AF | Scratch CNN–Transformer | 0.874 (0.854–0.894) | **0.987 (0.982–0.992)** | 0.645 (0.612–0.681) | 0.999 (0.999–1.000) | 0.767 (0.741–0.793) | 0.778 (0.755–0.802) |
| | ECG-FM | **0.906 (0.888–0.923)** | 0.986 (0.979–0.991) | 0.632 (0.599–0.668) | **1.000 (1.000–1.000)** | 0.770 (0.743–0.796) | 0.786 (0.764–0.809) |
| | DeepECG-SSL | 0.892 (0.873–0.911) | 0.982 (0.976–0.988) | **0.685 (0.653–0.718)** | **1.000 (1.000–1.000)** | **0.805 (0.781–0.828)** | **0.815 (0.794–0.836)** |
| | CardioLoom | 0.873 (0.850–0.894) | 0.985 (0.980–0.991) | 0.645 (0.612–0.679) | **1.000 (0.999–1.000)** | 0.774 (0.749–0.799) | 0.788 (0.766–0.810) |
| AFL | Scratch CNN–Transformer | **0.491 (0.356–0.619)** | 0.862 (0.806–0.917) | 0.492 (0.373–0.615) | **1.000 (1.000–1.000)** | 0.593 (0.476–0.692) | 0.605 (0.492–0.703) |
| | ECG-FM | 0.473 (0.335–0.630) | 0.839 (0.775–0.900) | **0.585 (0.462–0.700)** | **1.000 (1.000–1.000)** | **0.644 (0.535–0.736)** | **0.647 (0.540–0.739)** |
| | DeepECG-SSL | 0.457 (0.309–0.600) | **0.884 (0.828–0.934)** | 0.569 (0.449–0.688) | **1.000 (0.999–1.000)** | 0.617 (0.505–0.712) | 0.618 (0.510–0.715) |
| | CardioLoom | 0.473 (0.329–0.614) | 0.828 (0.757–0.892) | **0.585 (0.462–0.700)** | **1.000 (1.000–1.000)** | 0.639 (0.530–0.731) | 0.641 (0.533–0.734) |
| JPB | Scratch CNN–Transformer | **0.124 (0.105–0.147)** | **0.908 (0.891–0.925)** | 0.040 (0.023–0.058) | **0.998 (0.998–0.998)** | 0.063 (0.036–0.091) | 0.073 (0.041–0.106) |
| | ECG-FM | 0.075 (0.061–0.094) | 0.802 (0.776–0.829) | **0.066 (0.045–0.092)** | 0.997 (0.997–0.997) | **0.095 (0.066–0.128)** | **0.100 (0.069–0.136)** |
| | DeepECG-SSL | 0.094 (0.079–0.113) | 0.864 (0.842–0.884) | 0.051 (0.032–0.072) | **0.998 (0.997–0.998)** | 0.077 (0.048–0.109) | 0.085 (0.053–0.121) |
| | CardioLoom | 0.111 (0.094–0.131) | 0.884 (0.861–0.905) | 0.046 (0.028–0.067) | 0.997 (0.997–0.998) | 0.070 (0.044–0.099) | 0.077 (0.047–0.110) |
| AJR | Scratch CNN–Transformer | **0.021 (0.016–0.028)** | **0.826 (0.795–0.855)** | **0.141 (0.092–0.196)** | 0.989 (0.988–0.990) | **0.067 (0.043–0.093)** | **0.073 (0.045–0.102)** |
| | ECG-FM | 0.012 (0.009–0.018) | 0.747 (0.715–0.781) | 0.033 (0.010–0.059) | **0.997 (0.996–0.997)** | 0.033 (0.010–0.060) | 0.030 (0.007–0.057) |
| | DeepECG-SSL | 0.012 (0.009–0.016) | 0.769 (0.735–0.802) | 0.027 (0.006–0.051) | 0.996 (0.996–0.997) | 0.027 (0.006–0.051) | 0.024 (0.002–0.048) |
| | CardioLoom | 0.016 (0.012–0.022) | 0.787 (0.754–0.819) | 0.022 (0.005–0.045) | **0.997 (0.997–0.997)** | 0.024 (0.006–0.048) | 0.020 (0.002–0.045) |

| Disease | Model | AUPRC | AUROC | Sensitivity | Specificity | F1 Score | MCC |
|---|---|---|---|---|---|---|---|
| JTach | Scratch CNN–Transformer | **0.024 (0.017–0.035)** | **0.820 (0.787–0.850)** | **0.027 (0.006–0.053)** | 0.999 (0.999–0.999) | **0.044 (0.009–0.085)** | 0.053 (0.010–0.104) |
| | ECG-FM | 0.016 (0.010–0.032) | 0.722 (0.688–0.756) | 0.016 (0.000–0.037) | **1.000 (1.000–1.000)** | 0.030 (0.000–0.068) | **0.058 (-0.001–0.126)** |
| | DeepECG-SSL | 0.014 (0.010–0.026) | 0.752 (0.716–0.786) | **0.027 (0.005–0.053)** | 0.999 (0.999–1.000) | **0.044 (0.009–0.085)** | 0.056 (0.010–0.106) |
| | CardioLoom | 0.018 (0.012–0.032) | 0.761 (0.725–0.795) | 0.016 (0.000–0.037) | **1.000 (0.999–1.000)** | 0.029 (0.000–0.065) | 0.045 (-0.001–0.098) |
| JEB | Scratch CNN–Transformer | **0.026 (0.019–0.036)** | **0.842 (0.813–0.869)** | **0.027 (0.006–0.053)** | 0.998 (0.997–0.998) | **0.034 (0.007–0.065)** | **0.032 (0.004–0.064)** |
| | ECG-FM | 0.009 (0.007–0.013) | 0.704 (0.668–0.741) | 0.011 (0.000–0.028) | **0.999 (0.998–0.999)** | 0.016 (0.000–0.040) | 0.015 (-0.002–0.042) |
| | DeepECG-SSL | 0.016 (0.011–0.022) | 0.784 (0.751–0.818) | 0.011 (0.000–0.027) | **0.999 (0.998–0.999)** | 0.016 (0.000–0.040) | 0.016 (-0.002–0.042) |
| | CardioLoom | 0.013 (0.010–0.018) | 0.746 (0.709–0.780) | 0.005 (0.000–0.017) | 0.998 (0.998–0.999) | 0.008 (0.000–0.024) | 0.006 (-0.003–0.024) |
| RBBB | Scratch CNN–Transformer | 0.797 (0.783–0.811) | 0.975 (0.972–0.978) | **0.839 (0.824–0.854)** | 0.963 (0.961–0.965) | 0.661 (0.648–0.674) | 0.656 (0.643–0.670) |
| | ECG-FM | 0.789 (0.775–0.803) | 0.977 (0.975–0.979) | 0.705 (0.687–0.721) | **0.983 (0.982–0.984)** | **0.695 (0.681–0.708)** | 0.679 (0.664–0.693) |
| | DeepECG-SSL | 0.792 (0.777–0.807) | 0.974 (0.971–0.977) | 0.837 (0.823–0.852) | 0.960 (0.958–0.962) | 0.645 (0.632–0.660) | 0.641 (0.629–0.655) |
| | CardioLoom | **0.806 (0.792–0.819)** | **0.978 (0.976–0.981)** | 0.769 (0.752–0.784) | 0.977 (0.975–0.978) | **0.695 (0.681–0.709)** | **0.681 (0.666–0.695)** |
| 1° AVB | Scratch CNN–Transformer | 0.871 (0.861–0.881) | 0.988 (0.986–0.989) | 0.819 (0.805–0.833) | 0.984 (0.982–0.985) | 0.787 (0.775–0.798) | 0.774 (0.762–0.785) |
| | ECG-FM | 0.910 (0.901–0.919) | **0.993 (0.992–0.994)** | 0.859 (0.846–0.872) | **0.987 (0.986–0.988)** | 0.834 (0.824–0.843) | 0.823 (0.813–0.834) |
| | DeepECG-SSL | **0.914 (0.906–0.922)** | **0.993 (0.992–0.994)** | **0.918 (0.908–0.928)** | 0.980 (0.979–0.981) | 0.819 (0.809–0.829) | 0.812 (0.802–0.821) |
| | CardioLoom | **0.914 (0.907–0.923)** | 0.992 (0.991–0.993) | 0.867 (0.855–0.878) | **0.987 (0.986–0.988)** | **0.835 (0.825–0.844)** | **0.825 (0.815–0.834)** |
| QT Prolong | Scratch CNN–Transformer | 0.108 (0.087–0.133) | 0.880 (0.867–0.894) | 0.027 (0.015–0.043) | **1.000 (0.999–1.000)** | 0.051 (0.028–0.079) | 0.103 (0.060–0.151) |
| | ECG-FM | **0.218 (0.182–0.262)** | **0.897 (0.882–0.911)** | 0.029 (0.015–0.044) | **1.000 (1.000–1.000)** | 0.057 (0.029–0.084) | **0.147 (0.093–0.192)** |
| | DeepECG-SSL | 0.189 (0.157–0.225) | **0.897 (0.884–0.911)** | **0.039 (0.023–0.058)** | **1.000 (0.999–1.000)** | **0.073 (0.043–0.106)** | 0.138 (0.087–0.188) |
| | CardioLoom | 0.198 (0.165–0.240) | 0.893 (0.880–0.906) | 0.012 (0.004–0.022) | **1.000 (1.000–1.000)** | 0.023 (0.008–0.043) | 0.093 (0.039–0.135) |

| Disease | Model | AUPRC | AUROC | Sensitivity | Specificity | F1 Score | MCC |
|---|---|---|---|---|---|---|---|
| ER | Scratch CNN–Transformer | 0.229 (0.186–0.271) | 0.932 (0.920–0.943) | 0.044 (0.024–0.063) | **1.000 (1.000–1.000)** | 0.081 (0.045–0.116) | 0.157 (0.099–0.206) |
| | ECG-FM | 0.294 (0.246–0.337) | 0.949 (0.940–0.957) | 0.202 (0.162–0.238) | 0.998 (0.998–0.999) | 0.288 (0.237–0.333) | 0.314 (0.263–0.361) |
| | DeepECG-SSL | **0.319 (0.272–0.366)** | **0.960 (0.952–0.967)** | **0.320 (0.275–0.363)** | 0.996 (0.996–0.997) | **0.370 (0.320–0.411)** | **0.370 (0.321–0.412)** |
| | CardioLoom | 0.271 (0.223–0.312) | 0.947 (0.937–0.955) | 0.092 (0.066–0.117) | 0.999 (0.999–1.000) | 0.157 (0.115–0.197) | 0.222 (0.170–0.268) |
| LAFB | Scratch CNN–Transformer | 0.530 (0.492–0.566) | 0.973 (0.969–0.977) | 0.275 (0.246–0.307) | **0.998 (0.998–0.999)** | 0.396 (0.361–0.432) | 0.436 (0.401–0.472) |
| | ECG-FM | 0.602 (0.563–0.638) | 0.984 (0.981–0.987) | **0.435 (0.400–0.468)** | 0.997 (0.997–0.997) | **0.533 (0.499–0.563)** | **0.541 (0.509–0.573)** |
| | DeepECG-SSL | 0.586 (0.546–0.623) | **0.985 (0.983–0.988)** | 0.389 (0.352–0.423) | 0.997 (0.997–0.998) | 0.499 (0.461–0.532) | 0.515 (0.480–0.547) |
| | CardioLoom | **0.607 (0.566–0.645)** | 0.982 (0.977–0.985) | 0.328 (0.294–0.360) | **0.998 (0.998–0.998)** | 0.452 (0.415–0.485) | 0.484 (0.449–0.516) |
| LBBB | Scratch CNN–Transformer | 0.869 (0.834–0.900) | 0.980 (0.971–0.988) | **0.727 (0.688–0.766)** | **0.999 (0.999–1.000)** | **0.810 (0.782–0.837)** | **0.814 (0.787–0.840)** |
| | ECG-FM | 0.864 (0.832–0.893) | 0.981 (0.973–0.988) | 0.665 (0.621–0.707) | **0.999 (0.999–1.000)** | 0.772 (0.739–0.803) | 0.780 (0.749–0.809) |
| | DeepECG-SSL | 0.867 (0.834–0.897) | 0.977 (0.967–0.986) | 0.695 (0.654–0.736) | **0.999 (0.999–0.999)** | 0.781 (0.749–0.810) | 0.785 (0.755–0.813) |
| | CardioLoom | **0.876 (0.844–0.907)** | **0.984 (0.976–0.991)** | 0.697 (0.656–0.737) | **0.999 (0.999–0.999)** | 0.785 (0.754–0.814) | 0.790 (0.760–0.817) |
| IVB | Scratch CNN–Transformer | 0.062 (0.046–0.085) | 0.822 (0.796–0.845) | 0.088 (0.056–0.121) | **0.997 (0.997–0.998)** | 0.115 (0.075–0.156) | 0.117 (0.075–0.158) |
| | ECG-FM | 0.073 (0.055–0.098) | 0.804 (0.773–0.832) | 0.084 (0.054–0.118) | **0.997 (0.997–0.997)** | 0.108 (0.071–0.147) | 0.108 (0.070–0.149) |
| | DeepECG-SSL | 0.073 (0.056–0.093) | 0.819 (0.790–0.844) | 0.078 (0.048–0.108) | **0.997 (0.997–0.998)** | 0.100 (0.064–0.136) | 0.101 (0.062–0.137) |
| | CardioLoom | **0.093 (0.073–0.117)** | **0.862 (0.836–0.884)** | **0.114 (0.077–0.150)** | **0.997 (0.996–0.997)** | **0.137 (0.095–0.178)** | **0.136 (0.093–0.177)** |
| Short PR | Scratch CNN–Transformer | 0.226 (0.188–0.268) | 0.906 (0.890–0.919) | 0.105 (0.080–0.136) | **1.000 (0.999–1.000)** | 0.183 (0.141–0.231) | 0.271 (0.222–0.321) |
| | ECG-FM | 0.373 (0.322–0.420) | 0.913 (0.895–0.929) | **0.128 (0.097–0.161)** | **1.000 (1.000–1.000)** | **0.221 (0.172–0.271)** | **0.321 (0.266–0.368)** |
| | DeepECG-SSL | **0.383 (0.332–0.430)** | **0.938 (0.926–0.950)** | 0.121 (0.092–0.152) | **1.000 (1.000–1.000)** | 0.211 (0.166–0.258) | 0.313 (0.262–0.359) |
| | CardioLoom | **0.383 (0.335–0.430)** | 0.904 (0.885–0.921) | 0.050 (0.032–0.073) | **1.000 (1.000–1.000)** | 0.096 (0.061–0.136) | 0.223 (0.177–0.269) |

| Disease | Model | AUPRC | AUROC | Sensitivity | Specificity | F1 Score | MCC |
|---|---|---|---|---|---|---|---|
| LVH | Scratch CNN–Transformer | 0.162 (0.140–0.188) | 0.812 (0.798–0.826) | 0.133 (0.112–0.155) | 0.994 (0.994–0.995) | 0.189 (0.163–0.217) | 0.198 (0.170–0.228) |
| | ECG-FM | 0.235 (0.209–0.264) | **0.837 (0.824–0.851)** | 0.192 (0.168–0.217) | 0.995 (0.994–0.995) | 0.267 (0.238–0.297) | 0.281 (0.251–0.312) |
| | DeepECG-SSL | 0.253 (0.226–0.283) | 0.808 (0.791–0.824) | **0.258 (0.233–0.288)** | 0.991 (0.990–0.992) | **0.308 (0.281–0.338)** | **0.302 (0.275–0.332)** |
| | CardioLoom | **0.269 (0.242–0.299)** | 0.834 (0.819–0.848) | 0.142 (0.121–0.165) | **0.998 (0.998–0.999)** | 0.232 (0.201–0.264) | 0.293 (0.260–0.326) |
| LAH | Scratch CNN–Transformer | 0.486 (0.467–0.504) | 0.932 (0.929–0.935) | 0.270 (0.256–0.286) | 0.988 (0.987–0.989) | 0.376 (0.359–0.394) | 0.383 (0.366–0.401) |
| | ECG-FM | 0.491 (0.474–0.510) | 0.928 (0.925–0.932) | 0.261 (0.248–0.276) | 0.990 (0.989–0.991) | 0.371 (0.355–0.388) | 0.386 (0.368–0.403) |
| | DeepECG-SSL | 0.497 (0.479–0.516) | 0.935 (0.931–0.938) | **0.394 (0.378–0.411)** | 0.979 (0.977–0.980) | **0.464 (0.448–0.478)** | **0.441 (0.424–0.457)** |
| | CardioLoom | **0.514 (0.498–0.534)** | **0.938 (0.934–0.941)** | 0.166 (0.154–0.179) | **0.995 (0.994–0.995)** | 0.267 (0.250–0.284) | 0.316 (0.298–0.334) |
| RVH | Scratch CNN–Transformer | 0.019 (0.012–0.032) | **0.908 (0.887–0.929)** | 0.080 (0.024–0.141) | 0.997 (0.996–0.997) | 0.053 (0.016–0.093) | 0.054 (0.014–0.097) |
| | ECG-FM | 0.031 (0.015–0.063) | 0.823 (0.779–0.867) | **0.091 (0.033–0.156)** | 0.998 (0.998–0.999) | **0.089 (0.034–0.149)** | 0.088 (0.032–0.148) |
| | DeepECG-SSL | **0.035 (0.016–0.071)** | 0.767 (0.709–0.825) | 0.045 (0.010–0.093) | **1.000 (0.999–1.000)** | 0.073 (0.017–0.144) | **0.090 (0.020–0.177)** |
| | CardioLoom | **0.035 (0.017–0.073)** | 0.743 (0.687–0.803) | 0.045 (0.010–0.093) | 0.999 (0.999–1.000) | 0.065 (0.015–0.128) | 0.070 (0.015–0.141) |
| RAH | Scratch CNN–Transformer | 0.059 (0.033–0.103) | **0.886 (0.850–0.917)** | 0.018 (0.000–0.046) | **1.000 (1.000–1.000)** | 0.034 (0.000–0.084) | 0.084 (0.000–0.185) |
| | ECG-FM | 0.047 (0.024–0.095) | 0.842 (0.800–0.876) | 0.009 (0.000–0.032) | **1.000 (1.000–1.000)** | 0.018 (0.000–0.061) | 0.067 (0.000–0.163) |
| | DeepECG-SSL | 0.060 (0.031–0.109) | 0.866 (0.828–0.899) | **0.027 (0.000–0.059)** | **1.000 (1.000–1.000)** | **0.050 (0.000–0.108)** | **0.107 (0.000–0.200)** |
| | CardioLoom | **0.063 (0.023–0.113)** | 0.815 (0.775–0.851) | 0.000 (0.000–0.000) | **1.000 (1.000–1.000)** | 0.000 (0.000–0.000) | 0.000 (0.000–0.000) |

Values are estimates (95% cluster-bootstrap CIs). NE indicates fewer than 950 valid replicates among 1000. Sensitivity, specificity, F1 score, and MCC use the Zhongshan validation-F1 threshold. Bold type marks the highest displayed estimate within each disease.

**Supplementary Table 22: Detailed task-specific performance comparison on the external Chapman cohort.**

| Disease | Model | AUPRC | AUROC | Sensitivity | Specificity | F1 Score | MCC |
|---|---|---|---|---|---|---|---|
| STEMI | Scratch CNN–Transformer | 0.287 (0.206–0.369) | 0.954 (0.925–0.976) | **0.545 (0.457–0.628)** | 0.996 (0.995–0.996) | **0.343 (0.280–0.404)** | **0.367 (0.303–0.426)** |
| | ECG-FM | 0.260 (0.183–0.342) | 0.960 (0.934–0.978) | 0.390 (0.310–0.479) | 0.996 (0.996–0.997) | 0.276 (0.215–0.337) | 0.286 (0.226–0.348) |
| | DeepECG-SSL | **0.310 (0.230–0.390)** | 0.970 (0.952–0.983) | 0.520 (0.435–0.610) | 0.995 (0.994–0.995) | 0.300 (0.245–0.354) | 0.329 (0.273–0.384) |
| | CardioLoom | 0.255 (0.185–0.339) | **0.973 (0.955–0.985)** | 0.366 (0.289–0.455) | **0.997 (0.997–0.998)** | 0.305 (0.235–0.370) | 0.307 (0.240–0.373) |
| SBrad | Scratch CNN–Transformer | **0.998 (0.998–0.999)** | **0.999 (0.999–0.999)** | **0.982 (0.980–0.984)** | **0.994 (0.993–0.995)** | **0.985 (0.984–0.987)** | **0.977 (0.975–0.979)** |
| | ECG-FM | **0.998 (0.997–0.998)** | **0.999 (0.999–0.999)** | 0.975 (0.973–0.977) | **0.994 (0.993–0.995)** | 0.982 (0.981–0.983) | 0.972 (0.970–0.974) |
| | DeepECG-SSL | 0.997 (0.997–0.998) | **0.999 (0.999–0.999)** | 0.980 (0.978–0.982) | 0.993 (0.993–0.994) | 0.984 (0.983–0.986) | 0.975 (0.973–0.978) |
| | CardioLoom | **0.998 (0.997–0.998)** | **0.999 (0.999–0.999)** | **0.982 (0.980–0.984)** | 0.992 (0.991–0.993) | **0.985 (0.983–0.986)** | 0.976 (0.973–0.978) |
| STach | Scratch CNN–Transformer | 0.982 (0.980–0.984) | 0.996 (0.995–0.996) | 0.838 (0.829–0.846) | **0.998 (0.997–0.998)** | 0.906 (0.901–0.911) | 0.893 (0.888–0.899) |
| | ECG-FM | 0.982 (0.979–0.985) | **0.997 (0.997–0.998)** | **0.942 (0.937–0.947)** | 0.993 (0.993–0.994) | **0.953 (0.950–0.957)** | **0.945 (0.941–0.948)** |
| | DeepECG-SSL | 0.985 (0.983–0.988) | **0.997 (0.997–0.998)** | 0.911 (0.905–0.918) | 0.995 (0.994–0.996) | 0.940 (0.936–0.944) | 0.930 (0.925–0.935) |
| | CardioLoom | **0.986 (0.983–0.988)** | **0.997 (0.997–0.998)** | 0.891 (0.884–0.898) | 0.996 (0.995–0.996) | 0.932 (0.927–0.936) | 0.921 (0.916–0.925) |
| VPB | Scratch CNN–Transformer | 0.210 (0.182–0.245) | 0.970 (0.960–0.978) | 0.845 (0.806–0.884) | 0.972 (0.970–0.973) | 0.286 (0.260–0.313) | 0.374 (0.349–0.398) |
| | ECG-FM | 0.210 (0.181–0.249) | 0.966 (0.954–0.976) | 0.880 (0.845–0.919) | **0.973 (0.972–0.975)** | **0.305 (0.278–0.332)** | **0.396 (0.371–0.421)** |
| | DeepECG-SSL | 0.222 (0.193–0.256) | 0.965 (0.952–0.976) | 0.890 (0.856–0.924) | **0.973 (0.971–0.974)** | 0.303 (0.278–0.330) | **0.396 (0.374–0.421)** |
| | CardioLoom | **0.236 (0.204–0.276)** | **0.972 (0.962–0.980)** | **0.909 (0.878–0.941)** | 0.970 (0.969–0.972) | 0.294 (0.269–0.320) | 0.392 (0.369–0.416) |
| APB | Scratch CNN–Transformer | **0.001 (NE)** | 0.959 (NE) | **0.667 (NE)** | 0.965 (NE) | **0.003 (NE)** | **0.028 (NE)** |
| | ECG-FM | **0.001 (NE)** | **0.969 (NE)** | 0.333 (NE) | 0.968 (NE) | 0.001 (NE) | 0.014 (NE) |
| | DeepECG-SSL | 0.000 (NE) | 0.917 (NE) | 0.333 (NE) | **0.970 (NE)** | 0.001 (NE) | 0.015 (NE) |
| | CardioLoom | **0.001 (NE)** | 0.965 (NE) | 0.333 (NE) | 0.968 (NE) | 0.001 (NE) | 0.014 (NE) |

| Disease | Model | AUPRC | AUROC | Sensitivity | Specificity | F1 Score | MCC |
|---|---|---|---|---|---|---|---|
| SArr | Scratch CNN–Transformer | 0.766 (0.751–0.781) | 0.959 (0.955–0.963) | 0.385 (0.368–0.404) | **0.998 (0.998–0.999)** | 0.544 (0.525–0.563) | 0.584 (0.568–0.600) |
| | ECG-FM | 0.773 (0.758–0.788) | 0.963 (0.960–0.967) | **0.503 (0.484–0.522)** | 0.996 (0.996–0.997) | 0.642 (0.625–0.658) | 0.654 (0.639–0.669) |
| | DeepECG-SSL | **0.838 (0.825–0.851)** | 0.977 (0.974–0.979) | 0.501 (0.483–0.520) | **0.998 (0.997–0.998)** | **0.650 (0.634–0.667)** | **0.668 (0.654–0.684)** |
| | CardioLoom | 0.830 (0.817–0.843) | **0.980 (0.978–0.983)** | 0.479 (0.461–0.499) | **0.998 (0.997–0.998)** | 0.632 (0.615–0.649) | 0.654 (0.639–0.669) |
| AF | Scratch CNN–Transformer | 0.193 (0.182–0.206) | 0.902 (0.898–0.906) | 0.573 (0.551–0.597) | **0.908 (0.905–0.910)** | 0.301 (0.287–0.315) | 0.298 (0.284–0.314) |
| | ECG-FM | 0.187 (0.177–0.199) | 0.905 (0.901–0.909) | **0.596 (0.574–0.619)** | 0.907 (0.904–0.909) | **0.308 (0.294–0.323)** | **0.309 (0.294–0.324)** |
| | DeepECG-SSL | **0.203 (0.191–0.216)** | **0.911 (0.908–0.915)** | 0.592 (0.570–0.614) | 0.904 (0.902–0.907) | 0.302 (0.288–0.315) | 0.302 (0.288–0.316) |
| | CardioLoom | 0.200 (0.188–0.214) | 0.908 (0.905–0.912) | **0.596 (0.574–0.618)** | 0.902 (0.900–0.905) | 0.300 (0.287–0.314) | 0.302 (0.287–0.316) |
| AFL | Scratch CNN–Transformer | 0.621 (0.610–0.631) | 0.910 (0.907–0.913) | 0.123 (0.115–0.130) | 0.996 (0.995–0.996) | 0.215 (0.203–0.226) | 0.288 (0.277–0.299) |
| | ECG-FM | **0.716 (0.705–0.725)** | **0.942 (0.940–0.944)** | **0.146 (0.139–0.154)** | 0.997 (0.997–0.998) | **0.253 (0.241–0.264)** | **0.332 (0.321–0.343)** |
| | DeepECG-SSL | 0.610 (0.601–0.621) | 0.876 (0.873–0.880) | 0.133 (0.126–0.140) | **0.999 (0.999–0.999)** | 0.234 (0.223–0.245) | 0.327 (0.317–0.336) |
| | CardioLoom | 0.689 (0.679–0.699) | 0.934 (0.932–0.937) | 0.144 (0.136–0.151) | 0.998 (0.997–0.998) | 0.249 (0.237–0.260) | 0.330 (0.320–0.341) |
| JPB | Scratch CNN–Transformer | 0.009 (0.002–0.030) | 0.932 (0.875–0.984) | 0.182 (0.000–0.445) | **0.999 (0.998–0.999)** | **0.060 (0.000–0.146)** | 0.080 (-0.001–0.190) |
| | ECG-FM | 0.010 (0.002–0.030) | 0.895 (0.717–0.987) | **0.273 (0.000–0.571)** | 0.998 (0.997–0.998) | 0.052 (0.000–0.112) | **0.088 (-0.001–0.180)** |
| | DeepECG-SSL | 0.009 (0.003–0.024) | 0.976 (0.958–0.992) | 0.091 (0.000–0.300) | 0.998 (0.997–0.998) | 0.018 (0.000–0.061) | 0.030 (-0.001–0.095) |
| | CardioLoom | **0.022 (0.007–0.050)** | **0.977 (0.938–0.997)** | 0.182 (0.000–0.500) | 0.998 (0.998–0.999) | 0.042 (0.000–0.104) | 0.065 (-0.001–0.157) |
| ATach | Scratch CNN–Transformer | 0.254 (0.202–0.307) | 0.949 (0.940–0.958) | 0.294 (0.241–0.346) | **0.997 (0.997–0.998)** | 0.338 (0.283–0.391) | 0.338 (0.284–0.391) |
| | ECG-FM | 0.396 (0.326–0.458) | **0.966 (0.958–0.974)** | 0.490 (0.432–0.548) | 0.995 (0.994–0.996) | 0.436 (0.388–0.483) | 0.435 (0.386–0.482) |
| | DeepECG-SSL | **0.429 (0.365–0.492)** | 0.963 (0.953–0.972) | **0.514 (0.461–0.566)** | 0.996 (0.996–0.997) | **0.495 (0.447–0.541)** | **0.492 (0.444–0.538)** |
| | CardioLoom | 0.407 (0.345–0.474) | 0.948 (0.935–0.960) | 0.466 (0.408–0.519) | **0.997 (0.997–0.998)** | **0.495 (0.441–0.544)** | **0.492 (0.441–0.542)** |

| Disease | Model | AUPRC | AUROC | Sensitivity | Specificity | F1 Score | MCC |
|---|---|---|---|---|---|---|---|
| JEB | Scratch CNN–Transformer | 0.079 (0.052–0.119) | 0.976 (0.966–0.984) | 0.347 (0.244–0.461) | 0.995 (0.994–0.996) | 0.161 (0.109–0.218) | 0.189 (0.130–0.251) |
| | ECG-FM | 0.203 (0.139–0.291) | 0.992 (0.989–0.995) | 0.467 (0.361–0.582) | 0.996 (0.996–0.997) | 0.252 (0.184–0.317) | 0.282 (0.212–0.348) |
| | DeepECG-SSL | **0.210 (0.146–0.294)** | 0.993 (0.989–0.996) | 0.533 (0.419–0.641) | **0.997 (0.996–0.997)** | 0.303 (0.229–0.373) | 0.334 (0.258–0.405) |
| | CardioLoom | 0.200 (0.144–0.275) | **0.994 (0.990–0.996)** | **0.627 (0.509–0.732)** | 0.996 (0.996–0.997) | **0.322 (0.245–0.389)** | **0.367 (0.291–0.431)** |
| SVT | Scratch CNN–Transformer | 0.815 (0.783–0.844) | **0.993 (0.991–0.995)** | 0.511 (0.477–0.545) | **0.999 (0.999–0.999)** | 0.650 (0.618–0.679) | 0.672 (0.643–0.697) |
| | ECG-FM | **0.856 (0.828–0.883)** | 0.991 (0.987–0.995) | 0.662 (0.628–0.695) | **0.999 (0.998–0.999)** | 0.764 (0.738–0.788) | 0.770 (0.747–0.793) |
| | DeepECG-SSL | 0.847 (0.819–0.871) | 0.992 (0.988–0.994) | **0.744 (0.713–0.775)** | 0.998 (0.998–0.999) | **0.803 (0.780–0.826)** | **0.802 (0.779–0.825)** |
| | CardioLoom | 0.845 (0.814–0.872) | 0.992 (0.988–0.994) | 0.681 (0.649–0.716) | **0.999 (0.998–0.999)** | 0.777 (0.752–0.803) | 0.782 (0.759–0.806) |
| VEB | Scratch CNN–Transformer | 0.165 (0.094–0.275) | **0.963 (0.940–0.980)** | 0.270 (0.167–0.373) | **0.999 (0.998–0.999)** | 0.246 (0.153–0.336) | 0.246 (0.154–0.337) |
| | ECG-FM | 0.280 (0.177–0.399) | 0.942 (0.899–0.977) | **0.508 (0.389–0.629)** | 0.998 (0.997–0.998) | 0.322 (0.233–0.406) | 0.344 (0.260–0.433) |
| | DeepECG-SSL | 0.285 (0.172–0.406) | 0.944 (0.905–0.975) | 0.460 (0.340–0.587) | **0.999 (0.998–0.999)** | 0.395 (0.288–0.489) | 0.398 (0.292–0.492) |
| | CardioLoom | **0.376 (0.254–0.506)** | 0.918 (0.861–0.965) | 0.476 (0.348–0.596) | **0.999 (0.998–0.999)** | **0.400 (0.300–0.500)** | **0.404 (0.304–0.505)** |
| RBBB | Scratch CNN–Transformer | 0.352 (0.318–0.383) | **0.984 (0.982–0.986)** | **0.958 (0.941–0.972)** | 0.954 (0.952–0.956) | 0.375 (0.353–0.395) | 0.460 (0.442–0.477) |
| | ECG-FM | 0.328 (0.299–0.357) | 0.981 (0.978–0.984) | 0.949 (0.931–0.966) | **0.959 (0.957–0.960)** | **0.397 (0.373–0.417)** | **0.476 (0.457–0.493)** |
| | DeepECG-SSL | 0.341 (0.310–0.376) | 0.982 (0.978–0.984) | 0.949 (0.932–0.965) | 0.956 (0.954–0.958) | 0.382 (0.360–0.403) | 0.465 (0.447–0.482) |
| | CardioLoom | **0.355 (0.323–0.390)** | 0.983 (0.981–0.985) | 0.940 (0.921–0.956) | **0.959 (0.957–0.961)** | 0.396 (0.372–0.418) | 0.474 (0.455–0.491) |
| 1° AVB | Scratch CNN–Transformer | 0.666 (0.642–0.689) | 0.930 (0.922–0.939) | 0.590 (0.566–0.617) | 0.992 (0.991–0.993) | 0.641 (0.620–0.663) | 0.634 (0.612–0.656) |
| | ECG-FM | 0.689 (0.666–0.715) | **0.932 (0.924–0.939)** | 0.544 (0.519–0.572) | **0.996 (0.995–0.996)** | 0.649 (0.627–0.672) | 0.653 (0.632–0.675) |
| | DeepECG-SSL | 0.676 (0.653–0.701) | 0.909 (0.898–0.920) | **0.638 (0.614–0.666)** | 0.991 (0.990–0.992) | 0.664 (0.644–0.685) | 0.654 (0.634–0.676) |
| | CardioLoom | **0.690 (0.668–0.715)** | 0.898 (0.887–0.910) | 0.588 (0.561–0.613) | 0.994 (0.994–0.995) | **0.667 (0.646–0.689)** | **0.664 (0.644–0.685)** |

| Disease | Model | AUPRC | AUROC | Sensitivity | Specificity | F1 Score | MCC |
|---|---|---|---|---|---|---|---|
| QT Prolong | Scratch CNN–Transformer | 0.172 (0.137–0.211) | 0.922 (0.912–0.933) | 0.198 (0.157–0.235) | 0.995 (0.994–0.996) | 0.224 (0.182–0.264) | 0.221 (0.177–0.261) |
| | ECG-FM | 0.210 (0.170–0.254) | 0.950 (0.943–0.956) | 0.201 (0.159–0.241) | 0.996 (0.995–0.996) | 0.237 (0.190–0.278) | 0.236 (0.188–0.276) |
| | DeepECG-SSL | **0.269 (0.225–0.318)** | **0.951 (0.944–0.959)** | **0.305 (0.260–0.349)** | 0.994 (0.994–0.995) | **0.312 (0.270–0.355)** | **0.307 (0.264–0.349)** |
| | CardioLoom | 0.236 (0.194–0.280) | **0.951 (0.943–0.958)** | 0.157 (0.120–0.193) | **0.998 (0.997–0.998)** | 0.220 (0.172–0.264) | 0.235 (0.185–0.283) |
| ER | Scratch CNN–Transformer | 0.230 (0.187–0.273) | 0.957 (0.948–0.965) | 0.181 (0.143–0.220) | **0.998 (0.997–0.998)** | 0.246 (0.196–0.293) | 0.259 (0.207–0.308) |
| | ECG-FM | 0.256 (0.207–0.304) | 0.959 (0.950–0.967) | 0.197 (0.157–0.241) | 0.997 (0.997–0.998) | 0.262 (0.213–0.313) | 0.274 (0.221–0.326) |
| | DeepECG-SSL | **0.265 (0.222–0.314)** | **0.964 (0.955–0.971)** | **0.389 (0.337–0.440)** | 0.992 (0.991–0.993) | **0.323 (0.282–0.362)** | **0.322 (0.280–0.360)** |
| | CardioLoom | 0.252 (0.207–0.298) | 0.962 (0.953–0.970) | 0.162 (0.123–0.202) | **0.998 (0.997–0.998)** | 0.227 (0.179–0.275) | 0.245 (0.193–0.296) |
| LAFB | Scratch CNN–Transformer | 0.059 (0.042–0.081) | 0.852 (0.827–0.875) | 0.138 (0.096–0.178) | **0.994 (0.993–0.995)** | 0.121 (0.083–0.156) | 0.116 (0.079–0.152) |
| | ECG-FM | **0.071 (0.048–0.101)** | **0.866 (0.844–0.887)** | **0.179 (0.133–0.225)** | 0.993 (0.992–0.994) | **0.144 (0.107–0.179)** | **0.142 (0.103–0.178)** |
| | DeepECG-SSL | 0.062 (0.043–0.089) | 0.750 (0.710–0.789) | 0.171 (0.121–0.220) | 0.992 (0.991–0.993) | 0.130 (0.093–0.164) | 0.128 (0.090–0.164) |
| | CardioLoom | 0.064 (0.046–0.089) | 0.816 (0.782–0.849) | 0.146 (0.102–0.190) | **0.994 (0.993–0.995)** | 0.127 (0.089–0.166) | 0.123 (0.085–0.163) |
| IVB | Scratch CNN–Transformer | 0.048 (0.043–0.055) | 0.785 (0.772–0.798) | 0.029 (0.018–0.042) | **0.996 (0.995–0.996)** | 0.045 (0.028–0.065) | 0.047 (0.027–0.071) |
| | ECG-FM | **0.050 (0.044–0.057)** | **0.792 (0.779–0.804)** | 0.031 (0.020–0.043) | **0.996 (0.995–0.996)** | 0.049 (0.032–0.067) | **0.050 (0.029–0.072)** |
| | DeepECG-SSL | **0.050 (0.044–0.057)** | 0.768 (0.753–0.783) | 0.030 (0.018–0.043) | 0.995 (0.995–0.996) | 0.046 (0.029–0.067) | 0.047 (0.026–0.071) |
| | CardioLoom | 0.045 (0.039–0.052) | 0.751 (0.737–0.767) | **0.034 (0.022–0.048)** | 0.994 (0.994–0.995) | **0.050 (0.033–0.070)** | 0.047 (0.028–0.070) |
| VPE | Scratch CNN–Transformer | 0.603 (0.495–0.721) | 0.982 (0.970–0.992) | 0.528 (0.416–0.646) | **1.000 (1.000–1.000)** | 0.655 (0.552–0.750) | 0.675 (0.581–0.762) |
| | ECG-FM | 0.771 (0.680–0.859) | **0.989 (0.976–0.998)** | **0.667 (0.557–0.774)** | **1.000 (1.000–1.000)** | **0.733 (0.645–0.812)** | **0.736 (0.648–0.816)** |
| | DeepECG-SSL | 0.754 (0.655–0.848) | 0.982 (0.957–0.998) | 0.625 (0.513–0.739) | **1.000 (1.000–1.000)** | 0.726 (0.634–0.813) | 0.735 (0.650–0.818) |
| | CardioLoom | **0.783 (0.695–0.872)** | 0.980 (0.954–0.997) | 0.625 (0.517–0.731) | **1.000 (1.000–1.000)** | 0.726 (0.633–0.805) | 0.735 (0.650–0.810) |

| Disease | Model | AUPRC | AUROC | Sensitivity | Specificity | F1 Score | MCC |
|---|---|---|---|---|---|---|---|
| 3° AVB | Scratch CNN–Transformer | 0.101 (0.068–0.139) | **0.832 (0.811–0.851)** | 0.141 (0.100–0.180) | 0.996 (0.995–0.996) | 0.161 (0.116–0.202) | 0.158 (0.113–0.200) |
| | ECG-FM | 0.122 (0.082–0.165) | 0.805 (0.782–0.827) | 0.134 (0.095–0.173) | 0.997 (0.997–0.998) | 0.176 (0.126–0.222) | 0.181 (0.129–0.228) |
| | DeepECG-SSL | **0.155 (0.108–0.204)** | 0.822 (0.802–0.840) | **0.163 (0.120–0.205)** | 0.998 (0.997–0.998) | 0.221 (0.165–0.272) | 0.234 (0.175–0.286) |
| | CardioLoom | 0.150 (0.102–0.197) | 0.825 (0.807–0.843) | 0.159 (0.118–0.205) | **0.999 (0.998–0.999)** | **0.240 (0.180–0.298)** | **0.275 (0.211–0.331)** |
| 2° 1 Type AVB | Scratch CNN–Transformer | 0.073 (0.050–0.103) | **0.825 (0.808–0.843)** | 0.032 (0.015–0.054) | **1.000 (1.000–1.000)** | 0.062 (0.029–0.100) | 0.140 (0.077–0.203) |
| | ECG-FM | 0.091 (0.062–0.130) | 0.712 (0.687–0.739) | 0.032 (0.017–0.054) | **1.000 (1.000–1.000)** | 0.062 (0.032–0.100) | 0.144 (0.083–0.208) |
| | DeepECG-SSL | **0.136 (0.100–0.179)** | 0.744 (0.715–0.772) | **0.050 (0.029–0.074)** | **1.000 (1.000–1.000)** | **0.095 (0.055–0.137)** | **0.205 (0.146–0.262)** |
| | CardioLoom | 0.103 (0.072–0.143) | 0.619 (0.587–0.655) | 0.041 (0.021–0.065) | **1.000 (1.000–1.000)** | 0.078 (0.041–0.121) | 0.178 (0.113–0.236) |
| 2° 2 Type AVB | Scratch CNN–Transformer | 0.026 (0.016–0.044) | 0.595 (0.564–0.628) | 0.010 (0.000–0.022) | **1.000 (1.000–1.000)** | 0.019 (0.000–0.043) | **0.069 (-0.001–0.127)** |
| | ECG-FM | 0.034 (0.019–0.056) | 0.611 (0.583–0.643) | **0.013 (0.003–0.027)** | **1.000 (1.000–1.000)** | **0.025 (0.006–0.051)** | 0.060 (0.013–0.113) |
| | DeepECG-SSL | **0.036 (0.021–0.057)** | **0.653 (0.620–0.684)** | 0.010 (0.000–0.022) | **1.000 (1.000–1.000)** | 0.019 (0.000–0.042) | 0.056 (-0.001–0.111) |
| | CardioLoom | 0.022 (0.012–0.042) | 0.467 (0.431–0.504) | **0.013 (0.003–0.027)** | **1.000 (1.000–1.000)** | **0.025 (0.006–0.052)** | 0.068 (0.015–0.125) |
| LVH | Scratch CNN–Transformer | 0.263 (0.227–0.297) | 0.939 (0.930–0.946) | 0.628 (0.589–0.662) | 0.958 (0.957–0.960) | 0.279 (0.257–0.301) | 0.319 (0.296–0.340) |
| | ECG-FM | 0.284 (0.248–0.320) | 0.945 (0.938–0.951) | 0.636 (0.598–0.673) | 0.958 (0.956–0.960) | 0.281 (0.259–0.303) | 0.322 (0.299–0.344) |
| | DeepECG-SSL | 0.281 (0.243–0.317) | 0.936 (0.927–0.944) | **0.687 (0.649–0.722)** | 0.942 (0.940–0.945) | 0.242 (0.223–0.261) | 0.299 (0.278–0.319) |
| | CardioLoom | **0.299 (0.262–0.336)** | **0.950 (0.943–0.956)** | 0.659 (0.624–0.695) | **0.961 (0.959–0.963)** | **0.304 (0.280–0.327)** | **0.345 (0.320–0.368)** |
| RVH | Scratch CNN–Transformer | 0.236 (0.171–0.326) | 0.983 (0.977–0.989) | **0.573 (0.483–0.664)** | 0.995 (0.995–0.996) | **0.333 (0.273–0.393)** | **0.365 (0.304–0.426)** |
| | ECG-FM | **0.253 (0.179–0.339)** | **0.988 (0.983–0.992)** | 0.536 (0.440–0.630) | 0.995 (0.995–0.996) | 0.306 (0.245–0.360) | 0.336 (0.271–0.392) |
| | DeepECG-SSL | 0.175 (0.123–0.240) | 0.981 (0.975–0.987) | 0.473 (0.382–0.560) | **0.996 (0.995–0.996)** | 0.293 (0.235–0.351) | 0.314 (0.253–0.373) |
| | CardioLoom | 0.220 (0.154–0.298) | 0.985 (0.979–0.991) | 0.427 (0.333–0.520) | **0.996 (0.996–0.997)** | 0.294 (0.230–0.357) | 0.307 (0.239–0.368) |

| Disease | Model | AUPRC | AUROC | Sensitivity | Specificity | F1 Score | MCC |
|---|---|---|---|---|---|---|---|
| RAH | Scratch CNN–Transformer | 0.138 (0.066–0.242) | 0.963 (0.915–0.992) | 0.056 (0.000–0.140) | **1.000 (1.000–1.000)** | 0.083 (0.000–0.196) | 0.096 (0.000–0.226) |
| | ECG-FM | 0.217 (0.113–0.360) | 0.966 (0.909–0.994) | **0.444 (0.281–0.611)** | 0.999 (0.999–0.999) | **0.323 (0.205–0.438)** | **0.335 (0.214–0.452)** |
| | DeepECG-SSL | 0.192 (0.105–0.338) | 0.974 (0.926–0.996) | 0.250 (0.108–0.405) | **1.000 (0.999–1.000)** | 0.286 (0.129–0.435) | 0.288 (0.133–0.438) |
| | CardioLoom | **0.280 (0.146–0.438)** | **0.986 (0.962–0.998)** | 0.083 (0.000–0.184) | **1.000 (1.000–1.000)** | 0.136 (0.000–0.281) | 0.176 (0.000–0.344) |
| LK | Scratch CNN–Transformer | 0.029 (0.020–0.047) | 0.894 (0.874–0.915) | **0.096 (0.051–0.150)** | 0.995 (0.994–0.995) | 0.067 (0.035–0.105) | 0.066 (0.033–0.105) |
| | ECG-FM | 0.034 (0.022–0.052) | 0.875 (0.854–0.897) | 0.007 (0.000–0.025) | **1.000 (0.999–1.000)** | 0.013 (0.000–0.043) | 0.019 (-0.001–0.064) |
| | DeepECG-SSL | 0.054 (0.036–0.087) | 0.914 (0.888–0.936) | 0.051 (0.016–0.094) | 0.999 (0.999–0.999) | **0.079 (0.024–0.140)** | **0.092 (0.028–0.162)** |
| | CardioLoom | **0.061 (0.041–0.094)** | **0.928 (0.908–0.948)** | 0.037 (0.008–0.071) | 0.999 (0.999–0.999) | 0.057 (0.012–0.107) | 0.066 (0.013–0.127) |

Values are estimates (95% cluster-bootstrap CIs). NE indicates fewer than 950 valid replicates among 1000. Sensitivity, specificity, F1 score, and MCC use the Zhongshan validation-F1 threshold. Bold type marks the highest displayed estimate within each disease.

**Supplementary Table 23: Detailed task-specific performance comparison on the external PTB-XL cohort.**

| Disease | Model | AUPRC | AUROC | Sensitivity | Specificity | F1 Score | MCC |
|---|---|---|---|---|---|---|---|
| STEMI | Scratch CNN–Transformer | 0.642 (0.626–0.657) | 0.830 (0.823–0.837) | 0.036 (0.030–0.042) | 0.998 (0.998–0.999) | 0.068 (0.057–0.079) | 0.148 (0.132–0.164) |
| | ECG-FM | 0.731 (0.717–0.743) | 0.850 (0.843–0.857) | 0.036 (0.030–0.042) | **1.000 (1.000–1.000)** | 0.069 (0.059–0.081) | 0.163 (0.149–0.177) |
| | DeepECG-SSL | 0.711 (0.698–0.723) | 0.843 (0.837–0.850) | **0.046 (0.039–0.053)** | **1.000 (1.000–1.000)** | **0.087 (0.075–0.101)** | **0.185 (0.171–0.200)** |
| | CardioLoom | **0.745 (0.732–0.757)** | **0.855 (0.847–0.861)** | 0.025 (0.020–0.029) | **1.000 (1.000–1.000)** | 0.048 (0.039–0.057) | 0.135 (0.121–0.148) |
| NSR | Scratch CNN–Transformer | 0.533 (0.522–0.545) | 0.593 (0.585–0.601) | 0.743 (0.734–0.752) | 0.359 (0.349–0.369) | 0.578 (0.570–0.586) | 0.109 (0.095–0.122) |
| | ECG-FM | 0.539 (0.528–0.551) | 0.601 (0.593–0.609) | 0.744 (0.735–0.753) | 0.358 (0.348–0.368) | 0.578 (0.571–0.586) | 0.109 (0.095–0.123) |
| | DeepECG-SSL | **0.589 (0.577–0.600)** | **0.622 (0.614–0.630)** | **0.754 (0.745–0.763)** | 0.359 (0.350–0.370) | **0.584 (0.577–0.592)** | **0.122 (0.107–0.136)** |
| | CardioLoom | 0.549 (0.538–0.561) | 0.593 (0.585–0.602) | 0.736 (0.726–0.745) | **0.367 (0.357–0.377)** | 0.576 (0.569–0.584) | 0.109 (0.094–0.123) |
| SBrad | Scratch CNN–Transformer | 0.244 (0.221–0.271) | 0.944 (0.939–0.949) | 0.961 (0.945–0.975) | 0.855 (0.850–0.860) | 0.283 (0.266–0.301) | 0.366 (0.352–0.381) |
| | ECG-FM | 0.316 (0.285–0.353) | 0.950 (0.944–0.955) | 0.964 (0.949–0.979) | **0.856 (0.851–0.861)** | **0.285 (0.268–0.303)** | **0.369 (0.354–0.384)** |
| | DeepECG-SSL | **0.450 (0.407–0.491)** | **0.961 (0.956–0.966)** | 0.964 (0.949–0.978) | 0.850 (0.845–0.855) | 0.277 (0.261–0.294) | 0.362 (0.347–0.376) |
| | CardioLoom | 0.307 (0.274–0.343) | 0.948 (0.942–0.954) | **0.965 (0.951–0.980)** | 0.847 (0.843–0.853) | 0.275 (0.258–0.292) | 0.359 (0.345–0.374) |
| STach | Scratch CNN–Transformer | **0.894 (0.867–0.922)** | **0.992 (0.987–0.996)** | 0.910 (0.889–0.930) | 0.995 (0.994–0.996) | 0.889 (0.872–0.907) | 0.885 (0.867–0.903) |
| | ECG-FM | 0.887 (0.859–0.915) | 0.991 (0.986–0.994) | **0.921 (0.901–0.941)** | 0.994 (0.993–0.995) | 0.887 (0.871–0.905) | 0.884 (0.866–0.902) |
| | DeepECG-SSL | 0.893 (0.865–0.918) | **0.992 (0.987–0.995)** | 0.901 (0.878–0.923) | **0.996 (0.995–0.996)** | **0.895 (0.877–0.911)** | **0.891 (0.872–0.908)** |
| | CardioLoom | 0.885 (0.856–0.912) | 0.988 (0.981–0.993) | 0.912 (0.890–0.932) | 0.995 (0.994–0.996) | 0.893 (0.876–0.910) | 0.889 (0.871–0.907) |
| VPB | Scratch CNN–Transformer | 0.831 (0.807–0.855) | **0.990 (0.988–0.992)** | 0.920 (0.904–0.936) | **0.984 (0.982–0.986)** | 0.835 (0.819–0.851) | 0.829 (0.813–0.846) |
| | ECG-FM | **0.850 (0.826–0.872)** | **0.990 (0.987–0.992)** | 0.953 (0.940–0.965) | **0.984 (0.982–0.986)** | **0.850 (0.836–0.865)** | **0.847 (0.832–0.861)** |
| | DeepECG-SSL | 0.824 (0.798–0.851) | 0.985 (0.981–0.989) | 0.944 (0.930–0.957) | 0.983 (0.981–0.984) | 0.835 (0.820–0.852) | 0.831 (0.816–0.847) |
| | CardioLoom | 0.837 (0.812–0.863) | **0.990 (0.988–0.992)** | **0.954 (0.941–0.965)** | 0.982 (0.981–0.984) | 0.840 (0.824–0.856) | 0.837 (0.822–0.852) |

| Disease | Model | AUPRC | AUROC | Sensitivity | Specificity | F1 Score | MCC |
|---|---|---|---|---|---|---|---|
| APB | Scratch CNN–Transformer | 0.346 (0.304–0.393) | 0.975 (0.970–0.978) | **0.864 (0.832–0.897)** | 0.963 (0.961–0.966) | 0.451 (0.417–0.481) | 0.500 (0.472–0.527) |
| | ECG-FM | 0.358 (0.313–0.405) | 0.962 (0.952–0.970) | 0.849 (0.812–0.884) | 0.965 (0.962–0.968) | 0.456 (0.422–0.487) | 0.501 (0.471–0.529) |
| | DeepECG-SSL | **0.380 (0.336–0.429)** | **0.978 (0.972–0.982)** | **0.864 (0.827–0.897)** | **0.967 (0.965–0.970)** | **0.478 (0.444–0.511)** | **0.522 (0.492–0.551)** |
| | CardioLoom | 0.372 (0.326–0.421) | 0.975 (0.969–0.979) | 0.847 (0.810–0.881) | **0.967 (0.964–0.969)** | 0.465 (0.432–0.497) | 0.508 (0.479–0.537) |
| SArr | Scratch CNN–Transformer | 0.418 (0.378–0.456) | **0.904 (0.894–0.915)** | 0.439 (0.404–0.474) | **0.977 (0.975–0.979)** | 0.427 (0.397–0.457) | 0.405 (0.376–0.437) |
| | ECG-FM | 0.376 (0.337–0.411) | 0.863 (0.848–0.877) | 0.492 (0.455–0.527) | 0.968 (0.966–0.970) | 0.417 (0.388–0.443) | 0.397 (0.367–0.424) |
| | DeepECG-SSL | **0.443 (0.403–0.482)** | 0.889 (0.873–0.903) | **0.517 (0.481–0.553)** | 0.975 (0.973–0.977) | **0.470 (0.438–0.500)** | **0.450 (0.418–0.482)** |
| | CardioLoom | 0.432 (0.395–0.471) | 0.883 (0.866–0.898) | 0.503 (0.466–0.539) | 0.973 (0.970–0.975) | 0.446 (0.419–0.476) | 0.427 (0.398–0.457) |
| AF | Scratch CNN–Transformer | 0.838 (0.819–0.857) | 0.981 (0.979–0.984) | 0.589 (0.561–0.616) | **0.994 (0.993–0.995)** | 0.706 (0.683–0.728) | 0.705 (0.683–0.727) |
| | ECG-FM | 0.809 (0.785–0.832) | 0.977 (0.973–0.980) | **0.603 (0.578–0.630)** | **0.994 (0.993–0.996)** | **0.718 (0.698–0.740)** | **0.716 (0.697–0.737)** |
| | DeepECG-SSL | 0.839 (0.817–0.861) | **0.983 (0.979–0.986)** | 0.578 (0.551–0.605) | 0.993 (0.992–0.995) | 0.693 (0.671–0.715) | 0.691 (0.669–0.713) |
| | CardioLoom | **0.846 (0.826–0.865)** | 0.981 (0.978–0.984) | 0.602 (0.576–0.628) | **0.994 (0.992–0.995)** | 0.713 (0.693–0.735) | 0.710 (0.690–0.731) |
| AFL | Scratch CNN–Transformer | 0.589 (0.412–0.732) | 0.981 (0.954–0.998) | 0.849 (0.750–0.920) | **0.996 (0.995–0.997)** | 0.537 (0.422–0.630) | 0.575 (0.473–0.656) |
| | ECG-FM | 0.609 (0.413–0.753) | 0.987 (0.969–0.998) | 0.877 (0.793–0.944) | **0.996 (0.995–0.997)** | **0.554 (0.450–0.643)** | **0.594 (0.503–0.668)** |
| | DeepECG-SSL | 0.615 (0.445–0.737) | **0.990 (0.977–0.998)** | 0.781 (0.662–0.867) | **0.996 (0.995–0.997)** | 0.540 (0.425–0.635) | 0.566 (0.455–0.654) |
| | CardioLoom | **0.620 (0.430–0.758)** | 0.987 (0.965–0.999) | **0.890 (0.809–0.953)** | 0.995 (0.994–0.996) | 0.546 (0.444–0.635) | 0.590 (0.501–0.665) |
| AJR | Scratch CNN–Transformer | **0.008 (0.004–0.014)** | **0.908 (0.868–0.939)** | **0.000 (0.000–0.000)** | 0.993 (0.992–0.995) | **0.000 (0.000–0.000)** | -0.003 (-0.003–-0.002) |
| | ECG-FM | **0.008 (0.004–0.015)** | 0.866 (0.787–0.927) | **0.000 (0.000–0.000)** | 0.994 (0.993–0.995) | **0.000 (0.000–0.000)** | -0.003 (-0.003–-0.002) |
| | DeepECG-SSL | **0.008 (0.004–0.013)** | 0.892 (0.851–0.928) | **0.000 (0.000–0.000)** | **0.995 (0.995–0.996)** | **0.000 (0.000–0.000)** | **-0.002 (-0.003–-0.002)** |
| | CardioLoom | 0.007 (0.003–0.011) | 0.832 (0.753–0.897) | **0.000 (0.000–0.000)** | 0.994 (0.993–0.995) | **0.000 (0.000–0.000)** | -0.003 (-0.003–-0.002) |

| Disease | Model | AUPRC | AUROC | Sensitivity | Specificity | F1 Score | MCC |
|---|---|---|---|---|---|---|---|
| ATach | Scratch CNN–Transformer | 0.012 (0.006–0.024) | 0.928 (0.877–0.957) | 0.074 (0.000–0.192) | **0.995 (0.994–0.996)** | 0.030 (0.000–0.076) | 0.035 (-0.003–0.090) |
| | ECG-FM | 0.008 (0.004–0.016) | 0.832 (0.739–0.908) | 0.037 (0.000–0.121) | 0.994 (0.993–0.995) | 0.012 (0.000–0.040) | 0.014 (-0.003–0.052) |
| | DeepECG-SSL | **0.019 (0.007–0.082)** | **0.929 (0.895–0.956)** | **0.111 (0.000–0.243)** | 0.994 (0.993–0.995) | **0.038 (0.000–0.083)** | **0.048 (-0.002–0.107)** |
| | CardioLoom | 0.009 (0.004–0.024) | 0.792 (0.675–0.891) | 0.037 (0.000–0.121) | **0.995 (0.994–0.996)** | 0.015 (0.000–0.048) | 0.016 (-0.003–0.058) |
| JTach | Scratch CNN–Transformer | 0.022 (0.011–0.042) | **0.947 (0.910–0.974)** | 0.037 (0.000–0.120) | **0.998 (0.997–0.998)** | 0.024 (0.000–0.078) | 0.024 (-0.002–0.081) |
| | ECG-FM | **0.027 (0.011–0.054)** | 0.887 (0.800–0.954) | **0.074 (0.000–0.185)** | **0.998 (0.997–0.999)** | **0.056 (0.000–0.135)** | **0.056 (-0.002–0.140)** |
| | DeepECG-SSL | 0.023 (0.011–0.049) | 0.927 (0.859–0.973) | **0.074 (0.000–0.185)** | 0.997 (0.997–0.998) | 0.047 (0.000–0.112) | 0.048 (-0.002–0.122) |
| | CardioLoom | 0.021 (0.008–0.042) | 0.851 (0.748–0.932) | **0.074 (0.000–0.185)** | 0.997 (0.996–0.998) | 0.043 (0.000–0.105) | 0.045 (-0.002–0.116) |
| SVT | Scratch CNN–Transformer | 0.103 (0.059–0.152) | 0.682 (0.637–0.730) | 0.039 (0.015–0.069) | **1.000 (1.000–1.000)** | 0.074 (0.029–0.128) | 0.173 (0.084–0.251) |
| | ECG-FM | 0.100 (0.059–0.152) | **0.771 (0.734–0.809)** | 0.034 (0.011–0.063) | **1.000 (0.999–1.000)** | 0.062 (0.021–0.115) | 0.118 (0.042–0.200) |
| | DeepECG-SSL | **0.109 (0.064–0.159)** | 0.727 (0.685–0.768) | **0.061 (0.029–0.098)** | **1.000 (0.999–1.000)** | **0.111 (0.054–0.172)** | **0.187 (0.104–0.261)** |
| | CardioLoom | 0.105 (0.062–0.155) | 0.761 (0.723–0.800) | 0.050 (0.021–0.083) | **1.000 (1.000–1.000)** | 0.093 (0.040–0.149) | 0.178 (0.089–0.254) |
| RBBB | Scratch CNN–Transformer | 0.860 (0.845–0.875) | 0.971 (0.966–0.976) | **0.638 (0.614–0.665)** | 0.994 (0.993–0.995) | **0.746 (0.727–0.765)** | **0.741 (0.723–0.759)** |
| | ECG-FM | **0.873 (0.856–0.888)** | **0.981 (0.977–0.984)** | 0.628 (0.602–0.653) | **0.995 (0.994–0.996)** | 0.744 (0.724–0.763) | **0.741 (0.722–0.759)** |
| | DeepECG-SSL | 0.846 (0.829–0.862) | 0.972 (0.967–0.976) | 0.612 (0.588–0.639) | **0.995 (0.994–0.996)** | 0.731 (0.712–0.751) | 0.729 (0.711–0.747) |
| | CardioLoom | 0.848 (0.831–0.863) | 0.972 (0.967–0.976) | 0.603 (0.577–0.629) | **0.995 (0.993–0.996)** | 0.723 (0.702–0.743) | 0.721 (0.702–0.740) |
| 1° AVB | Scratch CNN–Transformer | **0.506 (0.467–0.548)** | 0.960 (0.954–0.966) | 0.665 (0.631–0.700) | **0.970 (0.968–0.973)** | **0.541 (0.512–0.570)** | **0.531 (0.502–0.561)** |
| | ECG-FM | 0.503 (0.465–0.544) | 0.961 (0.956–0.966) | 0.668 (0.632–0.702) | 0.969 (0.967–0.972) | 0.538 (0.506–0.565) | 0.528 (0.496–0.556) |
| | DeepECG-SSL | 0.489 (0.448–0.531) | 0.960 (0.955–0.966) | **0.692 (0.656–0.728)** | 0.965 (0.962–0.968) | 0.529 (0.499–0.556) | 0.523 (0.493–0.551) |
| | CardioLoom | 0.497 (0.457–0.538) | **0.962 (0.957–0.967)** | **0.692 (0.656–0.729)** | 0.967 (0.964–0.969) | 0.538 (0.509–0.567) | **0.531 (0.503–0.560)** |

| Disease | Model | AUPRC | AUROC | Sensitivity | Specificity | F1 Score | MCC |
|---|---|---|---|---|---|---|---|
| QT Prolong | Scratch CNN–Transformer | 0.083 (0.047–0.151) | 0.858 (0.824–0.890) | 0.214 (0.133–0.292) | 0.994 (0.993–0.995) | **0.181 (0.113–0.248)** | **0.178 (0.110–0.246)** |
| | ECG-FM | 0.092 (0.052–0.147) | 0.861 (0.821–0.896) | **0.222 (0.146–0.296)** | 0.993 (0.991–0.994) | 0.173 (0.114–0.231) | 0.172 (0.112–0.231) |
| | DeepECG-SSL | 0.076 (0.045–0.131) | 0.856 (0.816–0.893) | **0.222 (0.140–0.311)** | 0.991 (0.989–0.992) | 0.154 (0.097–0.215) | 0.156 (0.096–0.219) |
| | CardioLoom | **0.095 (0.053–0.149)** | **0.878 (0.839–0.909)** | 0.145 (0.088–0.208) | **0.995 (0.994–0.996)** | 0.145 (0.086–0.203) | 0.140 (0.082–0.199) |
| LAFB | Scratch CNN–Transformer | 0.649 (0.623–0.674) | 0.936 (0.930–0.941) | 0.198 (0.179–0.219) | **0.998 (0.997–0.998)** | 0.322 (0.296–0.350) | 0.398 (0.373–0.424) |
| | ECG-FM | 0.663 (0.637–0.685) | 0.936 (0.930–0.943) | **0.214 (0.195–0.236)** | **0.998 (0.997–0.998)** | **0.345 (0.319–0.374)** | **0.419 (0.396–0.444)** |
| | DeepECG-SSL | **0.716 (0.691–0.737)** | **0.952 (0.947–0.958)** | 0.209 (0.189–0.230) | **0.998 (0.997–0.999)** | 0.340 (0.312–0.367) | 0.417 (0.392–0.441) |
| | CardioLoom | 0.636 (0.610–0.662) | 0.908 (0.899–0.917) | 0.197 (0.178–0.219) | **0.998 (0.998–0.999)** | 0.324 (0.297–0.353) | 0.406 (0.382–0.431) |
| LBBB | Scratch CNN–Transformer | 0.925 (0.906–0.942) | 0.989 (0.984–0.994) | 0.803 (0.768–0.838) | 0.998 (0.998–0.999) | 0.865 (0.842–0.887) | 0.864 (0.842–0.886) |
| | ECG-FM | 0.925 (0.905–0.944) | **0.990 (0.985–0.995)** | 0.772 (0.733–0.806) | **0.999 (0.998–0.999)** | 0.852 (0.828–0.876) | 0.853 (0.830–0.877) |
| | DeepECG-SSL | 0.908 (0.886–0.929) | 0.984 (0.976–0.990) | **0.816 (0.783–0.847)** | 0.998 (0.998–0.999) | **0.872 (0.849–0.892)** | **0.871 (0.849–0.890)** |
| | CardioLoom | **0.930 (0.912–0.946)** | 0.989 (0.984–0.994) | 0.799 (0.761–0.834) | **0.999 (0.998–0.999)** | 0.866 (0.841–0.888) | 0.865 (0.842–0.887) |
| IVB | Scratch CNN–Transformer | 0.108 (0.092–0.128) | 0.715 (0.695–0.735) | 0.080 (0.061–0.100) | 0.990 (0.989–0.992) | 0.119 (0.092–0.148) | 0.119 (0.087–0.151) |
| | ECG-FM | 0.130 (0.111–0.152) | 0.736 (0.716–0.755) | 0.091 (0.070–0.114) | 0.990 (0.988–0.992) | 0.135 (0.105–0.164) | 0.135 (0.102–0.168) |
| | DeepECG-SSL | 0.127 (0.108–0.150) | 0.718 (0.698–0.738) | 0.080 (0.060–0.101) | **0.992 (0.991–0.993)** | 0.124 (0.094–0.154) | 0.132 (0.098–0.164) |
| | CardioLoom | **0.142 (0.123–0.165)** | **0.768 (0.749–0.786)** | **0.121 (0.096–0.147)** | 0.988 (0.986–0.989) | **0.167 (0.135–0.197)** | **0.160 (0.128–0.193)** |
| VPE | Scratch CNN–Transformer | 0.709 (0.594–0.810) | 0.957 (0.920–0.985) | 0.658 (0.537–0.767) | **1.000 (0.999–1.000)** | 0.743 (0.643–0.821) | 0.748 (0.655–0.824) |
| | ECG-FM | 0.715 (0.587–0.815) | 0.958 (0.924–0.982) | 0.696 (0.564–0.808) | 0.999 (0.999–1.000) | 0.738 (0.627–0.821) | 0.739 (0.629–0.822) |
| | DeepECG-SSL | 0.706 (0.584–0.809) | **0.974 (0.953–0.989)** | 0.696 (0.562–0.806) | 0.999 (0.999–1.000) | 0.753 (0.647–0.835) | 0.755 (0.650–0.836) |
| | CardioLoom | **0.731 (0.608–0.834)** | 0.964 (0.931–0.985) | **0.747 (0.625–0.852)** | 0.999 (0.999–1.000) | **0.771 (0.667–0.848)** | **0.771 (0.667–0.849)** |

| Disease | Model | AUPRC | AUROC | Sensitivity | Specificity | F1 Score | MCC |
|---|---|---|---|---|---|---|---|
| 3° AVB | Scratch CNN–Transformer | 0.302 (0.057–0.527) | **0.963 (0.907–0.994)** | 0.438 (0.214–0.696) | 0.998 (0.997–0.998) | 0.192 (0.075–0.320) | 0.231 (0.098–0.366) |
| | ECG-FM | 0.394 (0.128–0.682) | **0.963 (0.920–0.996)** | 0.500 (0.231–0.765) | **0.999 (0.998–0.999)** | 0.320 (0.136–0.500) | 0.342 (0.154–0.528) |
| | DeepECG-SSL | 0.350 (0.110–0.605) | 0.936 (0.871–0.987) | **0.625 (0.352–0.857)** | **0.999 (0.999–1.000)** | **0.465 (0.245–0.651)** | **0.481 (0.267–0.663)** |
| | CardioLoom | **0.424 (0.153–0.694)** | 0.937 (0.869–0.989) | **0.625 (0.352–0.857)** | **0.999 (0.999–1.000)** | 0.455 (0.232–0.643) | 0.472 (0.253–0.656) |
| 2° 1 Type AVB | Scratch CNN–Transformer | 0.128 (0.006–0.378) | **0.951 (0.918–0.983)** | 0.143 (0.000–0.357) | 0.999 (0.999–1.000) | 0.133 (0.000–0.323) | 0.133 (-0.001–0.322) |
| | ECG-FM | **0.155 (0.018–0.492)** | 0.776 (0.612–0.960) | 0.214 (0.000–0.444) | **1.000 (0.999–1.000)** | 0.250 (0.000–0.480) | 0.253 (0.000–0.500) |
| | DeepECG-SSL | 0.147 (0.007–0.419) | 0.881 (0.740–0.972) | **0.286 (0.000–0.625)** | **1.000 (0.999–1.000)** | **0.286 (0.000–0.546)** | **0.285 (0.000–0.554)** |
| | CardioLoom | 0.149 (0.005–0.434) | 0.773 (0.593–0.960) | 0.214 (0.000–0.556) | **1.000 (0.999–1.000)** | 0.222 (0.000–0.483) | 0.222 (-0.001–0.485) |
| 2° 2 Type AVB | Scratch CNN–Transformer | 0.013 (0.002–0.055) | 0.809 (0.686–0.946) | 0.000 (0.000–0.000) | **1.000 (1.000–1.000)** | 0.000 (0.000–0.000) | 0.000 (-0.001–0.000) |
| | ECG-FM | 0.027 (0.002–0.137) | 0.632 (0.380–0.914) | **0.071 (0.000–0.286)** | **1.000 (0.999–1.000)** | **0.095 (0.000–0.308)** | **0.101 (-0.001–0.327)** |
| | DeepECG-SSL | **0.047 (0.005–0.194)** | **0.851 (0.754–0.955)** | 0.000 (0.000–0.000) | **1.000 (1.000–1.000)** | 0.000 (0.000–0.000) | 0.000 (-0.001–0.000) |
| | CardioLoom | **0.047 (0.005–0.150)** | 0.694 (0.488–0.913) | 0.000 (0.000–0.000) | **1.000 (1.000–1.000)** | 0.000 (0.000–0.000) | 0.000 (0.000–0.000) |
| LVH | Scratch CNN–Transformer | 0.504 (0.481–0.530) | **0.845 (0.836–0.854)** | 0.264 (0.243–0.288) | **0.985 (0.983–0.987)** | 0.381 (0.357–0.408) | 0.387 (0.363–0.413) |
| | ECG-FM | 0.470 (0.446–0.496) | 0.828 (0.818–0.838) | **0.583 (0.562–0.604)** | 0.886 (0.881–0.891) | 0.462 (0.446–0.481) | 0.393 (0.375–0.412) |
| | DeepECG-SSL | 0.500 (0.475–0.526) | 0.835 (0.825–0.845) | 0.498 (0.476–0.521) | 0.935 (0.931–0.939) | 0.490 (0.471–0.510) | 0.427 (0.407–0.449) |
| | CardioLoom | **0.514 (0.489–0.539)** | 0.842 (0.832–0.852) | 0.443 (0.422–0.465) | 0.957 (0.955–0.961) | **0.493 (0.474–0.514)** | **0.444 (0.422–0.466)** |
| LAH | Scratch CNN–Transformer | **0.123 (0.096–0.156)** | **0.839 (0.822–0.854)** | 0.251 (0.209–0.292) | **0.974 (0.971–0.976)** | **0.196 (0.163–0.229)** | 0.181 (0.147–0.214) |
| | ECG-FM | 0.091 (0.074–0.116) | 0.813 (0.794–0.829) | 0.300 (0.256–0.348) | 0.949 (0.946–0.953) | 0.156 (0.132–0.183) | 0.151 (0.123–0.180) |
| | DeepECG-SSL | 0.114 (0.091–0.144) | 0.835 (0.816–0.851) | **0.315 (0.269–0.363)** | 0.960 (0.957–0.963) | 0.190 (0.162–0.219) | **0.183 (0.152–0.214)** |
| | CardioLoom | 0.113 (0.089–0.146) | 0.827 (0.808–0.843) | 0.291 (0.246–0.337) | 0.963 (0.960–0.966) | 0.185 (0.155–0.215) | 0.175 (0.143–0.206) |

| Disease | Model | AUPRC | AUROC | Sensitivity | Specificity | F1 Score | MCC |
|---|---|---|---|---|---|---|---|
| RVH | Scratch CNN–Transformer | **0.118 (0.071–0.176)** | **0.919 (0.902–0.935)** | **0.135 (0.077–0.197)** | **0.997 (0.996–0.998)** | **0.157 (0.091–0.223)** | **0.156 (0.088–0.222)** |
| | ECG-FM | 0.113 (0.067–0.171) | 0.909 (0.888–0.928) | 0.119 (0.068–0.176) | **0.997 (0.996–0.998)** | 0.142 (0.082–0.205) | 0.141 (0.080–0.206) |
| | DeepECG-SSL | 0.085 (0.048–0.143) | 0.859 (0.824–0.889) | 0.103 (0.053–0.161) | **0.997 (0.996–0.998)** | 0.130 (0.068–0.198) | 0.131 (0.066–0.201) |
| | CardioLoom | 0.071 (0.037–0.122) | 0.833 (0.804–0.862) | 0.103 (0.054–0.161) | **0.997 (0.997–0.998)** | 0.134 (0.070–0.201) | 0.137 (0.070–0.205) |
| VA | Scratch CNN–Transformer | 0.069 (0.046–0.098) | 0.913 (0.885–0.938) | 0.144 (0.083–0.211) | **0.995 (0.994–0.996)** | 0.128 (0.074–0.184) | 0.124 (0.069–0.181) |
| | ECG-FM | 0.064 (0.043–0.092) | 0.916 (0.892–0.939) | **0.231 (0.156–0.310)** | 0.992 (0.991–0.994) | **0.164 (0.113–0.222)** | **0.166 (0.113–0.225)** |
| | DeepECG-SSL | **0.076 (0.045–0.122)** | 0.926 (0.907–0.941) | 0.202 (0.121–0.286) | 0.993 (0.992–0.995) | 0.157 (0.091–0.221) | 0.156 (0.088–0.221) |
| | CardioLoom | 0.069 (0.046–0.102) | **0.933 (0.913–0.951)** | 0.163 (0.078–0.256) | 0.994 (0.993–0.995) | 0.138 (0.066–0.210) | 0.135 (0.063–0.209) |
| RAH | Scratch CNN–Transformer | 0.193 (0.120–0.283) | **0.952 (0.928–0.970)** | 0.020 (0.000–0.050) | **1.000 (1.000–1.000)** | 0.038 (0.000–0.090) | 0.081 (-0.001–0.179) |
| | ECG-FM | 0.166 (0.093–0.260) | 0.905 (0.871–0.935) | **0.091 (0.040–0.146)** | **1.000 (0.999–1.000)** | **0.157 (0.072–0.239)** | **0.225 (0.116–0.319)** |
| | DeepECG-SSL | **0.206 (0.127–0.292)** | 0.926 (0.893–0.951) | 0.071 (0.023–0.124) | **1.000 (0.999–1.000)** | 0.121 (0.042–0.200) | 0.169 (0.070–0.264) |
| | CardioLoom | 0.180 (0.096–0.278) | 0.909 (0.874–0.938) | 0.040 (0.009–0.082) | **1.000 (1.000–1.000)** | 0.075 (0.017–0.149) | 0.151 (0.036–0.248) |
| VVI | Scratch CNN–Transformer | 0.697 (0.633–0.755) | 0.963 (0.950–0.974) | 0.558 (0.490–0.622) | **0.999 (0.999–0.999)** | 0.685 (0.630–0.735) | 0.700 (0.653–0.746) |
| | ECG-FM | 0.659 (0.589–0.731) | 0.964 (0.952–0.974) | **0.667 (0.603–0.733)** | 0.998 (0.997–0.998) | 0.725 (0.673–0.774) | 0.724 (0.674–0.772) |
| | DeepECG-SSL | **0.701 (0.641–0.761)** | 0.953 (0.938–0.966) | 0.663 (0.600–0.726) | 0.998 (0.998–0.999) | **0.747 (0.697–0.793)** | **0.750 (0.701–0.794)** |
| | CardioLoom | 0.694 (0.633–0.760) | **0.970 (0.959–0.980)** | 0.646 (0.583–0.712) | **0.999 (0.998–0.999)** | 0.738 (0.689–0.787) | 0.742 (0.696–0.790) |
| VAT | Scratch CNN–Transformer | 0.273 (0.220–0.332) | 0.941 (0.930–0.952) | **0.129 (0.089–0.172)** | 0.998 (0.998–0.999) | 0.204 (0.146–0.261) | 0.246 (0.180–0.307) |
| | ECG-FM | 0.377 (0.317–0.442) | 0.961 (0.951–0.969) | 0.122 (0.085–0.165) | **1.000 (1.000–1.000)** | 0.215 (0.153–0.278) | **0.325 (0.260–0.385)** |
| | DeepECG-SSL | 0.225 (0.178–0.280) | 0.928 (0.919–0.937) | 0.119 (0.081–0.161) | 0.999 (0.998–0.999) | 0.198 (0.137–0.259) | 0.260 (0.189–0.327) |
| | CardioLoom | **0.394 (0.333–0.458)** | **0.975 (0.970–0.980)** | **0.129 (0.090–0.172)** | 0.999 (0.999–0.999) | **0.217 (0.156–0.278)** | 0.290 (0.219–0.354) |

| Disease | Model | AUPRC | AUROC | Sensitivity | Specificity | F1 Score | MCC |
|---|---|---|---|---|---|---|---|
| DDD | Scratch CNN–Transformer | **0.667 (0.604–0.731)** | 0.983 (0.975–0.990) | 0.109 (0.072–0.148) | **1.000 (1.000–1.000)** | 0.194 (0.133–0.255) | 0.309 (0.243–0.371) |
| | ECG-FM | 0.425 (0.361–0.486) | 0.978 (0.971–0.983) | **0.133 (0.091–0.182)** | **1.000 (0.999–1.000)** | **0.229 (0.163–0.301)** | 0.333 (0.264–0.395) |
| | DeepECG-SSL | 0.308 (0.255–0.366) | 0.961 (0.954–0.967) | 0.112 (0.073–0.158) | **1.000 (1.000–1.000)** | 0.200 (0.136–0.270) | 0.318 (0.253–0.381) |
| | CardioLoom | 0.576 (0.512–0.636) | **0.988 (0.982–0.992)** | 0.129 (0.089–0.175) | **1.000 (1.000–1.000)** | 0.228 (0.161–0.296) | **0.348 (0.284–0.410)** |
| AAI | Scratch CNN–Transformer | 0.095 (0.058–0.138) | 0.715 (0.685–0.747) | **0.061 (0.033–0.094)** | **1.000 (0.999–1.000)** | **0.112 (0.062–0.168)** | **0.199 (0.125–0.267)** |
| | ECG-FM | 0.073 (0.049–0.106) | 0.835 (0.818–0.854) | 0.048 (0.022–0.076) | 0.997 (0.996–0.998) | 0.076 (0.035–0.121) | 0.088 (0.039–0.140) |
| | DeepECG-SSL | 0.098 (0.065–0.137) | 0.773 (0.747–0.798) | 0.048 (0.022–0.078) | 0.999 (0.998–0.999) | 0.084 (0.041–0.134) | 0.129 (0.065–0.194) |
| | CardioLoom | **0.124 (0.089–0.164)** | **0.859 (0.839–0.878)** | 0.048 (0.022–0.078) | 0.999 (0.999–1.000) | 0.087 (0.042–0.138) | 0.148 (0.078–0.218) |
| VP | Scratch CNN–Transformer | **0.723 (0.662–0.783)** | **0.979 (0.971–0.985)** | 0.078 (0.046–0.114) | **1.000 (1.000–1.000)** | 0.144 (0.087–0.203) | 0.261 (0.193–0.324) |
| | ECG-FM | 0.655 (0.584–0.722) | 0.966 (0.954–0.976) | **0.503 (0.441–0.571)** | 0.999 (0.999–0.999) | **0.641 (0.583–0.700)** | **0.663 (0.612–0.716)** |
| | DeepECG-SSL | 0.681 (0.620–0.741) | 0.944 (0.927–0.960) | 0.235 (0.179–0.292) | **1.000 (1.000–1.000)** | 0.379 (0.303–0.451) | 0.478 (0.417–0.535) |
| | CardioLoom | 0.640 (0.573–0.707) | 0.974 (0.965–0.982) | 0.293 (0.238–0.347) | 0.999 (0.999–1.000) | 0.433 (0.369–0.497) | 0.491 (0.431–0.548) |

Values are estimates (95% cluster-bootstrap CIs). NE indicates fewer than 950 valid replicates among 1000. Sensitivity, specificity, F1 score, and MCC use the Zhongshan validation-F1 threshold. Bold type marks the highest displayed estimate within each disease.

**Supplementary Table 24: Detailed task-specific performance comparison on the external Georgia cohort.**

| Disease | Model | AUPRC | AUROC | Sensitivity | Specificity | F1 Score | MCC |
|---|---|---|---|---|---|---|---|
| STEMI | Scratch CNN–Transformer | 0.002 (0.001–0.007) | 0.765 (0.501–0.925) | 0.000 (0.000–0.000) | 0.993 (0.992–0.995) | 0.000 (0.000–0.000) | -0.002 (-0.003–-0.001) |
| | ECG-FM | 0.003 (0.001–0.009) | **0.838 (0.760–0.919)** | 0.000 (0.000–0.000) | 0.994 (0.992–0.996) | 0.000 (0.000–0.000) | -0.002 (-0.003–-0.001) |
| | DeepECG-SSL | **0.004 (0.001–0.016)** | 0.779 (0.611–0.917) | **0.143 (0.000–0.430)** | 0.993 (0.991–0.994) | **0.024 (0.000–0.078)** | **0.042 (-0.003–0.129)** |
| | CardioLoom | 0.003 (0.000–0.011) | 0.708 (0.469–0.911) | 0.000 (0.000–0.000) | **0.996 (0.995–0.997)** | 0.000 (0.000–0.000) | -0.002 (-0.002–-0.001) |
| SBrad | Scratch CNN–Transformer | **0.980 (0.975–0.984)** | **0.996 (0.994–0.997)** | 0.950 (0.939–0.960) | **0.988 (0.986–0.990)** | **0.944 (0.936–0.952)** | **0.933 (0.923–0.942)** |
| | ECG-FM | 0.975 (0.968–0.980) | 0.993 (0.990–0.995) | 0.951 (0.940–0.961) | 0.986 (0.983–0.988) | 0.939 (0.930–0.947) | 0.927 (0.917–0.937) |
| | DeepECG-SSL | 0.975 (0.968–0.981) | 0.995 (0.993–0.996) | **0.966 (0.956–0.974)** | 0.984 (0.981–0.986) | 0.943 (0.935–0.950) | 0.931 (0.922–0.941) |
| | CardioLoom | 0.973 (0.967–0.979) | 0.994 (0.992–0.995) | **0.966 (0.956–0.974)** | 0.981 (0.978–0.984) | 0.937 (0.928–0.945) | 0.924 (0.915–0.934) |
| STach | Scratch CNN–Transformer | **0.956 (0.945–0.965)** | 0.992 (0.990–0.994) | 0.877 (0.857–0.895) | **0.993 (0.991–0.995)** | 0.910 (0.897–0.922) | 0.899 (0.885–0.912) |
| | ECG-FM | 0.951 (0.938–0.962) | **0.993 (0.992–0.995)** | 0.897 (0.879–0.913) | 0.989 (0.987–0.991) | 0.908 (0.895–0.919) | 0.895 (0.880–0.908) |
| | DeepECG-SSL | 0.950 (0.937–0.962) | 0.992 (0.989–0.994) | 0.892 (0.874–0.909) | 0.991 (0.989–0.993) | **0.912 (0.900–0.924)** | **0.901 (0.887–0.914)** |
| | CardioLoom | **0.956 (0.945–0.966)** | **0.993 (0.991–0.995)** | **0.902 (0.883–0.918)** | 0.990 (0.987–0.992) | **0.912 (0.900–0.924)** | 0.900 (0.887–0.914) |
| VPB | Scratch CNN–Transformer | 0.377 (0.325–0.434) | 0.847 (0.824–0.868) | 0.424 (0.374–0.473) | **0.988 (0.986–0.990)** | 0.493 (0.445–0.539) | 0.483 (0.435–0.530) |
| | ECG-FM | **0.461 (0.415–0.521)** | **0.865 (0.844–0.884)** | 0.454 (0.403–0.503) | 0.986 (0.984–0.989) | 0.506 (0.460–0.552) | 0.492 (0.446–0.539) |
| | DeepECG-SSL | 0.419 (0.367–0.476) | 0.840 (0.816–0.865) | **0.461 (0.406–0.513)** | 0.986 (0.984–0.988) | **0.508 (0.463–0.555)** | **0.493 (0.448–0.542)** |
| | CardioLoom | 0.442 (0.387–0.495) | 0.860 (0.839–0.879) | **0.461 (0.409–0.512)** | 0.986 (0.983–0.988) | 0.507 (0.459–0.555) | 0.492 (0.446–0.540) |
| APB | Scratch CNN–Transformer | 0.591 (0.548–0.636) | 0.916 (0.903–0.930) | 0.649 (0.611–0.685) | 0.969 (0.965–0.972) | 0.612 (0.580–0.645) | 0.586 (0.553–0.620) |
| | ECG-FM | 0.588 (0.548–0.631) | 0.912 (0.897–0.926) | 0.677 (0.641–0.712) | **0.971 (0.968–0.974)** | 0.640 (0.609–0.668) | 0.615 (0.583–0.645) |
| | DeepECG-SSL | 0.646 (0.605–0.689) | **0.934 (0.922–0.945)** | **0.726 (0.692–0.760)** | 0.970 (0.967–0.974) | **0.667 (0.639–0.694)** | **0.646 (0.616–0.674)** |
| | CardioLoom | **0.653 (0.612–0.694)** | 0.927 (0.913–0.940) | 0.708 (0.672–0.741) | **0.971 (0.968–0.975)** | 0.661 (0.631–0.689) | 0.639 (0.608–0.668) |

| Disease | Model | AUPRC | AUROC | Sensitivity | Specificity | F1 Score | MCC |
|---|---|---|---|---|---|---|---|
| SArr | Scratch CNN–Transformer | 0.600 (0.547–0.646) | 0.929 (0.915–0.942) | 0.379 (0.335–0.424) | **0.995 (0.994–0.996)** | 0.510 (0.463–0.555) | 0.530 (0.484–0.574) |
| | ECG-FM | 0.548 (0.500–0.595) | 0.922 (0.908–0.936) | 0.421 (0.377–0.464) | 0.990 (0.988–0.992) | 0.513 (0.469–0.554) | 0.509 (0.465–0.550) |
| | DeepECG-SSL | **0.691 (0.646–0.731)** | 0.960 (0.951–0.969) | **0.535 (0.486–0.581)** | 0.992 (0.990–0.993) | **0.622 (0.577–0.661)** | **0.617 (0.572–0.655)** |
| | CardioLoom | 0.685 (0.637–0.727) | **0.966 (0.958–0.973)** | 0.515 (0.467–0.561) | 0.992 (0.991–0.994) | 0.613 (0.569–0.651) | 0.611 (0.567–0.649) |
| AF | Scratch CNN–Transformer | 0.750 (0.713–0.784) | 0.922 (0.906–0.938) | 0.611 (0.570–0.652) | 0.994 (0.993–0.996) | 0.715 (0.680–0.746) | 0.712 (0.679–0.744) |
| | ECG-FM | 0.776 (0.739–0.810) | 0.930 (0.916–0.945) | 0.625 (0.584–0.664) | **0.997 (0.996–0.998)** | 0.749 (0.717–0.778) | 0.754 (0.725–0.782) |
| | DeepECG-SSL | **0.788 (0.752–0.820)** | 0.925 (0.910–0.941) | **0.653 (0.612–0.690)** | 0.996 (0.994–0.997) | **0.757 (0.724–0.787)** | **0.756 (0.724–0.787)** |
| | CardioLoom | 0.787 (0.752–0.819) | **0.931 (0.917–0.945)** | 0.640 (0.602–0.678) | 0.996 (0.995–0.998) | 0.752 (0.720–0.780) | 0.753 (0.723–0.779) |
| AFL | Scratch CNN–Transformer | 0.554 (0.483–0.629) | 0.935 (0.912–0.956) | 0.548 (0.480–0.620) | 0.993 (0.992–0.995) | 0.573 (0.515–0.631) | 0.566 (0.508–0.625) |
| | ECG-FM | 0.600 (0.527–0.671) | 0.940 (0.914–0.962) | 0.629 (0.566–0.697) | 0.992 (0.990–0.994) | 0.609 (0.555–0.668) | 0.602 (0.547–0.663) |
| | DeepECG-SSL | **0.615 (0.536–0.698)** | 0.945 (0.919–0.966) | 0.608 (0.537–0.677) | **0.995 (0.993–0.996)** | **0.640 (0.582–0.698)** | **0.635 (0.577–0.695)** |
| | CardioLoom | 0.604 (0.528–0.679) | **0.953 (0.935–0.971)** | **0.640 (0.576–0.707)** | 0.993 (0.991–0.994) | 0.626 (0.573–0.680) | 0.619 (0.566–0.674) |
| JPB | Scratch CNN–Transformer | 0.001 (NE) | 0.936 (NE) | **0.000 (NE)** | **0.997 (NE)** | **0.000 (NE)** | **-0.001 (NE)** |
| | ECG-FM | 0.002 (NE) | 0.978 (NE) | **0.000 (NE)** | 0.994 (NE) | **0.000 (NE)** | **-0.001 (NE)** |
| | DeepECG-SSL | 0.001 (NE) | 0.945 (NE) | **0.000 (NE)** | 0.994 (NE) | **0.000 (NE)** | **-0.001 (NE)** |
| | CardioLoom | **0.003 (NE)** | **0.981 (NE)** | **0.000 (NE)** | 0.996 (NE) | **0.000 (NE)** | **-0.001 (NE)** |
| AJR | Scratch CNN–Transformer | 0.009 (0.004–0.016) | 0.857 (0.782–0.918) | 0.053 (0.000–0.176) | 0.984 (0.981–0.987) | 0.011 (0.000–0.036) | 0.012 (-0.006–0.055) |
| | ECG-FM | **0.023 (0.009–0.053)** | **0.930 (0.886–0.964)** | **0.105 (0.000–0.250)** | 0.992 (0.991–0.994) | **0.040 (0.000–0.096)** | 0.047 (-0.004–0.117) |
| | DeepECG-SSL | 0.012 (0.005–0.024) | 0.881 (0.819–0.929) | 0.000 (0.000–0.000) | **0.994 (0.992–0.995)** | 0.000 (0.000–0.000) | -0.003 (-0.004–-0.002) |
| | CardioLoom | 0.012 (0.006–0.026) | 0.897 (0.844–0.938) | **0.105 (0.000–0.250)** | 0.992 (0.991–0.994) | **0.040 (0.000–0.100)** | **0.048 (-0.004–0.119)** |

| Disease | Model | AUPRC | AUROC | Sensitivity | Specificity | F1 Score | MCC |
|---|---|---|---|---|---|---|---|
| ATach | Scratch CNN–Transformer | 0.057 (0.021–0.135) | 0.910 (0.864–0.949) | 0.286 (0.121–0.462) | **0.993 (0.991–0.995)** | 0.148 (0.062–0.239) | 0.165 (0.069–0.263) |
| | ECG-FM | 0.053 (0.021–0.157) | 0.936 (0.910–0.959) | 0.250 (0.100–0.412) | 0.990 (0.988–0.992) | 0.103 (0.036–0.176) | 0.123 (0.043–0.206) |
| | DeepECG-SSL | **0.131 (0.048–0.262)** | **0.974 (0.964–0.983)** | **0.357 (0.178–0.533)** | 0.992 (0.990–0.994) | **0.165 (0.080–0.256)** | **0.192 (0.093–0.290)** |
| | CardioLoom | 0.106 (0.033–0.224) | 0.944 (0.918–0.966) | 0.286 (0.125–0.458) | **0.993 (0.991–0.994)** | 0.143 (0.059–0.238) | 0.161 (0.067–0.263) |
| JTach | Scratch CNN–Transformer | 0.014 (0.000–0.066) | **0.769 (0.197–0.998)** | 0.250 (0.000–1.000) | 0.996 (0.995–0.997) | 0.043 (0.000–0.152) | 0.076 (-0.002–0.233) |
| | ECG-FM | **0.098 (0.000–0.476)** | 0.607 (0.178–1.000) | **0.500 (0.000–1.000)** | **0.997 (0.996–0.998)** | **0.121 (0.000–0.286)** | **0.185 (-0.001–0.373)** |
| | DeepECG-SSL | 0.049 (0.000–0.277) | 0.590 (0.141–0.999) | **0.500 (0.000–1.000)** | 0.995 (0.994–0.997) | 0.073 (0.000–0.180) | 0.139 (-0.001–0.285) |
| | CardioLoom | 0.021 (0.000–0.129) | 0.677 (0.182–0.999) | 0.250 (0.000–1.000) | 0.995 (0.994–0.996) | 0.036 (0.000–0.124) | 0.068 (-0.002–0.212) |
| JEB | Scratch CNN–Transformer | 0.074 (0.001–0.334) | 0.860 (0.625–0.999) | **0.600 (0.000–1.000)** | 0.996 (0.995–0.997) | 0.122 (0.000–0.265) | **0.201 (-0.001–0.365)** |
| | ECG-FM | 0.098 (0.004–0.515) | 0.873 (0.584–1.000) | 0.400 (0.000–1.000) | **0.998 (0.997–0.999)** | **0.138 (0.000–0.345)** | 0.182 (-0.001–0.400) |
| | DeepECG-SSL | **0.230 (0.001–0.710)** | **0.904 (0.721–0.999)** | 0.400 (0.000–1.000) | **0.998 (0.997–0.999)** | **0.138 (0.000–0.323)** | 0.182 (-0.001–0.387) |
| | CardioLoom | 0.051 (0.002–0.314) | 0.894 (0.663–0.999) | 0.400 (0.000–1.000) | 0.997 (0.996–0.998) | 0.105 (0.000–0.250) | 0.155 (-0.001–0.333) |
| SVT | Scratch CNN–Transformer | 0.288 (0.138–0.480) | 0.952 (0.894–0.991) | 0.156 (0.037–0.292) | **0.999 (0.999–1.000)** | 0.233 (0.059–0.392) | 0.265 (0.076–0.433) |
| | ECG-FM | 0.359 (0.184–0.542) | 0.923 (0.862–0.970) | 0.281 (0.128–0.435) | **0.999 (0.999–1.000)** | 0.383 (0.200–0.539) | 0.410 (0.219–0.567) |
| | DeepECG-SSL | 0.349 (0.190–0.562) | **0.970 (0.947–0.987)** | **0.406 (0.240–0.571)** | **0.999 (0.998–0.999)** | **0.456 (0.293–0.607)** | **0.458 (0.293–0.612)** |
| | CardioLoom | **0.364 (0.199–0.542)** | 0.943 (0.907–0.974) | 0.281 (0.129–0.444) | **0.999 (0.999–1.000)** | 0.375 (0.192–0.542) | 0.396 (0.206–0.563) |
| VEB | Scratch CNN–Transformer | **0.000 (NE)** | 0.384 (NE) | **0.000 (NE)** | **0.999 (NE)** | **0.000 (NE)** | **0.000 (NE)** |
| | ECG-FM | **0.000 (NE)** | 0.675 (NE) | **0.000 (NE)** | **0.999 (NE)** | **0.000 (NE)** | **0.000 (NE)** |
| | DeepECG-SSL | **0.000 (NE)** | 0.636 (NE) | **0.000 (NE)** | **0.999 (NE)** | **0.000 (NE)** | **0.000 (NE)** |
| | CardioLoom | **0.000 (NE)** | **0.825 (NE)** | **0.000 (NE)** | **0.999 (NE)** | **0.000 (NE)** | **0.000 (NE)** |

| Disease | Model | AUPRC | AUROC | Sensitivity | Specificity | F1 Score | MCC |
|---|---|---|---|---|---|---|---|
| RBBB | Scratch CNN–Transformer | **0.805 (0.780–0.828)** | **0.948 (0.940–0.955)** | **0.760 (0.732–0.784)** | 0.972 (0.968–0.975) | **0.750 (0.728–0.771)** | **0.723 (0.700–0.746)** |
| | ECG-FM | 0.767 (0.737–0.795) | 0.923 (0.910–0.935) | 0.692 (0.662–0.720) | **0.975 (0.972–0.978)** | 0.718 (0.693–0.740) | 0.690 (0.663–0.714) |
| | DeepECG-SSL | 0.784 (0.757–0.807) | 0.941 (0.931–0.949) | 0.750 (0.721–0.777) | 0.967 (0.963–0.970) | 0.727 (0.703–0.746) | 0.697 (0.671–0.718) |
| | CardioLoom | 0.789 (0.760–0.815) | 0.941 (0.931–0.950) | 0.740 (0.712–0.767) | 0.973 (0.970–0.976) | 0.742 (0.719–0.764) | 0.715 (0.690–0.738) |
| 1° AVB | Scratch CNN–Transformer | 0.735 (0.699–0.767) | 0.935 (0.923–0.946) | 0.689 (0.659–0.721) | 0.979 (0.976–0.981) | 0.715 (0.688–0.739) | 0.691 (0.663–0.716) |
| | ECG-FM | **0.773 (0.739–0.805)** | **0.940 (0.930–0.951)** | 0.692 (0.660–0.721) | **0.985 (0.982–0.987)** | 0.742 (0.717–0.764) | 0.723 (0.696–0.746) |
| | DeepECG-SSL | 0.764 (0.730–0.796) | 0.937 (0.925–0.948) | **0.734 (0.702–0.763)** | 0.979 (0.976–0.982) | **0.745 (0.721–0.769)** | 0.723 (0.698–0.748) |
| | CardioLoom | 0.764 (0.729–0.797) | 0.921 (0.905–0.935) | 0.713 (0.683–0.742) | 0.982 (0.979–0.985) | **0.745 (0.722–0.769)** | **0.725 (0.699–0.749)** |
| QT Prolong | Scratch CNN–Transformer | 0.552 (0.523–0.579) | 0.860 (0.851–0.870) | 0.233 (0.211–0.256) | 0.989 (0.987–0.991) | 0.357 (0.330–0.386) | 0.381 (0.355–0.409) |
| | ECG-FM | 0.641 (0.614–0.667) | 0.894 (0.885–0.903) | **0.298 (0.274–0.325)** | 0.989 (0.987–0.991) | **0.435 (0.407–0.465)** | **0.450 (0.422–0.478)** |
| | DeepECG-SSL | 0.603 (0.574–0.630) | 0.881 (0.872–0.890) | 0.278 (0.252–0.301) | 0.987 (0.985–0.989) | 0.408 (0.377–0.436) | 0.421 (0.392–0.447) |
| | CardioLoom | **0.654 (0.628–0.680)** | **0.901 (0.893–0.910)** | 0.212 (0.189–0.233) | **0.993 (0.991–0.995)** | 0.337 (0.307–0.365) | 0.383 (0.354–0.410) |
| ER | Scratch CNN–Transformer | 0.325 (0.246–0.412) | 0.926 (0.903–0.946) | 0.121 (0.071–0.182) | **0.999 (0.999–1.000)** | 0.209 (0.125–0.300) | 0.296 (0.203–0.387) |
| | ECG-FM | 0.361 (0.278–0.451) | 0.937 (0.916–0.956) | 0.307 (0.235–0.388) | 0.995 (0.994–0.997) | 0.376 (0.296–0.455) | 0.379 (0.297–0.459) |
| | DeepECG-SSL | **0.383 (0.296–0.466)** | 0.937 (0.914–0.957) | **0.336 (0.263–0.416)** | 0.996 (0.995–0.997) | **0.409 (0.327–0.485)** | **0.412 (0.336–0.490)** |
| | CardioLoom | 0.348 (0.267–0.433) | **0.939 (0.917–0.958)** | 0.200 (0.135–0.270) | 0.998 (0.997–0.999) | 0.290 (0.208–0.371) | 0.320 (0.236–0.402) |
| LAFB | Scratch CNN–Transformer | 0.208 (0.166–0.266) | 0.935 (0.921–0.948) | 0.217 (0.157–0.278) | **0.992 (0.991–0.994)** | 0.264 (0.196–0.331) | 0.260 (0.190–0.328) |
| | ECG-FM | 0.231 (0.185–0.288) | 0.938 (0.920–0.952) | 0.233 (0.177–0.301) | 0.991 (0.989–0.993) | 0.267 (0.205–0.333) | 0.258 (0.195–0.326) |
| | DeepECG-SSL | **0.263 (0.214–0.325)** | 0.938 (0.921–0.955) | **0.272 (0.205–0.333)** | **0.992 (0.990–0.993)** | **0.313 (0.246–0.374)** | **0.306 (0.240–0.369)** |
| | CardioLoom | 0.227 (0.185–0.282) | **0.941 (0.925–0.956)** | 0.200 (0.145–0.257) | 0.991 (0.989–0.992) | 0.232 (0.169–0.294) | 0.223 (0.160–0.287) |

| Disease | Model | AUPRC | AUROC | Sensitivity | Specificity | F1 Score | MCC |
|---|---|---|---|---|---|---|---|
| LBBB | Scratch CNN–Transformer | 0.664 (0.613–0.711) | 0.937 (0.922–0.951) | 0.425 (0.369–0.480) | **0.998 (0.998–0.999)** | **0.576 (0.520–0.624)** | **0.608 (0.559–0.649)** |
| | ECG-FM | 0.644 (0.583–0.702) | 0.917 (0.896–0.938) | 0.391 (0.339–0.445) | 0.997 (0.996–0.998) | 0.532 (0.474–0.585) | 0.562 (0.509–0.610) |
| | DeepECG-SSL | 0.635 (0.576–0.692) | **0.939 (0.924–0.954)** | **0.434 (0.381–0.491)** | 0.997 (0.996–0.998) | 0.567 (0.512–0.619) | 0.586 (0.536–0.637) |
| | CardioLoom | **0.667 (0.616–0.715)** | 0.935 (0.918–0.951) | 0.410 (0.357–0.462) | 0.997 (0.996–0.998) | 0.547 (0.494–0.600) | 0.572 (0.521–0.619) |
| IVB | Scratch CNN–Transformer | 0.050 (0.039–0.065) | 0.721 (0.689–0.755) | 0.064 (0.034–0.099) | **0.985 (0.983–0.988)** | 0.071 (0.038–0.110) | 0.055 (0.021–0.094) |
| | ECG-FM | 0.063 (0.046–0.083) | 0.742 (0.712–0.774) | 0.079 (0.044–0.119) | **0.985 (0.982–0.987)** | 0.085 (0.048–0.127) | 0.068 (0.031–0.111) |
| | DeepECG-SSL | 0.059 (0.046–0.075) | 0.761 (0.730–0.791) | 0.084 (0.047–0.127) | 0.984 (0.981–0.986) | 0.088 (0.050–0.132) | 0.071 (0.033–0.115) |
| | CardioLoom | **0.075 (0.060–0.095)** | **0.814 (0.787–0.839)** | **0.128 (0.082–0.175)** | 0.981 (0.977–0.983) | **0.122 (0.079–0.167)** | **0.104 (0.060–0.149)** |
| Short PR | Scratch CNN–Transformer | 0.001 (NE) | 0.908 (NE) | **0.000 (NE)** | 0.996 (NE) | **0.000 (NE)** | **-0.001 (NE)** |
| | ECG-FM | **0.004 (NE)** | **0.977 (NE)** | **0.000 (NE)** | 0.995 (NE) | **0.000 (NE)** | **-0.001 (NE)** |
| | DeepECG-SSL | 0.002 (NE) | 0.963 (NE) | **0.000 (NE)** | 0.993 (NE) | **0.000 (NE)** | **-0.001 (NE)** |
| | CardioLoom | 0.002 (NE) | 0.951 (NE) | **0.000 (NE)** | **0.997 (NE)** | **0.000 (NE)** | **-0.001 (NE)** |
| VPE | Scratch CNN–Transformer | **0.063 (NE)** | 0.877 (NE) | **0.500 (NE)** | **0.999 (NE)** | 0.143 (NE) | 0.204 (NE) |
| | ECG-FM | 0.042 (NE) | **0.907 (NE)** | **0.500 (NE)** | **0.999 (NE)** | 0.125 (NE) | 0.189 (NE) |
| | DeepECG-SSL | **0.063 (NE)** | 0.594 (NE) | **0.500 (NE)** | **0.999 (NE)** | 0.111 (NE) | 0.176 (NE) |
| | CardioLoom | 0.042 (NE) | 0.715 (NE) | **0.500 (NE)** | **0.999 (NE)** | **0.154 (NE)** | **0.213 (NE)** |
| 3° AVB | Scratch CNN–Transformer | 0.047 (0.020–0.108) | 0.735 (0.685–0.782) | 0.073 (0.024–0.132) | 0.998 (0.996–0.998) | 0.106 (0.035–0.183) | 0.115 (0.036–0.203) |
| | ECG-FM | 0.081 (0.031–0.150) | 0.788 (0.745–0.829) | 0.073 (0.024–0.133) | **0.999 (0.998–0.999)** | 0.120 (0.040–0.207) | 0.153 (0.053–0.260) |
| | DeepECG-SSL | 0.109 (0.048–0.174) | 0.804 (0.771–0.839) | 0.085 (0.029–0.147) | **0.999 (0.998–1.000)** | 0.140 (0.050–0.236) | 0.179 (0.070–0.291) |
| | CardioLoom | **0.116 (0.055–0.186)** | **0.840 (0.811–0.870)** | **0.098 (0.037–0.163)** | **0.999 (0.999–1.000)** | **0.167 (0.067–0.265)** | **0.234 (0.110–0.346)** |

| Disease | Model | AUPRC | AUROC | Sensitivity | Specificity | F1 Score | MCC |
|---|---|---|---|---|---|---|---|
| 2° 1 Type AVB | Scratch CNN–Transformer | 0.102 (0.048–0.175) | **0.811 (0.777–0.842)** | 0.073 (0.026–0.130) | 0.999 (0.998–1.000) | 0.125 (0.046–0.215) | 0.176 (0.068–0.282) |
| | ECG-FM | 0.126 (0.058–0.207) | 0.690 (0.649–0.735) | 0.094 (0.042–0.160) | **1.000 (0.999–1.000)** | 0.165 (0.076–0.268) | 0.253 (0.139–0.360) |
| | DeepECG-SSL | **0.142 (0.071–0.225)** | 0.700 (0.655–0.751) | **0.115 (0.056–0.188)** | **1.000 (0.999–1.000)** | **0.196 (0.101–0.300)** | **0.278 (0.166–0.382)** |
| | CardioLoom | 0.119 (0.051–0.204) | 0.580 (0.523–0.641) | 0.094 (0.041–0.157) | **1.000 (0.999–1.000)** | 0.164 (0.074–0.263) | 0.243 (0.128–0.355) |
| 2° 2 Type AVB | Scratch CNN–Transformer | 0.020 (0.012–0.046) | **0.603 (0.549–0.660)** | 0.010 (0.000–0.033) | **1.000 (0.999–1.000)** | 0.020 (0.000–0.063) | 0.049 (-0.002–0.135) |
| | ECG-FM | 0.024 (0.012–0.059) | 0.551 (0.497–0.605) | **0.021 (0.000–0.056)** | **1.000 (0.999–1.000)** | **0.039 (0.000–0.099)** | 0.075 (-0.003–0.176) |
| | DeepECG-SSL | **0.031 (0.013–0.074)** | 0.528 (0.472–0.587) | **0.021 (0.000–0.056)** | **1.000 (0.999–1.000)** | **0.039 (0.000–0.099)** | 0.075 (-0.003–0.176) |
| | CardioLoom | 0.029 (0.012–0.073) | 0.424 (0.366–0.492) | **0.021 (0.000–0.056)** | **1.000 (0.999–1.000)** | **0.039 (0.000–0.100)** | **0.081 (-0.002–0.186)** |
| LVH | Scratch CNN–Transformer | **0.350 (0.321–0.377)** | **0.729 (0.714–0.745)** | **0.200 (0.178–0.222)** | 0.978 (0.975–0.981) | **0.295 (0.267–0.322)** | **0.287 (0.256–0.317)** |
| | ECG-FM | 0.328 (0.301–0.355) | 0.711 (0.695–0.727) | 0.185 (0.165–0.207) | 0.973 (0.970–0.977) | 0.269 (0.243–0.296) | 0.248 (0.220–0.277) |
| | DeepECG-SSL | 0.317 (0.293–0.344) | 0.700 (0.683–0.717) | 0.191 (0.169–0.214) | 0.972 (0.968–0.975) | 0.275 (0.246–0.301) | 0.249 (0.218–0.277) |
| | CardioLoom | 0.340 (0.313–0.367) | 0.706 (0.689–0.723) | 0.168 (0.148–0.188) | **0.982 (0.980–0.985)** | 0.259 (0.232–0.287) | 0.264 (0.235–0.294) |
| LAH | Scratch CNN–Transformer | **0.464 (0.432–0.501)** | **0.891 (0.881–0.902)** | **0.532 (0.499–0.564)** | 0.945 (0.940–0.949) | **0.508 (0.481–0.536)** | **0.458 (0.429–0.488)** |
| | ECG-FM | 0.450 (0.419–0.487) | 0.877 (0.867–0.889) | 0.477 (0.445–0.511) | 0.951 (0.946–0.955) | 0.484 (0.456–0.512) | 0.434 (0.405–0.464) |
| | DeepECG-SSL | 0.443 (0.410–0.478) | 0.879 (0.869–0.890) | 0.491 (0.462–0.523) | 0.942 (0.938–0.947) | 0.474 (0.449–0.503) | 0.420 (0.392–0.450) |
| | CardioLoom | 0.462 (0.428–0.498) | 0.887 (0.877–0.897) | 0.447 (0.416–0.479) | **0.956 (0.952–0.960)** | 0.474 (0.444–0.505) | 0.426 (0.394–0.458) |
| RVH | Scratch CNN–Transformer | 0.042 (0.020–0.077) | 0.724 (0.658–0.787) | 0.183 (0.094–0.277) | 0.988 (0.986–0.990) | 0.123 (0.062–0.183) | 0.122 (0.059–0.183) |
| | ECG-FM | 0.049 (0.024–0.095) | 0.783 (0.731–0.835) | **0.254 (0.155–0.364)** | 0.987 (0.985–0.989) | 0.163 (0.095–0.234) | 0.166 (0.095–0.238) |
| | DeepECG-SSL | 0.054 (0.022–0.103) | 0.708 (0.647–0.778) | 0.169 (0.086–0.259) | **0.990 (0.989–0.992)** | 0.133 (0.067–0.198) | 0.128 (0.061–0.195) |
| | CardioLoom | **0.056 (0.028–0.100)** | **0.798 (0.743–0.851)** | **0.254 (0.156–0.364)** | **0.990 (0.988–0.992)** | **0.186 (0.114–0.263)** | **0.185 (0.111–0.266)** |

| Disease | Model | AUPRC | AUROC | Sensitivity | Specificity | F1 Score | MCC |
|---|---|---|---|---|---|---|---|
| RAH | Scratch CNN–Transformer | **0.146 (0.100–0.199)** | 0.838 (0.800–0.871) | **0.021 (0.000–0.047)** | **1.000 (1.000–1.000)** | **0.042 (0.000–0.089)** | **0.145 (0.000–0.214)** |
| | ECG-FM | 0.115 (0.078–0.167) | 0.828 (0.791–0.862) | 0.014 (0.000–0.037) | 0.999 (0.999–1.000) | 0.027 (0.000–0.069) | 0.057 (-0.003–0.138) |
| | DeepECG-SSL | 0.098 (0.068–0.144) | 0.814 (0.773–0.853) | 0.014 (0.000–0.036) | 0.999 (0.999–1.000) | 0.027 (0.000–0.068) | 0.057 (-0.003–0.146) |
| | CardioLoom | 0.128 (0.091–0.186) | **0.841 (0.806–0.875)** | 0.000 (0.000–0.000) | **1.000 (1.000–1.000)** | 0.000 (0.000–0.000) | 0.000 (0.000–0.000) |
| AAI | Scratch CNN–Transformer | 0.230 (0.127–0.380) | 0.917 (0.865–0.960) | 0.269 (0.159–0.405) | **0.998 (0.997–0.999)** | 0.322 (0.203–0.454) | 0.325 (0.205–0.458) |
| | ECG-FM | 0.479 (0.339–0.644) | 0.946 (0.900–0.983) | **0.769 (0.647–0.878)** | 0.995 (0.994–0.996) | 0.567 (0.471–0.663) | 0.585 (0.490–0.676) |
| | DeepECG-SSL | **0.586 (0.441–0.722)** | 0.954 (0.914–0.987) | 0.731 (0.600–0.844) | 0.996 (0.994–0.997) | 0.559 (0.450–0.654) | 0.572 (0.467–0.666) |
| | CardioLoom | 0.493 (0.362–0.668) | **0.959 (0.918–0.990)** | 0.731 (0.605–0.840) | 0.996 (0.995–0.997) | **0.585 (0.477–0.681)** | **0.594 (0.486–0.690)** |
| VP | Scratch CNN–Transformer | **0.197 (0.096–0.330)** | 0.903 (0.842–0.952) | 0.089 (0.021–0.180) | **1.000 (0.999–1.000)** | 0.148 (0.036–0.280) | **0.197 (0.050–0.344)** |
| | ECG-FM | 0.111 (0.060–0.195) | **0.926 (0.885–0.957)** | **0.200 (0.091–0.319)** | 0.996 (0.995–0.997) | **0.186 (0.084–0.292)** | 0.182 (0.081–0.288) |
| | DeepECG-SSL | 0.120 (0.062–0.213) | 0.880 (0.821–0.927) | 0.022 (0.000–0.077) | 0.999 (0.998–1.000) | 0.035 (0.000–0.113) | 0.041 (-0.003–0.133) |
| | CardioLoom | 0.112 (0.061–0.182) | 0.917 (0.869–0.955) | 0.044 (0.000–0.111) | 0.998 (0.997–0.999) | 0.057 (0.000–0.141) | 0.057 (-0.003–0.147) |

Values are estimates (95% cluster-bootstrap CIs). NE indicates fewer than 950 valid replicates among 1000. Sensitivity, specificity, F1 score, and MCC use the Zhongshan validation-F1 threshold. Bold type marks the highest displayed estimate within each disease.

**Supplementary Table 25: Detailed task-specific performance comparison on the external CPSC2018 cohort.**

| Disease | Model | AUPRC | AUROC | Sensitivity | Specificity | F1 Score | MCC |
|---|---|---|---|---|---|---|---|
| STEMI | Scratch CNN–Transformer | 0.125 (0.093–0.170) | **0.743 (0.711–0.777)** | 0.050 (0.023–0.078) | **0.998 (0.997–0.999)** | 0.091 (0.042–0.138) | 0.147 (0.071–0.214) |
| STEMI | ECG-FM | 0.129 (0.094–0.179) | 0.726 (0.692–0.759) | **0.064 (0.033–0.096)** | **0.998 (0.997–0.999)** | **0.112 (0.059–0.165)** | **0.167 (0.092–0.233)** |
| STEMI | DeepECG-SSL | 0.113 (0.083–0.154) | 0.729 (0.698–0.760) | 0.059 (0.029–0.089) | 0.997 (0.995–0.998) | 0.102 (0.052–0.150) | 0.140 (0.072–0.204) |
| STEMI | CardioLoom | **0.138 (0.101–0.188)** | 0.719 (0.684–0.757) | 0.045 (0.020–0.073) | **0.998 (0.998–0.999)** | 0.083 (0.039–0.132) | 0.144 (0.071–0.212) |
| NSR | Scratch CNN–Transformer | 0.347 (0.317–0.376) | 0.769 (0.753–0.783) | **0.961 (0.947–0.973)** | 0.382 (0.369–0.394) | **0.322 (0.306–0.337)** | **0.247 (0.232–0.260)** |
| NSR | ECG-FM | 0.405 (0.370–0.436) | 0.795 (0.781–0.809) | 0.959 (0.945–0.971) | 0.375 (0.363–0.388) | 0.319 (0.303–0.334) | 0.241 (0.227–0.254) |
| NSR | DeepECG-SSL | **0.497 (0.463–0.530)** | **0.805 (0.789–0.819)** | 0.960 (0.946–0.971) | **0.383 (0.370–0.395)** | **0.322 (0.306–0.337)** | **0.247 (0.232–0.260)** |
| NSR | CardioLoom | 0.390 (0.356–0.424) | 0.770 (0.754–0.784) | 0.956 (0.942–0.969) | 0.382 (0.370–0.394) | 0.320 (0.305–0.336) | 0.244 (0.229–0.257) |
| VPB | Scratch CNN–Transformer | 0.692 (0.652–0.731) | 0.893 (0.875–0.910) | 0.779 (0.748–0.810) | 0.968 (0.963–0.972) | 0.756 (0.731–0.779) | 0.728 (0.700–0.753) |
| VPB | ECG-FM | **0.754 (0.717–0.788)** | **0.908 (0.893–0.923)** | 0.786 (0.756–0.815) | **0.971 (0.967–0.975)** | 0.771 (0.744–0.793) | 0.744 (0.716–0.768) |
| VPB | DeepECG-SSL | 0.735 (0.696–0.770) | 0.904 (0.888–0.920) | **0.793 (0.764–0.824)** | **0.971 (0.967–0.975)** | **0.774 (0.749–0.797)** | **0.748 (0.720–0.773)** |
| VPB | CardioLoom | 0.741 (0.704–0.778) | 0.906 (0.890–0.922) | 0.791 (0.762–0.820) | 0.969 (0.964–0.973) | 0.765 (0.739–0.789) | 0.738 (0.710–0.764) |
| APB | Scratch CNN–Transformer | 0.587 (0.541–0.629) | 0.901 (0.886–0.916) | 0.680 (0.645–0.716) | 0.957 (0.952–0.962) | 0.642 (0.611–0.672) | 0.605 (0.573–0.637) |
| APB | ECG-FM | 0.567 (0.522–0.614) | 0.900 (0.885–0.915) | **0.688 (0.652–0.724)** | 0.959 (0.954–0.964) | **0.655 (0.624–0.685)** | **0.620 (0.587–0.652)** |
| APB | DeepECG-SSL | 0.583 (0.537–0.629) | 0.910 (0.896–0.924) | 0.685 (0.649–0.721) | 0.956 (0.951–0.961) | 0.643 (0.612–0.672) | 0.607 (0.574–0.639) |
| APB | CardioLoom | **0.608 (0.563–0.653)** | **0.913 (0.899–0.927)** | 0.677 (0.642–0.714) | **0.960 (0.955–0.965)** | 0.650 (0.619–0.681) | 0.614 (0.581–0.648) |
| AF | Scratch CNN–Transformer | 0.876 (0.860–0.892) | 0.967 (0.963–0.971) | 0.478 (0.452–0.507) | 0.996 (0.994–0.997) | 0.639 (0.614–0.664) | 0.638 (0.617–0.661) |
| AF | ECG-FM | 0.886 (0.870–0.901) | 0.968 (0.963–0.972) | 0.509 (0.483–0.536) | **0.997 (0.996–0.999)** | 0.668 (0.645–0.692) | 0.667 (0.646–0.688) |
| AF | DeepECG-SSL | **0.914 (0.899–0.927)** | **0.978 (0.975–0.982)** | 0.503 (0.477–0.531) | **0.997 (0.996–0.999)** | 0.663 (0.640–0.688) | 0.662 (0.642–0.684) |
| AF | CardioLoom | 0.902 (0.888–0.916) | 0.974 (0.970–0.978) | **0.520 (0.494–0.548)** | 0.996 (0.995–0.998) | **0.677 (0.655–0.701)** | **0.672 (0.653–0.694)** |

| Disease | Model | AUPRC | AUROC | Sensitivity | Specificity | F1 Score | MCC |
|---|---|---|---|---|---|---|---|
| RBBB | Scratch CNN–Transformer | 0.953 (0.943–0.962) | 0.982 (0.978–0.985) | 0.805 (0.786–0.822) | 0.981 (0.978–0.985) | 0.867 (0.855–0.879) | 0.828 (0.813–0.842) |
| | ECG-FM | **0.962 (0.955–0.968)** | **0.985 (0.982–0.988)** | **0.810 (0.793–0.828)** | **0.984 (0.981–0.988)** | **0.875 (0.863–0.887)** | **0.838 (0.824–0.853)** |
| | DeepECG-SSL | 0.946 (0.937–0.955) | 0.975 (0.970–0.979) | 0.792 (0.774–0.810) | 0.982 (0.979–0.986) | 0.861 (0.848–0.873) | 0.821 (0.806–0.835) |
| | CardioLoom | 0.954 (0.944–0.963) | 0.981 (0.978–0.984) | 0.792 (0.774–0.810) | 0.983 (0.980–0.987) | 0.862 (0.850–0.874) | 0.823 (0.808–0.838) |
| 1° AVB | Scratch CNN–Transformer | 0.862 (0.836–0.886) | 0.974 (0.967–0.979) | 0.722 (0.688–0.754) | 0.987 (0.984–0.990) | 0.789 (0.764–0.812) | 0.771 (0.745–0.796) |
| | ECG-FM | 0.884 (0.860–0.907) | 0.979 (0.973–0.984) | 0.688 (0.655–0.720) | **0.991 (0.988–0.993)** | 0.780 (0.755–0.805) | 0.766 (0.740–0.791) |
| | DeepECG-SSL | **0.887 (0.863–0.910)** | **0.980 (0.974–0.985)** | **0.744 (0.713–0.777)** | 0.988 (0.986–0.991) | **0.808 (0.785–0.829)** | **0.791 (0.767–0.814)** |
| | CardioLoom | 0.885 (0.861–0.906) | 0.979 (0.974–0.984) | 0.711 (0.678–0.742) | 0.990 (0.987–0.992) | 0.790 (0.765–0.814) | 0.774 (0.748–0.799) |
| LBBB | Scratch CNN–Transformer | 0.875 (0.833–0.911) | 0.976 (0.963–0.987) | 0.602 (0.544–0.662) | 0.999 (0.998–1.000) | 0.742 (0.693–0.786) | 0.756 (0.712–0.796) |
| | ECG-FM | **0.896 (0.861–0.927)** | 0.977 (0.963–0.989) | 0.636 (0.576–0.698) | 0.999 (0.999–1.000) | 0.769 (0.724–0.813) | 0.781 (0.742–0.820) |
| | DeepECG-SSL | 0.892 (0.856–0.924) | 0.974 (0.960–0.986) | **0.653 (0.595–0.712)** | 0.999 (0.999–1.000) | **0.782 (0.737–0.824)** | **0.792 (0.753–0.830)** |
| | CardioLoom | 0.894 (0.858–0.926) | **0.978 (0.967–0.988)** | 0.631 (0.569–0.692) | **1.000 (0.999–1.000)** | 0.768 (0.719–0.813) | 0.781 (0.739–0.821) |

Values are estimates (95% cluster-bootstrap CIs). NE indicates fewer than 950 valid replicates among 1000. Sensitivity, specificity, F1 score, and MCC use the Zhongshan validation-F1 threshold. Bold type marks the highest displayed estimate within each disease.

**Supplementary Table 26: Detailed task-specific performance comparison on the external Shiyuan cohort.**

| Disease | Model | AUPRC | AUROC | Sensitivity | Specificity | F1 Score | MCC |
|---|---|---|---|---|---|---|---|
| NER | Scratch CNN–Transformer | 0.003 (0.003–0.004) | 0.621 (0.593–0.648) | 0.675 (0.627–0.719) | 0.482 (0.479–0.486) | 0.005 (0.005–0.006) | 0.014 (0.010–0.018) |
| | ECG-FM | **0.004 (0.003–0.005)** | 0.736 (0.715–0.758) | **0.900 (0.866–0.932)** | 0.448 (0.444–0.451) | **0.007 (0.006–0.007)** | 0.031 (0.028–0.035) |
| | DeepECG-SSL | 0.003 (0.002–0.003) | 0.632 (0.606–0.658) | 0.784 (0.742–0.827) | 0.410 (0.406–0.413) | 0.005 (0.005–0.006) | 0.018 (0.013–0.022) |
| | CardioLoom | **0.004 (0.004–0.005)** | **0.750 (0.727–0.772)** | 0.866 (0.826–0.901) | **0.490 (0.487–0.493)** | **0.007 (0.006–0.008)** | **0.032 (0.028–0.036)** |
| LAE | Scratch CNN–Transformer | 0.328 (0.319–0.338) | 0.793 (0.789–0.797) | 0.659 (0.652–0.667) | 0.754 (0.751–0.757) | 0.377 (0.371–0.383) | 0.293 (0.286–0.299) |
| | ECG-FM | 0.334 (0.325–0.343) | 0.795 (0.791–0.798) | 0.671 (0.664–0.679) | 0.755 (0.752–0.757) | 0.383 (0.377–0.389) | 0.301 (0.295–0.307) |
| | DeepECG-SSL | 0.353 (0.343–0.363) | 0.804 (0.800–0.808) | 0.629 (0.621–0.637) | **0.794 (0.791–0.796)** | **0.396 (0.390–0.403)** | 0.312 (0.305–0.319) |
| | CardioLoom | **0.366 (0.357–0.376)** | **0.813 (0.810–0.817)** | **0.700 (0.692–0.707)** | 0.753 (0.750–0.756) | 0.394 (0.388–0.400) | **0.318 (0.312–0.324)** |
| HCM | Scratch CNN–Transformer | 0.335 (0.324–0.348) | 0.820 (0.816–0.825) | 0.548 (0.538–0.559) | 0.865 (0.862–0.867) | 0.355 (0.348–0.363) | 0.301 (0.293–0.309) |
| | ECG-FM | 0.375 (0.363–0.388) | 0.841 (0.838–0.845) | **0.575 (0.565–0.584)** | 0.872 (0.870–0.874) | 0.380 (0.372–0.388) | 0.330 (0.322–0.337) |
| | DeepECG-SSL | **0.398 (0.386–0.411)** | **0.850 (0.846–0.854)** | 0.522 (0.512–0.532) | **0.908 (0.906–0.910)** | **0.407 (0.399–0.415)** | **0.353 (0.344–0.361)** |
| | CardioLoom | 0.381 (0.369–0.395) | 0.840 (0.836–0.844) | 0.555 (0.545–0.565) | 0.883 (0.881–0.885) | 0.384 (0.376–0.392) | 0.332 (0.324–0.340) |
| nHCM | Scratch CNN–Transformer | 0.294 (0.284–0.306) | 0.815 (0.810–0.819) | 0.549 (0.539–0.560) | 0.858 (0.856–0.860) | 0.342 (0.334–0.350) | 0.289 (0.281–0.297) |
| | ECG-FM | 0.315 (0.305–0.327) | 0.834 (0.831–0.839) | **0.599 (0.590–0.609)** | 0.853 (0.850–0.855) | 0.361 (0.353–0.368) | 0.315 (0.307–0.323) |
| | DeepECG-SSL | **0.345 (0.334–0.357)** | **0.845 (0.841–0.849)** | 0.541 (0.531–0.551) | **0.894 (0.892–0.896)** | **0.389 (0.380–0.397)** | **0.337 (0.328–0.345)** |
| | CardioLoom | 0.330 (0.319–0.344) | 0.836 (0.832–0.840) | 0.542 (0.532–0.553) | 0.883 (0.881–0.885) | 0.372 (0.364–0.380) | 0.320 (0.311–0.328) |
| BAE | Scratch CNN–Transformer | 0.360 (0.338–0.380) | 0.970 (0.968–0.972) | 0.641 (0.621–0.660) | 0.978 (0.977–0.978) | 0.431 (0.415–0.447) | 0.444 (0.429–0.459) |
| | ECG-FM | 0.394 (0.370–0.417) | 0.974 (0.972–0.976) | 0.724 (0.704–0.742) | 0.977 (0.977–0.978) | 0.472 (0.456–0.488) | 0.492 (0.477–0.507) |
| | DeepECG-SSL | 0.415 (0.392–0.439) | 0.978 (0.976–0.980) | 0.741 (0.721–0.759) | **0.979 (0.978–0.980)** | 0.499 (0.483–0.515) | 0.518 (0.501–0.533) |
| | CardioLoom | **0.439 (0.417–0.463)** | **0.979 (0.977–0.981)** | **0.779 (0.763–0.797)** | **0.979 (0.978–0.980)** | **0.515 (0.499–0.530)** | **0.537 (0.523–0.552)** |

| Disease | Model | AUPRC | AUROC | Sensitivity | Specificity | F1 Score | MCC |
|---|---|---|---|---|---|---|---|
| HFrEF | Scratch CNN–Transformer | 0.483 (0.460–0.505) | 0.950 (0.946–0.955) | 0.449 (0.429–0.469) | 0.993 (0.992–0.993) | 0.489 (0.470–0.507) | 0.483 (0.463–0.501) |
| | ECG-FM | 0.616 (0.595–0.636) | **0.969 (0.965–0.972)** | 0.411 (0.392–0.431) | **0.997 (0.997–0.997)** | 0.526 (0.507–0.545) | 0.542 (0.524–0.560) |
| | DeepECG-SSL | 0.616 (0.594–0.640) | 0.966 (0.962–0.970) | 0.444 (0.424–0.466) | 0.996 (0.996–0.997) | 0.544 (0.524–0.565) | 0.552 (0.533–0.573) |
| | CardioLoom | **0.641 (0.620–0.662)** | **0.969 (0.965–0.973)** | **0.498 (0.478–0.517)** | 0.996 (0.996–0.996) | **0.578 (0.559–0.596)** | **0.580 (0.561–0.597)** |
| RMWSSFLV | Scratch CNN–Transformer | 0.001 (0.000–0.004) | 0.872 (0.710–0.996) | 0.400 (0.000–1.000) | 0.984 (0.983–0.985) | 0.001 (0.000–0.003) | 0.015 (-0.001–0.034) |
| | ECG-FM | 0.002 (0.000–0.009) | 0.952 (0.866–0.998) | 0.400 (0.000–1.000) | 0.982 (0.981–0.983) | 0.001 (0.000–0.003) | 0.014 (-0.001–0.031) |
| | DeepECG-SSL | **0.003 (0.000–0.015)** | 0.955 (0.886–0.999) | **0.600 (0.000–1.000)** | 0.982 (0.981–0.983) | **0.002 (0.000–0.004)** | 0.021 (0.000–0.039) |
| | CardioLoom | 0.001 (0.000–0.005) | **0.959 (0.886–0.998)** | **0.600 (0.000–1.000)** | **0.985 (0.984–0.985)** | **0.002 (0.000–0.004)** | **0.023 (0.000–0.042)** |
| DCM | Scratch CNN–Transformer | 0.001 (0.000–0.002) | 0.984 (0.965–0.997) | 0.333 (0.000–1.000) | 0.994 (0.993–0.994) | 0.001 (0.000–0.005) | 0.015 (0.000–0.041) |
| | ECG-FM | 0.001 (0.000–0.004) | 0.986 (0.975–0.999) | 0.333 (0.000–1.000) | 0.993 (0.992–0.993) | 0.001 (0.000–0.004) | 0.014 (0.000–0.039) |
| | DeepECG-SSL | **0.003 (0.000–0.011)** | **0.996 (0.993–1.000)** | **0.667 (0.000–1.000)** | **0.995 (0.994–0.995)** | **0.004 (0.000–0.009)** | 0.034 (0.000–0.064) |
| | CardioLoom | 0.001 (0.000–0.003) | 0.993 (0.987–0.997) | **0.667 (0.000–1.000)** | **0.995 (0.995–0.995)** | **0.004 (0.000–0.009)** | **0.035 (0.000–0.066)** |
| oHCM | Scratch CNN–Transformer | 0.026 (0.020–0.035) | 0.836 (0.815–0.857) | 0.150 (0.117–0.186) | **0.995 (0.994–0.995)** | 0.092 (0.071–0.116) | 0.097 (0.074–0.121) |
| | ECG-FM | **0.047 (0.036–0.062)** | 0.882 (0.865–0.899) | 0.179 (0.144–0.214) | **0.995 (0.995–0.996)** | 0.117 (0.092–0.143) | 0.121 (0.095–0.147) |
| | DeepECG-SSL | 0.039 (0.030–0.052) | 0.867 (0.847–0.885) | **0.207 (0.168–0.249)** | 0.994 (0.993–0.994) | 0.112 (0.090–0.139) | 0.123 (0.098–0.152) |
| | CardioLoom | **0.047 (0.036–0.061)** | **0.892 (0.875–0.909)** | 0.186 (0.151–0.224) | **0.995 (0.995–0.996)** | **0.120 (0.096–0.146)** | **0.125 (0.100–0.152)** |
| LVDD | Scratch CNN–Transformer | 0.823 (0.819–0.826) | 0.737 (0.734–0.740) | 0.016 (0.015–0.017) | **0.993 (0.993–0.994)** | 0.032 (0.030–0.034) | 0.040 (0.036–0.044) |
| | ECG-FM | **0.839 (0.836–0.842)** | **0.763 (0.760–0.766)** | **0.030 (0.029–0.031)** | 0.990 (0.989–0.991) | **0.058 (0.056–0.060)** | **0.063 (0.059–0.066)** |
| | DeepECG-SSL | 0.826 (0.822–0.829) | 0.741 (0.738–0.745) | 0.020 (0.019–0.021) | **0.993 (0.992–0.994)** | 0.039 (0.037–0.041) | 0.050 (0.046–0.053) |
| | CardioLoom | 0.816 (0.813–0.820) | 0.737 (0.734–0.741) | 0.029 (0.028–0.030) | 0.986 (0.985–0.988) | 0.056 (0.054–0.058) | 0.047 (0.042–0.051) |

| Disease | Model | AUPRC | AUROC | Sensitivity | Specificity | F1 Score | MCC |
|---|---|---|---|---|---|---|---|
| VA | Scratch CNN–Transformer | 0.346 (0.296–0.400) | 0.981 (0.974–0.987) | 0.344 (0.297–0.393) | **0.999 (0.998–0.999)** | 0.392 (0.346–0.437) | 0.394 (0.348–0.439) |
| | ECG-FM | 0.412 (0.360–0.472) | 0.982 (0.975–0.989) | **0.516 (0.466–0.576)** | 0.998 (0.998–0.998) | **0.477 (0.436–0.522)** | **0.476 (0.436–0.521)** |
| | DeepECG-SSL | **0.442 (0.384–0.507)** | **0.987 (0.982–0.992)** | 0.378 (0.330–0.435) | **0.999 (0.999–0.999)** | 0.443 (0.395–0.495) | 0.448 (0.402–0.499) |
| | CardioLoom | 0.441 (0.386–0.504) | **0.987 (0.981–0.992)** | 0.459 (0.406–0.518) | 0.998 (0.998–0.999) | 0.471 (0.424–0.520) | 0.470 (0.423–0.520) |
| RAE | Scratch CNN–Transformer | 0.026 (0.018–0.037) | 0.946 (0.932–0.957) | 0.197 (0.135–0.269) | **0.996 (0.995–0.996)** | **0.074 (0.050–0.101)** | 0.093 (0.064–0.127) |
| | ECG-FM | 0.022 (0.016–0.033) | 0.946 (0.931–0.960) | 0.216 (0.154–0.287) | 0.993 (0.993–0.994) | 0.055 (0.038–0.073) | 0.080 (0.057–0.106) |
| | DeepECG-SSL | **0.032 (0.020–0.050)** | 0.950 (0.933–0.965) | **0.376 (0.290–0.467)** | 0.988 (0.987–0.988) | 0.054 (0.039–0.070) | **0.102 (0.075–0.129)** |
| | CardioLoom | 0.031 (0.022–0.046) | **0.958 (0.943–0.970)** | 0.179 (0.121–0.247) | **0.996 (0.996–0.997)** | 0.071 (0.047–0.099) | 0.087 (0.058–0.120) |
| AM | Scratch CNN–Transformer | 0.000 (0.000–0.001) | 0.698 (0.612–0.780) | 0.000 (0.000–0.000) | 0.999 (0.999–0.999) | 0.000 (0.000–0.000) | 0.000 (0.000–0.000) |
| | ECG-FM | **0.035 (0.000–0.116)** | 0.753 (0.648–0.846) | **0.033 (0.000–0.111)** | **1.000 (1.000–1.000)** | **0.029 (0.000–0.100)** | **0.029 (0.000–0.101)** |
| | DeepECG-SSL | 0.002 (0.000–0.006) | **0.760 (0.659–0.838)** | 0.000 (0.000–0.000) | **1.000 (1.000–1.000)** | 0.000 (0.000–0.000) | 0.000 (0.000–0.000) |
| | CardioLoom | 0.002 (0.000–0.005) | 0.754 (0.652–0.836) | 0.000 (0.000–0.000) | **1.000 (1.000–1.000)** | 0.000 (0.000–0.000) | 0.000 (0.000–0.000) |
| LAM | Scratch CNN–Transformer | 0.001 (0.000–0.004) | 0.696 (0.594–0.755) | 0.036 (0.000–0.127) | 0.999 (0.998–0.999) | 0.011 (0.000–0.034) | 0.015 (-0.001–0.050) |
| | ECG-FM | 0.006 (0.000–0.032) | 0.704 (0.628–0.765) | **0.091 (0.000–0.242)** | **1.000 (1.000–1.000)** | **0.076 (0.000–0.168)** | **0.077 (0.000–0.176)** |
| | DeepECG-SSL | **0.011 (0.001–0.070)** | **0.785 (0.623–0.868)** | 0.055 (0.000–0.163) | **1.000 (1.000–1.000)** | 0.063 (0.000–0.159) | 0.064 (0.000–0.160) |
| | CardioLoom | 0.007 (0.000–0.046) | 0.690 (0.606–0.819) | 0.018 (0.000–0.064) | **1.000 (1.000–1.000)** | 0.029 (0.000–0.092) | 0.035 (0.000–0.109) |
| AVC | Scratch CNN–Transformer | 0.256 (0.249–0.263) | 0.755 (0.751–0.759) | 0.670 (0.662–0.679) | 0.697 (0.694–0.700) | 0.321 (0.316–0.327) | 0.240 (0.235–0.247) |
| | ECG-FM | 0.252 (0.246–0.259) | 0.756 (0.751–0.760) | 0.673 (0.665–0.681) | 0.693 (0.690–0.697) | 0.320 (0.314–0.325) | 0.239 (0.233–0.245) |
| | DeepECG-SSL | **0.271 (0.264–0.280)** | **0.770 (0.766–0.774)** | 0.684 (0.676–0.694) | **0.706 (0.703–0.709)** | **0.333 (0.327–0.339)** | **0.257 (0.250–0.263)** |
| | CardioLoom | **0.271 (0.264–0.279)** | 0.768 (0.763–0.772) | **0.726 (0.718–0.734)** | 0.664 (0.661–0.667) | 0.322 (0.317–0.328) | 0.249 (0.244–0.255) |

| Disease | Model | AUPRC | AUROC | Sensitivity | Specificity | F1 Score | MCC |
|---|---|---|---|---|---|---|---|
| TR | Scratch CNN–Transformer | 0.366 (0.349–0.383) | 0.889 (0.883–0.894) | 0.530 (0.514–0.546) | 0.962 (0.961–0.963) | 0.407 (0.396–0.419) | 0.393 (0.381–0.405) |
| | ECG-FM | 0.438 (0.420–0.454) | **0.910 (0.904–0.916)** | **0.606 (0.590–0.621)** | 0.961 (0.960–0.962) | 0.446 (0.435–0.457) | 0.438 (0.427–0.449) |
| | DeepECG-SSL | **0.439 (0.420–0.455)** | 0.909 (0.904–0.915) | 0.589 (0.572–0.604) | 0.965 (0.964–0.966) | 0.458 (0.445–0.470) | **0.447 (0.434–0.459)** |
| | CardioLoom | 0.433 (0.416–0.450) | 0.907 (0.900–0.912) | 0.566 (0.549–0.582) | **0.969 (0.968–0.970)** | **0.461 (0.447–0.473)** | 0.446 (0.433–0.459) |
| MR | Scratch CNN–Transformer | 0.221 (0.206–0.236) | 0.870 (0.860–0.878) | 0.434 (0.415–0.453) | 0.966 (0.965–0.967) | 0.308 (0.296–0.322) | 0.300 (0.286–0.314) |
| | ECG-FM | **0.283 (0.266–0.301)** | 0.894 (0.886–0.901) | 0.502 (0.483–0.522) | **0.967 (0.966–0.968)** | **0.353 (0.341–0.367)** | **0.349 (0.336–0.363)** |
| | DeepECG-SSL | 0.277 (0.260–0.296) | **0.898 (0.891–0.905)** | **0.504 (0.483–0.524)** | 0.965 (0.964–0.966) | 0.345 (0.332–0.359) | 0.342 (0.328–0.357) |
| | CardioLoom | 0.271 (0.253–0.291) | 0.896 (0.889–0.904) | 0.501 (0.481–0.521) | 0.966 (0.965–0.967) | 0.348 (0.335–0.363) | 0.344 (0.331–0.359) |
| AR | Scratch CNN–Transformer | 0.092 (0.083–0.103) | 0.800 (0.790–0.810) | 0.167 (0.153–0.184) | 0.985 (0.984–0.985) | 0.160 (0.146–0.175) | 0.146 (0.132–0.161) |
| | ECG-FM | **0.133 (0.120–0.146)** | 0.831 (0.822–0.840) | **0.184 (0.168–0.201)** | 0.987 (0.987–0.988) | 0.188 (0.173–0.204) | 0.175 (0.160–0.191) |
| | DeepECG-SSL | 0.127 (0.114–0.140) | 0.834 (0.825–0.843) | 0.162 (0.147–0.177) | **0.990 (0.990–0.991)** | 0.185 (0.169–0.201) | 0.175 (0.159–0.191) |
| | CardioLoom | 0.131 (0.119–0.145) | **0.836 (0.827–0.845)** | 0.173 (0.158–0.189) | **0.990 (0.989–0.990)** | **0.192 (0.177–0.208)** | **0.181 (0.166–0.198)** |
| MVPLAC | Scratch CNN–Transformer | 0.057 (0.052–0.062) | 0.803 (0.794–0.811) | 0.328 (0.309–0.349) | 0.926 (0.925–0.928) | 0.107 (0.100–0.115) | 0.116 (0.107–0.126) |
| | ECG-FM | **0.078 (0.070–0.085)** | 0.826 (0.818–0.835) | 0.406 (0.385–0.428) | **0.931 (0.929–0.932)** | **0.138 (0.129–0.146)** | **0.157 (0.147–0.167)** |
| | DeepECG-SSL | 0.076 (0.069–0.084) | 0.833 (0.825–0.840) | 0.406 (0.385–0.431) | 0.930 (0.929–0.932) | 0.137 (0.128–0.146) | 0.156 (0.146–0.167) |
| | CardioLoom | **0.078 (0.071–0.086)** | **0.835 (0.827–0.843)** | **0.419 (0.395–0.441)** | 0.926 (0.924–0.928) | 0.135 (0.126–0.143) | 0.156 (0.145–0.166) |
| MS | Scratch CNN–Transformer | 0.124 (0.084–0.174) | 0.942 (0.926–0.956) | 0.279 (0.220–0.341) | 0.997 (0.997–0.997) | 0.155 (0.122–0.191) | 0.172 (0.135–0.209) |
| | ECG-FM | **0.275 (0.211–0.336)** | **0.981 (0.976–0.985)** | **0.453 (0.385–0.522)** | **0.998 (0.998–0.998)** | **0.307 (0.257–0.355)** | **0.323 (0.271–0.372)** |
| | DeepECG-SSL | 0.167 (0.115–0.229) | 0.974 (0.966–0.981) | 0.397 (0.339–0.459) | 0.997 (0.997–0.997) | 0.214 (0.176–0.253) | 0.240 (0.202–0.280) |
| | CardioLoom | 0.180 (0.127–0.239) | 0.971 (0.960–0.980) | 0.366 (0.299–0.435) | **0.998 (0.997–0.998)** | 0.229 (0.182–0.272) | 0.245 (0.198–0.290) |

| Disease | Model | AUPRC | AUROC | Sensitivity | Specificity | F1 Score | MCC |
|---|---|---|---|---|---|---|---|
| AS | Scratch CNN–Transformer | 0.029 (0.020–0.043) | 0.823 (0.800–0.843) | 0.170 (0.135–0.207) | 0.990 (0.990–0.991) | 0.068 (0.053–0.084) | 0.081 (0.063–0.099) |
| | ECG-FM | **0.087 (0.061–0.119)** | 0.878 (0.858–0.896) | **0.188 (0.149–0.229)** | 0.997 (0.997–0.998) | **0.172 (0.134–0.209)** | **0.170 (0.133–0.208)** |
| | DeepECG-SSL | 0.081 (0.057–0.112) | 0.888 (0.872–0.903) | 0.144 (0.111–0.178) | **0.998 (0.998–0.999)** | 0.159 (0.123–0.193) | 0.158 (0.122–0.192) |
| | CardioLoom | 0.076 (0.054–0.104) | **0.889 (0.872–0.904)** | 0.155 (0.121–0.193) | 0.997 (0.997–0.998) | 0.141 (0.108–0.175) | 0.139 (0.108–0.173) |
| BAV | Scratch CNN–Transformer | 0.003 (0.002–0.005) | 0.623 (0.577–0.670) | **0.040 (0.014–0.078)** | 0.993 (0.993–0.994) | 0.011 (0.004–0.022) | 0.013 (0.003–0.029) |
| | ECG-FM | 0.004 (0.002–0.005) | **0.692 (0.642–0.737)** | 0.028 (0.008–0.052) | **0.998 (0.997–0.998)** | 0.018 (0.005–0.033) | 0.018 (0.004–0.034) |
| | DeepECG-SSL | 0.004 (0.003–0.006) | 0.667 (0.621–0.712) | **0.040 (0.018–0.065)** | **0.998 (0.998–0.998)** | **0.027 (0.011–0.043)** | **0.027 (0.011–0.044)** |
| | CardioLoom | **0.005 (0.003–0.011)** | 0.688 (0.644–0.733) | 0.032 (0.008–0.062) | **0.998 (0.998–0.999)** | **0.027 (0.007–0.052)** | 0.026 (0.006–0.052) |
| PS | Scratch CNN–Transformer | 0.002 (0.000–0.010) | 0.682 (0.553–0.790) | 0.065 (0.000–0.152) | **1.000 (0.999–1.000)** | 0.031 (0.000–0.075) | 0.036 (0.000–0.083) |
| | ECG-FM | **0.010 (0.000–0.047)** | 0.707 (0.567–0.839) | **0.161 (0.030–0.333)** | **1.000 (0.999–1.000)** | **0.080 (0.016–0.162)** | **0.092 (0.018–0.185)** |
| | DeepECG-SSL | 0.002 (0.000–0.012) | 0.704 (0.571–0.834) | 0.032 (0.000–0.115) | 0.999 (0.999–0.999) | 0.012 (0.000–0.037) | 0.015 (0.000–0.049) |
| | CardioLoom | 0.003 (0.000–0.014) | **0.770 (0.644–0.872)** | 0.000 (0.000–0.000) | **1.000 (1.000–1.000)** | 0.000 (0.000–0.000) | 0.000 (0.000–0.000) |
| EA | Scratch CNN–Transformer | **0.005 (0.000–0.029)** | **0.945 (0.877–0.998)** | **0.000 (0.000–0.000)** | **1.000 (1.000–1.000)** | **0.000 (0.000–0.000)** | **0.000 (0.000–0.000)** |
| | ECG-FM | 0.000 (0.000–0.001) | 0.867 (0.815–0.938) | **0.000 (0.000–0.000)** | 0.998 (0.998–0.999) | **0.000 (0.000–0.000)** | **0.000 (0.000–0.000)** |
| | DeepECG-SSL | 0.001 (0.000–0.003) | 0.872 (0.754–0.979) | **0.000 (0.000–0.000)** | **1.000 (1.000–1.000)** | **0.000 (0.000–0.000)** | **0.000 (0.000–0.000)** |
| | CardioLoom | 0.002 (0.000–0.007) | 0.878 (0.702–0.984) | **0.000 (0.000–0.000)** | **1.000 (1.000–1.000)** | **0.000 (0.000–0.000)** | **0.000 (0.000–0.000)** |
| PR | Scratch CNN–Transformer | 0.003 (0.002–0.005) | 0.668 (0.603–0.725) | 0.012 (0.000–0.032) | 0.999 (0.999–0.999) | 0.012 (0.000–0.030) | 0.011 (-0.001–0.030) |
| | ECG-FM | 0.005 (0.003–0.012) | 0.716 (0.662–0.764) | 0.000 (0.000–0.000) | **1.000 (1.000–1.000)** | 0.000 (0.000–0.000) | 0.000 (0.000–0.000) |
| | DeepECG-SSL | 0.005 (0.003–0.009) | **0.775 (0.726–0.819)** | 0.000 (0.000–0.000) | **1.000 (1.000–1.000)** | 0.000 (0.000–0.000) | 0.000 (0.000–0.000) |
| | CardioLoom | **0.007 (0.003–0.018)** | 0.727 (0.678–0.777) | **0.037 (0.007–0.076)** | 0.999 (0.999–0.999) | **0.034 (0.006–0.070)** | **0.033 (0.005–0.069)** |

| Disease | Model | AUPRC | AUROC | Sensitivity | Specificity | F1 Score | MCC |
|---|---|---|---|---|---|---|---|
| CHD | Scratch CNN–Transformer | 0.179 (0.107–0.281) | 0.862 (0.834–0.888) | 0.339 (0.257–0.433) | 0.994 (0.993–0.994) | 0.212 (0.153–0.285) | 0.224 (0.164–0.298) |
| | ECG-FM | **0.247 (0.179–0.333)** | 0.890 (0.866–0.913) | **0.346 (0.271–0.428)** | **0.998 (0.997–0.998)** | 0.333 (0.263–0.416) | 0.331 (0.260–0.414) |
| | DeepECG-SSL | 0.227 (0.178–0.289) | **0.899 (0.875–0.920)** | 0.322 (0.263–0.381) | **0.998 (0.998–0.998)** | 0.329 (0.269–0.396) | 0.327 (0.268–0.394) |
| | CardioLoom | 0.244 (0.174–0.338) | 0.889 (0.862–0.912) | 0.340 (0.264–0.429) | **0.998 (0.998–0.998)** | **0.355 (0.278–0.448)** | **0.353 (0.276–0.446)** |
| ASD | Scratch CNN–Transformer | 0.161 (0.098–0.259) | 0.849 (0.816–0.880) | 0.314 (0.227–0.415) | 0.995 (0.994–0.995) | 0.204 (0.142–0.287) | 0.215 (0.150–0.300) |
| | ECG-FM | 0.273 (0.189–0.377) | **0.894 (0.868–0.917)** | **0.376 (0.293–0.469)** | **0.998 (0.997–0.998)** | 0.351 (0.267–0.444) | 0.350 (0.266–0.442) |
| | DeepECG-SSL | 0.239 (0.158–0.342) | 0.881 (0.852–0.907) | 0.323 (0.248–0.407) | **0.998 (0.997–0.998)** | 0.307 (0.232–0.392) | 0.305 (0.231–0.389) |
| | CardioLoom | **0.293 (0.197–0.409)** | 0.874 (0.844–0.904) | 0.350 (0.264–0.452) | **0.998 (0.998–0.999)** | **0.378 (0.288–0.481)** | **0.377 (0.288–0.480)** |
| VSD | Scratch CNN–Transformer | 0.008 (0.004–0.015) | 0.835 (0.791–0.874) | 0.151 (0.088–0.226) | 0.993 (0.992–0.993) | 0.024 (0.013–0.037) | 0.043 (0.024–0.064) |
| | ECG-FM | 0.045 (0.019–0.086) | 0.862 (0.814–0.904) | **0.209 (0.130–0.295)** | 0.996 (0.996–0.997) | 0.062 (0.039–0.088) | 0.086 (0.053–0.120) |
| | DeepECG-SSL | 0.037 (0.018–0.076) | 0.881 (0.840–0.921) | 0.129 (0.075–0.196) | **0.999 (0.999–0.999)** | 0.094 (0.055–0.136) | 0.097 (0.057–0.142) |
| | CardioLoom | **0.070 (0.032–0.121)** | **0.886 (0.852–0.919)** | **0.209 (0.136–0.296)** | 0.998 (0.998–0.999) | **0.110 (0.071–0.152)** | **0.124 (0.080–0.171)** |
| PFO | Scratch CNN–Transformer | **0.006 (0.005–0.009)** | 0.586 (0.552–0.618) | 0.003 (0.000–0.009) | **0.999 (0.999–1.000)** | 0.005 (0.000–0.016) | 0.005 (-0.001–0.021) |
| | ECG-FM | **0.006 (0.005–0.007)** | 0.592 (0.555–0.626) | 0.027 (0.015–0.042) | 0.994 (0.994–0.995) | 0.019 (0.010–0.029) | 0.016 (0.007–0.026) |
| | DeepECG-SSL | **0.006 (0.005–0.007)** | **0.617 (0.588–0.644)** | 0.001 (0.000–0.005) | **0.999 (0.999–0.999)** | 0.002 (0.000–0.006) | 0.000 (-0.002–0.005) |
| | CardioLoom | **0.006 (0.005–0.008)** | 0.615 (0.584–0.644) | **0.032 (0.017–0.047)** | 0.995 (0.994–0.995) | **0.023 (0.013–0.035)** | **0.020 (0.009–0.032)** |
| PDA | Scratch CNN–Transformer | 0.004 (0.001–0.021) | 0.729 (0.652–0.811) | 0.046 (0.000–0.104) | **1.000 (0.999–1.000)** | 0.037 (0.000–0.083) | 0.038 (0.000–0.084) |
| | ECG-FM | 0.008 (0.002–0.027) | 0.796 (0.736–0.861) | **0.077 (0.014–0.163)** | 0.999 (0.999–0.999) | 0.046 (0.009–0.096) | 0.050 (0.009–0.104) |
| | DeepECG-SSL | 0.009 (0.003–0.026) | **0.838 (0.751–0.911)** | 0.046 (0.000–0.107) | **1.000 (1.000–1.000)** | 0.048 (0.000–0.101) | 0.047 (0.000–0.101) |
| | CardioLoom | **0.018 (0.005–0.052)** | 0.826 (0.753–0.896) | 0.062 (0.013–0.131) | **1.000 (1.000–1.000)** | **0.087 (0.019–0.173)** | **0.095 (0.022–0.184)** |

| Disease | Model | AUPRC | AUROC | Sensitivity | Specificity | F1 Score | MCC |
|---|---|---|---|---|---|---|---|
| TOF | Scratch CNN–Transformer | 0.001 (0.001–0.003) | 0.799 (0.677–0.899) | 0.000 (0.000–0.000) | **0.998 (0.997–0.998)** | 0.000 (0.000–0.000) | -0.001 (-0.001–0.000) |
| | ECG-FM | 0.001 (0.000–0.002) | **0.901 (0.860–0.937)** | **1.000 (1.000–1.000)** | 0.088 (0.086–0.090) | 0.000 (0.000–0.000) | 0.003 (0.003–0.004) |
| | DeepECG-SSL | 0.001 (0.000–0.005) | 0.663 (0.541–0.769) | 0.481 (0.250–0.692) | 0.698 (0.696–0.701) | 0.000 (0.000–0.001) | 0.004 (-0.001–0.010) |
| | CardioLoom | **0.005 (0.000–0.025)** | 0.780 (0.676–0.879) | 0.630 (0.414–0.857) | 0.816 (0.814–0.818) | **0.001 (0.000–0.001)** | **0.013 (0.006–0.019)** |
| PH | Scratch CNN–Transformer | 0.368 (0.359–0.377) | 0.763 (0.759–0.767) | 0.467 (0.458–0.475) | 0.877 (0.875–0.879) | 0.396 (0.389–0.403) | 0.303 (0.296–0.310) |
| | ECG-FM | **0.412 (0.403–0.421)** | **0.798 (0.794–0.802)** | 0.459 (0.451–0.467) | **0.908 (0.907–0.910)** | **0.433 (0.425–0.440)** | **0.350 (0.342–0.358)** |
| | DeepECG-SSL | 0.408 (0.399–0.417) | 0.791 (0.787–0.795) | **0.476 (0.467–0.485)** | 0.896 (0.895–0.898) | 0.428 (0.421–0.435) | 0.342 (0.334–0.349) |
| | CardioLoom | 0.404 (0.395–0.414) | 0.787 (0.783–0.791) | **0.476 (0.467–0.484)** | 0.895 (0.893–0.897) | 0.426 (0.418–0.433) | 0.339 (0.331–0.347) |
| AD | Scratch CNN–Transformer | 0.311 (0.303–0.319) | 0.756 (0.751–0.760) | 0.436 (0.428–0.444) | 0.856 (0.854–0.858) | 0.360 (0.353–0.367) | 0.253 (0.245–0.260) |
| | ECG-FM | 0.358 (0.348–0.366) | 0.790 (0.786–0.794) | 0.400 (0.392–0.409) | **0.903 (0.901–0.905)** | 0.389 (0.381–0.396) | 0.296 (0.288–0.304) |
| | DeepECG-SSL | 0.363 (0.353–0.372) | **0.793 (0.789–0.797)** | 0.434 (0.425–0.443) | 0.888 (0.886–0.890) | 0.395 (0.387–0.403) | 0.299 (0.290–0.307) |
| | CardioLoom | **0.367 (0.358–0.375)** | 0.791 (0.787–0.795) | **0.451 (0.443–0.461)** | 0.882 (0.880–0.884) | **0.400 (0.393–0.407)** | **0.303 (0.295–0.311)** |
| Dex | Scratch CNN–Transformer | **0.023 (0.011–0.045)** | 0.881 (0.824–0.925) | **0.115 (0.044–0.188)** | **1.000 (0.999–1.000)** | **0.092 (0.033–0.154)** | **0.094 (0.033–0.155)** |
| | ECG-FM | 0.001 (0.001–0.002) | 0.763 (0.699–0.822) | 0.000 (0.000–0.000) | **1.000 (1.000–1.000)** | 0.000 (0.000–0.000) | 0.000 (0.000–0.000) |
| | DeepECG-SSL | 0.003 (0.001–0.007) | 0.852 (0.799–0.900) | 0.000 (0.000–0.000) | **1.000 (1.000–1.000)** | 0.000 (0.000–0.000) | 0.000 (0.000–0.000) |
| | CardioLoom | 0.019 (0.009–0.036) | **0.923 (0.890–0.952)** | 0.013 (0.000–0.043) | **1.000 (1.000–1.000)** | 0.021 (0.000–0.069) | 0.027 (0.000–0.085) |
| PE | Scratch CNN–Transformer | 0.177 (0.168–0.186) | 0.742 (0.736–0.748) | **0.296 (0.284–0.306)** | 0.941 (0.939–0.942) | 0.253 (0.243–0.262) | 0.206 (0.197–0.216) |
| | ECG-FM | 0.212 (0.202–0.223) | 0.760 (0.754–0.766) | 0.240 (0.230–0.251) | 0.969 (0.969–0.970) | 0.270 (0.259–0.281) | 0.236 (0.225–0.247) |
| | DeepECG-SSL | 0.218 (0.208–0.228) | 0.766 (0.761–0.772) | 0.282 (0.271–0.293) | 0.958 (0.957–0.959) | **0.279 (0.269–0.289)** | 0.237 (0.227–0.248) |
| | CardioLoom | **0.226 (0.215–0.237)** | **0.769 (0.764–0.775)** | 0.204 (0.194–0.214) | **0.979 (0.978–0.980)** | 0.259 (0.248–0.270) | **0.238 (0.227–0.250)** |

| Disease | Model | AUPRC | AUROC | Sensitivity | Specificity | F1 Score | MCC |
|---|---|---|---|---|---|---|---|
| CP | Scratch CNN–Transformer | 0.003 (0.000–0.019) | 0.985 (0.960–1.000) | **0.250 (0.000–1.000)** | 0.999 (0.999–0.999) | **0.009 (0.000–0.029)** | **0.033 (0.000–0.097)** |
| | ECG-FM | 0.004 (0.000–0.014) | 0.987 (0.951–0.999) | 0.000 (0.000–0.000) | **1.000 (1.000–1.000)** | 0.000 (0.000–0.000) | 0.000 (0.000–0.000) |
| | DeepECG-SSL | 0.008 (0.001–0.025) | 0.998 (0.997–1.000) | 0.000 (0.000–0.000) | **1.000 (1.000–1.000)** | 0.000 (0.000–0.000) | 0.000 (0.000–0.000) |
| | CardioLoom | **0.025 (0.003–0.073)** | **0.999 (0.998–1.000)** | 0.000 (0.000–0.000) | **1.000 (1.000–1.000)** | 0.000 (0.000–0.000) | 0.000 (0.000–0.000) |

Values are estimates (95% cluster-bootstrap CIs). NE indicates fewer than 950 valid replicates among 1000. Sensitivity, specificity, F1 score, and MCC use the Zhongshan validation-F1 threshold. Bold type marks the highest displayed estimate within each disease.

**Supplementary Table 27: Detailed task-specific performance comparison on the external BIDMC cohort.**

| Disease | Model | AUPRC | AUROC | Sensitivity | Specificity | F1 Score | MCC |
|---|---|---|---|---|---|---|---|
| LAE | Scratch CNN–Transformer | 0.614 (0.609–0.619) | 0.616 (0.613–0.620) | **0.556 (0.551–0.561)** | 0.590 (0.585–0.595) | 0.577 (0.573–0.581) | 0.146 (0.140–0.152) |
| | ECG-FM | 0.608 (0.603–0.613) | 0.596 (0.592–0.600) | 0.526 (0.521–0.532) | 0.595 (0.590–0.600) | 0.556 (0.551–0.560) | 0.121 (0.114–0.128) |
| | DeepECG-SSL | **0.634 (0.629–0.639)** | **0.635 (0.631–0.639)** | 0.498 (0.493–0.503) | **0.669 (0.665–0.674)** | 0.554 (0.549–0.558) | **0.169 (0.162–0.176)** |
| | CardioLoom | 0.630 (0.624–0.635) | **0.635 (0.631–0.639)** | 0.554 (0.549–0.560) | 0.607 (0.602–0.612) | **0.580 (0.576–0.585)** | 0.161 (0.154–0.168) |
| BAE | Scratch CNN–Transformer | 0.590 (0.583–0.598) | 0.727 (0.723–0.731) | 0.231 (0.225–0.237) | 0.946 (0.944–0.948) | 0.347 (0.339–0.355) | 0.263 (0.256–0.271) |
| | ECG-FM | 0.634 (0.627–0.641) | 0.764 (0.761–0.768) | 0.242 (0.235–0.248) | 0.958 (0.956–0.960) | 0.366 (0.359–0.374) | 0.302 (0.294–0.309) |
| | DeepECG-SSL | 0.643 (0.636–0.650) | 0.771 (0.768–0.775) | 0.243 (0.237–0.250) | 0.959 (0.957–0.960) | 0.368 (0.361–0.377) | 0.305 (0.298–0.312) |
| | CardioLoom | **0.651 (0.644–0.658)** | **0.775 (0.772–0.779)** | **0.248 (0.241–0.254)** | **0.961 (0.960–0.963)** | **0.376 (0.368–0.384)** | **0.316 (0.309–0.324)** |
| HFrEF | Scratch CNN–Transformer | 0.507 (0.482–0.531) | 0.879 (0.872–0.887) | 0.353 (0.333–0.375) | 0.970 (0.968–0.973) | 0.439 (0.417–0.460) | 0.405 (0.383–0.428) |
| | ECG-FM | 0.661 (0.638–0.682) | 0.928 (0.923–0.933) | 0.442 (0.420–0.465) | 0.983 (0.981–0.985) | 0.555 (0.534–0.576) | 0.539 (0.518–0.561) |
| | DeepECG-SSL | 0.665 (0.641–0.686) | 0.929 (0.924–0.935) | **0.495 (0.472–0.516)** | 0.976 (0.974–0.979) | **0.582 (0.561–0.601)** | **0.554 (0.532–0.574)** |
| | CardioLoom | **0.677 (0.655–0.698)** | **0.931 (0.925–0.936)** | 0.439 (0.417–0.460) | **0.984 (0.982–0.986)** | 0.556 (0.535–0.576) | 0.544 (0.523–0.563) |
| oHCM | Scratch CNN–Transformer | 0.059 (0.052–0.067) | 0.644 (0.631–0.657) | 0.072 (0.059–0.086) | 0.985 (0.984–0.986) | 0.091 (0.075–0.108) | 0.075 (0.058–0.092) |
| | ECG-FM | 0.081 (0.072–0.092) | 0.694 (0.681–0.706) | 0.089 (0.075–0.101) | 0.989 (0.989–0.990) | 0.122 (0.104–0.139) | 0.114 (0.096–0.132) |
| | DeepECG-SSL | 0.077 (0.067–0.089) | 0.669 (0.655–0.681) | **0.098 (0.082–0.115)** | 0.987 (0.986–0.988) | **0.126 (0.105–0.146)** | 0.112 (0.092–0.134) |
| | CardioLoom | **0.087 (0.076–0.100)** | **0.701 (0.689–0.713)** | 0.067 (0.054–0.079) | **0.994 (0.993–0.995)** | 0.105 (0.087–0.124) | **0.116 (0.094–0.137)** |
| VA | Scratch CNN–Transformer | 0.924 (0.895–0.948) | 0.639 (0.560–0.708) | 0.047 (0.032–0.064) | 0.963 (0.915–1.000) | 0.089 (0.061–0.120) | 0.014 (-0.059–0.068) |
| | ECG-FM | **0.941 (0.920–0.960)** | **0.681 (0.607–0.750)** | **0.097 (0.074–0.124)** | 0.975 (0.937–1.000) | **0.176 (0.137–0.219)** | **0.077 (0.028–0.114)** |
| | DeepECG-SSL | 0.939 (0.918–0.959) | 0.673 (0.599–0.744) | 0.072 (0.052–0.095) | 0.975 (0.933–1.000) | 0.134 (0.099–0.173) | 0.057 (0.002–0.097) |
| | CardioLoom | 0.939 (0.918–0.959) | 0.674 (0.597–0.742) | 0.067 (0.048–0.089) | **0.988 (0.960–1.000)** | 0.126 (0.091–0.164) | 0.070 (0.030–0.099) |

| Disease | Model | AUPRC | AUROC | Sensitivity | Specificity | F1 Score | MCC |
|---|---|---|---|---|---|---|---|
| RAE | Scratch CNN–Transformer | 0.634 (0.627–0.641) | 0.677 (0.673–0.680) | 0.025 (0.023–0.026) | **0.997 (0.996–0.997)** | 0.048 (0.044–0.051) | 0.093 (0.087–0.099) |
| | ECG-FM | 0.640 (0.633–0.646) | 0.681 (0.676–0.685) | 0.050 (0.048–0.053) | 0.992 (0.991–0.992) | 0.095 (0.090–0.100) | 0.129 (0.123–0.135) |
| | DeepECG-SSL | 0.654 (0.647–0.661) | 0.685 (0.681–0.689) | **0.079 (0.076–0.082)** | 0.990 (0.989–0.991) | **0.145 (0.140–0.151)** | **0.173 (0.167–0.179)** |
| | CardioLoom | **0.661 (0.654–0.667)** | **0.707 (0.704–0.711)** | 0.030 (0.028–0.032) | 0.996 (0.996–0.997) | 0.057 (0.054–0.061) | 0.105 (0.099–0.110) |
| TR | Scratch CNN–Transformer | 0.328 (0.318–0.339) | 0.788 (0.782–0.793) | 0.546 (0.535–0.558) | **0.841 (0.838–0.844)** | 0.394 (0.386–0.403) | 0.307 (0.297–0.316) |
| | ECG-FM | 0.395 (0.383–0.407) | **0.822 (0.817–0.827)** | 0.628 (0.617–0.639) | 0.836 (0.833–0.839) | **0.434 (0.426–0.442)** | **0.359 (0.350–0.367)** |
| | DeepECG-SSL | 0.391 (0.379–0.404) | 0.819 (0.814–0.823) | **0.654 (0.643–0.665)** | 0.818 (0.814–0.821) | 0.428 (0.419–0.436) | 0.354 (0.345–0.362) |
| | CardioLoom | **0.402 (0.390–0.414)** | 0.821 (0.816–0.825) | 0.615 (0.604–0.626) | 0.839 (0.836–0.843) | 0.432 (0.423–0.440) | 0.354 (0.345–0.363) |
| MR | Scratch CNN–Transformer | 0.269 (0.260–0.278) | 0.751 (0.745–0.756) | 0.501 (0.490–0.512) | **0.823 (0.819–0.826)** | 0.344 (0.336–0.352) | 0.249 (0.240–0.257) |
| | ECG-FM | 0.305 (0.296–0.315) | 0.789 (0.784–0.794) | 0.569 (0.558–0.580) | 0.821 (0.817–0.824) | **0.380 (0.372–0.389)** | 0.295 (0.287–0.304) |
| | DeepECG-SSL | **0.308 (0.298–0.319)** | 0.791 (0.786–0.797) | 0.616 (0.605–0.627) | 0.793 (0.789–0.797) | 0.377 (0.369–0.385) | 0.296 (0.288–0.305) |
| | CardioLoom | **0.308 (0.298–0.319)** | **0.793 (0.787–0.798)** | **0.625 (0.614–0.636)** | 0.791 (0.787–0.794) | **0.380 (0.372–0.388)** | **0.300 (0.292–0.309)** |
| AR | Scratch CNN–Transformer | 0.098 (0.088–0.108) | 0.635 (0.619–0.650) | 0.234 (0.213–0.256) | 0.899 (0.895–0.902) | 0.157 (0.142–0.173) | 0.097 (0.081–0.113) |
| | ECG-FM | **0.138 (0.123–0.157)** | **0.690 (0.677–0.703)** | **0.247 (0.225–0.270)** | 0.924 (0.921–0.927) | **0.194 (0.176–0.212)** | **0.140 (0.121–0.159)** |
| | DeepECG-SSL | 0.126 (0.113–0.142) | 0.675 (0.661–0.690) | 0.212 (0.191–0.235) | 0.938 (0.936–0.941) | 0.187 (0.170–0.206) | 0.134 (0.117–0.155) |
| | CardioLoom | 0.132 (0.117–0.151) | 0.677 (0.664–0.691) | 0.208 (0.186–0.231) | **0.943 (0.940–0.945)** | 0.190 (0.171–0.211) | 0.139 (0.119–0.161) |
| MS | Scratch CNN–Transformer | 0.109 (0.080–0.143) | 0.748 (0.717–0.777) | 0.172 (0.127–0.214) | 0.985 (0.983–0.987) | 0.188 (0.140–0.230) | 0.172 (0.124–0.215) |
| | ECG-FM | **0.201 (0.139–0.265)** | **0.818 (0.793–0.842)** | 0.254 (0.184–0.327) | 0.987 (0.985–0.989) | **0.278 (0.207–0.339)** | **0.264 (0.195–0.327)** |
| | DeepECG-SSL | 0.176 (0.124–0.232) | 0.812 (0.786–0.837) | **0.258 (0.198–0.314)** | 0.985 (0.983–0.987) | 0.268 (0.211–0.319) | 0.253 (0.197–0.303) |
| | CardioLoom | 0.177 (0.128–0.227) | 0.810 (0.783–0.833) | 0.210 (0.155–0.263) | **0.989 (0.987–0.991)** | 0.246 (0.185–0.299) | 0.236 (0.174–0.288) |

| Disease | Model | AUPRC | AUROC | Sensitivity | Specificity | F1 Score | MCC |
|---|---|---|---|---|---|---|---|
| AS | Scratch CNN–Transformer | 0.502 (0.487–0.516) | 0.584 (0.573–0.594) | **0.094 (0.086–0.102)** | 0.944 (0.938–0.949) | **0.160 (0.148–0.173)** | **0.072 (0.056–0.089)** |
| | ECG-FM | 0.510 (0.495–0.524) | 0.588 (0.577–0.598) | 0.043 (0.037–0.048) | 0.979 (0.975–0.982) | 0.080 (0.070–0.089) | 0.062 (0.046–0.078) |
| | DeepECG-SSL | **0.524 (0.509–0.538)** | **0.603 (0.593–0.613)** | 0.025 (0.021–0.029) | **0.991 (0.989–0.993)** | 0.048 (0.041–0.056) | 0.064 (0.047–0.079) |
| | CardioLoom | 0.521 (0.506–0.535) | 0.600 (0.589–0.611) | 0.036 (0.030–0.041) | 0.986 (0.983–0.988) | 0.068 (0.058–0.077) | 0.069 (0.053–0.085) |
| PS | Scratch CNN–Transformer | 0.003 (0.001–0.005) | 0.471 (0.369–0.595) | 0.000 (0.000–0.000) | 0.999 (0.998–0.999) | 0.000 (0.000–0.000) | -0.001 (-0.002–-0.001) |
| | ECG-FM | **0.004 (0.002–0.007)** | **0.653 (0.551–0.748)** | 0.000 (0.000–0.000) | 0.999 (0.999–0.999) | 0.000 (0.000–0.000) | -0.001 (-0.002–-0.001) |
| | DeepECG-SSL | 0.002 (0.001–0.004) | 0.538 (0.476–0.598) | **0.043 (0.000–0.102)** | 0.990 (0.989–0.991) | **0.014 (0.000–0.030)** | **0.014 (-0.004–0.039)** |
| | CardioLoom | 0.002 (0.001–0.003) | 0.524 (0.447–0.612) | 0.000 (0.000–0.000) | **1.000 (1.000–1.000)** | 0.000 (0.000–0.000) | -0.001 (-0.001–0.000) |
| PR | Scratch CNN–Transformer | 0.025 (0.023–0.028) | 0.515 (0.498–0.532) | 0.002 (0.001–0.005) | 0.999 (0.999–0.999) | 0.005 (0.001–0.009) | 0.008 (-0.001–0.019) |
| | ECG-FM | **0.032 (0.028–0.038)** | **0.599 (0.584–0.614)** | 0.005 (0.001–0.009) | **1.000 (0.999–1.000)** | 0.010 (0.003–0.018) | **0.026 (0.007–0.049)** |
| | DeepECG-SSL | 0.030 (0.027–0.034) | 0.595 (0.579–0.610) | 0.001 (0.000–0.003) | **1.000 (0.999–1.000)** | 0.003 (0.000–0.007) | 0.007 (-0.003–0.019) |
| | CardioLoom | 0.030 (0.027–0.034) | 0.592 (0.578–0.607) | **0.015 (0.008–0.021)** | 0.995 (0.995–0.996) | **0.024 (0.014–0.034)** | 0.020 (0.007–0.033) |
| VSD | Scratch CNN–Transformer | 0.771 (0.705–0.848) | **0.566 (0.491–0.636)** | **0.106 (0.055–0.167)** | 0.934 (0.880–0.978) | **0.188 (0.103–0.282)** | 0.060 (-0.053–0.158) |
| | ECG-FM | 0.747 (0.671–0.826) | 0.525 (0.452–0.604) | 0.055 (0.012–0.107) | 0.945 (0.896–0.988) | 0.103 (0.024–0.191) | 0.000 (-0.144–0.114) |
| | DeepECG-SSL | **0.775 (0.700–0.850)** | 0.549 (0.472–0.632) | 0.031 (0.007–0.066) | **0.989 (0.965–1.000)** | 0.061 (0.014–0.122) | 0.057 (-0.054–0.129) |
| | CardioLoom | 0.759 (0.685–0.834) | 0.523 (0.447–0.605) | 0.051 (0.020–0.091) | 0.978 (0.927–1.000) | 0.097 (0.038–0.164) | **0.063 (-0.041–0.145)** |
| PH | Scratch CNN–Transformer | 0.314 (0.305–0.323) | 0.747 (0.742–0.752) | 0.766 (0.759–0.775) | 0.600 (0.596–0.605) | 0.374 (0.367–0.380) | 0.260 (0.253–0.267) |
| | ECG-FM | **0.387 (0.377–0.398)** | **0.789 (0.785–0.794)** | 0.723 (0.715–0.732) | **0.711 (0.707–0.715)** | **0.424 (0.417–0.431)** | **0.321 (0.314–0.328)** |
| | DeepECG-SSL | 0.373 (0.362–0.383) | 0.783 (0.778–0.787) | 0.751 (0.743–0.759) | 0.674 (0.669–0.678) | 0.411 (0.404–0.418) | 0.308 (0.302–0.315) |
| | CardioLoom | 0.383 (0.373–0.393) | 0.783 (0.779–0.788) | **0.792 (0.785–0.799)** | 0.630 (0.625–0.635) | 0.400 (0.394–0.407) | 0.301 (0.294–0.307) |

| Disease | Model | AUPRC | AUROC | Sensitivity | Specificity | F1 Score | MCC |
|---|---|---|---|---|---|---|---|
| AD | Scratch CNN–Transformer | 0.300 (0.292–0.308) | 0.685 (0.680–0.690) | **0.524 (0.515–0.533)** | 0.721 (0.717–0.725) | 0.359 (0.352–0.366) | 0.196 (0.188–0.204) |
| | ECG-FM | 0.321 (0.313–0.331) | 0.707 (0.702–0.713) | 0.448 (0.440–0.458) | 0.811 (0.807–0.814) | 0.374 (0.367–0.382) | 0.228 (0.220–0.237) |
| | DeepECG-SSL | 0.336 (0.327–0.346) | **0.718 (0.713–0.723)** | 0.481 (0.472–0.491) | 0.800 (0.797–0.804) | **0.388 (0.381–0.396)** | **0.243 (0.236–0.252)** |
| | CardioLoom | **0.341 (0.333–0.352)** | **0.718 (0.713–0.724)** | 0.457 (0.447–0.466) | **0.817 (0.814–0.820)** | 0.385 (0.377–0.393) | **0.243 (0.234–0.251)** |
| PE | Scratch CNN–Transformer | 0.106 (0.096–0.117) | 0.683 (0.669–0.696) | 0.647 (0.621–0.671) | 0.619 (0.612–0.626) | 0.166 (0.155–0.177) | 0.128 (0.114–0.140) |
| | ECG-FM | 0.121 (0.108–0.134) | 0.711 (0.698–0.724) | 0.584 (0.557–0.610) | **0.697 (0.690–0.704)** | 0.181 (0.169–0.193) | 0.142 (0.127–0.156) |
| | DeepECG-SSL | 0.129 (0.118–0.143) | 0.727 (0.715–0.740) | **0.749 (0.726–0.772)** | 0.604 (0.596–0.611) | 0.184 (0.173–0.195) | 0.168 (0.155–0.180) |
| | CardioLoom | **0.138 (0.125–0.153)** | **0.738 (0.724–0.751)** | 0.653 (0.626–0.681) | 0.693 (0.685–0.699) | **0.198 (0.185–0.210)** | **0.173 (0.158–0.187)** |

Values are estimates (95% cluster-bootstrap CIs). NE indicates fewer than 950 valid replicates among 1000. Sensitivity, specificity, F1 score, and MCC use the Zhongshan validation-F1 threshold. Bold type marks the highest displayed estimate within each disease.

**Supplementary Table 28: Label-efficiency analysis for the final four models.**

| Training patients (%) | Model | Internal macro-AUPRC | External 203-combination mean |
|---|---|---|---|
| 1% | Scratch CNN–Transformer | 0.206 (0.202–0.212) | 0.197 |
| 1% | ECG-FM | 0.367 (0.361–0.376) | 0.317 |
| 1% | DeepECG-SSL | 0.374 (0.367–0.382) | 0.314 |
| 1% | CardioLoom | **0.420 (0.413–0.428)** | **0.330** |
| 10% | Scratch CNN–Transformer | 0.366 (0.359–0.375) | 0.310 |
| 10% | ECG-FM | 0.515 (0.506–0.524) | 0.370 |
| 10% | DeepECG-SSL | 0.527 (0.519–0.536) | 0.394 |
| 10% | CardioLoom | **0.559 (0.551–0.568)** | **0.401** |
| 100% | Scratch CNN–Transformer | 0.549 (0.541–0.560) | 0.394 |
| 100% | ECG-FM | 0.608 (0.600–0.618) | 0.414 |
| 100% | DeepECG-SSL | 0.615 (0.607–0.625) | 0.424 |
| 100% | CardioLoom | **0.624 (0.616–0.634)** | **0.425** |

Internal macro-AUPRC is the mean (95% patient-bootstrap CI) across 45 ECG targets. External values are equal-weight means across 203 combinations of tasks and cohorts. Bold type marks the highest displayed estimate within each training fraction.

**Supplementary Table 29: Internal report-aware versus ordinary-CLIP macro comparisons.**

| Target set | Budget | Metric | Difference (95% CI) | q | Valid draws |
|---|---|---|---|---|---|
| ECG interpretation | 100% | AUPRC | 0.0029 (-0.0038–0.0094) | 1.000 | 1000 |
| ECHO diagnosis | 100% | AUPRC | 0.0026 (CI NE) | NE | 839 |
| Rare disease | 100% | AUPRC | 0.0075 (CI NE) | NE | 847 |
| Overall | 100% | AUPRC | 0.0030 (CI NE) | NE | 711 |
| ECG interpretation | 1% | AUPRC | 0.0103 (0.0044–0.0158) | 0.001 | 1000 |
| ECG interpretation | 10% | AUPRC | 0.0131 (0.0083–0.0183) | 0.001 | 1000 |
| ECG interpretation | 100% | AUROC | 0.0002 (-0.0008–0.0015) | 1.000 | 1000 |
| ECHO diagnosis | 100% | AUROC | 0.0018 (CI NE) | NE | 839 |
| Rare disease | 100% | AUROC | -0.0448 (CI NE) | NE | 847 |
| Overall | 100% | AUROC | -0.0016 (CI NE) | NE | 711 |
| ECG interpretation | 1% | AUROC | 0.0147 (0.0081–0.0206) | 0.001 | 1000 |
| ECG interpretation | 10% | AUROC | 0.0047 (0.0035–0.0059) | 0.001 | 1000 |

Differences are CardioLoom minus ordinary CLIP. NE denotes fewer than 950 valid complete-target bootstrap replicates. Benjamini–Hochberg adjustment is described in the Methods. Source Data provides comparisons for 89 targets at full budget and 45 ECG targets at each of the 1% and 10% budgets.

**Supplementary Table 30: Operating characteristics for rare disease prediction at validation-targeted 99% specificity thresholds.**

| Disease | Pos./N (%) | Model | TP | FP | Sens. | Spec. | PPV | NPV | NNS | N/FP |
|---|---|---|---|---|---|---|---|---|---|---|
| CA | 115/108,173 (0.106%) | Scratch | 43 | 1,233 | 0.374 | 0.989 | 0.034 | **0.999** | 2,516 | 88 |
| | 115/108,173 (0.106%) | ECG-FM | 39 | **1,121** | 0.339 | **0.990** | 0.034 | **0.999** | 2,774 | **96** |
| | 115/108,173 (0.106%) | DeepECG-SSL | 34 | 1,241 | 0.296 | 0.989 | 0.027 | **0.999** | 3,182 | 87 |
| | 115/108,173 (0.106%) | CardioLoom | **57** | 1,220 | **0.496** | 0.989 | **0.045** | **0.999** | **1,898** | 89 |
| ARVC | 4/108,173 (0.004%) | Scratch | 1 | 1,123 | 0.250 | **0.990** | 0.001 | **1.000** | 108,173 | 96 |
| | 4/108,173 (0.004%) | ECG-FM | 1 | 1,154 | 0.250 | 0.989 | 0.001 | **1.000** | 108,173 | 94 |
| | 4/108,173 (0.004%) | DeepECG-SSL | 0 | 1,135 | 0.000 | **0.990** | 0.000 | **1.000** | — | 95 |
| | 4/108,173 (0.004%) | CardioLoom | **3** | **1,104** | **0.750** | **0.990** | **0.003** | **1.000** | **36,058** | **98** |
| TC | 14/108,173 (0.013%) | Scratch | 0 | **217** | 0.000 | **0.998** | 0.000 | **1.000** | — | **498** |
| | 14/108,173 (0.013%) | ECG-FM | 1 | 1,064 | 0.071 | 0.990 | 0.001 | **1.000** | 108,173 | 102 |
| | 14/108,173 (0.013%) | DeepECG-SSL | 2 | 1,082 | 0.143 | 0.990 | 0.002 | **1.000** | 54,086 | 100 |
| | 14/108,173 (0.013%) | CardioLoom | **3** | 1,071 | **0.214** | 0.990 | **0.003** | **1.000** | **36,058** | 101 |
| AC | 6/108,173 (0.006%) | Scratch | 0 | **131** | 0.000 | **0.999** | 0.000 | **1.000** | — | **826** |
| | 6/108,173 (0.006%) | ECG-FM | 0 | 1,085 | 0.000 | 0.990 | 0.000 | **1.000** | — | 100 |
| | 6/108,173 (0.006%) | DeepECG-SSL | 0 | 1,016 | 0.000 | 0.991 | 0.000 | **1.000** | — | 106 |
| | 6/108,173 (0.006%) | CardioLoom | **2** | 1,124 | **0.333** | 0.990 | **0.002** | **1.000** | **54,086** | 96 |
| NCVM | 16/108,173 (0.015%) | Scratch | 0 | **1,028** | 0.000 | **0.990** | 0.000 | **1.000** | — | **105** |
| | 16/108,173 (0.015%) | ECG-FM | 0 | 1,156 | 0.000 | 0.989 | 0.000 | **1.000** | — | 94 |
| | 16/108,173 (0.015%) | DeepECG-SSL | 0 | 1,133 | 0.000 | **0.990** | 0.000 | **1.000** | — | 95 |
| | 16/108,173 (0.015%) | CardioLoom | **2** | 1,130 | **0.125** | **0.990** | **0.002** | **1.000** | **54,086** | 96 |

Thresholds target at least 99% specificity in Zhongshan validation data and are fixed for testing. Scratch denotes the supervised control. Pos./N denotes positive/evaluated ECGs. Sens. and Spec. denote sensitivity and specificity, respectively. PPV and NPV denote positive and negative predictive values. NNS = N/TP is the number evaluated per detected positive ECG, and N/FP is the number evaluated per false-positive ECG. An em dash denotes a zero denominator. Bold type marks the most favorable displayed value within each disease.

**Supplementary Table 31: CA uncertainty at validation-targeted 99% specificity.**

| Metric | Scratch | ECG-FM | DeepECG-SSL | CardioLoom |
|---|---|---|---|---|
| Sensitivity | 0.374 (0.200–0.621) | 0.339 (0.081–0.747) | 0.296 (0.049–0.596) | **0.496 (0.221–0.750)** |
| Specificity | 0.989 (0.988–0.990) | **0.990 (0.989–0.991)** | 0.989 (0.987–0.989) | 0.989 (0.988–0.990) |
| PPV | 0.034 (0.008–0.065) | 0.034 (0.005–0.075) | 0.027 (0.002–0.060) | **0.045 (0.008–0.089)** |
| NPV | 0.999 (0.999–1.000) | 0.999 (0.998–1.000) | 0.999 (0.998–1.000) | **0.999 (0.999–1.000)** |
| Positive calls /1000 ECGs | 11.80 (10.70–12.94) | 10.72 (9.55–11.87) | 11.79 (10.69–12.96) | 11.81 (10.63–13.02) |
| False-positive calls /1000 ECGs | 11.40 (10.34–12.42) | **10.36 (9.31–11.38)** | 11.47 (10.49–12.51) | 11.28 (10.28–12.30) |
| ECGs per detected positive | 2515.65 (1271.86–10648.29) | 2773.67 (1212.00–17844.80) | 3181.56 (1359.96–36532.67) | **1897.77 (895.71–10727.07)** |
| ECGs per false positive | 87.73 (80.50–96.72) | **96.50 (87.84–107.47)** | 87.17 (79.96–95.31) | 88.67 (81.30–97.26) |

Operating points and ECG counts are given in Supplementary Table 30. Estimates are accompanied by percentile 95% confidence intervals from paired patient-cluster bootstrap resampling with fixed classifiers and validation-selected thresholds. There were 1000 valid replicates for 31 of the 32 model-metric estimates. ECGs per detected positive for DeepECG-SSL had 996 valid replicates. The test population contains 115 CA-positive ECGs from 11 patients among 108,173 ECGs from 35,049 patients. Paired differences are provided in Source Data.